\documentclass[11pt, twoside, a4paper, openright]{book}
\usepackage[utf8]{inputenc}
\usepackage[T1]{fontenc}
\usepackage[english]{babel}

\title{causal-learning-additive-noise-KAP-Benjamin}
\author{Kap Benjamin\\016060274F}
\date{February 2021}

\usepackage{mathtools}

\usepackage{xcolor}

\usepackage{subcaption}

\usepackage[section]{placeins}

\usepackage{hyperref}
\usepackage[capitalise]{cleveref}

\usepackage{float}

\usepackage[margin=2.9cm,inner=3.4cm]{geometry}

\usepackage{graphicx}
\graphicspath{ {graphics/} }

\usepackage[backend=bibtex,natbib, style=authoryear-comp, sorting=none]{biblatex}

\bibliography{references}

\usepackage{afterpage}

\usepackage{pdflscape}

\usepackage{amssymb}
\usepackage{amsmath}

\usepackage{algorithm}
\usepackage[noend]{algpseudocode}

\usepackage{ulem}
\usepackage{cancel}

\usepackage{adjustbox}
\usepackage[graphicx]{realboxes}
\usepackage{rotating}

\newcommand{\indep}{\perp \!\!\! \perp}

\usepackage{fancyhdr}
\begin{document}

\pagenumbering{gobble}

\frontmatter
\newcommand{\HRule}{\rule{\linewidth}{0.5mm}}

\begin{titlepage}

\begin{center}

\includegraphics{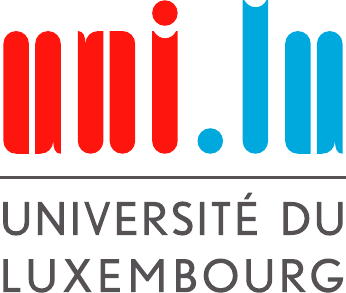}

\vspace{1cm}

\textsc{\Large Faculty of Science, Technology and Communication}\\[1.0cm]

\vspace{1cm}

\HRule \\[0.4cm]
{\huge \bfseries The Effect of Noise Level on Causal Identification with Additive Noise Models}\\
\HRule \\[1.5cm]

\begin{minipage}{0.8\textwidth}
\begin{center}
{\Large Thesis Submitted in Partial Fulfillment of the Requirements
for the Degree of Master in Information and Computer Sciences}
\end{center}
\end{minipage}

\vspace{4cm}
\begin{minipage}[t]{0.4\textwidth}
\begin{flushleft} \large
\emph{Author:}\\
Benjamin \textsc{Kap}
\end{flushleft}
\end{minipage}
\begin{minipage}[t]{0.4\textwidth}
\begin{flushright} \large
\emph{Supervisor:} \\
Prof.~Thomas \textsc{Engel} \\
\vspace{.5em}
\emph{Reviewer:} \\
Prof.~Nicolas \textsc{Navet} \\
\vspace{.5em}
\emph{Advisor:} \\
Dr.~Marharyta \textsc{Aleksandrova}
\end{flushright}
\end{minipage}

\vfill

{\large May 2021}

\end{center}

\end{titlepage}

\clearpage\mbox{}\clearpage

\chapter{Abstract}
In recent years a lot of research has been conducted within the area
of causal inference and causal learning. Many methods have been developed
to identify the cause-effect pairs in models and have been successfully
applied to observational real-world data in order to determine
the direction of causal relationships. Many of these methods require simplifying assumptions,
such as absence of confounding, cycles, and selection bias. Yet in bivariate
situations causal discovery problems remain challenging. One class of such
methods, that also allows tackling the bivariate case, is based on Additive Noise Models (ANMs). Unfortunately, one aspect
of these methods has not received much attention until now: what is the impact of different noise levels 
on the ability of these methods to identify the direction of the causal relationship.
This work aims to bridge this gap with the help of an empirical study. For this work, we considered
bivariate cases, which is the most elementary form of a causal discovery problem
where one needs to decide whether $X$ causes $Y$ or $Y$ causes $X$, given
joint distributions of two variables $X$, $Y$. Furthermore, two specific
methods have been selected, \textit{Regression with Subsequent Independence Test}
and \textit{Identification using Conditional Variances}, which have been tested
with an exhaustive range of ANMs where the additive noises' levels gradually change
from 1\% to 10000\% of the causes' noise level (the latter remains fixed). 
Additionally, the experiments in this work 
consider several different types of distributions as well as linear and non-linear
ANMs. The results of the experiments show that these methods can fail to capture
the true causal direction for some levels of noise.
\clearpage\mbox{}\clearpage

\tableofcontents

\listoffigures

\mainmatter

\pagenumbering{arabic}


\chapter{Introduction}
Due to the technological and computational advances during the last decades,
scientist were able to tackle non-trivial problems from different
research areas successfully. One of these research areas is causality.
The fundamentals of causality is to determine causal relationship
between two or more variables in a system. For example, given altitude and
temperature we want to answer the question if temperature has an effect on altitude, or
if altitude has an effect on temperature. This is of particular interest
since if such a causal relationship is known then one can predict
the effects on a system in case of intervention or perturbation. 
One method to determine causal relationships in a system is controlled
experimentation (A/B tests) in which there are two identical
groups with only one variation. The only variable that is varied (intervened on) is the potential cause.
This procedure allows estimating causal effect of this variable in the given system.
For example, testing the efficacy of medications is done within A/B tests. 
The control group
receives no medication or a placebo, and the intervention group receives the 
real medication. The results often show the true effect (if any) of the
medication on the human health.
However, such tests are often too expensive, unethical or even
technically impossible to execute. Therefore, it is of great interest
to determine causal relationships from observational data only (e.g., structure learning).

\section{State of the Art}
Structure learning is the procedure to determine causal relationship directions
from observational data only and representing these as a (causal) graph. The basic idea
emerged from \citet{wright1921correlation} as \textit{path analysis}. This is used
to describe directed dependencies among a set of variables and includes various models
such as ANOVA, ANCOVA, etc. In his work, Wright made a distinction between three possible
types of causal substructures which were allowed in a directed acyclic (no cycles) graph:
1) $X \to Y \to Z$, 2) $X \gets Y \to Z$, and 3) $X \to Y \gets Z$. \citet{rebane} developed 
an algorithm to recover directed acyclic graphs from statistical data, which relied
on this distinction of the previously 3 mentioned substructures. In general, one can easily
identify the skeleton of a graph (that is the graph without arrows on the edges) and then
partially identify the arrows (partially, because the three substructures' skeletons are identical
but only 3) is distinguishable form others).
\citet{bookspirtes1993,spirtes2000causation} used Bayes networks to axiomatize the connection
between causal structure and probabilistic independence and formalized under what assumptions
one could draw causal knowledge from observational data only. Furthermore, they also formalized
how incomplete causal knowledge could be used for causal intervention.
Judea Pearl presented in his work \parencite[]{judea2000causality} a comprehensive theory
of causality and unified the probabilistic, manipulative, counterfactual, and structural approaches
to causation. Judea also introduced precise mathematical definitions
of causal analysis for the standard curricula of statistics.
From the work \citet{judea2000causality} we have the following key point: 
if there is a statistical association,
e.g. two variables $X,Y$ are dependent, then one of the following is true: 1) there is a causal relationship, 
either $X$ has an effect on $Y$ or $Y$ has an effect on $X$; 2) there is a common cause 
(confounder) that has effect on both $X$ and $Y$; 3) there is a possibly unobserved common effect of $X$ and $Y$ that is conditioned upon
data acquisition (selection bias); or 4) there can be a combination of these.
From there on a lot of research has been conducted to develop
theoretical approaches and methods for identifying causal relationships from 
observational data only. Before we present some major works from the last two decades,
we introduce the common concept behind all these approaches in a short formal manner.

In general, all these methods exploit the complexity of the marginal and conditional probability
distributions in some way (e.g., \citet{janzing2012information, sgouritsa2015inference}) and under 
certain assumptions these methods are then able to solve the task of causal discovery.
Let $C$ denote the cause and $E$ the effect. In a system with two or more variables
we might have cause-effect pairs and then their joint density can be expressed with $p_{C,E}(c,e)$.
This joint density can be factorized into either 1) $p_C(c)\cdot P_{E|C}(e|c)$ or 2) $p_E(e)\cdot P_{C|E}(c|e)$.
The idea is then that 1 gives models of lower total complexity than 2 and this allows
us to draw conclusions about the causal relationship direction. Intuitively this makes sense, because
the effect contains information from the cause but not vice-versa (of course under the assumption
that there are no cycles aka feedback loops). Therefore, 2 has at least as much complexity as 1.
However, the definition of complexity is ambiguous. For example, one can say that
"\textit{$p_C$ contains no information about $P_{E|C}(e|c)$}" and then is able
to draw partial conclusions about the causal direction of the given system.
This complexity question is often colloquially referred to as "\textit{breaking
the symmetry}" (that is $p_C(c)\cdot P_{E|C}(e|c) \neq p_E(e)\cdot P_{C|E}(c|e)$).\\

\citet{DBLP:journals/corr/abs-1301-3857} addressed the problem of learning
the structure of a Bayesian network in domains which contain continuous
variables. In their work they showed that in probabilistic networks with continuous variables
one can use Gaussian Process priors to compute marginal likelihoods for structure learning.\\
\citet{kano2003causal} developed a model for causal inference
using non-normality of observed data and improved path analysis \parencite[]{wright1921correlation}
using non-normal data. \citet{shimizu2006linear} proposed a method on how
to determine the complete causal graph of continuous data under three
assumptions: the data generating process is linear, no unobserved confounders,
and noise variables have non-Gaussian distributions of non-zero variances.
This method was not scale-invariant, but in a later work \parencite[]{shimizu2014direct}
this problem had been addressed and a new method was proposed which was
guaranteed to converge to the right solution within a small fixed number of steps
if the data strictly followed the model.\\ \citet{sun2006causal} introduced a method
based on comparing the conditional distributions of variables given their direct
causes for all hypothetical causal directions and choosing the most plausible one (Markov kernels).
Those Markov kernels which maximize the conditional entropies constrained by their observed
expectation, variance and covariance with its direct causes based on their given domain
are considered as plausible kernels. \citet{sun2008causal} continued the work on kernels
by using the concept of reproducing kernel Hilbert spaces.\\
\citet{hoyer2008nonlinear} generalized the linear framework of additive noise models
to nonlinear models. Additive noise models are models in which the effect is a function
of the cause and some random and non-observed additive noise term.
For continuous variables, earlier methods often assumed linear models
for the independence tests. However, if data contained non-Gaussian variables, then this
can help in distinguishing the causal directions and identify the causal graph.\\
\citet{janzing2009telling} proposed a method for inferring linear causal
relationships among multi-dimensional variables by factorizing the joint
distribution into products with marginal and conditional distributions (as seen above with 1 and 2).
Then, in one of these products the factors (e.g., $P(E)$ and $P(C|E)$) satisfy non-generic
relations indicating that $E \to C$ is wrong.\\
\citet{mooij2009regression} introduced a method which minimizes the statistical
dependence between the regressors and residuals. If residuals (the difference between actual output
and predicted output) are no longer
dependent on the input, then regression can successfully model the dependence of the output
on the input. This method does not need to 
assume a particular distribution of the noise because any form of regression
can be used (e.g., Linear Regression) and is well suited for the task
of causal inference in additive noise models.\\
\citet{stegle2010probabilistic} created a method to model observed data by using
probabilistic latent (hidden) variable models, which incorporate the effects of
unobserved noise. To analyze the joint density of cause and effect, 
the effect is modeled as a function of the cause and some independent noise (not necessarily additive).
With general non-parametric priors on this function and on the distribution of the cause
the causal direction is then determined by using standard Bayesian model selection.\\
\citet{mooij2011causal} introduced a method to determine causal relationship
in cyclic additive noise models and state that such models are generally
identifiable in the bivariate, Gaussian-noise case. Their method works
for continuous data and can be seen as a special case of nonlinear
independent component analysis.\\
\citet{zhang2012identifiability} tested the stability of the identifiability
of post-nonlinear models with two variables and listed all cases in which this 
model is not identifiable anymore. Furthermore, they showed how to approach
multivariate cases with post-nonlinear models.\\
\citet{daniusis2012inferring} showed that even in deterministic cases (noise-free
cases) there are asymmetries that can be exploited for causal inference.
Their method is based on the idea that if $X \to Y$ then the distribution of
$X$ and the function mapping $X$ to $Y$ must be independent since they correspond
to independent mechanisms of nature.\\
\citet{hyvarinen2013pairwise} proposed a method which is based
on the likelihood ratio under the linear non-Gaussian acyclic
model (LiNGAM, \citet{shimizu2014lingam}) and therefore not resorting to
independent component analysis algorithm as previous methods did.\\
\citet{10.1093/biomet/ast043} proved full identifiability of linear Gaussian 
structural equation models if all the noise variables have the same variance
(\textit{full identifiability} means that not only the skeleton of the causal graph
is recoverable but also the arrows).\\
\citet{peters2014causal} proposed a method that can identify the directed acyclic
graph from the distribution under mild conditions. In contrast, other methods
assumed faithfulness and could only identify the Markov equivalence class of the graph
(\textit{Markov equivalence class} refers to the class of graphs where all graphs have the same
skeletons).\\
\citet{nowzohour2016score} proposed to use penalized likelihood scores
instead of independence scores to determine the true causal graph from the
Markov equivalence class.\\
\citet{park2019identifiability} and \citet{chen2019causal} proved
that linear Gaussian models where noise variables have different error variance
can be identifiable by ordering variables according to the law of total variances
and then performing independence tests between variables.\\

Despite all the research in the past years one small but nonetheless important
aspect of causal discovery methods has not received much attention: can
different noise levels have an impact on the correctness of these methods.
In real world, observational data often differs in terms of noise level. Usually,
these levels do not differ significantly but it can occur that noise levels
change drastically from cause to effect. For example, if the data collection
process has a lot of interference (e.g., in outer space) then such noise levels
can differ a lot.\\

In this work we will focus on the \textit{Additive Noise Models} (ANMs)
as they are well established and yielded many good results (\citet{kpotufe2014consistency}). 
ANMs are heavily based on the presence of noise and thus our research question is then formulated as follows: 
\textit{how do different noise levels of the additive noise impact
the correctness of ANM methods?}
For this work, two specific methods have been selected to test this question:
\textit{Regression with Subsequent Independence Test (RESIT)} \parencite[]{peters2014causal}
and \textit{Identification using Conditional Variances (Uncertainty Scoring)} \parencite[]{park2020identifiability}.
Both methods yielded good results and can be used even when variables have different distribution types 
(e.g., Laplace).
The next chapter introduces Additive Noise Models in a formal manner, followed
by standard notation definition. Chapters 3 and 4 contain the theory and experiments
of the first and the second methods respectively. The last chapter draws conclusions and discusses possible future work.


\chapter{Additive Noise Models and Notation}
The two selected methods for this work can be used for additive noise models, 
although the second method is not limited to it (\citet{park2019identifiability}).
In general, such causal discovery methods exploit the additivity of the noise in
order to determine the causal relationship between two or more variables.
We will give now a short definition of additive noise models for the bivariate
case only. For more details and for multivariate cases please refer to \citet{hoyer2008nonlinear} and
\citet{peters2014causal}.\\

Let $X, Y \in \mathbb{R}$ be the cause and effect, respectively. Let there be also  $m$
latent (hidden) causes $U =(U_1,\dots,U_m) \in \mathbb{R}^m$.
Then the causal relationship can be modeled as
\[\begin{cases}
Y = f(X,U_1,\cdots,U_m)\\
X \indep U, X \sim p_X(x), U \sim p_U(u_1,\cdots, u_m)
\end{cases}
\]
where $f:\mathbb{R} \times \mathbb{R}^m \to \mathbb{R}$ is a linear or nonlinear
function, and $p_X(x)$ and $p_U(u_1,\cdots, u_m)$ are the joint densities of the
observed cause $X$ and the latent causes $U$. We are assuming that there is
no confounding, no selection bias, and no feedback loop between $X$ and $Y$ and
therefore $X$ and $U$ are independent, which is denoted as $X \indep U$.
Since the latent causes $U$ are unobserved, their influence can be summarized by
a single noise variable $N_y \in \mathbb{R}$, and the model can be defined as follows:
\[\begin{cases}
Y = f(X,N_y)\\
X \indep N_y, X \sim p_X(x), N_y \sim p_{N_y}(n_y).
\end{cases}\]

\subsubsection*{Notations}

In the experiments, we are considering both linear and nonlinear additive noise models:
$$Y = \beta X + N_y \text{ with } \beta \in \mathbb{R} \text{, for the linear case}$$ and
$$Y = \beta X^{\alpha} + N_y \text{ with } \beta,\alpha \in \mathbb{R} \text{, for the nonlinear case}.$$

\begin{figure}[h]
  \centering
  \includegraphics[scale=0.2]{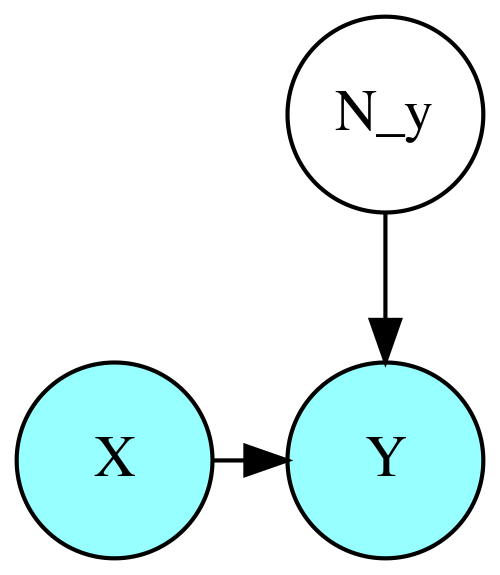}
\caption{A simple directed acyclic graph showing causal relationships between variables.}
\label{fig:intro1}
\end{figure}

\begin{figure}[h]
  \centering
  \includegraphics[scale=0.2]{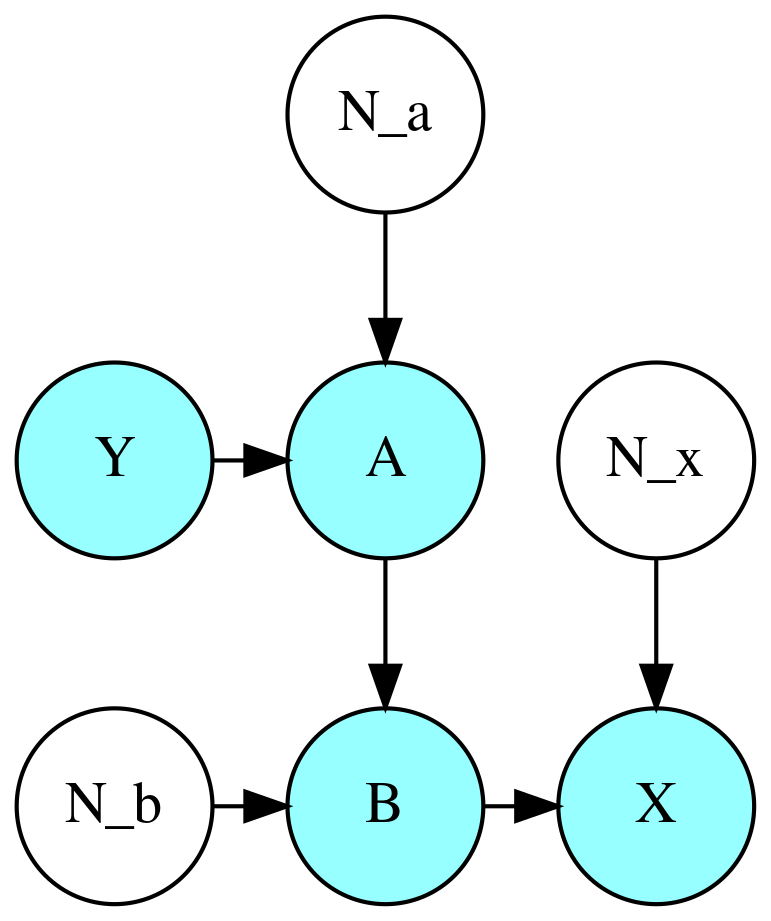}
\caption{Another directed acyclic graph showing causal relationships between variables.}
\label{fig:intro2}
\end{figure}

The arrow in "$X \to Y$" signifies $X$ has an effect on $Y$, or in other words, $X$ is a cause of $Y$. $X$ and $N_y$ can be
drawn from one of the following distributions: the normal distribution denoted by the
calligraphic letter $\mathcal{N}$, the uniform distribution denoted by the 
calligraphic letter $\mathcal{U}$, or the laplace distribution denoted by
the calligraphic letter $\mathcal{L}$. For example, throughout this work 
"\textit{X is drawn from a normal distribution}" is denoted as
$X \sim \mathcal{N}$ or $X \sim \mathcal{N}(\mu_x, \sigma^2_x)$ with
$\mu_x$ standing for the mean and $\sigma^2_x$ for the variance.
The \textit{Structural causal model} of a directed acyclic graph (DAG) refers
to the equations of the cause and effects. For example, the equation $Y = f(X, N_y)$ 
refers to the DAG in \cref{fig:intro1} and also composes
the entire structural equation model of that DAG (since there are no other variables
in that DAG). In \cref{fig:intro2} the structural equation model looks as follows:
\[
\begin{cases}
A = f(Y, N_a) \\
B = f(A, N_b) \\
X = f(B, N_x)
\end{cases}
\]

As we are only considering bivariate cases, we only have one equation and therefore
\textit{structural equation model}, \textit{structural equation} and \textit{equation}
are used interchangeably. Lastly, we introduce the four following terms (with abbreviations): parent (Pa), antecedent (An), descendent (De) and non-descendent (Nd). These notations are used to describe the relationship between two variables in a causal graph. Parent is the direct antecedent of a node (e.g., $Pa(A) = Y$ in \cref{fig:intro2}). A descendent of a (starting) variable is any variable which can be reached from the
starting variable following a directed path (e.g., $De(Y) = \{A, B, X\}$ in \cref{fig:intro2}). Non-descendent
is the opposite, i.e. cannot be reached following a directed path (e.g., $Nd(Y) = \{N_a, N_b, N_x\}$). Finally,
the antecedent is any starting variable for a descendent (e.g., $An(X) = \{Y, A, B, N_x, N_a, N_b\}$).

\newpage
\section{Experiment Layout}
One additional part of the experiments is to see whether splitting
data into training and test data is beneficial or not. However, as some experiments
did not allow us to split data and/or the results were too small to be contained in an own
subsection it is not quite obvious from the contents table where data splitting occurred or not.
The following \cref{layouttable} thus shall provide an overview where we applied splitting data and where not.
Furthermore, throughout this work splitting data is referred to as \textit{decoupled estimation}
and not splitting data is referred to as \textit{coupled estimation}, both terms originate from
\citet{kpotufe2014consistency}.

\begin{table}[h]
\begin{center}
\begin{tabular}{|c|c|c|c|}
     \hline
     \textbf{Section} & \textbf{Type} & \textbf{Decoupled} & \textbf{Coupled} \\\hline
     \cref{res1}& RESIT \& different noise levels & \checkmark & \\\hline
     \cref{res3}& RESIT \& different noise levels & & \checkmark \\\hline
     \cref{res2}& \parbox[t]{5cm}{RESIT \& different noise levels \\\& no prior assumption} & \checkmark & \checkmark \\\hline
     \cref{res4}& RESIT \& different means & \checkmark & \\\hline
     \cref{park}& Uncertainty Scoring & & \checkmark \\\hline
\end{tabular}
\end{center}
\caption{Experiments Layout for Decoupled and Coupled estimation.}
\label{layouttable}
\end{table}

We have uploaded all source codes and results on a Gitlab repository:
\url{https://gitlab.com/Shinkaiika/noise-level-causal-identification-additive-noise-models/}

\chapter{Experiments on Regression with Subsequent Independence Test}\label{resit}
This chapter is an empirical study which involves testing the performance
of \textit{Regression with Subsequent Independence Test (RESIT)} methods \parencite[]{peters2014causal} 
for additive noise models (ANM) \parencite[]{hoyer2008nonlinear,kpotufe2014consistency, mooij2016distinguishing} given i.i.d data from a joint distribution to try to estimate
the corresponding directed acyclic graph (DAG). This chapter is divided into four sections. The first
three sections (\cref{res1} to \cref{res2}) contain the main work of this chapter. In \cref{res1} we test
RESIT with different noise levels and decoupled estimation. In \cref{res3} we test RESIT with different noise levels
and coupled estimation. In \cref{res2} we test RESIT with different noise levels and both decoupled and coupled estimation,
but without the prior assumption (which says that exactly one causal direction must be present in the bi-variate case).
In \cref{res4} we perform tests on RESIT with different means (for both the cause variable and the noise term).
For all experiments we generate artificial data using linear and non-linear functions.
While both linear and non-linear data can be identifiable in causal models,
non-linearity helps in identifying the causal direction as was shown by \citet{hoyer2008nonlinear}.
Therefore, we consider both cases in our experiments in order to investigate how
different noise levels affect causal discovery for linear and on non-linear data. In all
experiments we use the equation $Y = X + N_Y$ for the linear cases 
and $Y = X^3 + N_Y$ for the non-linear cases. These two structural causal models
have been selected arbitrarily for simplicity. For the consistency of the identifiability
of linear and non-linear data in additive noise models, the reader is referred to
\citet{kpotufe2014consistency,shimizu2006linear,hoyer2008nonlinear,zhang2012identifiability}.
Finally, in both sections
we will also measure performance difference when deploying decoupled estimation (splitting data into
training and test data) and coupled estimation (no splitting), see \citet{kpotufe2014consistency},
\citet{mooij2016distinguishing}.

\section{RESIT with Different Noise Levels} \label{res1}

The first section of experiments involves testing different noise levels for the additive noise term
and to see whether it has an impact on the accuracy of RESIT methods or not.
The RESIT method is based on the fact that for each
node $X_i$ the corresponding noise variable $N_i$ is independent of all non-descendants of $X_i$.
For example, in a DAG, if we have $Y = X_1 + N_1$ then $X_1 \indep N_1$.
Following this idea we visit every single node in a specific order. To determine this order
we do an iterative process where in each step we determine the next node in the order. 
To determine the next node we look for a \textit{sink} node.
This is done by regressing each single variable on all other variables
and then measuring the independence between the residuals and those other variables, and finally
selecting the single variable which led to the least dependent residuals. 
After this iterative process, the order is established and every node is visited again. 
We then eliminate incoming edges until the residuals are not independent anymore.
We will restrict our experiments to bivariate cases only to reduce runtimes. 
The authors in \citet{peters2014causal} generalized RESIT to multivariate cases and proved identifiability.
In our experiments we then have two variables, $X$ 
and $Y$, and the task is to determine whether $X$ causes $Y$ ($X \to Y$) or $Y$ causes $X$ ($Y \to
X$).

\subsection{Setup} \label{1Setup}
For all empirical tests we assume $X$ to be a cause of $Y$, that is  $X \to Y$. In the sense of additive noise models,
the equation is then:
$$Y = \beta X + N_y$$ 
$$(Y = \beta X^3 + N_y \text{ for the non-linear case})$$ 
where
$$\beta = 1,$$ and
$$X \sim \begin{cases}
	    \mathcal{N}(0, 1)& \text{or} \\
	    \mathcal{U}(-1, 1)& \text{or}\\
	    \mathcal{L}(0, 1)
	 \end{cases}$$ and

$$N_y \sim \begin{cases}
            \mathcal{N}(0, 1 \cdot i)& \text{or} \\
            \mathcal{U}(-1 \cdot i, 1 \cdot i)& \text{or}\\
            \mathcal{L}(0, 1 \cdot i)
         \end{cases}$$ with $i$ being a scaling factor for the noise level in $N_y$.
         The goal is to analyze how different standard deviations (boundaries for the uniform case)
         in the noise term $N_y$ relative to the standard deviation (or boundaries for the uniform case) in the $X$ term impact 
         the RESIT method.\\

We apply the same algorithm as Algorithm 1 in \citet{mooij2016distinguishing} which requires inputs
$X$ and $Y$, a regression method and a score estimator $\hat{C}: \mathbb{R}^N \times \mathbb{R}^N \to \mathbb{R}$
and outputs $dir$ (casual relationship \textbf{dir}ection).
First, the data is split into training data (80\%) and test data (20\%). \citet{kpotufe2014consistency}
refers to this as "decoupled estimation".
The training data is used to fit the regression model
and the test data is used for the estimator. The idea is to regress $Y$ on $X$ with the training data, 
predict $\hat{Y}$ with the
test data and then calculate residuals $Y_{res} = \hat{Y} - Y_{Test}$. $Y_{res}$ and
$X_{Test}$ are then used in the criterion step, where we use several estimators independently and
receive $\hat{C}_{X \to Y}$, a score for the assumed case $X \to Y$. Similarly, to test
the other case ($Y \to X$), we regress $X$ on $Y$, 
calculate residuals $X_{res} = \hat{X} - X_{Test}$ and estimate $\hat{C}_{Y \to X}$.
In our test scenario our generated data always follows $X \to Y$. This verifies the \textbf{assumption}
that only one direction in our data is correct (and not both) and therefore we can compare both
scores directly in order to make a decision on the cause-effect direction. Thus, for independence
tests, we can compare estimates directly and we do not need
to determine the value of $\alpha$ for the independence tests. \cref{algo1} shows
pseudo code of the procedure explained above.\\

\begin{algorithm}
    \caption{General procedure to decide whether $p(x,y)$ satisfies Additive
    Noise Model $X \to Y$ or $Y \to X$ with decoupled estimation.}\label{algo1}
    \begin{algorithmic}[1]
        \State \textbf{Input:}
        \State \hspace{2.5mm} 1) I.i.d. sample data $X$ and $Y$
        \State \hspace{2.5mm} 2) Regression method
        \State \hspace{2.5mm} 3) Score estimator $\hat{C}: \mathbb{R}^N \times \mathbb{R}^N \to \mathbb{R}$
        \State \textbf{Output:}
        \State \hspace{2.5mm} \textit{dir}
        \State
        \State \textbf{Procedure}
        \State 1) Split data into training and test data:
        \State \hspace{2.5mm} $X_{Train}, X_{Test} \gets X$
        \State \hspace{2.5mm} $Y_{Train}, Y_{Test} \gets Y$
        \State
        \State 2) Train regression models
        \State \hspace{2.5mm} $reg_1 \gets$ Regress $Y_{Train} \text{ on } X_{Train}$
        \State \hspace{2.5mm} $reg_2 \gets$ Regress $X_{Train} \text{ on } Y_{Train}$
        \State
        \State 3) Calculate Residuals:
        \State \hspace{2.5mm} $Y_{res} = reg_1.predict(X_{Test}) - Y_{Test}$
        \State \hspace{2.5mm} $X_{res} = reg_2.predict(Y_{Test}) - X_{Test}$
        \State
        \State 4) Calculate Scores:
        \State \hspace{2.5mm} $\hat{C}_{X \to Y} = \hat{C}(X_{Test}, Y_{res})$
        \State \hspace{2.5mm} $\hat{C}_{Y \to X} = \hat{C}(Y_{Test}, X_{res})$
        \State
        \State 5) Finally, output direction \textit{dir}:
            \begin{gather}
                dir = \begin{cases}
	            X \to Y & \text{if } \hat{C}_{X \to Y} < \hat{C}_{Y \to X},\\
	            Y \to X & \text{if } \hat{C}_{X \to Y} > \hat{C}_{Y \to X},\\
	            ? & \text{if } \hat{C}_{X \to Y} = \hat{C}_{Y \to X}.
	            \end{cases}
            \label{eqn:dir}
            \end{gather}

    \end{algorithmic}
\end{algorithm}

\begin{figure}[h]
\centering
\begin{subfigure}{.5\textwidth}
  \centering
  \includegraphics[scale=0.5]{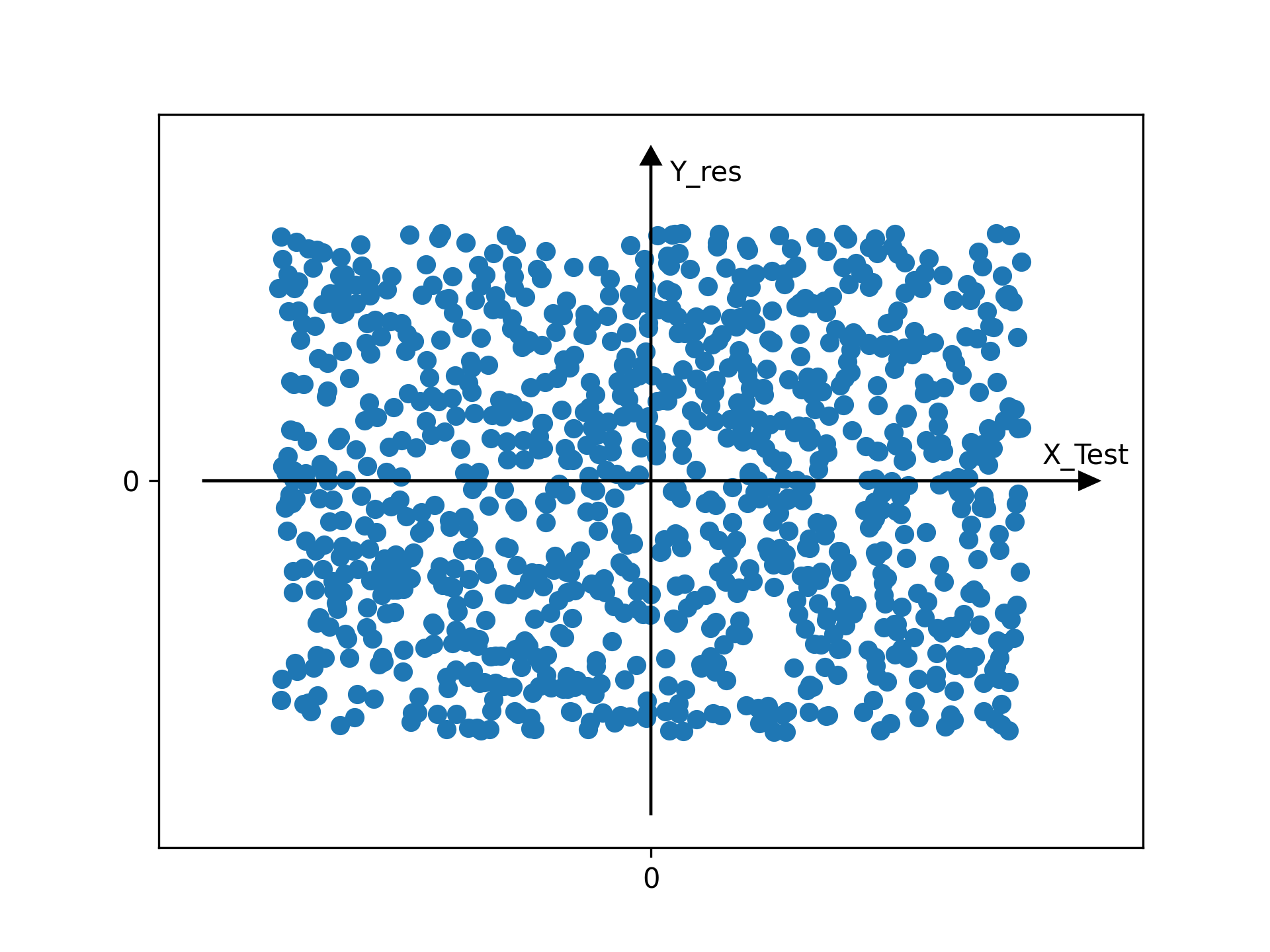}
  \caption{Regress $Y$ on $X$.}
\end{subfigure}%
\begin{subfigure}{.5\textwidth}
  \centering
  \includegraphics[scale=0.5]{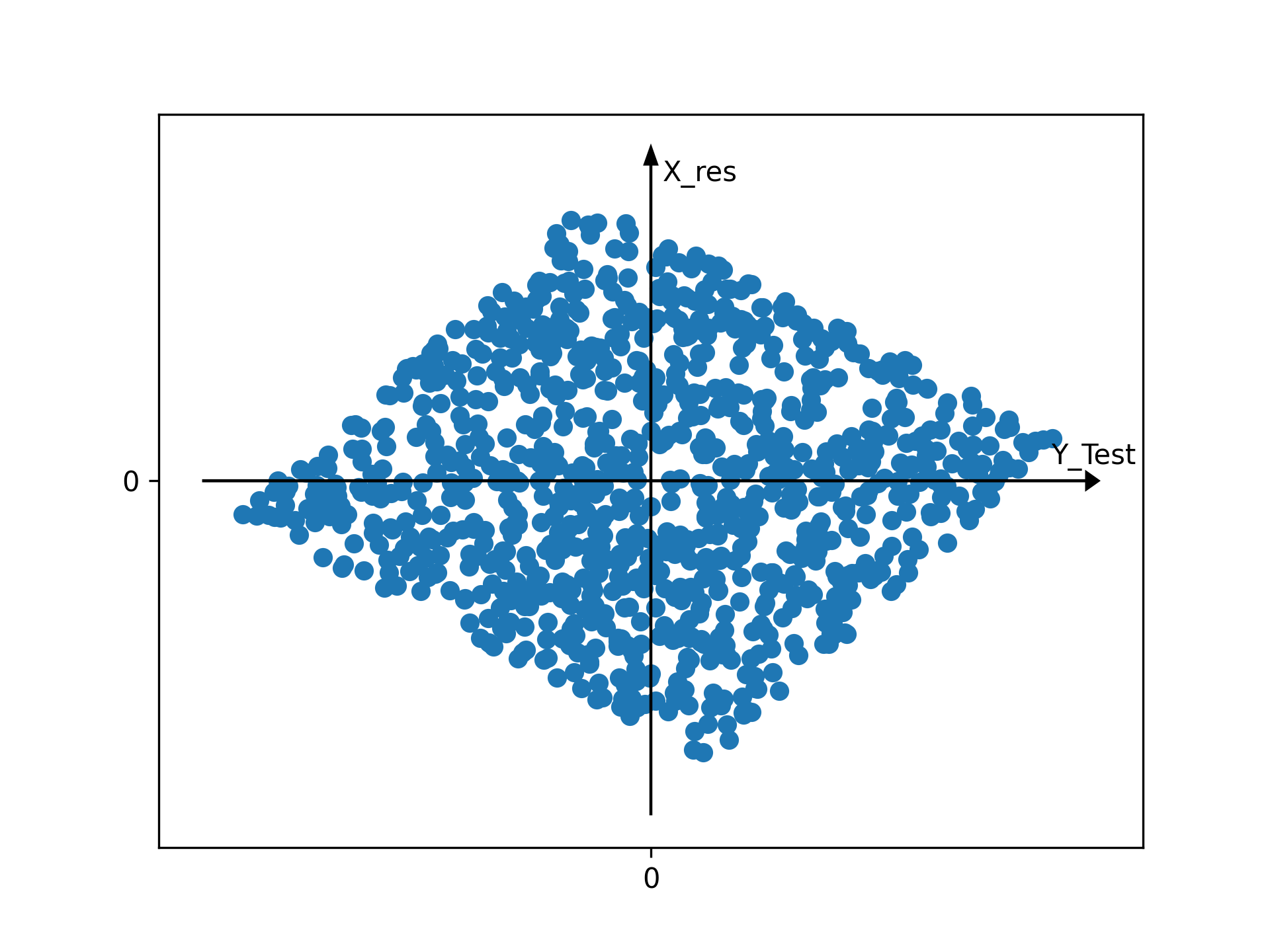}
  \caption{Regress $X$ on $Y$.}
\end{subfigure}
\caption{Data is generated following $Y = X + N_y$ with uniform distributions only.}
\label{fig:resgraphs}
\end{figure}

\cref{fig:resgraphs} explains the concept graphically. The left-hand side of the
figure shows a scatter plot of $X_{Test} \text { and } Y_{res}$. As one can see,
given any input for $X_{Test}$ the range does not change, and therefore $X_{Test}$
and $Y_{res}$ are independent. The right-hand side of the figure shows the opposite case.
For any given input, the range does change. This means that
$Y_{Test} \text{ and } X_{res}$ are dependent and $Y_{Test}$ contains some information
of $X_{res}$ (which contradicts our assumption of independent noise!). 
This allows us to draw conclusions about the causal relationship direction and therefore we know
that $X$ is the cause and not vice-versa.
This illustrates the estimation step \cref{eqn:dir} in \cref{algo1}.\\

For the regression, Linear Regression is used. Linear regression can also be used in the non-linear case
when an appropriate coordinates transformation is applied.
For the estimator several different scores can be used. In our experiments we used
six different independence tests and six different entropy measures for the estimation
criterion. In general, for the independence tests we have:
$$\hat{C}(X_{Test},Y_{res}) = I(X_{Test},Y_{res})$$ with $I(\cdot,\cdot)$ being any independence test.\\\\
In the case of entropy estimators we have:
$$\hat{C}(X_{Test},Y_{res}) = H(X_{Test}) + H(Y_{res}),$$ with $H(\cdot)$ being any entropy measure. The estimator
score for entropy is derived from Lemma 1 in \citet{kpotufe2014consistency}:\\

\textbf{Lemma 1} \textit{Consider a joint distribution of} $X,Y$ \textit{with density p(x,y). For arbitrary functions
f,g:} $\mathbb{R} \to \mathbb{R}$ \textit{we have:}
\begin{gather}
H(X) + H(Y-f(X)) = H(Y) + H(X-g(Y)) - I(X-g(Y),Y) + I(Y-f(X),X),
\label{eqn:entropy-lemma}
\end{gather}
\textit{where} $H(\cdot)$ \textit{denotes differential Shannon entropy, and} $I(\cdot,\cdot)$ \textit{denotes
differential mutual information} (\citet{cover1999elements}).\\

The above Lemma can be proven with the chain rule of differential entropy (see \citet{kpotufe2014consistency}).
If the density $p(x,y)$ satisfies an identifiable ANM $X \to Y$, then there exists a function $f$ with $I(Y-f(X),X) = 0$
(e.g., the regression function $x \mapsto \mathbb{E}(Y | X = x)$), 
but $I(X-g(Y),Y) > 0$ for any function $g$. Therefore, in the case of $X \to Y$, in \ref{eqn:entropy-lemma} we have:\\
$$H(X) + H(Y-f(X)) < H(Y) + H(X-g(Y))$$ \\which is equivalent to the first row in \ref{eqn:dir}:
$$\hat{C}_{X \to Y} < \hat{C}_{Y \to X}.$$\\

The following estimators were used in this work. The implementation of estimators with numbers 2 - 12 was taken from
the \textit{information theoretical estimators} toolbox \citet{szabo14information}:\footnote{More details on the estimators is given in the documentation
of the toolbox.}

\begin{enumerate}
  \item \textbf{HSIC}: Hilbert-Schmidt Independence Criterion with RBF Kernel \footnote{Source: https://github.com/amber0309/HSIC}\\
                       $$I_{HSIC}(x,y) := ||C_{xy}||^2_{HS}$$ where $C_{xy}$ is the cross-covariance
                       operator and $HS$ the squared Hilbert-Schmidt norm.
  \item \textbf{HSIC\_IC}: Hilbert-Schmidt Independence Criterion using incomplete
	                   Cholesky decomposition (low rank decomposition of the Gram matrices, 
	                   which permits an accurate approximation
	                   to HSIC as long as the kernel has a fast decaying spectrum) which has $\eta = 1*10^{-6}$
	                   precision in the incomplete cholesky decomposition.
  \item \textbf{HSIC\_IC2}: Same as HSIC\_IC but with $\eta = 1*10^{-2}$.
  \item \textbf{DISTCOV}: Distance covariance estimator using pairwise distances. This is simply
  the $L^2_w$ norm of the characteristic functions $\varphi_{12}$ and $\varphi_1 \varphi_2$ of input $x,y$:
  $$\varphi_{12}(\boldsymbol{u}^1,\boldsymbol{u}^2) = \mathbb{E}[e^{i\langle \boldsymbol{u}^1, \boldsymbol{x} \rangle +
  i\langle \boldsymbol{u}^2, \boldsymbol{y} \rangle}],$$
  $$\varphi_1(\boldsymbol{u}^1) = \mathbb{E}[e^{i\langle \boldsymbol{u}^1, \boldsymbol{x} \rangle}],$$
  $$\varphi_2(\boldsymbol{u}^2) = \mathbb{E}[e^{i\langle \boldsymbol{u}^2, \boldsymbol{y} \rangle}].$$
  With $i = \sqrt{-1}$, $\langle \cdot, \cdot \rangle$ the standard Euclidean inner product and $\mathbb{E}$ the 
  expectation. Finally, we have:
  $$I_{dCov}(x,y) = ||\varphi_{12} - \varphi_1 \varphi_2||_{L^2_w}$$
  
  \item \textbf{DISTCORR}: Distance correlation estimator using pairwise distances. It is simply the standardized
                        version of the distance covariance:
  $$I_{dCor}(x,y) = \begin{cases}
              \frac{I_{dCov}(x,y)}{\sqrt{I_{dVar}(x,x)I_{dVar}(y,y)}}, &\text{if } I_{dVar}(x,x)I_{dVar}(y,y) > 0 \\
              0, & \text{otherwise,}
  \end{cases}$$ with
  $$I_{dVar}(x,x) = ||\varphi_{11} - \varphi_1 \varphi_1||_{L^2_w},\: I_{dVar}(y,y) = ||\varphi_{22} - \varphi_2
  \varphi_2||_{L^2_w}$$
  (see characteristic functions under 3. DISTCOV)

  \item \textbf{HOEFFDING}: Hoeffding's Phi
  $$I_{\Phi}(x,y) = I_{\Phi}(C) = \left(h_2(d) \int_{[0,1]^d} [C(\boldsymbol{u}) - \Pi (\boldsymbol{u})]^2d\boldsymbol{u}\right)^{\frac{1}{2}}$$ with
  $C$ standing for the copula of the input and $\Pi$ standing for the product copula.
  \item \textbf{SH\_KNN}: Shannon differential entropy estimator using kNNs (k-nearest neighbors)
  $$H(\boldsymbol{Y}_{1:T}) = log(T-1) - \psi(k) + log(V_d) + \frac{d}{T}\sum^T_{t=1}log(\rho_k(t))$$
  with $T$ standing for the number of samples, $\rho_k(t)$ -  the Euclidean distance of the $k^{th}$ nearest neighbour of $\boldsymbol{y}_t$
  in the sample $\boldsymbol{Y}_{1:T}\backslash\{\boldsymbol{y}_t\}$ and $V \subseteq \mathbb{R}^d$ a finite set.
  \item \textbf{SH\_KNN\_2}: Shannon differential entropy estimator using kNNs
	                     with k=3 and kd-tree for quick nearest-neighbour lookup
  \item \textbf{SH\_KNN\_3}: Shannon differential entropy estimator using kNNs with k=5
  \item \textbf{SH\_MAXENT1}: Maximum entropy distribution-based Shannon entropy estimator
  $$H(\boldsymbol{Y}_{1:T}) = H(n) - \left[k_1 \left(\frac{1}{T}\sum^T_{t=1}G_1(y'_t)\right)^2
  + k_2 \left(\frac{1}{T}\sum^T_{t=1}G_2(y'_t)-\sqrt{\frac{2}{\pi}}\right)^2\right] + log(\hat{\sigma}),$$ with
  $$\hat{\sigma} = \hat{\sigma}(\boldsymbol{Y}_{1:T}) = \sqrt{\frac{1}{T-1}\sum^T_{t=1}(y_t)^2},$$
  $$y'_t = \frac{y_t}{\hat{\sigma}}, (t= 1, \dots, T)$$
  $$G_1(z) = ze^{\frac{-z^2}{2}},$$
  $$G_2(z) = |z|,$$
  $$k_1 = \frac{36}{8\sqrt{3}-9},$$
  $$k_2 = \frac{1}{2-\frac{6}{\pi}},$$
  \item \textbf{SH\_MAXENT2}: Maximum entropy distribution-based Shannon entropy estimator, same as SH\_MAXENT1 with
  the following changes:
  $$G_2(z) = e^{\frac{-z^2}{2}},$$
  $$k_2 = \frac{24}{16\sqrt{3}-27},$$
  \item \textbf{SH\_SPACING\_V}: Shannon entropy estimator using Vasicek's spacing method.
  $$H(\boldsymbol{Y}_{1:T}) = \frac{1}{T}\sum^T_{t=1}log\left(\frac{T}{2m}[y_{(t+m)}-y_{(t-m)}]\right)$$
  with $T$ number of samples, the convention that $y_{(t)} := y_{(1)}$ if $t < 1$ and $y_{(t)} := y_{(T)}$ if
  $t > T$ and $m = \lfloor \sqrt{T} \rfloor$.
\end{enumerate}

\subsection{Execution} \label{1Init}
First, 199 different $i$ factors have been generated:
$$i \in \{0.01, 0.02, \dots, 1.00\} \cup \{1, 2, \dots, 100\}$$
For each $i$ every linear and non-linear combination of distribution types have been tested
(18 in total). That is, we have the general structures $Y = X + N_y$ and $Y = X^3 + N_y$ where $X \text{ and } N_y$
are drawn from the three different distributions, $\mathcal{N}, \mathcal{U} \text{ or } \mathcal{L}$.
$$Y = X \sim \mathcal{N}+ N_y \sim \mathcal{N},$$
$$Y = X \sim \mathcal{N}+ N_y \sim \mathcal{U},$$
$$\vdots$$
$$Y = X \sim \mathcal{L}^3+ N_y \sim \mathcal{U}.$$ 
Note that $\mathcal{L}^3$ here signifies the non-linear case $Y = X^3 + N_y$.\\
Next, for each of the 18 combinations we perform 100 tests each time.
For every test we generate 1000 new samples for the distribution of $X$ and the distribution of $N_y$.\\
Lastly, we simply count how many tests are successful in these 100 tests and define this
ratio as our accuracy measure.

\subsection{Experimental Results}\label{results1}
In the following figures the y-axis shows the accuracy ($\frac{\text{\#successful tests}}{100}$) and 
the x-axis shows the range of the $i$ factor. Each figure contains two subfigures, the left with
$i \in \{0.01, 0.02, \dots, 1.00\}$ and the right figure with $i \in \{1, 2, \dots, 100\}$.
If the values of the estimators shown in the plots are close to $0.5$, then this means that in 50\% of the 
tests the algorithm decided the correct direction (and vice versa 50\% the wrong direction) and
thus indicates \textbf{unidentifiability.} If plots are closer to accuracy $1$ then we have
very good/consistent \textbf{identifiability}. The legend for both subfigures
is placed on the right side of the right subfigure and for the decoupled experiments
the estimator names are suffixed with "\_S", S(plit). However, in the analysis text
the suffix is dropped for better readability. The next subsection describes each figure
individually and the following subsection thereafter provides a summary and draws conclusions.
Note that the first group of results are obtained with the assumption "$X\to Y \oplus Y\to X$" and using
decoupled estimation. Also the plots for DISTCOV (dark green) and DISTCORR (medium purple) 
often overlap (more than in 90\% of cases), resulting in a dark purple line.

\newpage

\subsubsection*{Individual Analysis}\label{ia.res}
This section can be skipped as we provide a summary table in the next section (\cref{summarytable1}).

\begin{figure}[h]
\centering
\begin{subfigure}{.5\textwidth}
  \centering
  \includegraphics[scale=0.5]{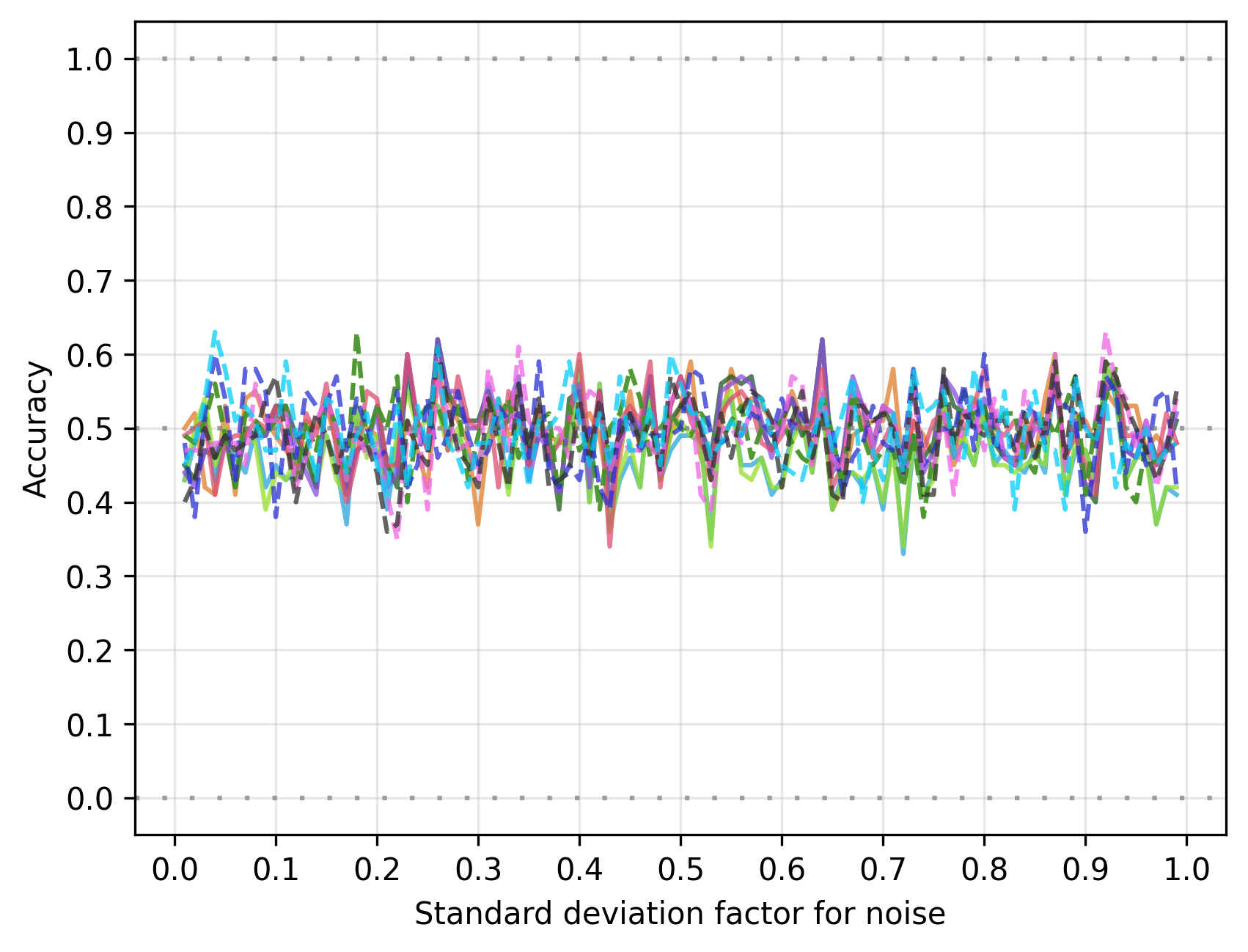}
\end{subfigure}%
\begin{subfigure}{.5\textwidth}
  \centering
  \includegraphics[scale=0.5]{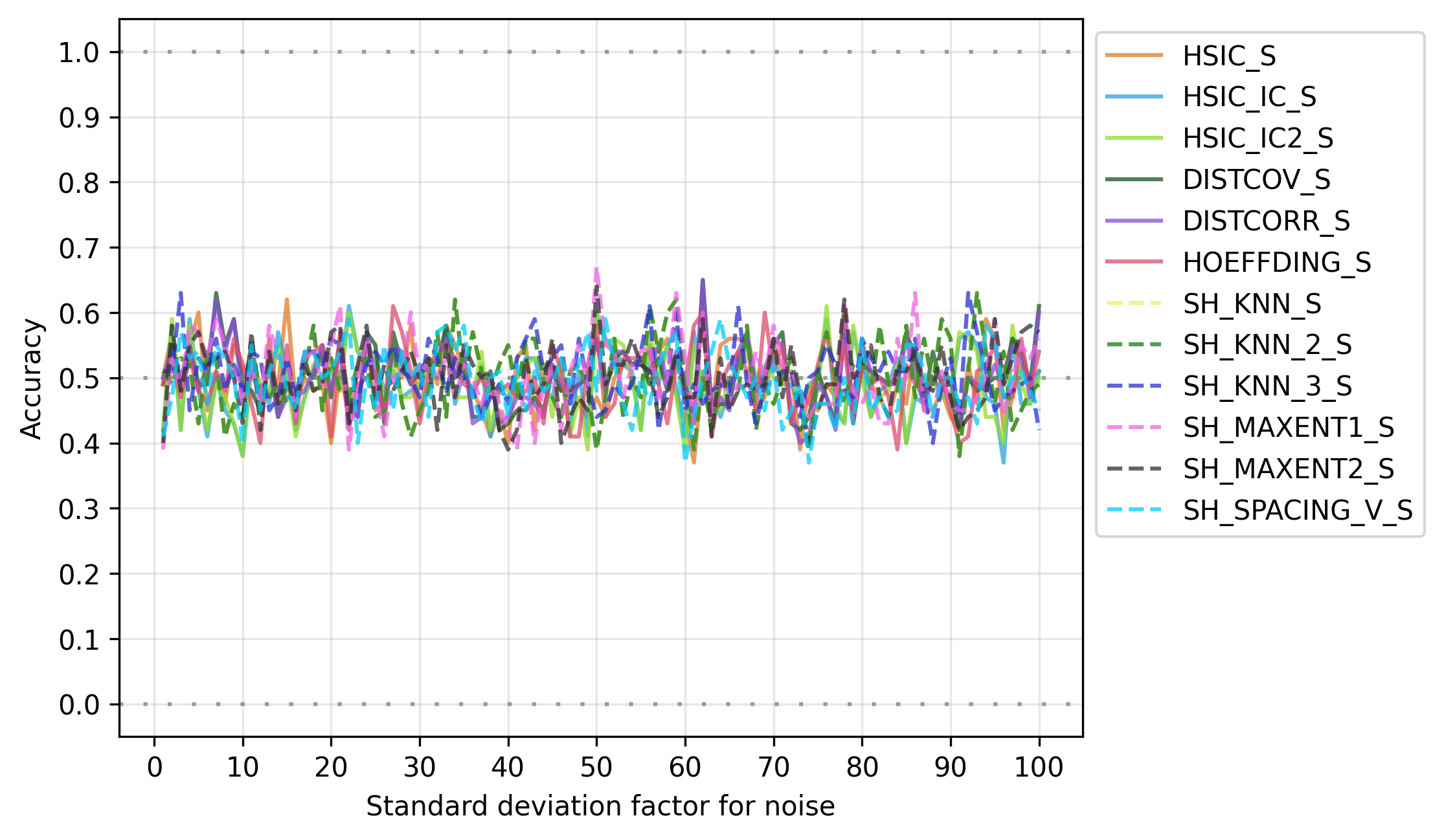}
\end{subfigure}
\caption{RESIT \& different noise levels \& decoupled estimation \& $Y = \mathcal{N}+\mathcal{N}$}
\label{fig:1}
\vspace{5mm}
\centering
\begin{subfigure}{.5\textwidth}
  \centering
  \includegraphics[scale=0.5]{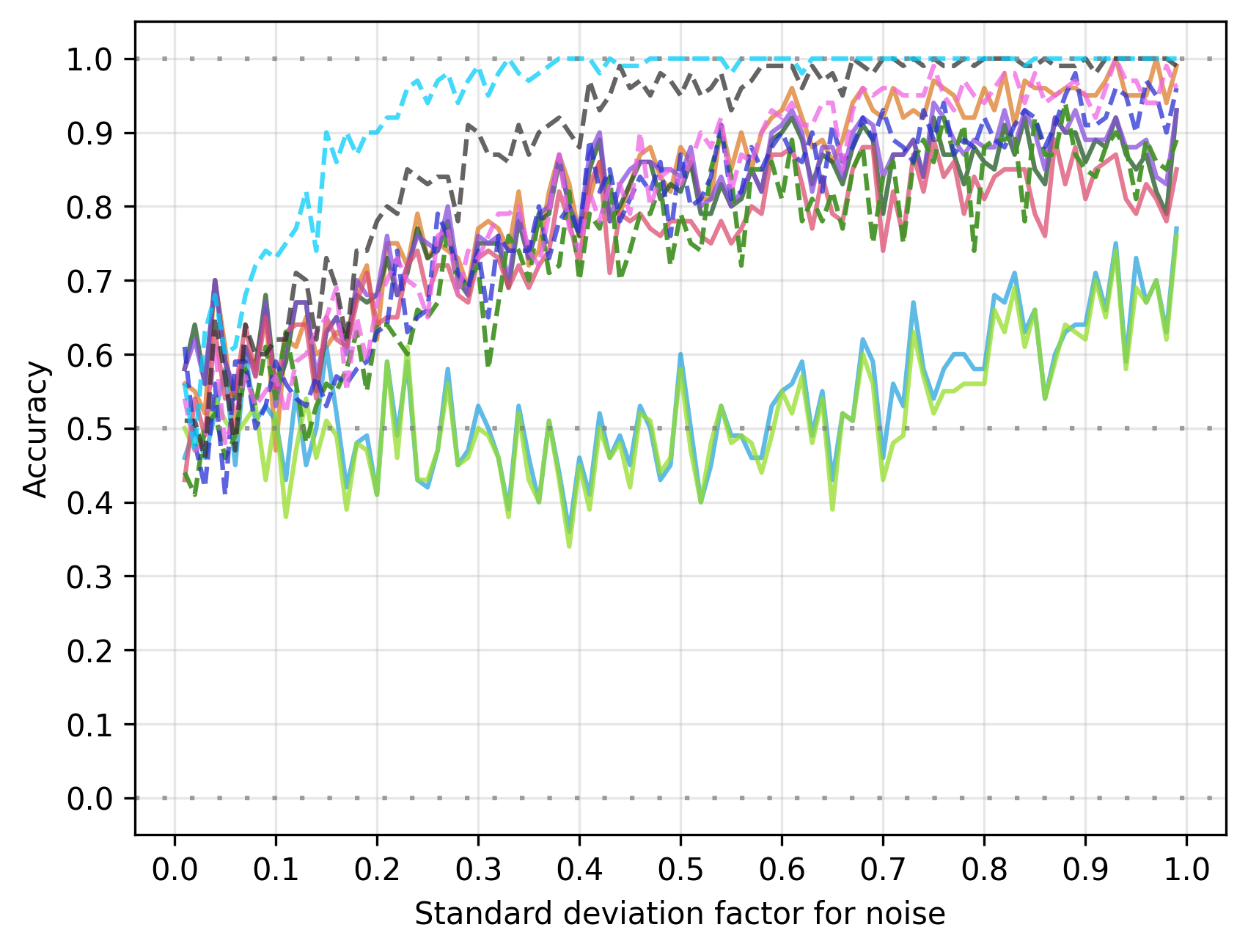}
\end{subfigure}%
\begin{subfigure}{.5\textwidth}
  \centering
  \includegraphics[scale=0.5]{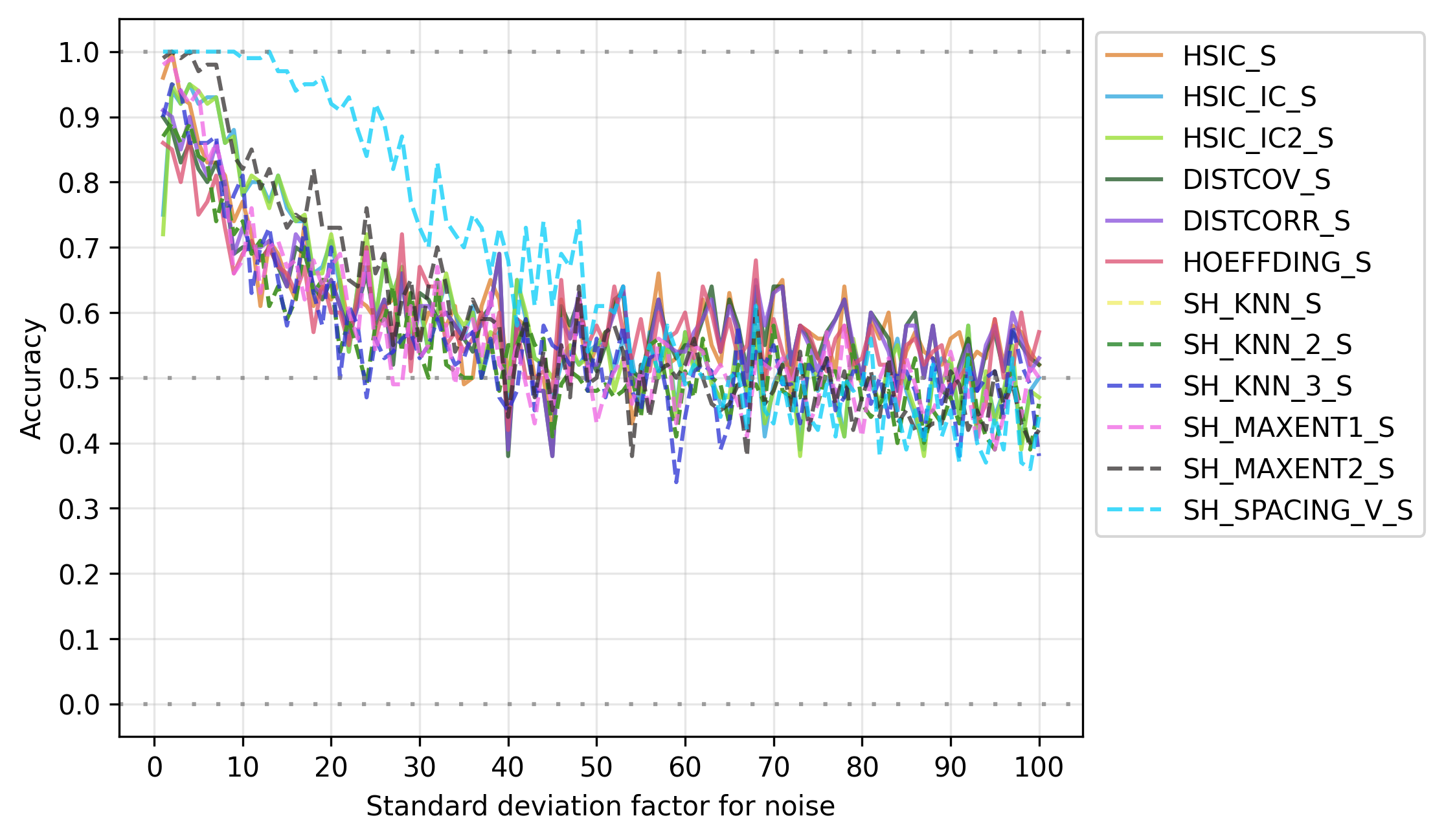}
\end{subfigure}
\caption{RESIT \& different noise levels \& decoupled estimation \& $Y = \mathcal{N}+\mathcal{U}$}
\label{fig:2}
\end{figure}

\cref{fig:1} shows the only case where we never achieve identifiability. This is the well known
linear Gaussian structural causal model and has only recently been tackled successfully in 
\citet{chen2019causal} and \citet{park2019identifiability}.\\

\cref{fig:2} shows the linear model with $Y= \mathcal{N} + \mathcal{U}$, that is independent variable $X$ being
distributed according to the normal distribution and noise - according to uniform distribution.
SH\_SPACING\_V performs the best with the accuracy of 100\% for $i \in [0.55; 7]$. 
HSIC\_IC and HSIC\_IC2 perform the worst here with only an accuracy above 90\% for $i \in [3;7]$. 
All other estimators perform mediocre with an accuracy $>$ 80\% for $i \in [0.5; 7]$.\\

\newpage

\begin{figure}[h]
\centering
\begin{subfigure}{.5\textwidth}
  \centering
  \includegraphics[scale=0.5]{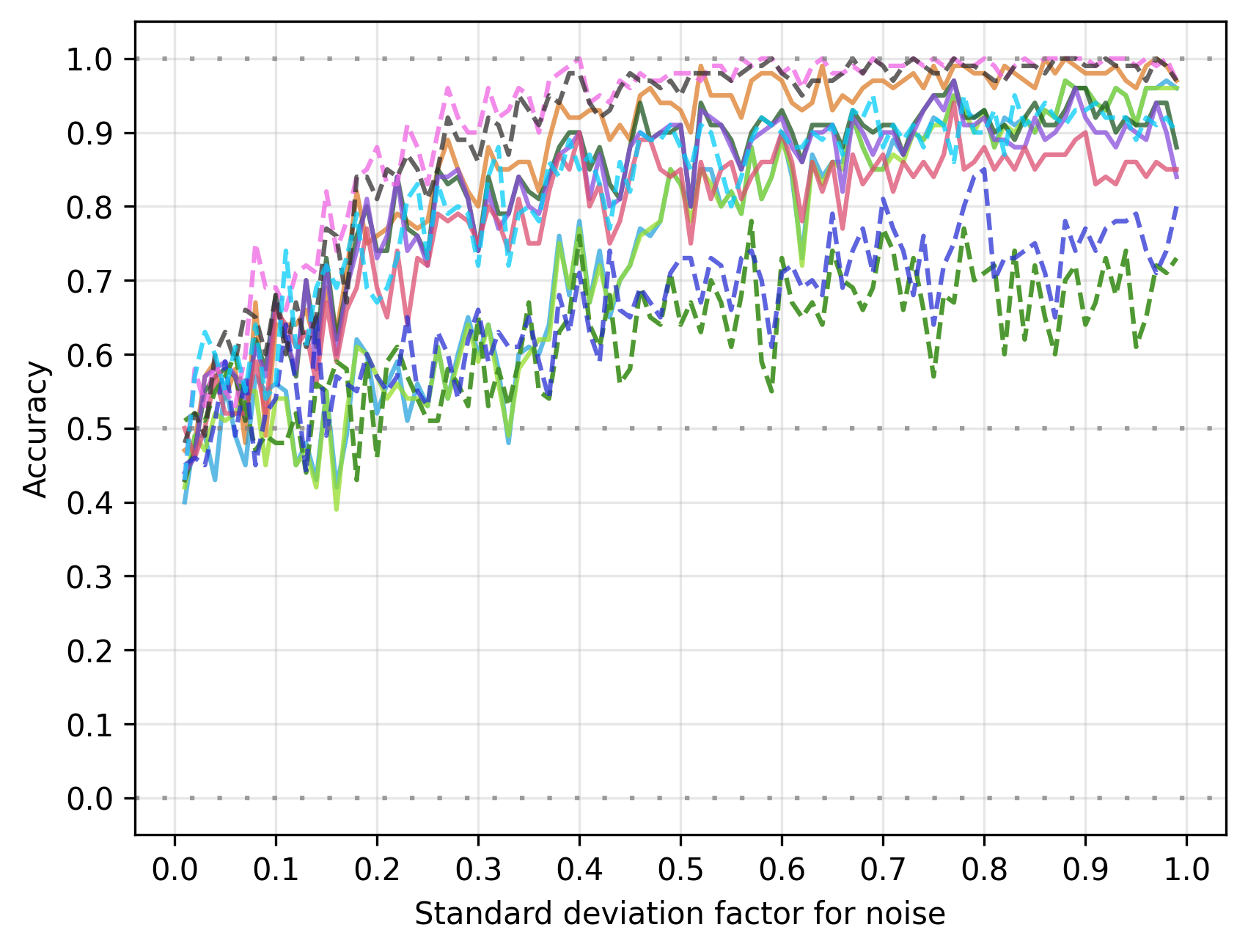}
\end{subfigure}%
\begin{subfigure}{.5\textwidth}
  \centering
  \includegraphics[scale=0.5]{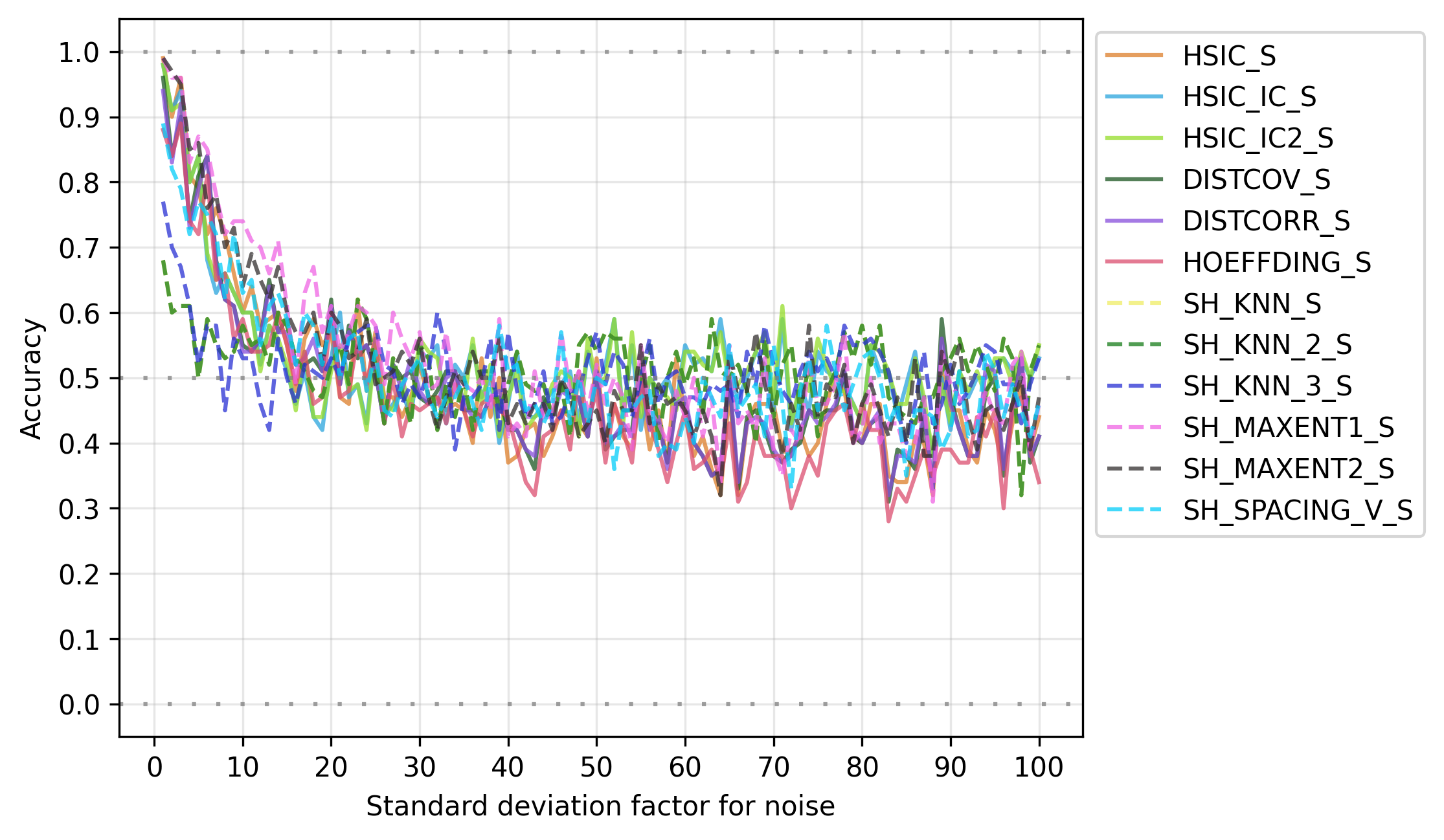}
\end{subfigure}
\caption{RESIT \& different noise levels \& decoupled estimation \& $Y = \mathcal{N}+\mathcal{L}$}
\label{fig:3}
\vspace{5mm}
\centering
\begin{subfigure}{.5\textwidth}
  \centering
  \includegraphics[scale=0.5]{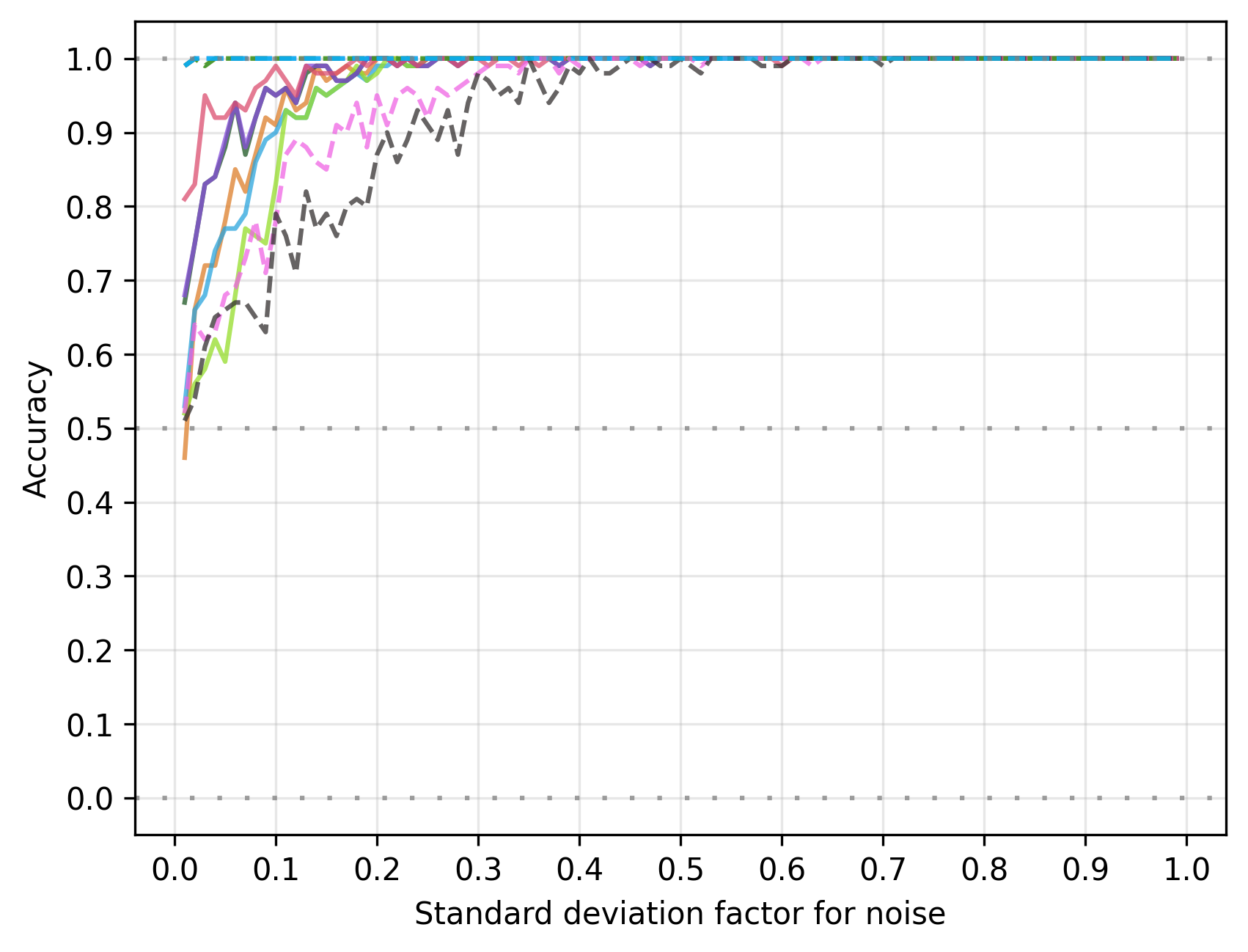}
\end{subfigure}%
\begin{subfigure}{.5\textwidth}
  \centering
  \includegraphics[scale=0.5]{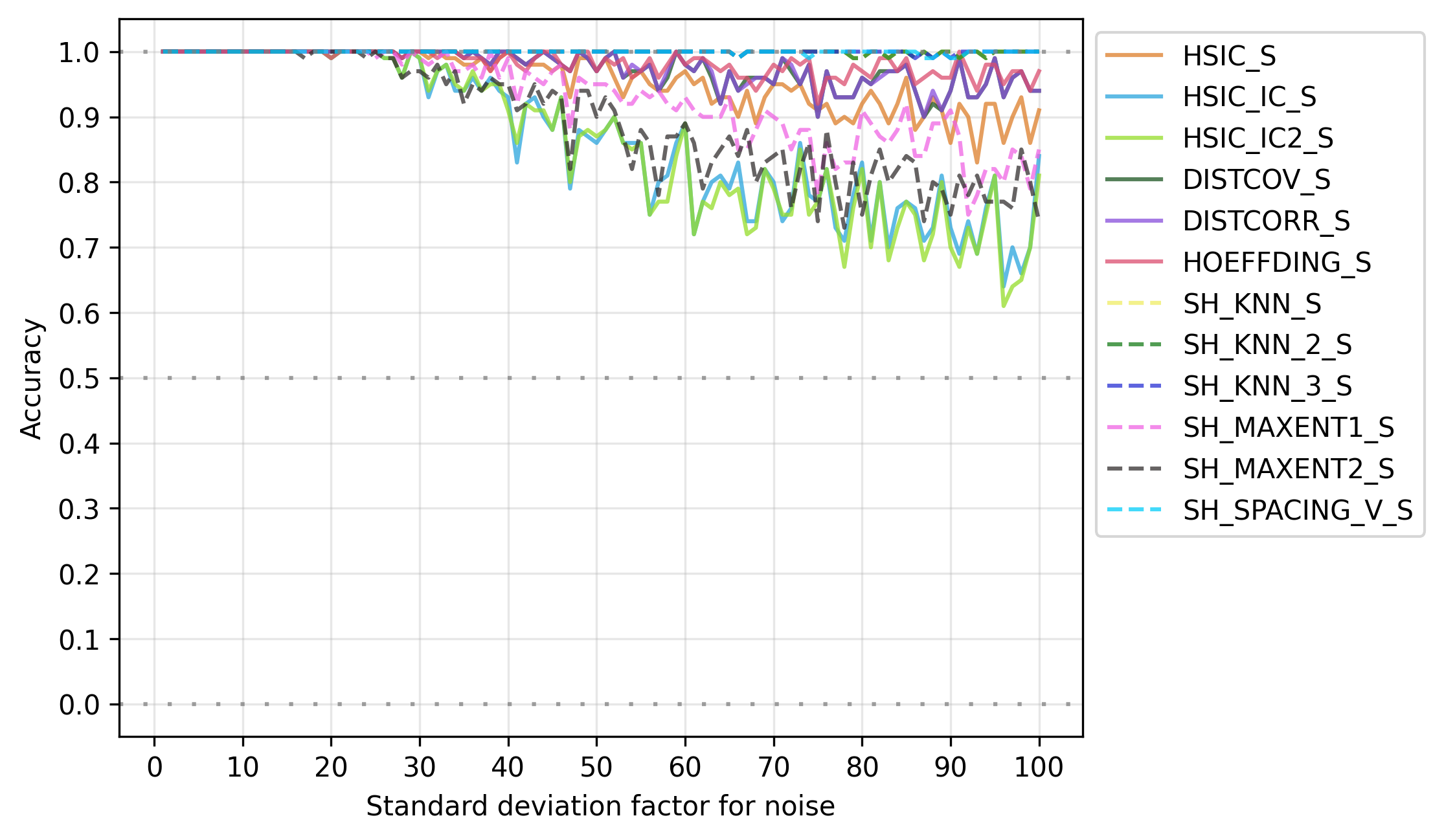}
\end{subfigure}
\caption{RESIT \& different noise levels \& decoupled estimation \& $Y = \mathcal{N}^3+\mathcal{N}$}
\label{fig:4}
\end{figure}

\cref{fig:3} shows the linear model $Y= \mathcal{N} + \mathcal{L}$. The best estimators are SH\_MAXENT1 and SH\_MAXENT2
with accuracy above 90\% for $i \in [0.35; 2]$. HSIC also performs good with accuracy over 90\% for $i \in [0.38; 2]$.
The worst estimators are the three Shannon differential
entropy estimators using kNNs which never remain consistently above 80\% accuracy. The remaining estimators
lie within the range 90\% $\pm$ 8\% accuracy with $i \in [0.3; 3]$. \cref{fig:4} shows the non-linear model $Y= \mathcal{N}^3 + \mathcal{N}$. Here all estimators perform very good
with $i \in [0.4; 25]$ having an accuracy of almost 100\%. With $i < 0.3$ most estimators drop fast below 90\% 
accuracy.
With $i \in [20;100]$ all estimators remain above 90\% accuracy, except for HSIC\_IC, HSIC\_IC2,
SH\_MAXENT1 and SH\_MAXENT2 which drop below 90\% after $i = 45$. DISTCOV, SH\_SPACING\_V and the three
Shannon kNN estimators remain close to 100\% in $i \in [0.01; 100]$.

\newpage

\begin{figure}[h]
\centering
\begin{subfigure}{.5\textwidth}
  \centering
  \includegraphics[scale=0.5]{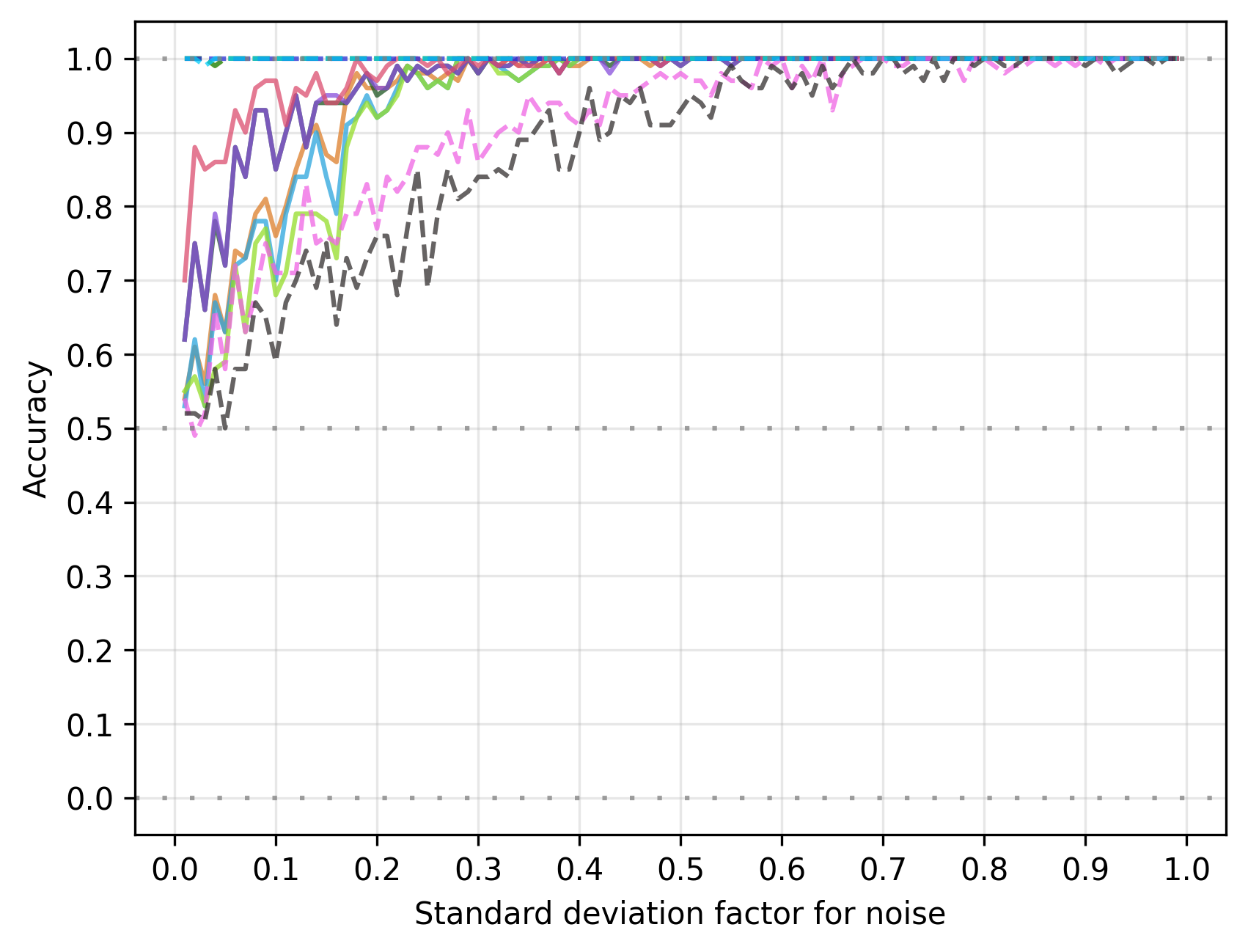}
\end{subfigure}%
\begin{subfigure}{.5\textwidth}
  \centering
  \includegraphics[scale=0.5]{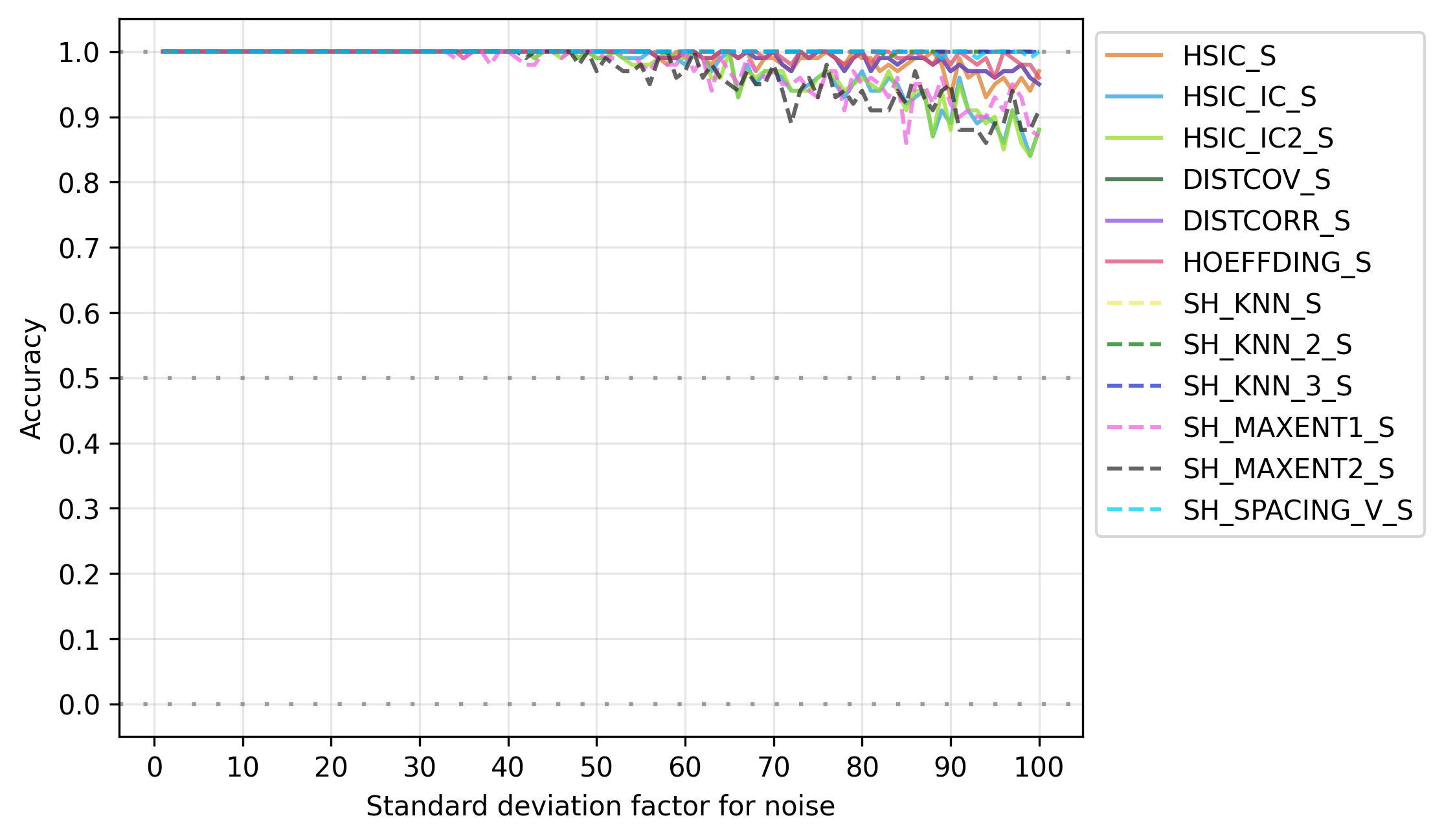}
\end{subfigure}
\caption{RESIT \& different noise levels \& decoupled estimation \& $Y = \mathcal{N}^3+\mathcal{U}$}
\label{fig:5}
\vspace{5mm} 
\centering
\begin{subfigure}{.5\textwidth}
  \centering
  \includegraphics[scale=0.5]{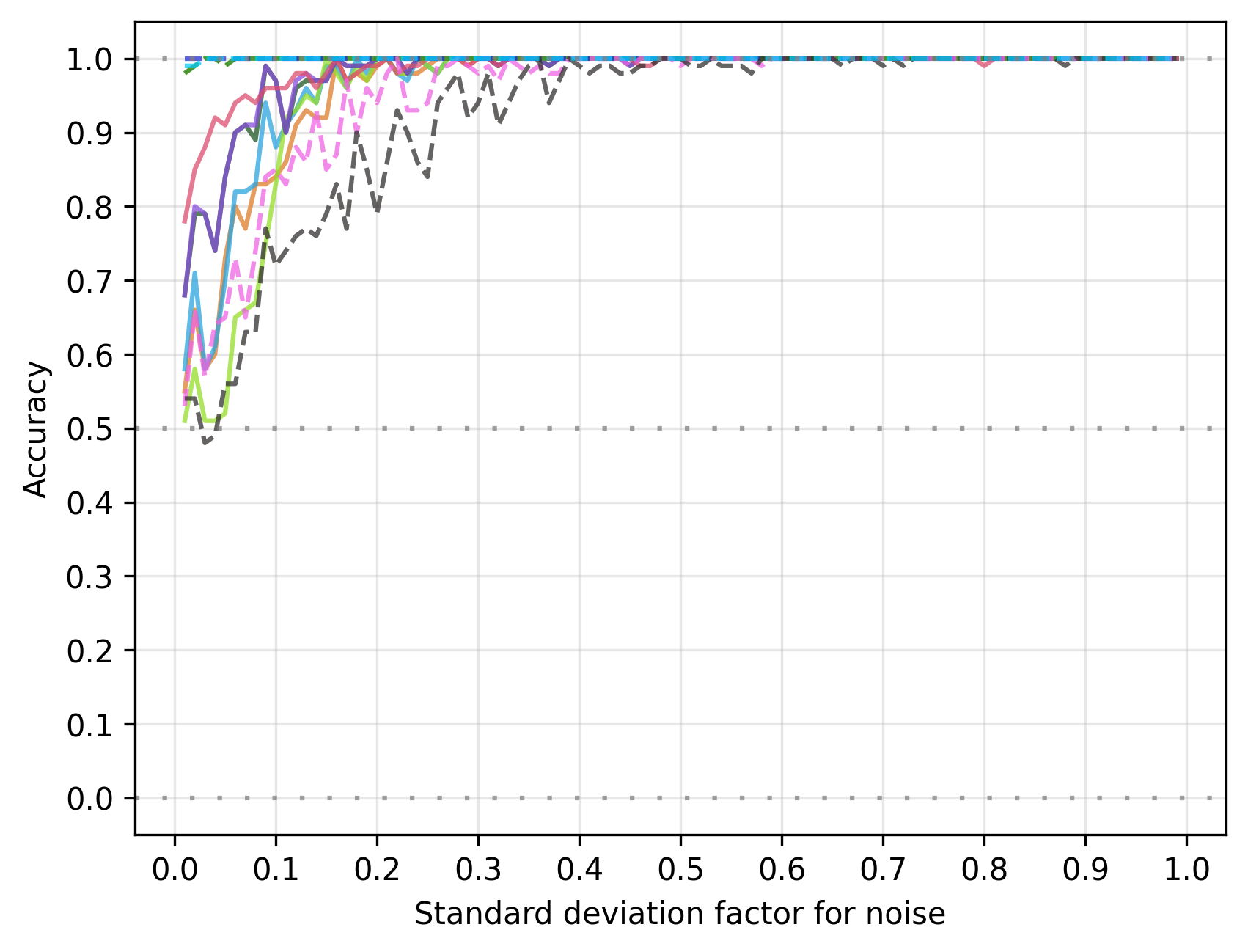}
\end{subfigure}%
\begin{subfigure}{.5\textwidth}
  \centering
  \includegraphics[scale=0.5]{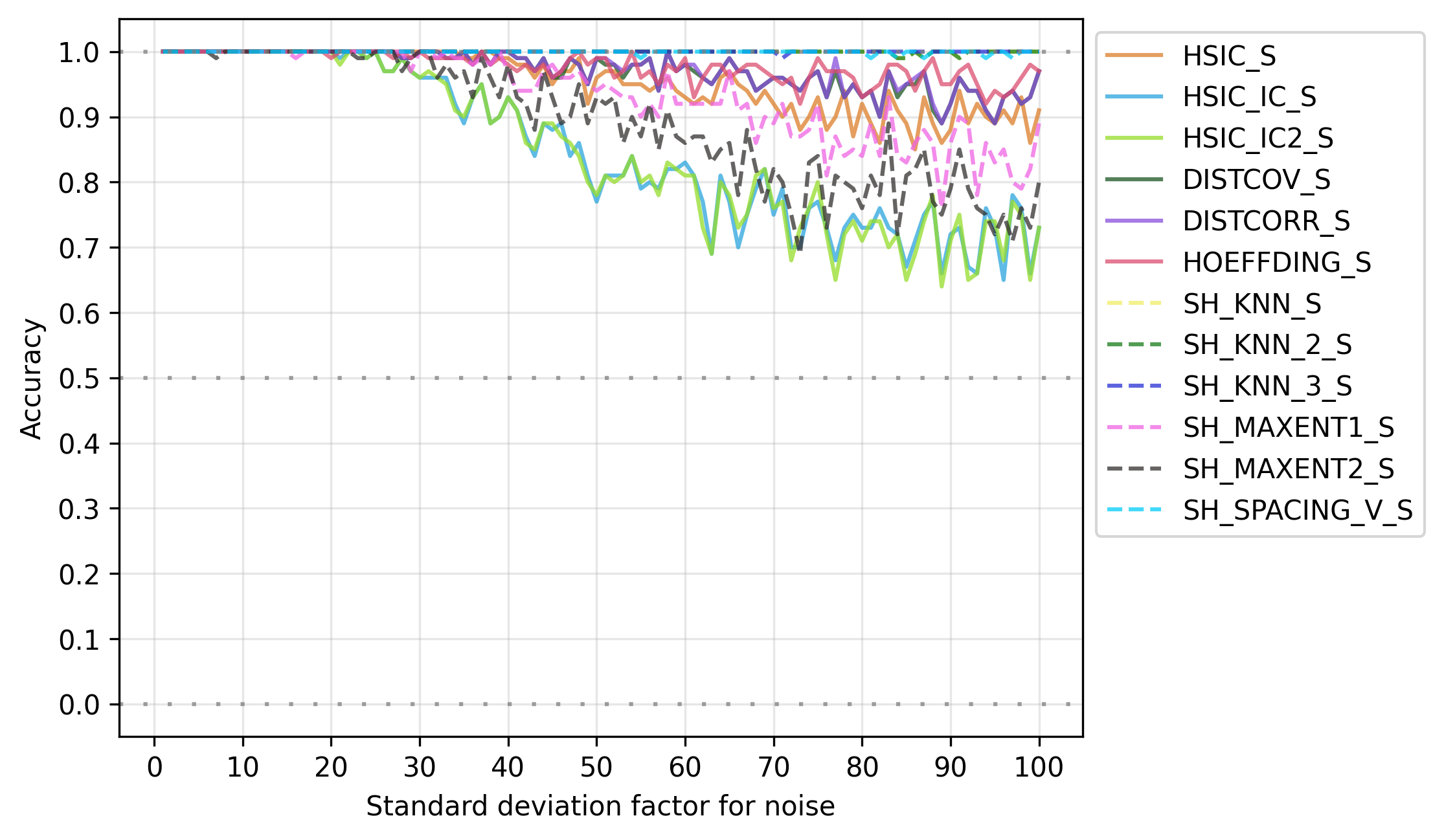}
\end{subfigure}
\caption{RESIT \& different noise levels \& decoupled estimation \& $Y = \mathcal{N}^3+\mathcal{L}$}
\label{fig:6}
\end{figure}

\cref{fig:5} shows the non-linear model $Y= \mathcal{N}^3 + \mathcal{U}$. This result is similar as the previous one.
With $i \in [0.45; 80]$ we have 90\% or higher accuracy for all estimators. DISTCOV, SH\_SPACING\_V and the three
Shannon kNN estimators remain close to 100\% in $i \in [0.01; 100]$.
\cref{fig:6} shows the non-linear model $Y= \mathcal{N}^3 + \mathcal{L}$. Here all estimators perform very good
with $i \in [0.4; 30]$ having an accuracy close to 100\%.
With $i < 0.25$ all estimator drop rapidly and with $i > 30$ HSIC\_IC, HSIC\_IC2, SH\_MAXENT1 and SH\_MAXENT2 drop
below 90\% accuracy for higher $i$.
All others remain over 90\% accuracy while HSIC remains around 90\%.

\newpage

\begin{figure}[h]
\centering
\begin{subfigure}{.5\textwidth}
  \centering
  \includegraphics[scale=0.5]{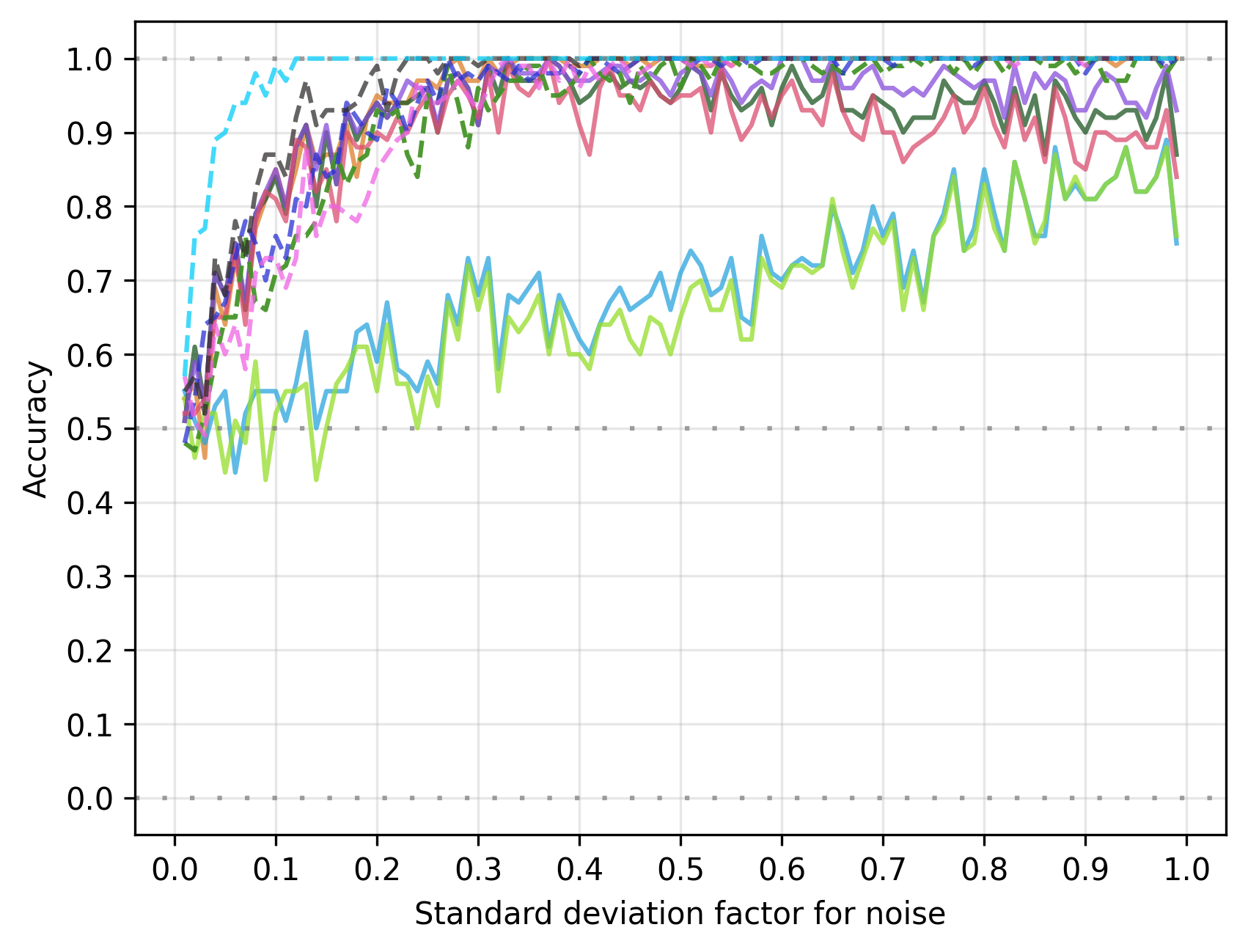}
\end{subfigure}%
\begin{subfigure}{.5\textwidth}
  \centering
  \includegraphics[scale=0.5]{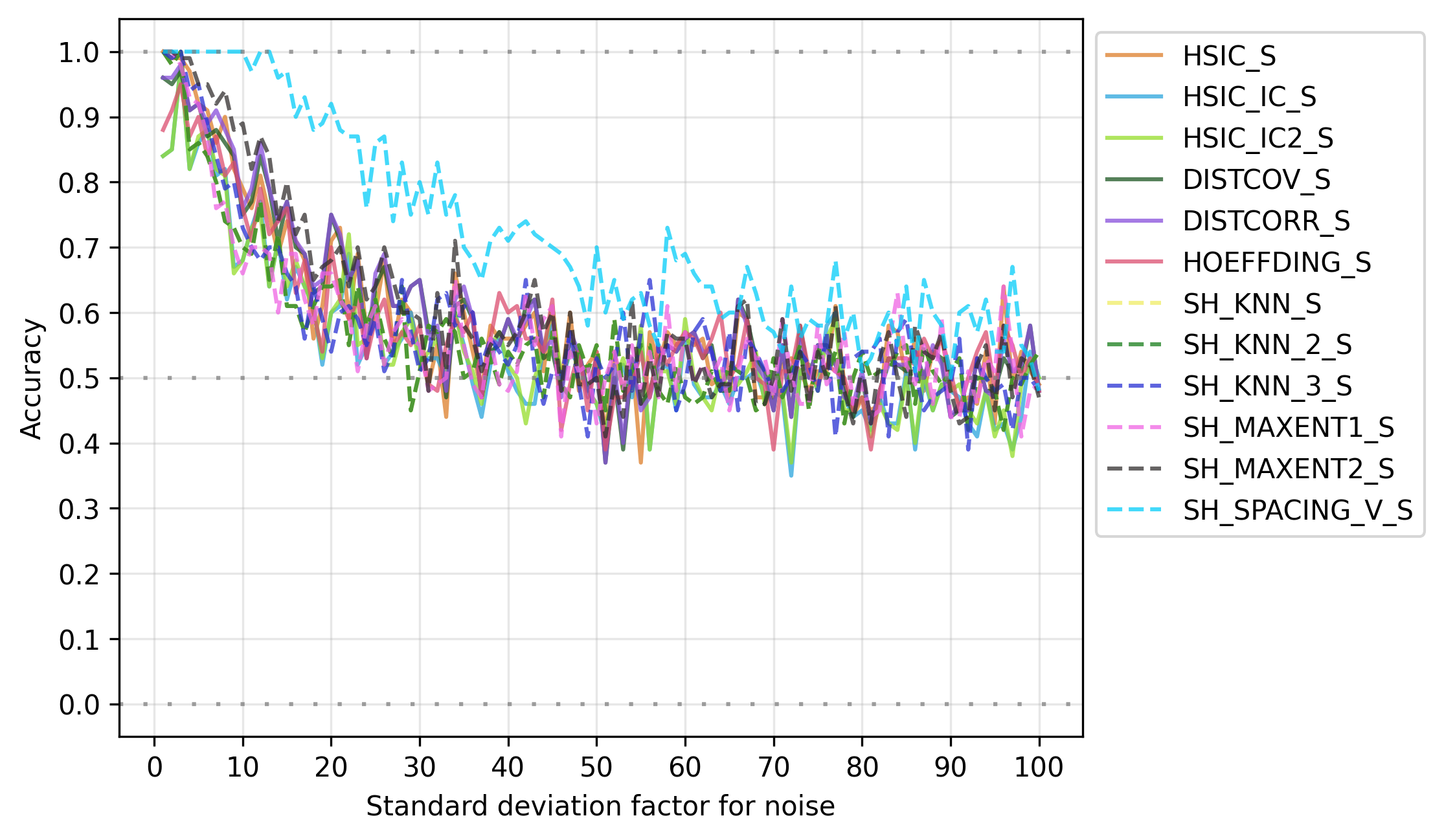}
\end{subfigure}
\caption{RESIT \& different noise levels \& decoupled estimation \& $Y = \mathcal{U}+\mathcal{U}$}
\label{fig:7}
\centering
\begin{subfigure}{.5\textwidth}
  \centering
  \includegraphics[scale=0.5]{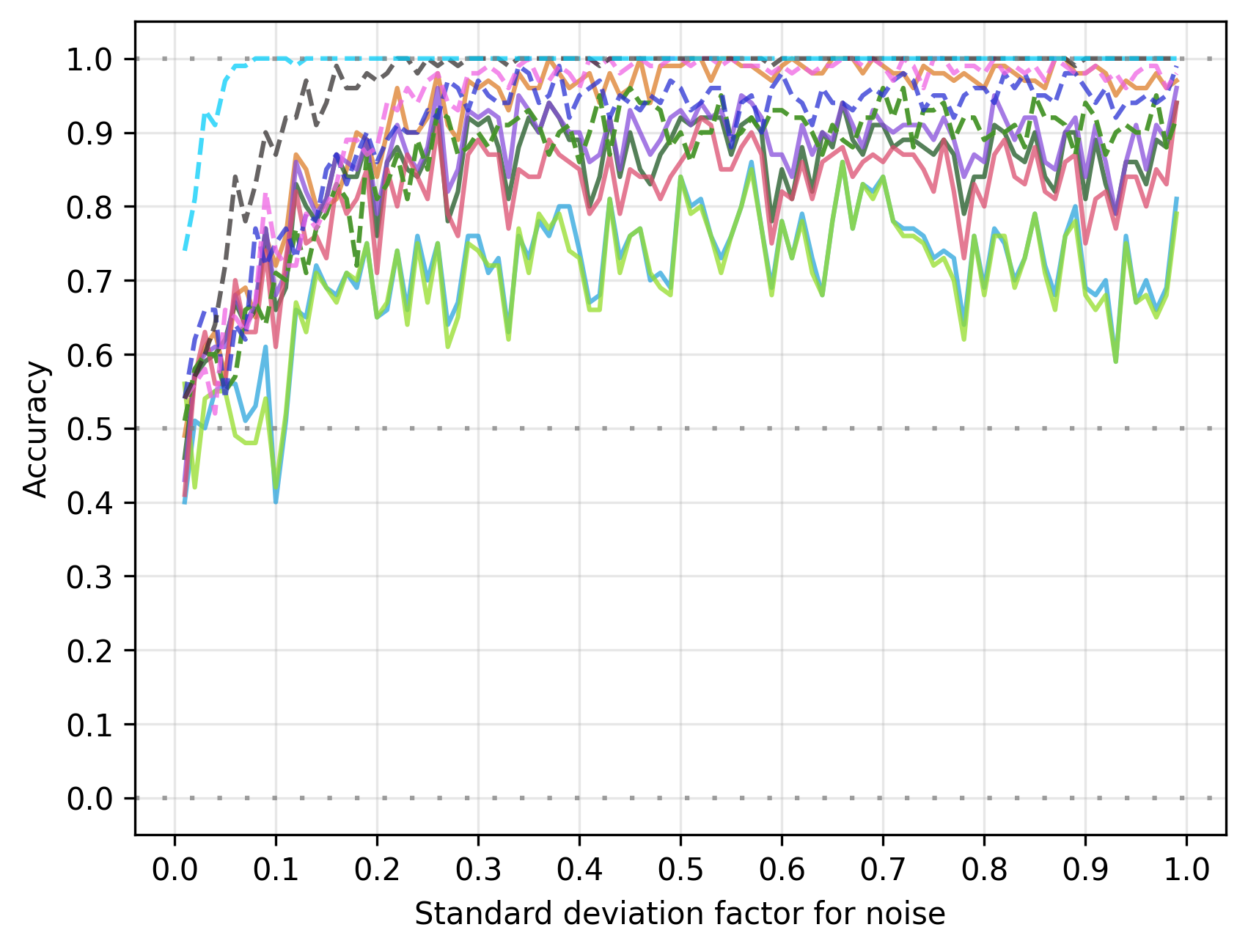}
\end{subfigure}%
\begin{subfigure}{.5\textwidth}
  \centering
  \includegraphics[scale=0.5]{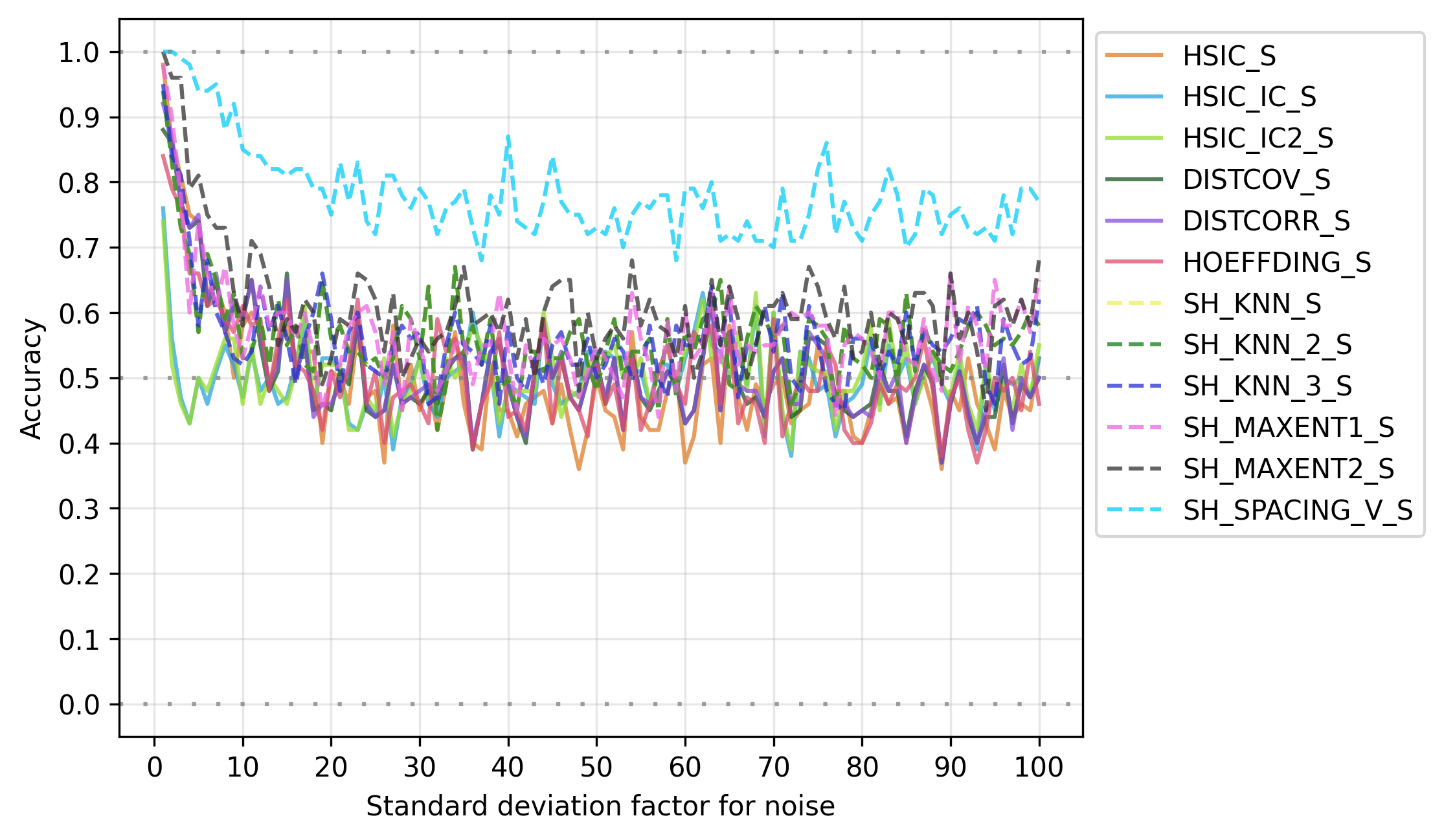}
\end{subfigure}
\caption{RESIT \& different noise levels \& decoupled estimation \& $Y = \mathcal{U}+\mathcal{N}$}
\label{fig:8}
\end{figure}

\cref{fig:7} shows the linear model $Y= \mathcal{U} + \mathcal{U}$. HSIC\_IC and HSIC\_IC2 only reach
accuracy above 90\% around $i = 3$. All other estimator perform quite good with $i \in [0.3; 5]$ while only HOEFFDING and DISTCOV
drop slightly below 90\% accuracy for $i = 1$. SH\_SPACING\_V has 100\% accuracy for $i \in [0.12;10]$ and has
on the remaining $i$ factors better accuracy than all other estimators.
\cref{fig:8} shows the linear model $Y= \mathcal{U} + \mathcal{N}$. Here all estimators differ stronger than in the previous
experiments. First, SH\_SPACING\_V is performing the best with 100\% accuracy for $i \in [0.08; 2]$.
With $i = 1$ all other estimators remain above 90\%, expect HOEFFDING ($\sim$88\%) and HSIC\_IC and HSIC\_IC2 (both $\sim 75\%$). 
After $i = 2$ all estimators drop drastically towards 50\% accuracy except SH\_SPACING\_V which remains above 70\%.
For $i \in [0.2; 1]$ some estimators remain between 80\% and 95\% while HSIC is above 95\%. Only HSIC\_IC and HSIC\_IC2 perform worse
than all other estimators.

\newpage

\begin{figure}[h]
\centering
\begin{subfigure}{.5\textwidth}
  \centering
  \includegraphics[scale=0.5]{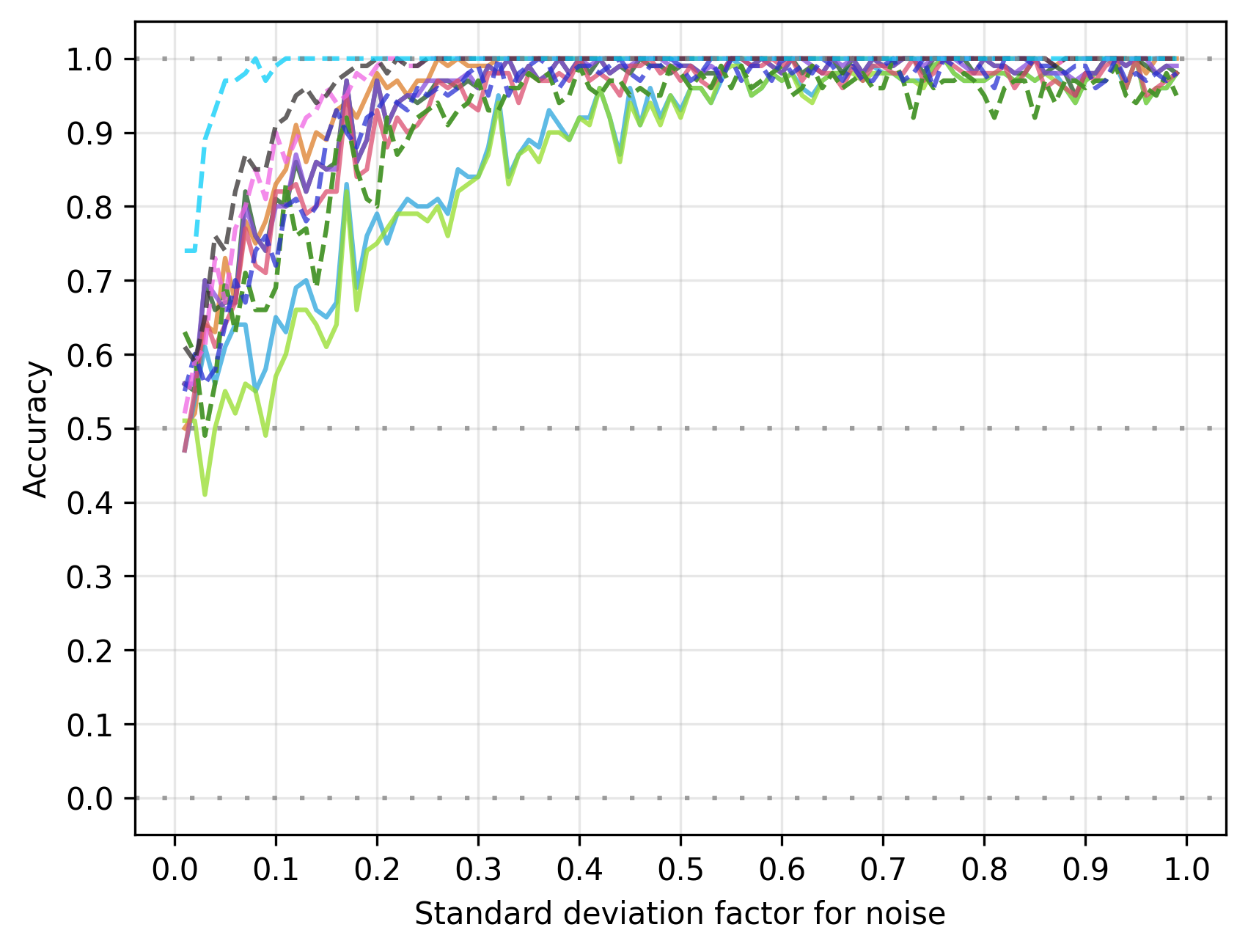}
\end{subfigure}%
\begin{subfigure}{.5\textwidth}
  \centering
  \includegraphics[scale=0.5]{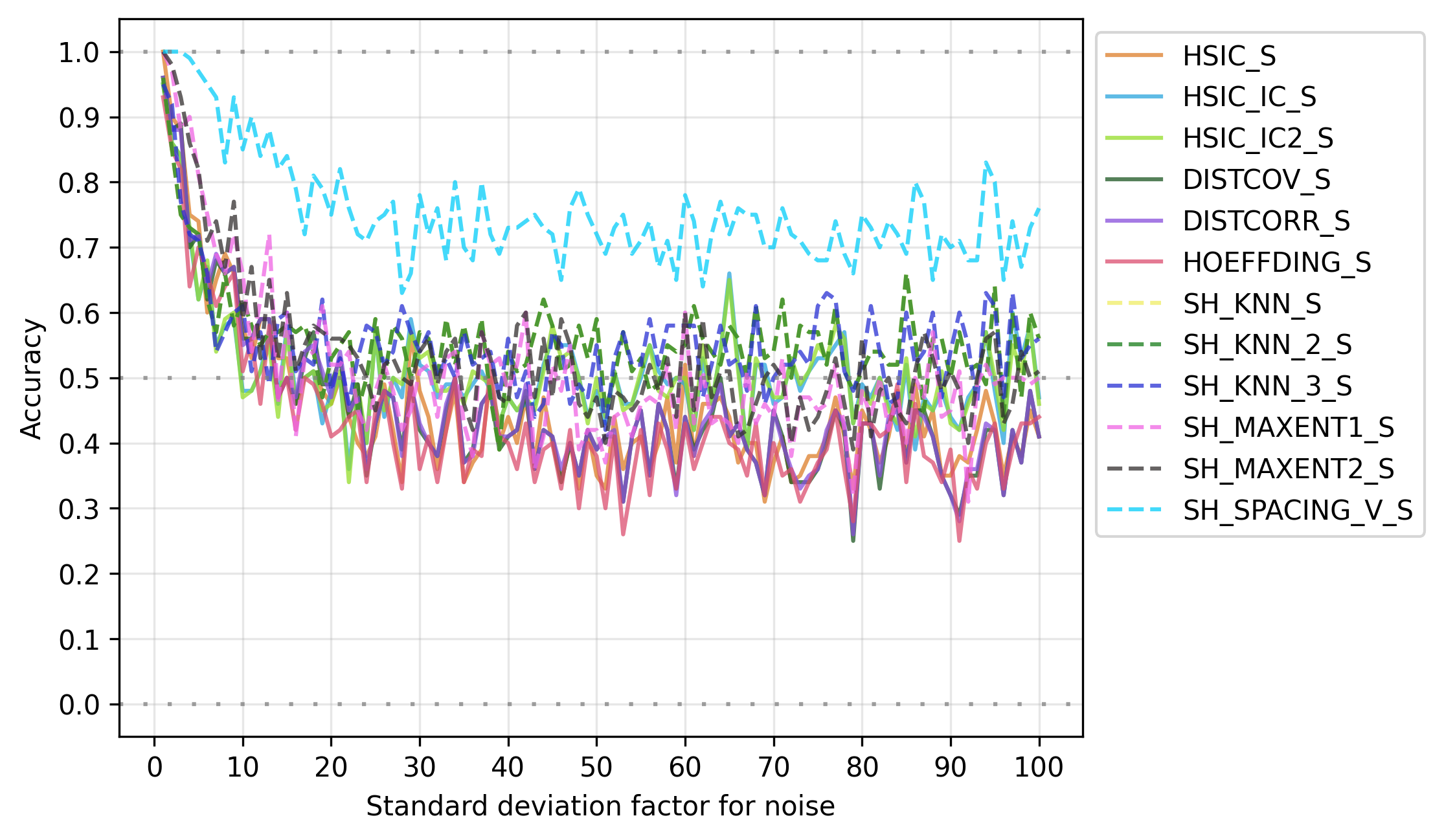}
\end{subfigure}
\caption{RESIT \& different noise levels \& decoupled estimation \& $Y = \mathcal{U}+\mathcal{L}$}
\label{fig:9}
\centering
\begin{subfigure}{.5\textwidth}
  \centering
  \includegraphics[scale=0.5]{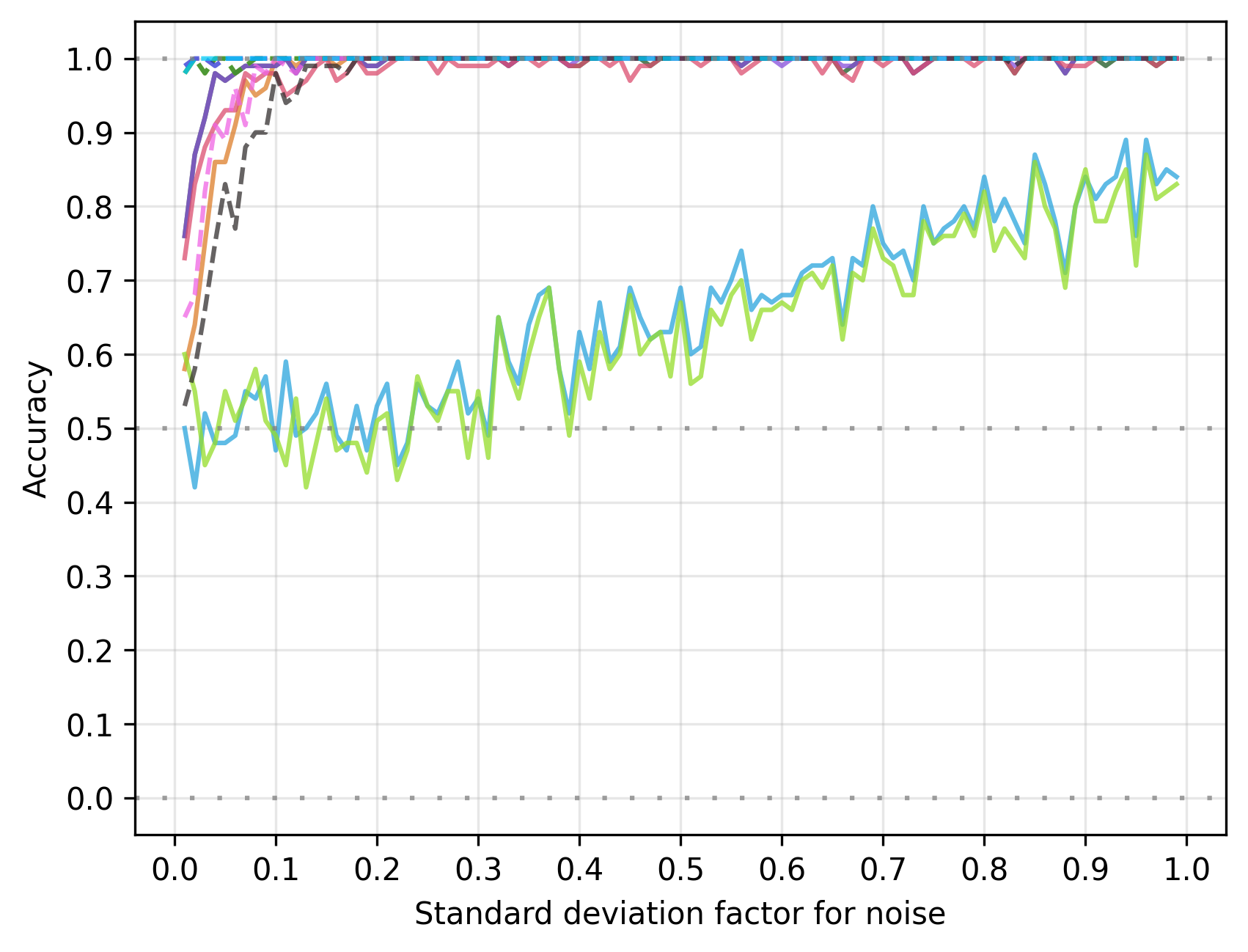}
\end{subfigure}%
\begin{subfigure}{.5\textwidth}
  \centering
  \includegraphics[scale=0.5]{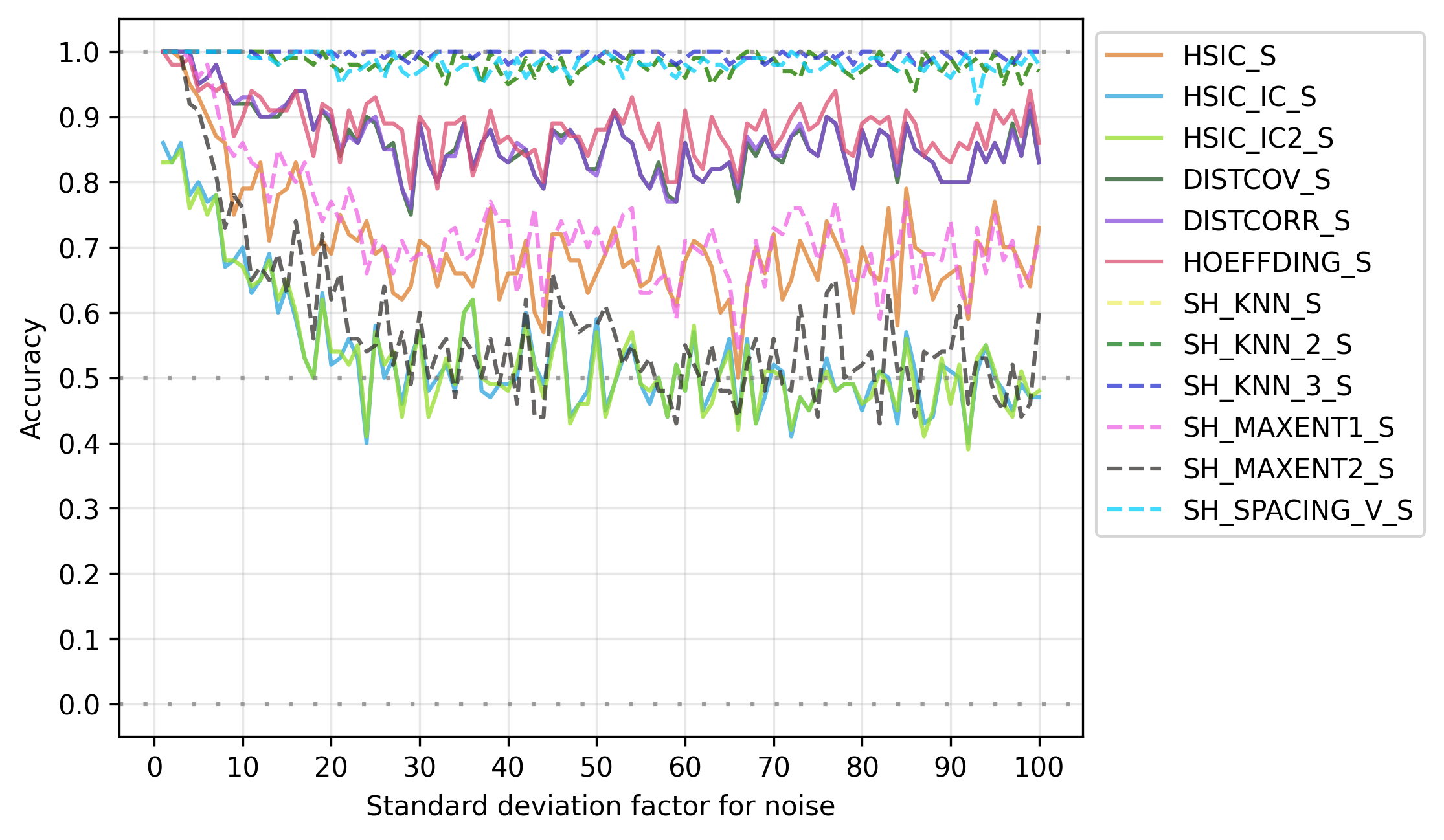}
\end{subfigure}
\caption{RESIT \& different noise levels \& decoupled estimation \& $Y = \mathcal{U}^3+\mathcal{U}$}
\label{fig:10}
\end{figure}

\cref{fig:9} shows the linear model $Y= \mathcal{U} + \mathcal{L}$. For $i \in [0.3; 1]$ all estimators
perform well with 90\% or higher accuracy, except HSIC\_IC and HSIC\_IC2 which remain above 90\% accuracy only after $i= 0.45$. 
After $i = 1$ each estimator drops drastically and all converge towards 50\%
accuracy, except for SH\_SPACING\_V which remains with a mean of 70\% accuracy higher than other estimators.
For $i \in [0.08; 1]$ SH\_SPACING\_V also has accuracy 100\%. For $i < 0.3$ all other
estimators drop fast towards 50\%.
\cref{fig:10} shows the non-linear model $Y= \mathcal{U}^3 + \mathcal{U}$. For $i \in [0.09; 1]$ all
estimators (except HSIC\_IC and HSIC\_IC2) remain above 95\% accuracy, while SH\_KNN, SH\_KNN\_2, and SH\_SPACING\_V
continue to do so for $i \in [1; 100]$. DISTCOV, DISCORR and HOEFFDING remain between
80\% and 90\%. HSIC and SH\_MAXENT1 drop to $> 60$\% after $i = 20$ and remain above 60\%
for $i < 100$. SH\_MAXENT2, HSIC\_IC and HSIC\_IC2 drop to 50\% for $i \in [20; 100]$.

\newpage

\begin{figure}[h]
\centering
\begin{subfigure}{.5\textwidth}
  \centering
  \includegraphics[scale=0.5]{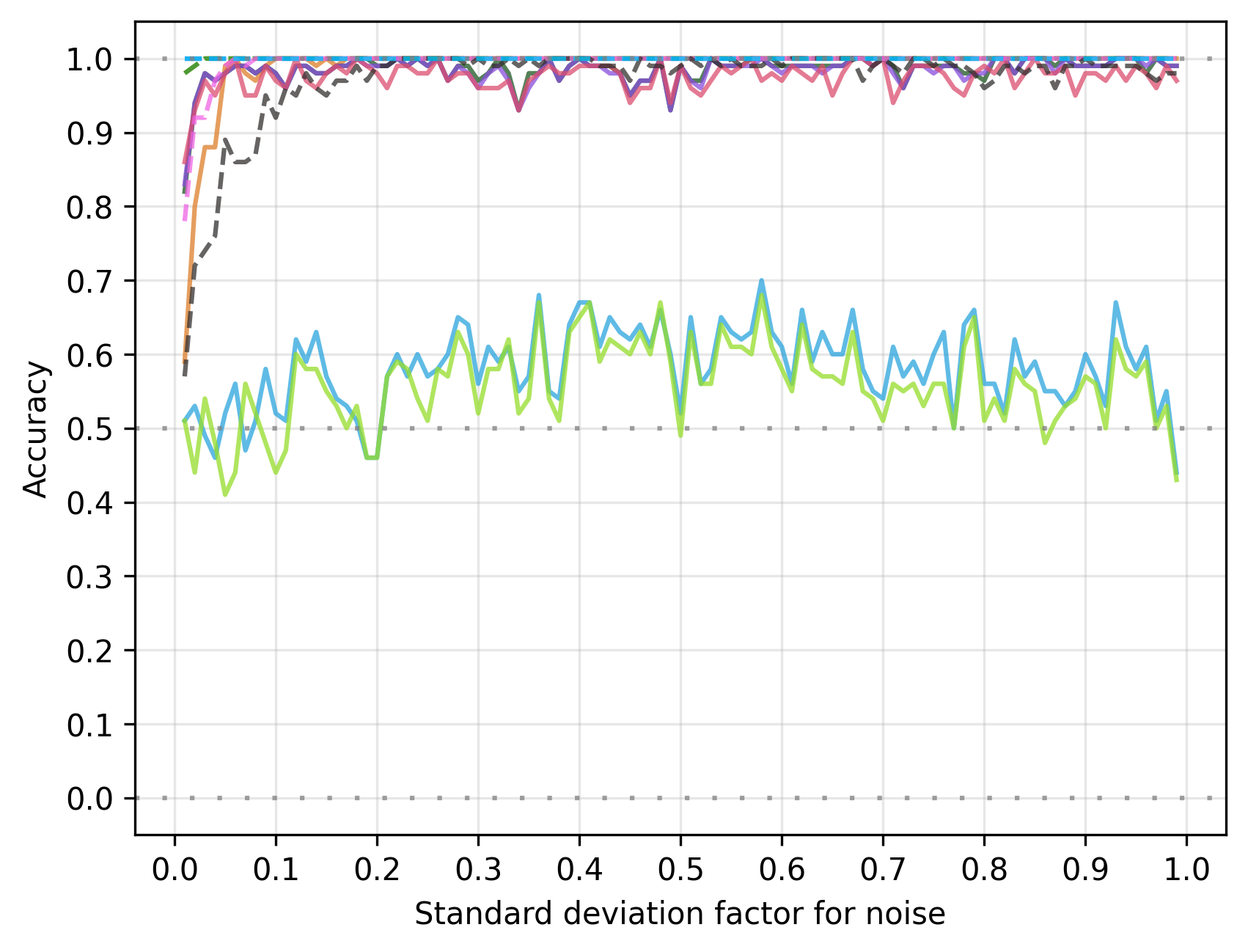}
\end{subfigure}%
\begin{subfigure}{.5\textwidth}
  \centering
  \includegraphics[scale=0.5]{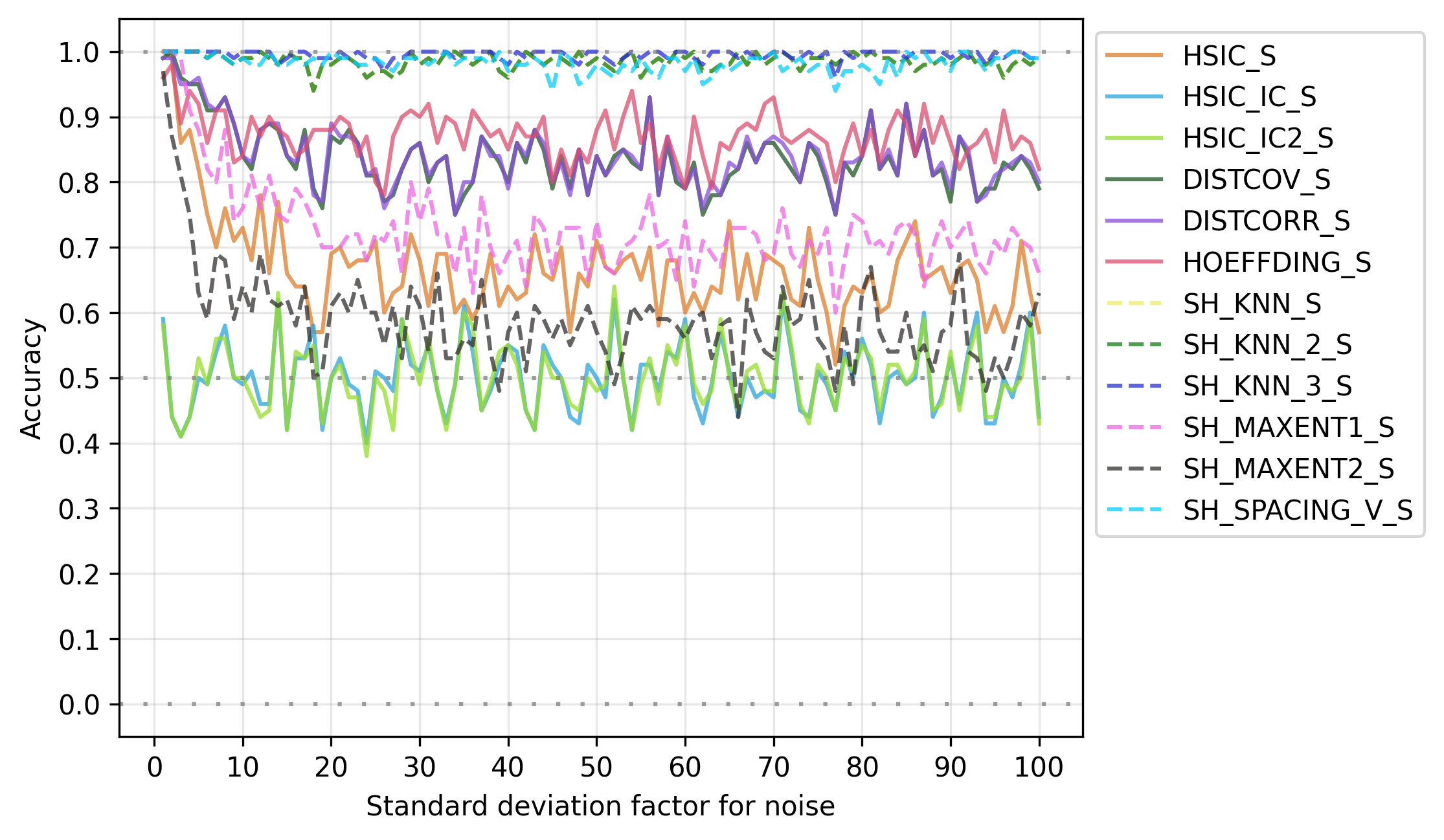}
\end{subfigure}
\caption{RESIT \& different noise levels \& decoupled estimation \& $Y = \mathcal{U}^3+\mathcal{N}$}
\label{fig:11}
\vspace{5mm}
\centering
\begin{subfigure}{.5\textwidth}
  \centering
  \includegraphics[scale=0.5]{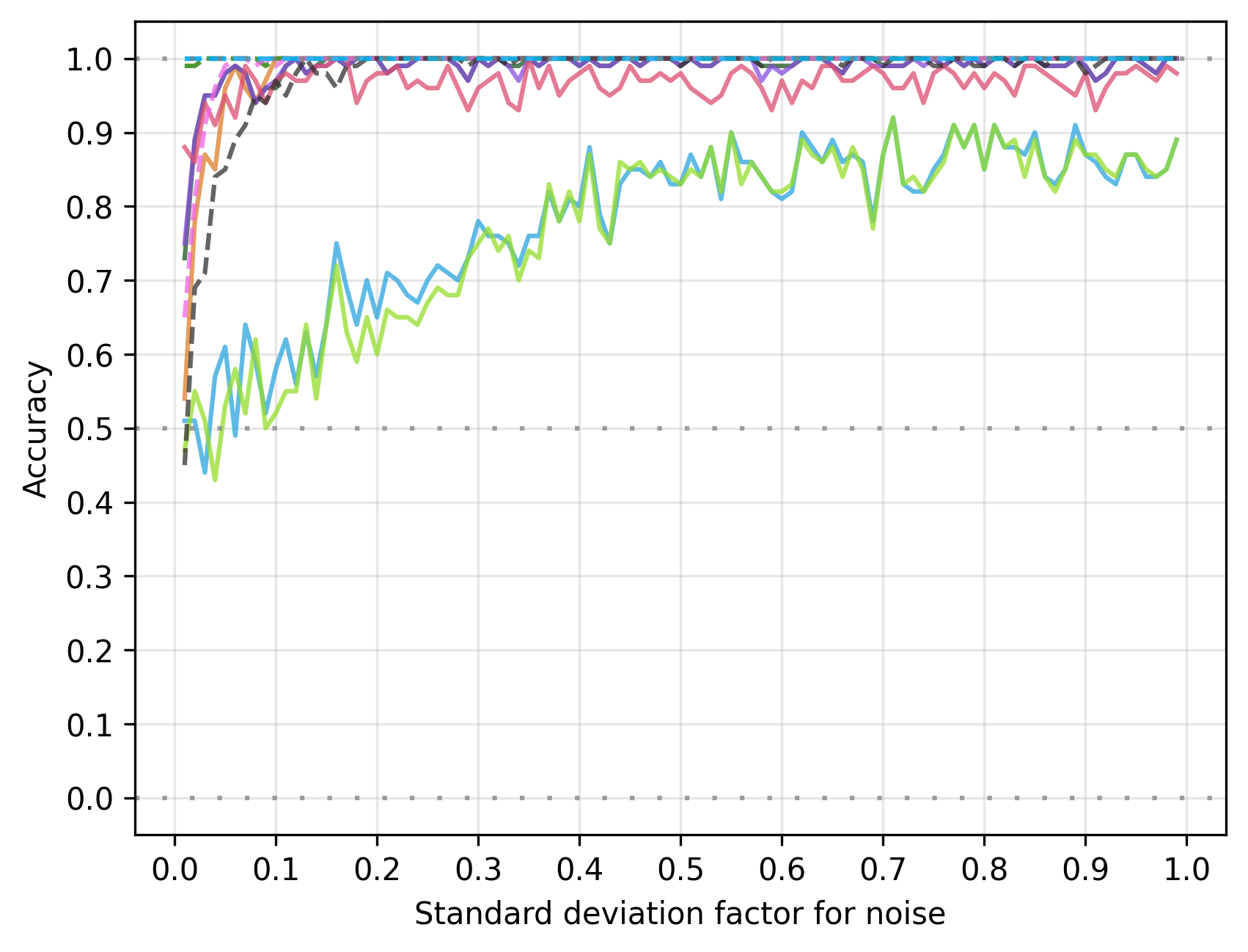}
\end{subfigure}%
\begin{subfigure}{.5\textwidth}
  \centering
  \includegraphics[scale=0.5]{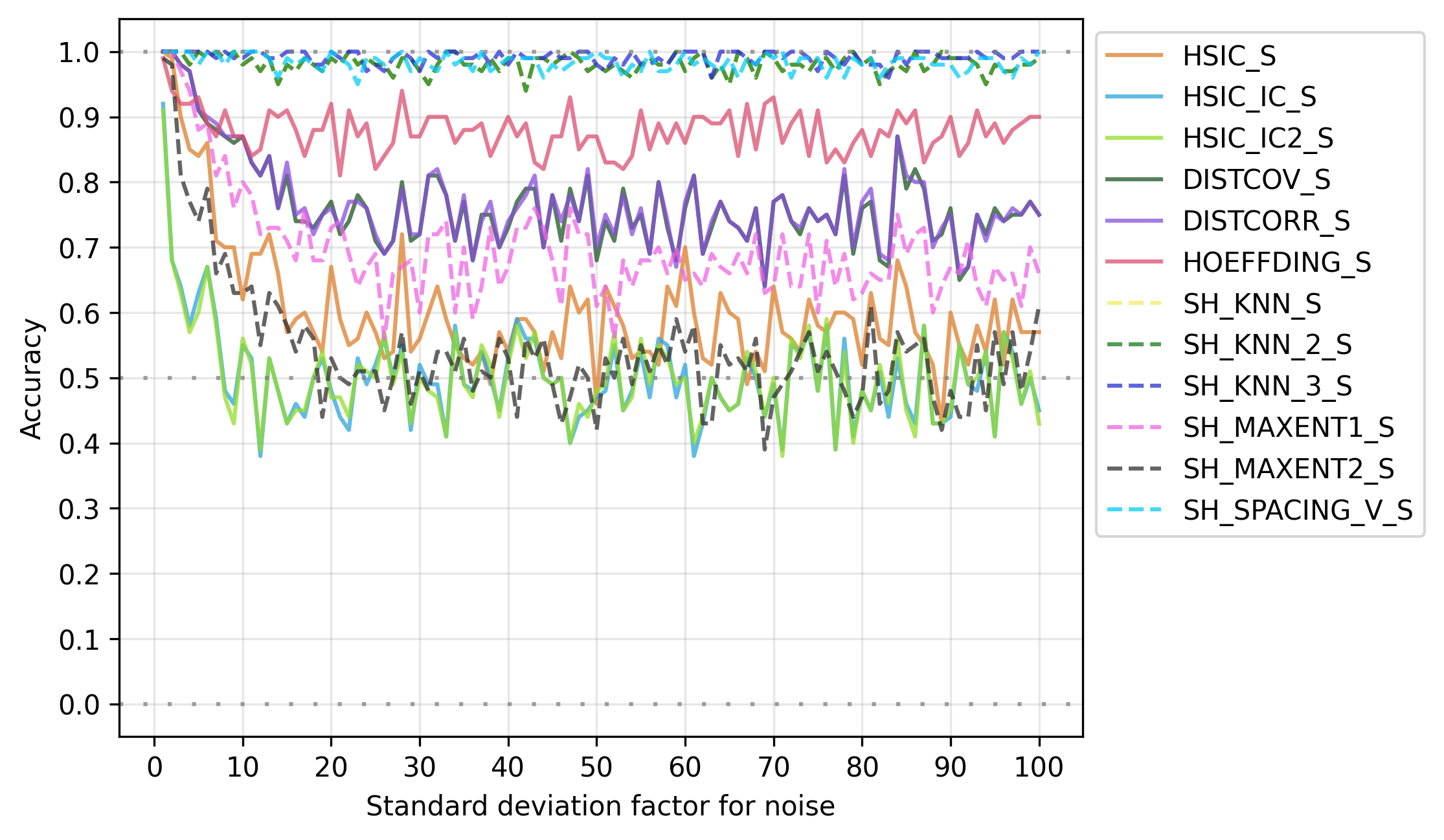}
\end{subfigure}
\caption{RESIT \& different noise levels \& decoupled estimation \& $Y = \mathcal{U}^3+\mathcal{L}$}
\label{fig:12}
\end{figure}

\cref{fig:11} shows the non-linear model $Y= \mathcal{U}^3 + \mathcal{N}$. All estimators, except
HSIC\_IC and HSIC\_IC2, remain
above 95\% for $i \in [0.05; 1]$ and for $i \in [1; 100]$ the estimators converge differently. All three
Shannon differential entropy measures with kNNs and SH\_SPACING\_V remain above 95\% accuracy.
DISTCOV, DISCORR and HOEFFDING keep a mean of $\sim$85\% accuracy. HSIC and SH\_MAXENT1 remain above 60\% accuracy. SH\_MAXENT2 is pretty much unidentifiable and HSIC\_IC and HSIC\_IC2
are unidentifiable for $i \in [0.01; 100]$.
\cref{fig:12} shows the non-linear model $Y= \mathcal{U}^3 + \mathcal{L}$. The behaviour of different estimators is almost the same
as for $Y= \mathcal{U}^3 + \mathcal{N}$. The only differences are that HSIC\_IC performs slightly better
for $i \in [0.2;1]$ and DISTCOV, DISTCORR, HSIC and SH\_MAXENT2 perform worse.

\newpage

\begin{figure}[h]
\centering
\begin{subfigure}{.5\textwidth}
  \centering
  \includegraphics[scale=0.5]{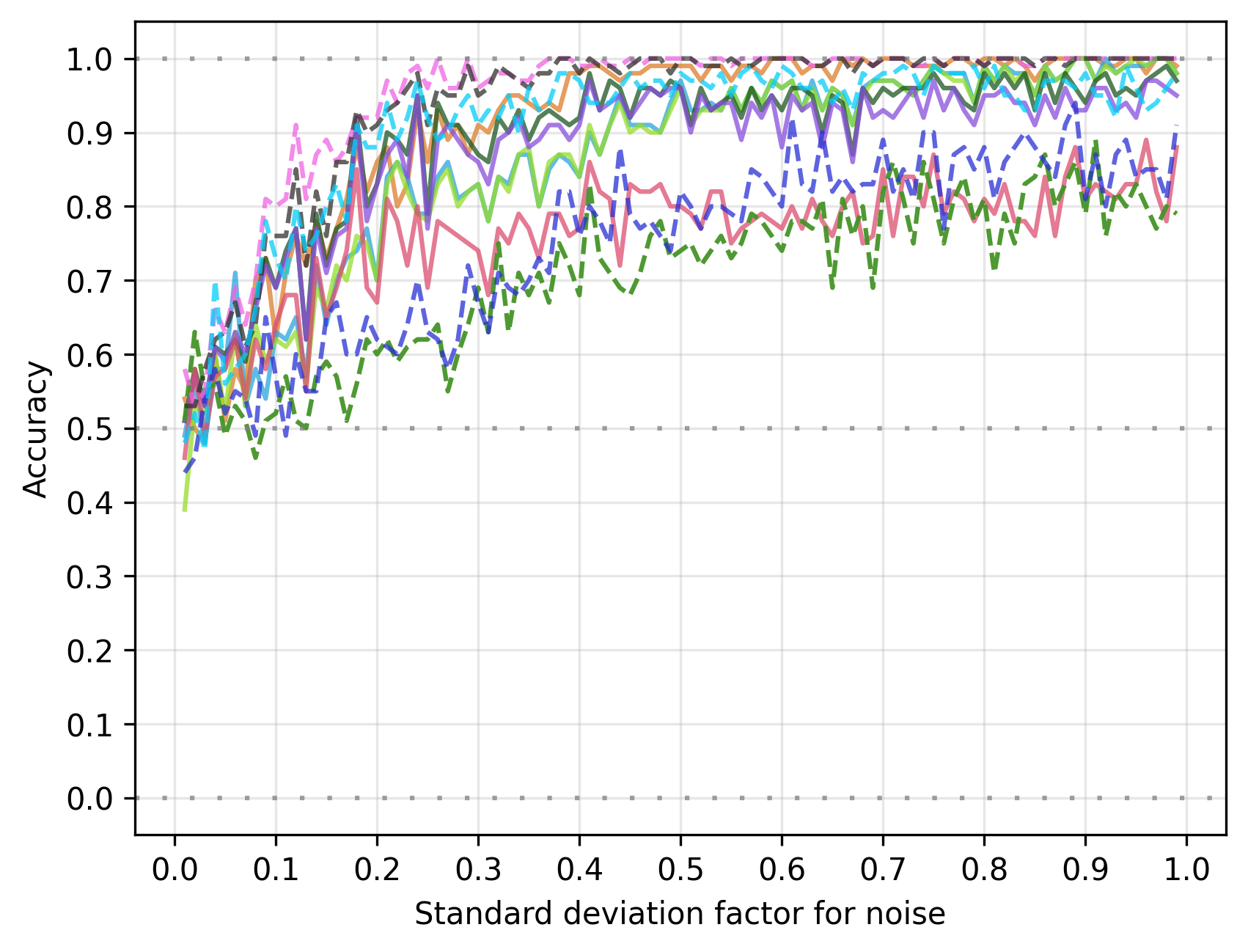}
\end{subfigure}%
\begin{subfigure}{.5\textwidth}
  \centering
  \includegraphics[scale=0.5]{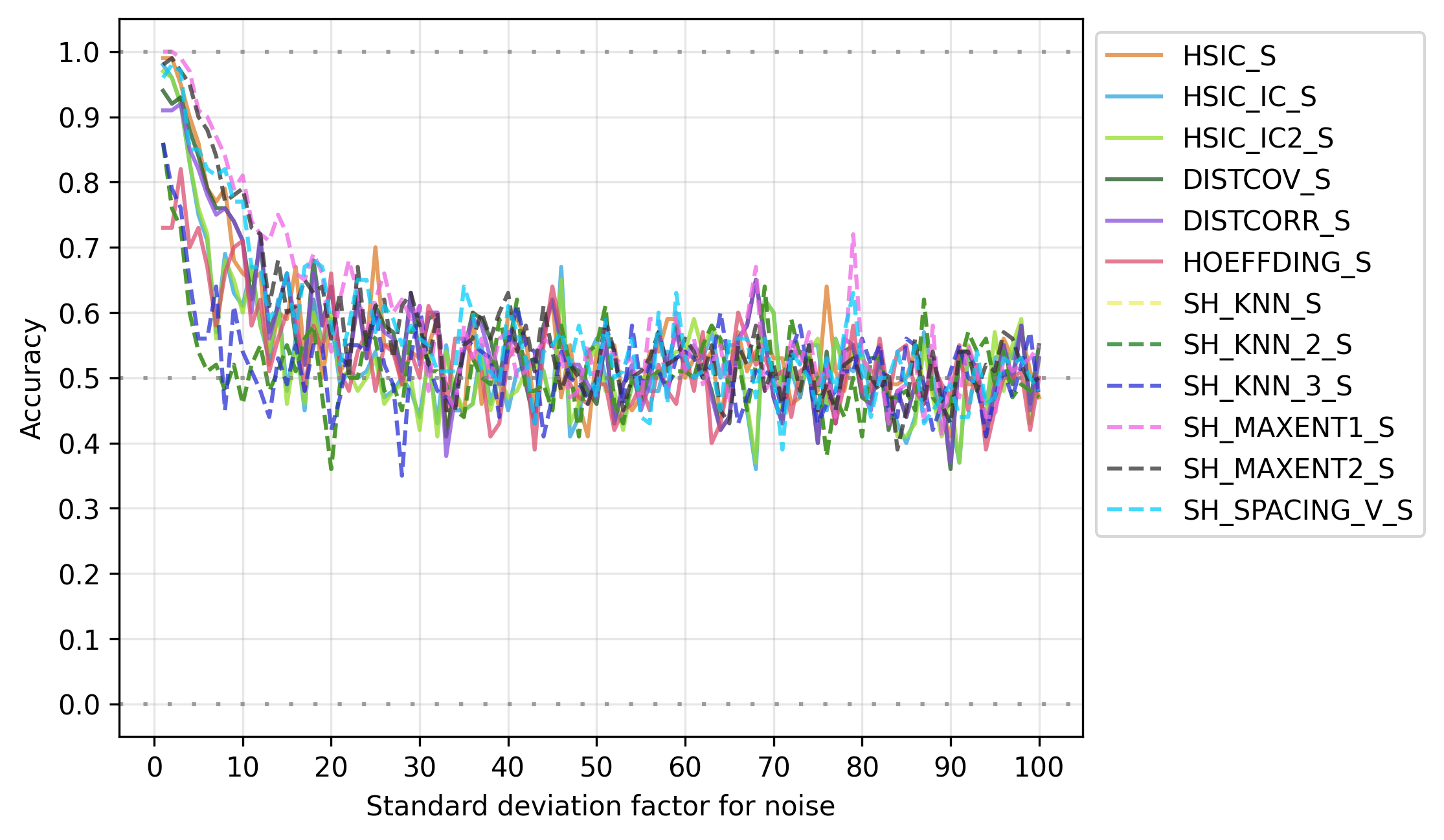}
\end{subfigure}
\caption{RESIT \& different noise levels \& decoupled estimation \& $Y = \mathcal{L}+\mathcal{L}$}
\label{fig:13}
\centering
\begin{subfigure}{.5\textwidth}
  \centering
  \includegraphics[scale=0.5]{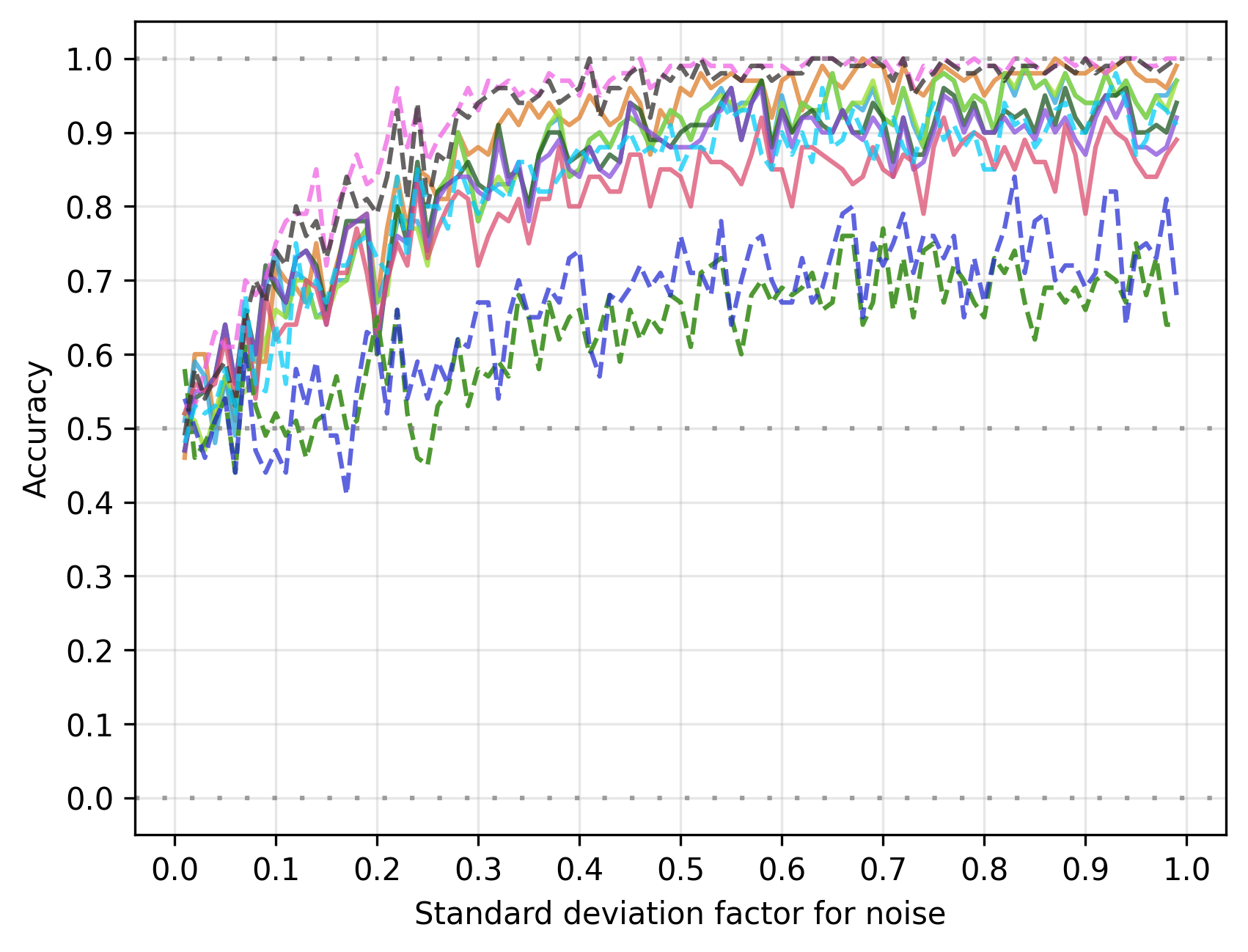}
\end{subfigure}%
\begin{subfigure}{.5\textwidth}
  \centering
  \includegraphics[scale=0.5]{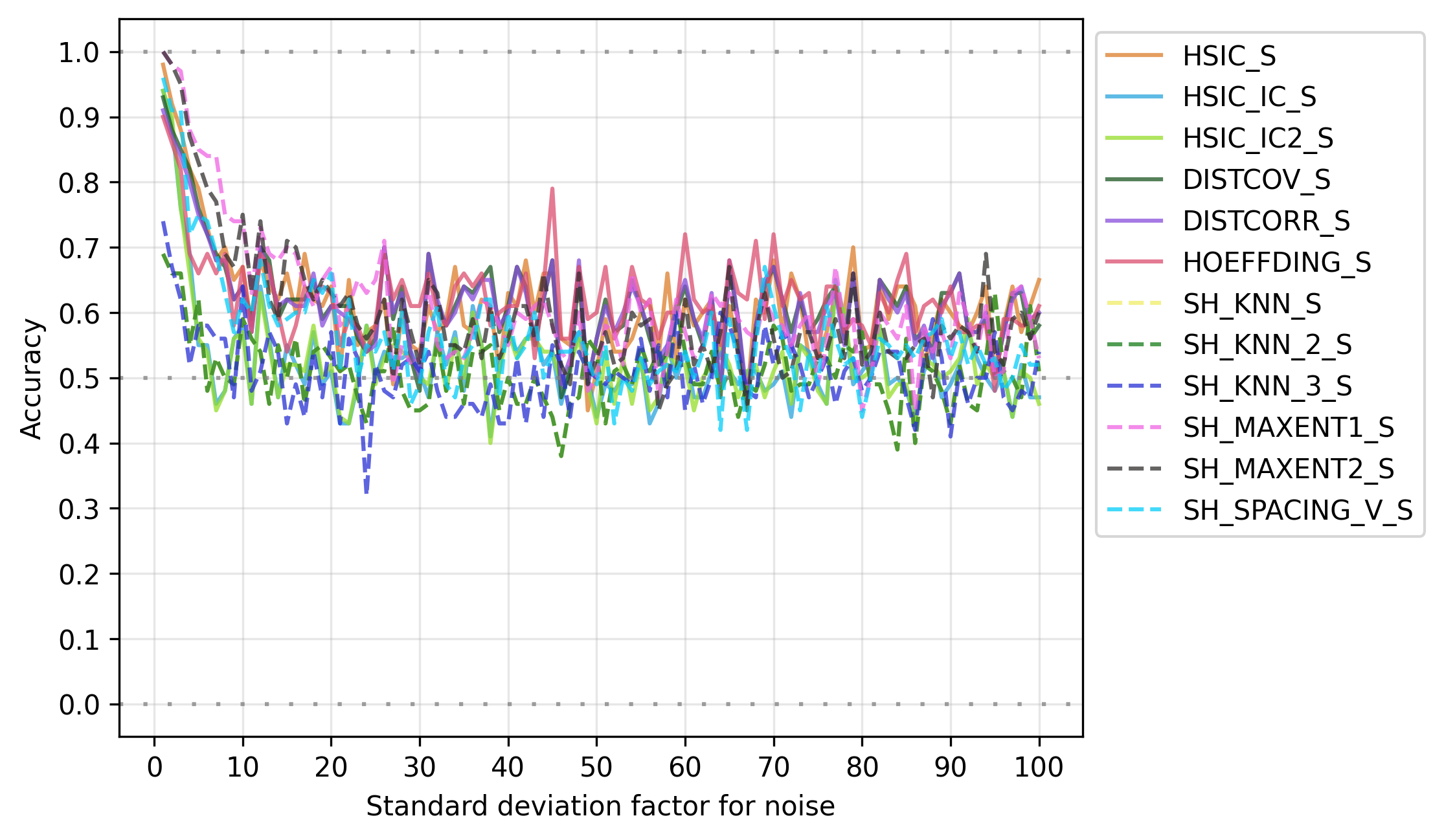}
\end{subfigure}
\caption{RESIT \& different noise levels \& decoupled estimation \& $Y = \mathcal{L}+\mathcal{N}$}
\label{fig:14}
\end{figure}

\cref{fig:13} shows the linear model $Y= \mathcal{L} + \mathcal{L}$. First, for $i \in [0.4;2]$ SH\_MAXENT1 and
SH\_MAXENT2 have accuracy close to 100\%. Next, for $i \in [0.4;1]$ HSIC, HSIC\_IC, HSIC\_IC2, SH\_SPACING\_V, DISTCOV and DISTCORR
remain above 90\% accuracy. HOEFFDING, and the three Shannon kNN estimators never reach an accuracy above 90\%.
After $i = 1$ all estimators drop fast towards 50\%.
\cref{fig:14} shows the linear model $Y= \mathcal{L} + \mathcal{N}$. This has a similar pattern as the previous one.
For $i \in [0.3; 1]$ HSIC, SH\_MAXENT1 and SH\_MAXENT2 have accuracy greater than 90\%.
SH\_SPACING\_V, HSIC\_IC, HSIC\_IC2, DISTCOV and DISTCORR lie between 85\% and 95\% accuracy
for $i \in [0.4; 1]$ and HOEFFDING remains between 80\% and 90\%.
Again, the three Shannon kNN estimators never reach an accuracy higher than 80\%.
After $i = 1$ all estimators drop fairly fast towards unidentifiability.
Note: in \cref{fig:13} and \cref{fig:14} SH\_KNN and SH\_KNN\_2 are overlapping completely.

\newpage

\begin{figure}[h]
\centering
\begin{subfigure}{.5\textwidth}
  \centering
  \includegraphics[scale=0.5]{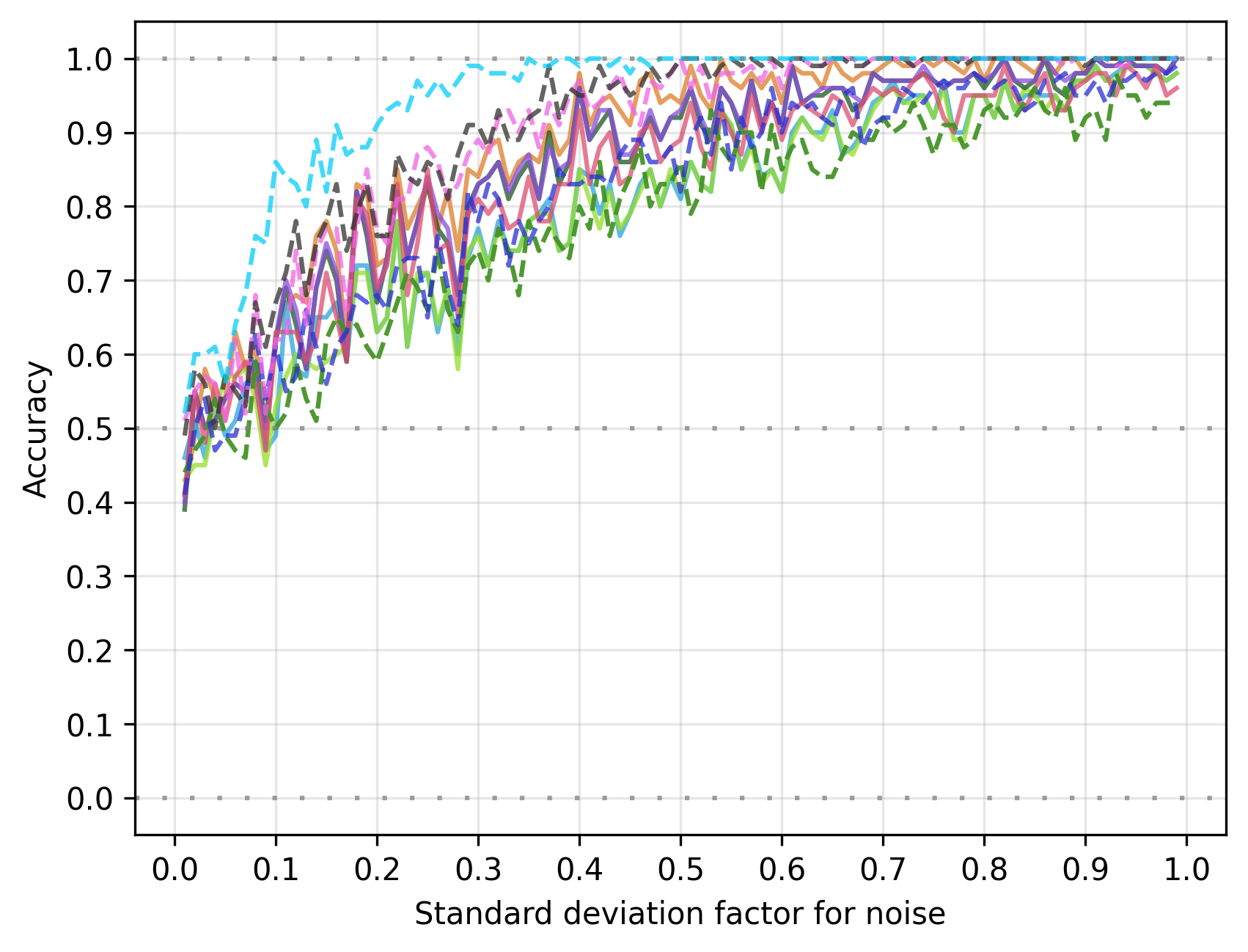}
\end{subfigure}%
\begin{subfigure}{.5\textwidth}
  \centering
  \includegraphics[scale=0.5]{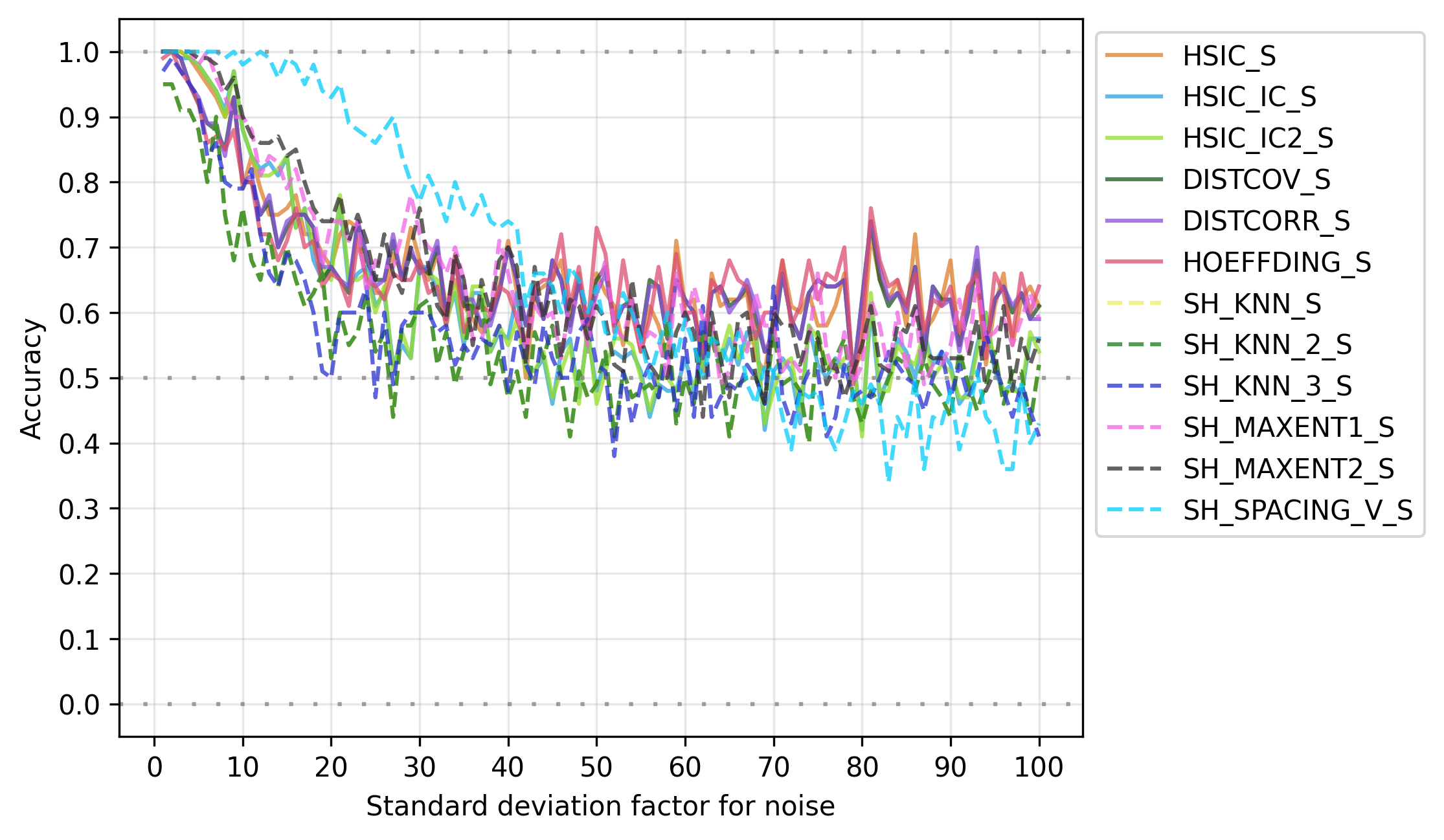}
\end{subfigure}
\caption{RESIT \& different noise levels \& decoupled estimation \& $Y = \mathcal{L}+\mathcal{U}$}
\label{fig:15}
\vspace{5mm} 
\centering
\begin{subfigure}{.5\textwidth}
  \centering
  \includegraphics[scale=0.5]{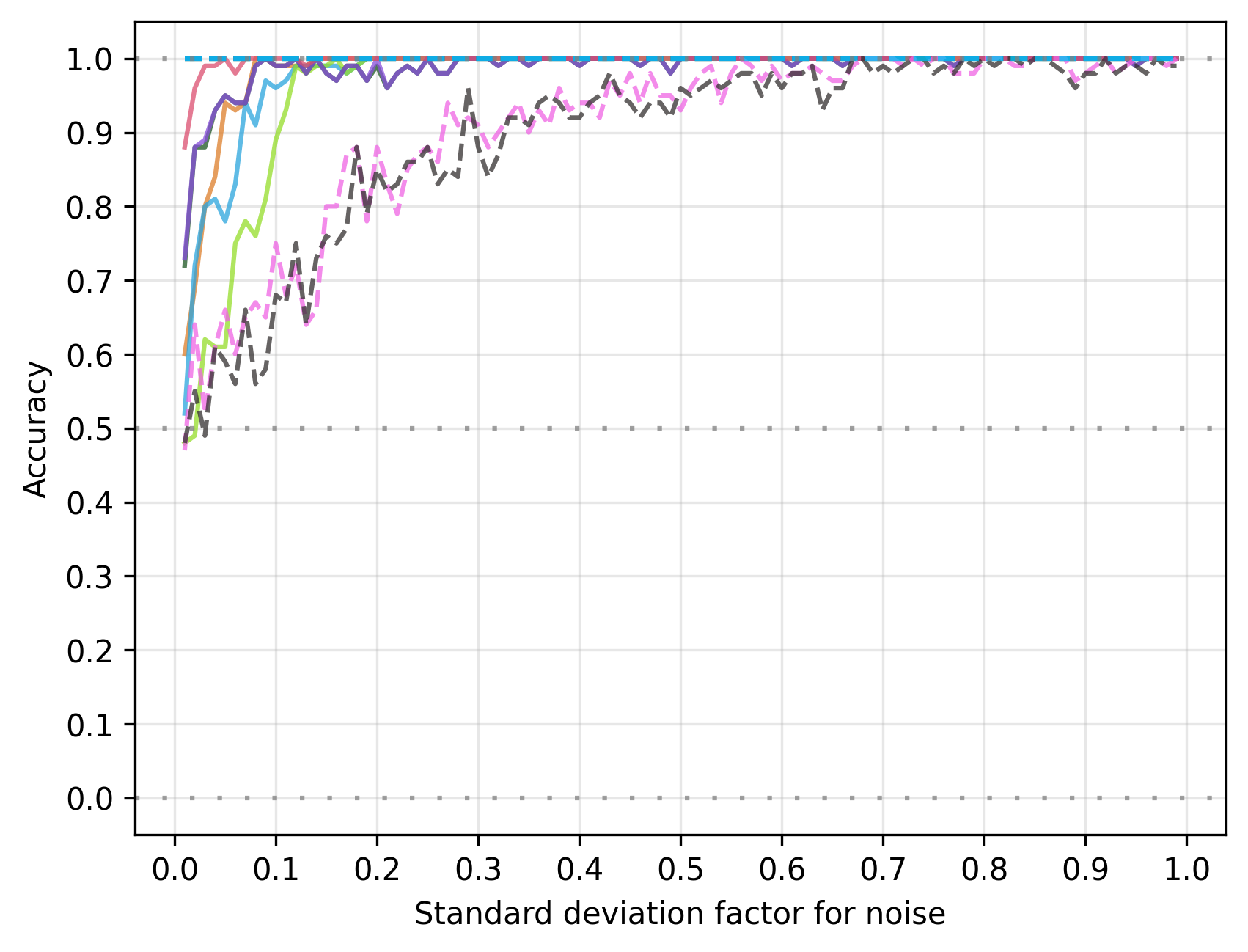}
\end{subfigure}%
\begin{subfigure}{.5\textwidth}
  \centering
  \includegraphics[scale=0.5]{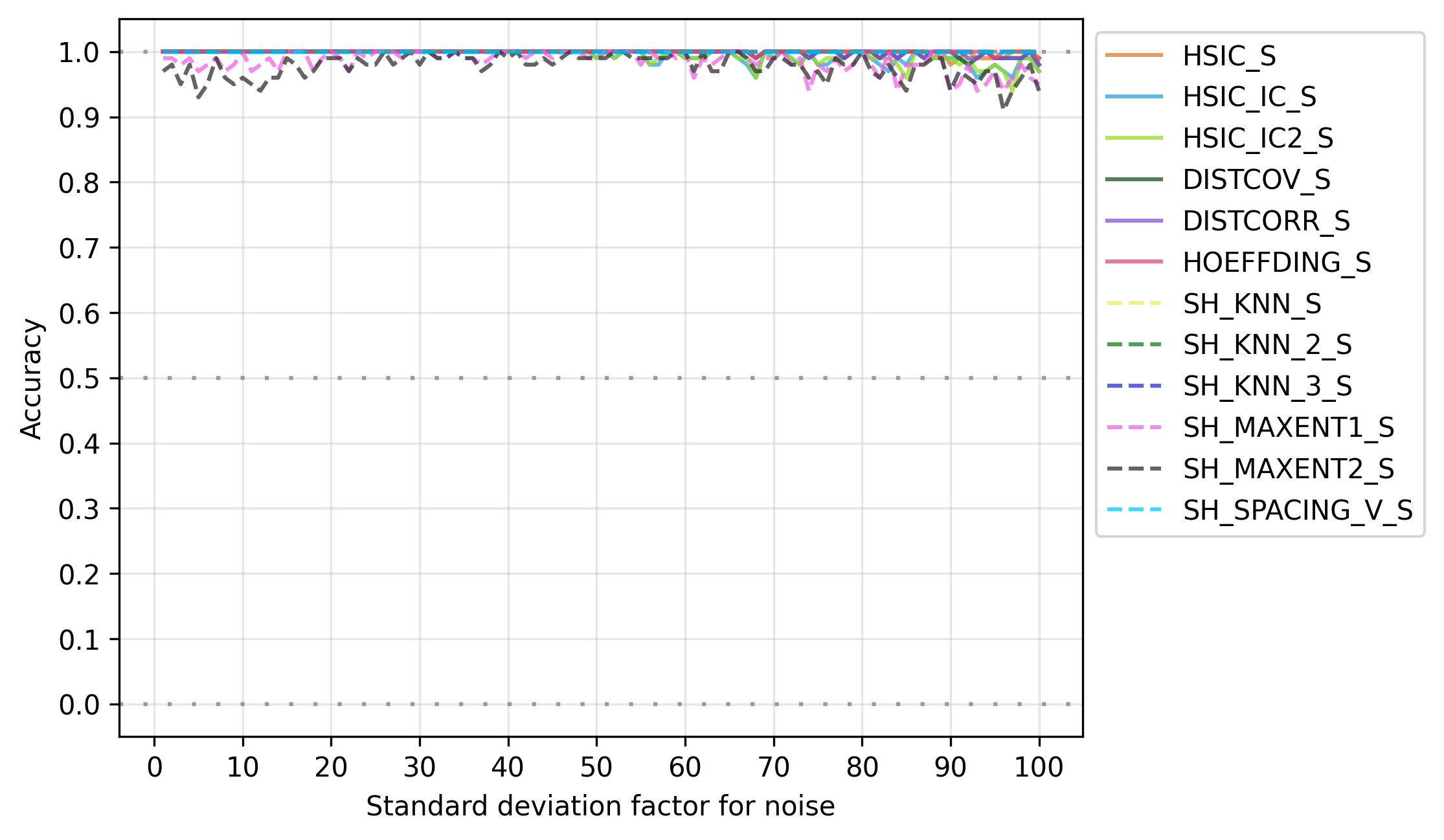}
\end{subfigure}
\caption{RESIT \& different noise levels \& decoupled estimation \& $Y = \mathcal{L}^3+\mathcal{L}$}
\label{fig:16}
\end{figure}

\cref{fig:15} shows the linear model $Y= \mathcal{L} + \mathcal{U}$. SH\_SPACING\_V performs the best of all
estimators and has an accuracy of 100\% for $i \in [0.5; 5]$. All other estimators slowly climb
towards good identifiability and for $i \in [0.7; 7]$ they remain above 90\% accuracy. Afterwards,
all other estimators drop with similar pace towards unidentifiability.
\cref{fig:16} shows the non-linear model $Y= \mathcal{L}^3 + \mathcal{L}$. This experiments shows the best
results of all. For $i \in [0.1;100]$ all estimators (except SH\_MAXENT1 and SH\_MAXENT2) have an accuracy 90\% or higher
SH\_SPACING\_V and the three Shannon kNN estimators have an accuracy of 100\% for $i \in [0.01; 100]$.
Only SH\_MAXENT1 and SH\_MAXENT2 perform bad at the beginning but still have an accuracy of 90\% or higher
for $i \in [0.35;100]$.

\newpage

\begin{figure}[h]
\centering
\begin{subfigure}{.5\textwidth}
  \centering
  \includegraphics[scale=0.5]{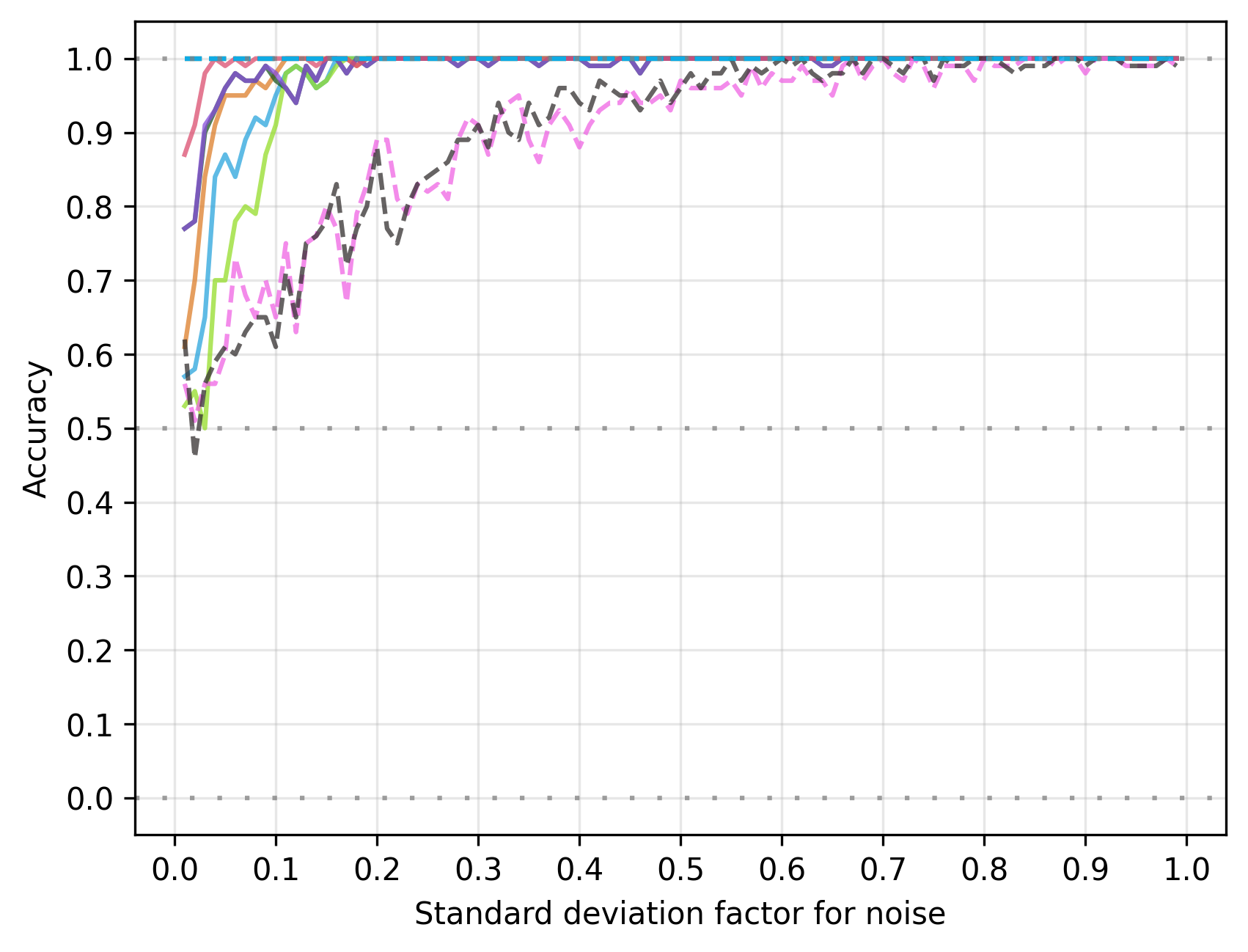}
\end{subfigure}%
\begin{subfigure}{.5\textwidth}
  \centering
  \includegraphics[scale=0.5]{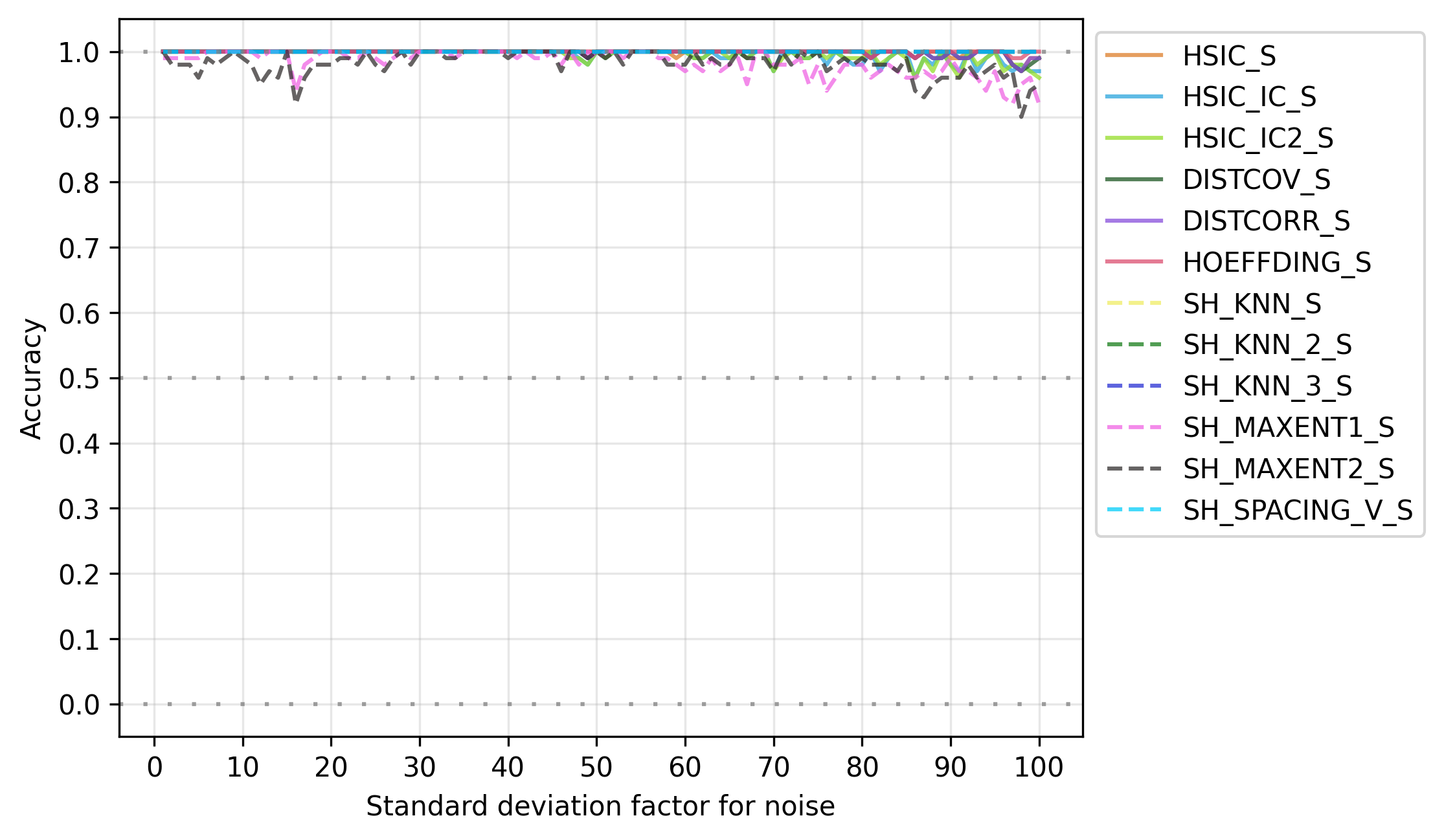}
\end{subfigure}
\caption{RESIT \& different noise levels \& decoupled estimation \& $Y = \mathcal{L}^3+\mathcal{N}$}
\label{fig:17}
\centering
\begin{subfigure}{.5\textwidth}
  \centering
  \includegraphics[scale=0.5]{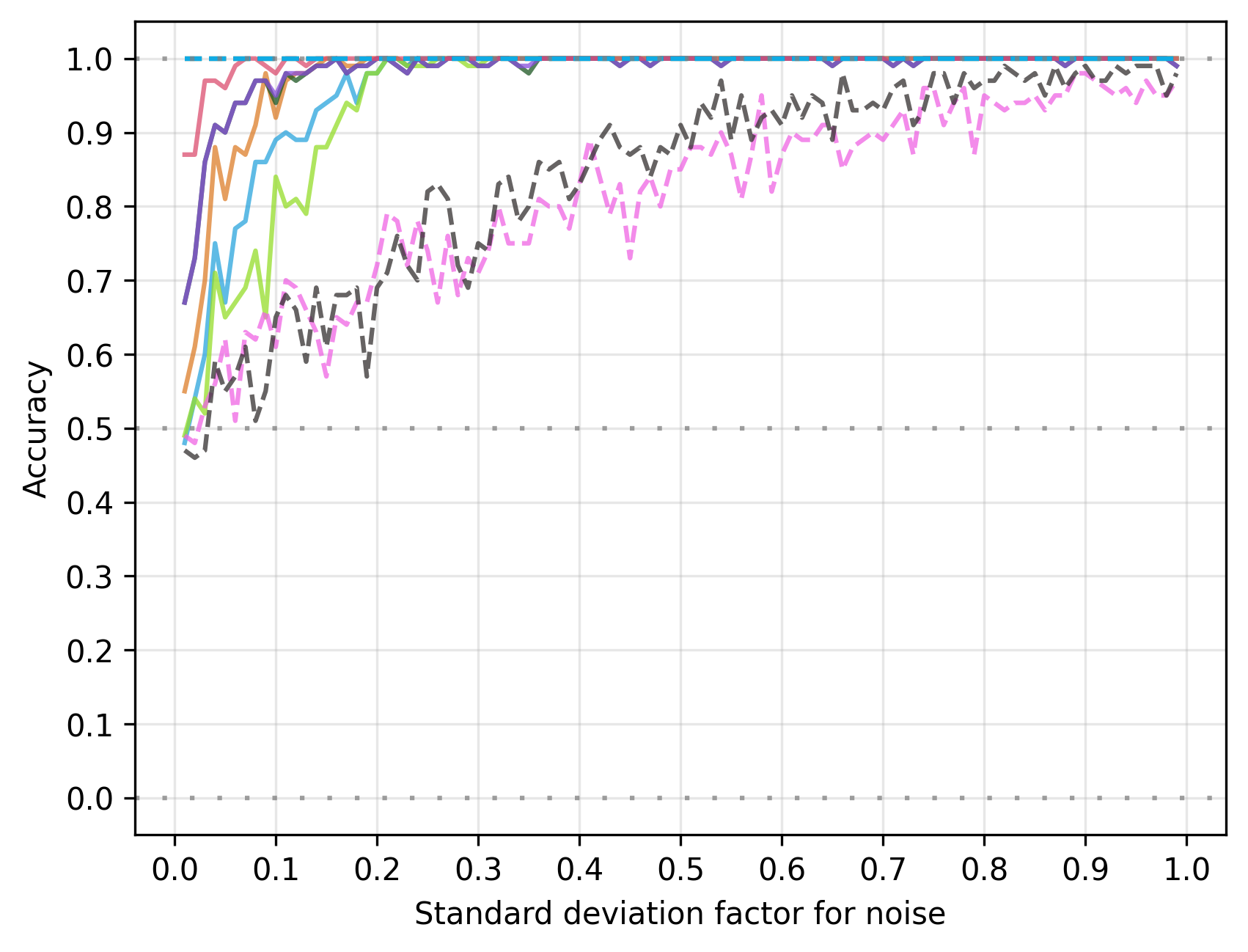}
\end{subfigure}%
\begin{subfigure}{.5\textwidth}
  \centering
  \includegraphics[scale=0.5]{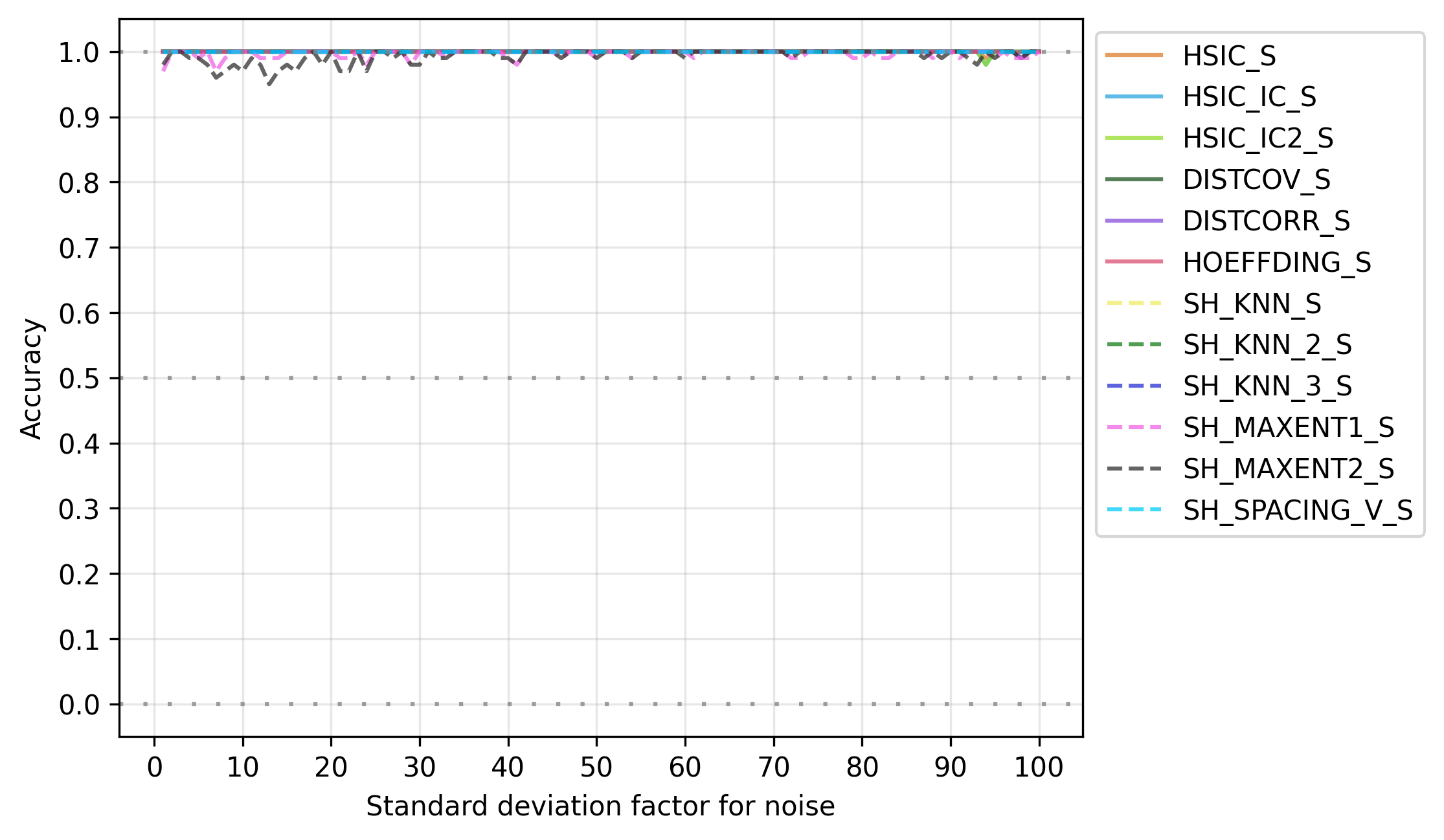}
\end{subfigure}
\caption{RESIT \& different noise levels \& decoupled estimation \& $Y = \mathcal{L}^3+\mathcal{U}$}
\label{fig:18}
\end{figure}

\cref{fig:17} shows the non-linear model $Y= \mathcal{L}^3 + \mathcal{N}$. This shows similar results as \cref{fig:16}.
For $i \in [0.1;100]$ all estimators (except SH\_MAXENT1 and SH\_MAXENT2) have an accuracy 90\% or higher
SH\_SPACING\_V and the three Shannon kNN estimators have an accuracy of 100\% for $i \in [0.01; 100]$.
Only SH\_MAXENT1 and SH\_MAXENT2 perform bad at the beginning but still have an accuracy of 90\% or higher
for $i \in [0.35;100]$.
\cref{fig:18} shows the non-linear model $Y= \mathcal{L}^3 + \mathcal{U}$. Similar as the two previous results.
For $i \in [0.15;100]$ all estimators (except SH\_MAXENT1 and SH\_MAXENT2) have an accuracy 90\% or higher
SH\_SPACING\_V, and the three Shannon kNN estimators have an accuracy of 100\% for $i \in [0.01; 100]$.
Only SH\_MAXENT1 and SH\_MAXENT2 perform bad at the beginning but still have an accuracy of 90\% or higher
for $i \in [0.7;100]$. For $i \in [1;100]$ all estimators are very close to 100\% accuracy.

\subsubsection*{Summary and Conclusion}
\cref{summarytable1} contains an overview of the results. The rows represent each structural equation
and each column represents the estimator used. The values in the cells describe on what range of $i$ the
estimator \textit{can} reach over 90\%. Estimators have some variance in the results and thus on some intervals they fall below 90\% accuracy. The limits in the cells were chosen as follows: the lower limit designates
where estimators reaches the first time 90\% or higher, and the upper limit designates the last time where it
reaches 90\% or higher. In between, most of the time estimators remain above 90\% or rarely fall below, but not
more than 10\%. Empty cells mean that in that case the estimator never reached 90\% or higher.\\
As the results show, different noise levels do have an impact on the identifiability performance 
in RESIT methods.
In general the linear equation models are more fragile in RESIT than the non-linear equation models
because non-linear relationships tend to break the symmetry between the variables easier, see
\citet{hoyer2008nonlinear}. Furthermore, in all cases the test results themselves
have a standard deviation between 0.05 to 0.1 as one can see in the sharp wiggles in the
plots.\\\\
Looking now only at the best estimation function and assuming a strong identifiability equal 
or greater than 90\% accuracy then for all linear cases factor $i$ is smaller than 10 (even
smaller than 5 in some cases, e.g., \cref{fig:3}) and bigger than 0.5. In other words,
if $i \notin [0.5; 5]$ then the accuracy is below 90\%.
This looks different for the non-linear cases. For \cref{fig:5} e.g., accuracy is greater
than 90\% if $i \in [0.05; 100]$. Accuracy is even equal to 100\% in the cases where $X \sim \mathcal{L}$
(\cref{fig:16} - \cref{fig:18}) with $i \in [0.01; 100]$.\\\\
Some estimators perform differently depending on the setup. For example, in \cref{fig:10} - \cref{fig:12}, the three 
Shannon differential entropy estimators with kNNs always perform above 90\% accuracy for $i \in [0.01; 100]$ 
and even with 100\% accuracy for $i \in [0.01; 100]$ in \cref{fig:16} - \cref{fig:18}. At the same time, the HSIC 
with incomplete cholesky (HSIC\_IC and HSIC\_IC2) has the worst performance. There is even one case where HSIC\_IC never
achieves identifiability (\cref{fig:11}). Overall, SH\_SPACING\_V performs the best in almost all
cases, and is only outperformed by SH\_MAXENT1 and SH\_MAXENT2 in \cref{fig:3}, \cref{fig:13} and \cref{fig:14}.
Some independence tests lose some
of the accuracy while entropy estimators retain an accuracy over 90\% (e.g., \cref{fig:10}, \cref{fig:11} and
\cref{fig:12}).\\
Additionally, it is worth to mention that entropy estimators are less computationally demanding than 
independence tests but can be quite sensitive to discretization effects \citet{mooij2016distinguishing}.
However, entropy estimators can only be used with the prior assumption we made:
\textit{there is only one causal direction and it is present in the model}. 
\cref{res2} contains experiments with the same setup but without this prior assumption.

In the next section, we compare decoupled estimation and coupled estimation.

\begin{sidewaystable}[p]
\begin{center}
\resizebox{\textheight}{!}{%
\begin{tabular}{|c|c|c|c|c|c|c|c|c|c|c|c|c|}
     \hline
     \textbf{Equation} & \textbf{HSIC} & \textbf{HISC\_IC} & \textbf{HSIC\_IC2} & \textbf{DISTCOV} & \textbf{DISTCORR} & \textbf{HOEFFDING} & \textbf{SH\_KNN} & \textbf{SH\_KNN\_2} & \textbf{SH\_KNN\_3} & \textbf{SH\_MAXENT1} &  \textbf{SH\_MAXENT2} & \textbf{SH\_SPACING\_V}\\\hline
     
     \textbf{$Y=\mathcal{N}+\mathcal{N}$} & & & & & & & & & & & & \\\hline
     
     \textbf{$Y=\mathcal{N}+\mathcal{U}$} & 0.6 - 4 & 3 - 7 & 3 - 7 & 1 & 1 &  &  &  & 0.85 - 4 & 0.65 - 5 & 0.4 - 8 & 0.2 - 22\\\hline
     
     \textbf{$Y=\mathcal{N}+\mathcal{L}$} & 0.38 - 2 & 0.75 - 2 & 0.75 - 2 & 0.8 - 1 & 0.87 - 1 & & & & & 0.3 - 3 & 0.33 - 3 & 0.82 - 1 \\\hline
     
     \textbf{$Y=\mathcal{N}^3+\mathcal{N}$} & 0.09 - 68 & 0.1 - 40 & 0.1 - 40 & 0.07 - 100 & 0.07 - 100 & 0.02 - 100 & 0.01 - 100 & 0.01 - 100 & 0.01 - 100 & 0.2 - 45 & 0.28 - 45 & 0.01 - 100 \\\hline
     
     \textbf{$Y=\mathcal{N}^3+\mathcal{U}$} & 0.16 - 100 & 0.16 - 88 & 0.17 - 88 & 0.14 - 100 & 0.14 - 100 & 0.05 - 100 & 0.01 - 100 & 0.01 - 100 & 0.01 - 100 & 0.33 - 85 & 0.43 - 91 & 0.01 - 100 \\\hline
     
     \textbf{$Y=\mathcal{N}^3+\mathcal{L}$} & 0.11 - 70 & 0.1 - 35 & 0.1 - 35 & 0.08 - 88 & 0.05 - 88 & 0.04 - 100 & 0.01 - 100 & 0.01 - 100 & 0.01 - 100 & 0.18 - 67 & 0.25 - 65 & 0.01 - 100\\\hline
     
     \textbf{$Y=\mathcal{U}+\mathcal{U}$} & 0.18 - 6 & 3 & 3 & 0.2 - 3 & 0.2 - 3 & 0.2 - 3 & 0.2 - 3 & 0.2 - 2 & 0.2 - 3 & 0.23 - 3 & 0.12 - 8 & 0.05 - 21 \\\hline
     
     \textbf{$Y=\mathcal{U}+\mathcal{N}$} & 0.21 - 1 & & & 0.25 - 1 & 0.25 - 1 & 1 & 0.26 - 1 & 0.26 - 1 & 0.2 - 1 & 0.2 - 1 & 0.1 - 4 & 0.04 - 9 \\\hline
     
     \textbf{$Y=\mathcal{U}+\mathcal{L}$} & 0.15 - 1 & 0.35 - 1 & 0.35 - 1 & 0.2 - 1 & 0.2 - 1 & 0.15 - 1 & 0.21 - 1 & 0.21 - 1 & 0.16 - 1 & 0.12 - 3 & 0.1 - 3 & 0.03 - 8 \\\hline
     
     \textbf{$Y=\mathcal{U}^3+\mathcal{U}$} & 0.05 - 6 & & & 0.03 - 20 & 0.03 - 20 & 0.03 - 100 & 0.01 - 100 & 0.01 - 100 & 0.01 - 100 & 0.05 - 7 & 0.09 - 5 & 0.01 - 100 \\\hline
     
     \textbf{$Y=\mathcal{U}^3+\mathcal{N}$} & 0.04 - 3 & & & 0.01 - 85 & 0.01 - 85 & 0.01 - 95 & 0.01 - 100 & 0.01 - 100 & 0.01 - 100 & 0.02 - 4 & 0.08 - 2 & 0.01 - 100 \\\hline
     
     \textbf{$Y=\mathcal{U}^3+\mathcal{L}$} & 0.05 - 3 & 0.7 - 1 & 0.7 - 1 & 0.02 - 5 & 0.02 - 5 & 0.02 - 100 & 0.01 - 100 & 0.01 - 100 & 0.01 - 100 & 0.03 - 4 & 0.06 - 3 & 0.01 - 100 \\\hline
     
     \textbf{$Y=\mathcal{L}+\mathcal{L}$} & 0.3 - 5 & 0.4 - 4 & 0.4 - 4 & 0.23 - 4 & 0.23 - 4 & & & & 0.6 - 1 & 0.2 - 6 & 0.17 - 5 & 0.17 - 4 \\\hline
     
     \textbf{$Y=\mathcal{L}+\mathcal{N}$} & 0.32 - 3 & 0.36 - 3 & 0.36 - 3 & 0.33 - 2 & 0.33 - 2 & 0.57 - 1 & & & & 0.21 - 4 & 0.21 - 4 & 0.49 - 4 \\\hline
     
     \textbf{$Y=\mathcal{L}+\mathcal{U}$} & 0.4 - 8 & 0.6 - 7 & 0.6 - 7 & 0.4 - 5 & 0.4 - 5 & 0.4 - 5 & 0.53 - 4 & 0.53 - 4 & 0.51 - 5 & 0.32 - 10 & 0.32 - 10 & 0.16 - 27 \\\hline
     
     \textbf{$Y=\mathcal{L}^3+\mathcal{L}$} & 0.05 - 100 & 0.06 - 100 & 0.1 - 100 & 0.03 - 100 & 0.03 - 100 & 0.01 - 100 & 0.01 - 100 & 0.01 - 100 & 0.01 - 100 & 0.26 - 100 & 0.28 - 100 & 0.01 - 100 \\\hline
     
     \textbf{$Y=\mathcal{L}^3+\mathcal{N}$} & 0.04 - 100 & 0.07 - 100 & 0.1 - 100 & 0.03 - 100 & 0.03 - 100 & 0.02 - 100 & 0.01 - 100 & 0.01 - 100 & 0.01 - 100 & 0.28 - 100 & 0.3 - 100 & 0.01 - 100 \\\hline
     
     \textbf{$Y=\mathcal{L}^3+\mathcal{U}$} & 0.07 - 100 & 0.12 - 100 & 0.15 - 100 & 0.04 - 100 & 0.04 - 100 & 0.02 - 100 & 0.01 - 100 & 0.01 - 100 & 0.01 - 100 & 0.53 - 100 & 0.43 - 100 & 0.01 - 100\\\hline
\end{tabular}}
\caption{Summary Table for RESIT \& different noise levels \& Decoupled estimation. The numbers reflect the ranges of noise that allow identifiability with accuracy around 90\%.}
\label{summarytable1}
\end{center}
\end{sidewaystable}

\section{RESIT with Different Noise Levels and coupled estimation}\label{res3}

In the previous section, \cref{res1}, we discussed decoupled estimation when data was split into training (80\%) an testing (20\%) sets. The reason one would want to split
data is of computational nature: smaller sets of samples allow for faster
computation of the estimates, notably the independence estimators, but might decrease the accuracy
of the algorithm. 
The same data can be analyzed using the coupled estimation method when the entire set of 1000 samples
is used for both the regression step and the estimation step, and in this case we achieve higher
accuracy (or overall performance of the algorithm). Therefore, choosing between decoupled and coupled
estimation is a trade-off between identifiability performance and computation speed.
In this section we analyze the performance of RESIT for this setup.
As the experimental procedure and the setup stay the same as in \cref{res1}, we directly proceed to the discussion of results.

\subsection{Results}
The figures for the results here are to be interpreted in the same way as in the section before (see \cref{results1}).

\newpage
\subsubsection*{Individual Analysis}
\begin{figure}[h]
\centering
\begin{subfigure}{.5\textwidth}
  \centering
  \includegraphics[scale=0.5]{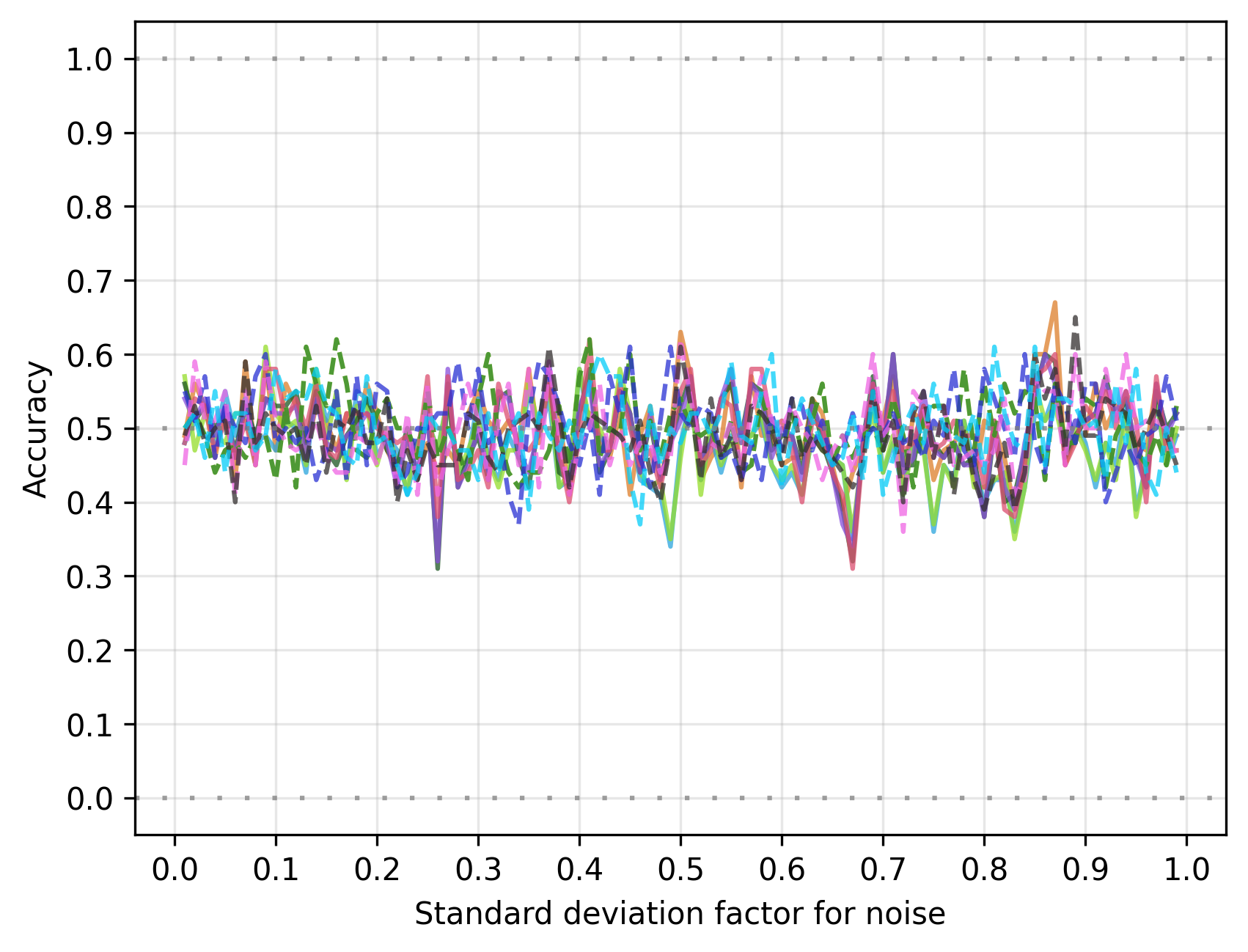}
\end{subfigure}%
\begin{subfigure}{.5\textwidth}
  \centering
  \includegraphics[scale=0.5]{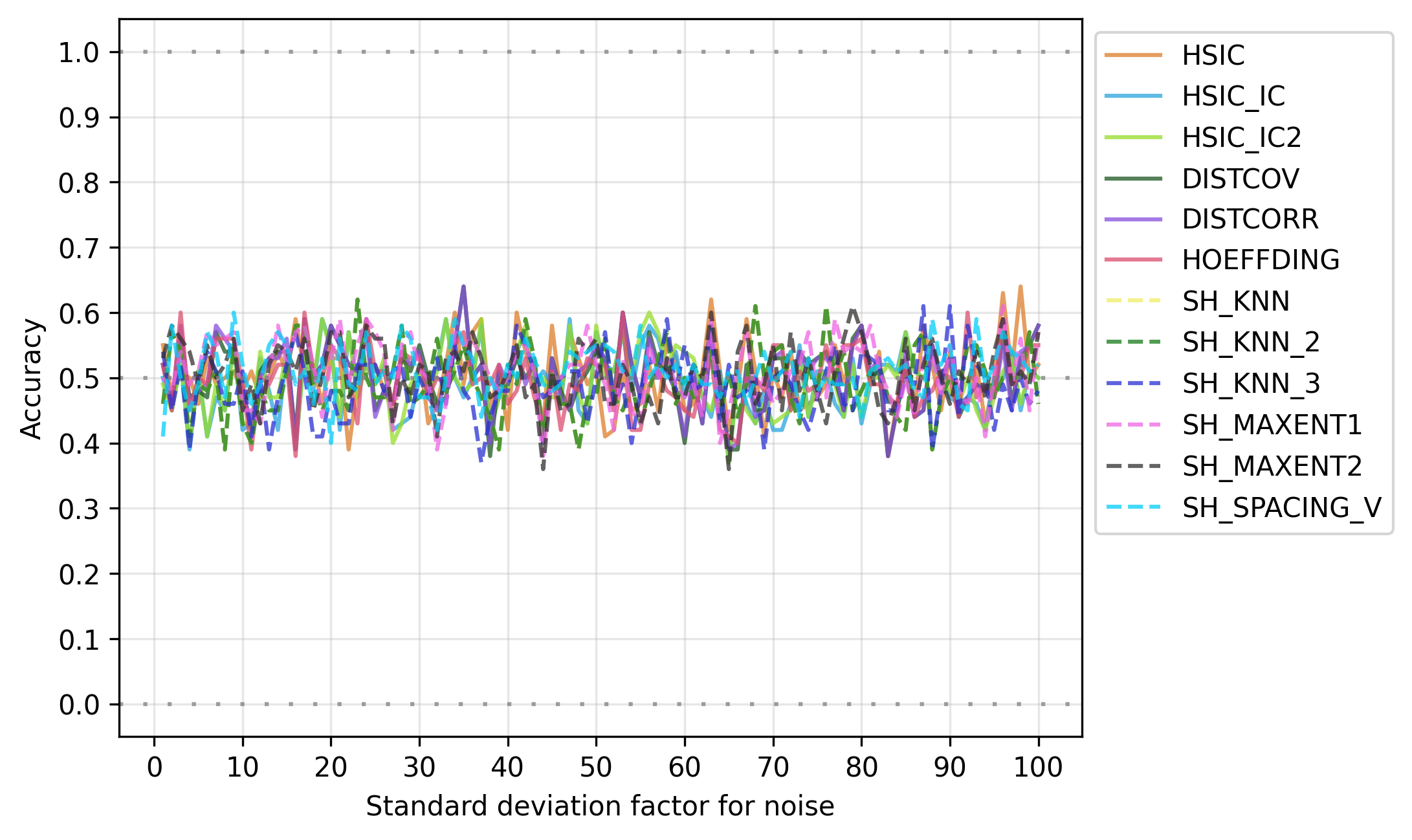}
\end{subfigure}
\caption{RESIT \& different noise levels \& coupled estimation \& $Y = \mathcal{N}+\mathcal{N}$}
\label{fig:19}

\centering
\begin{subfigure}{.5\textwidth}
  \centering
  \includegraphics[scale=0.5]{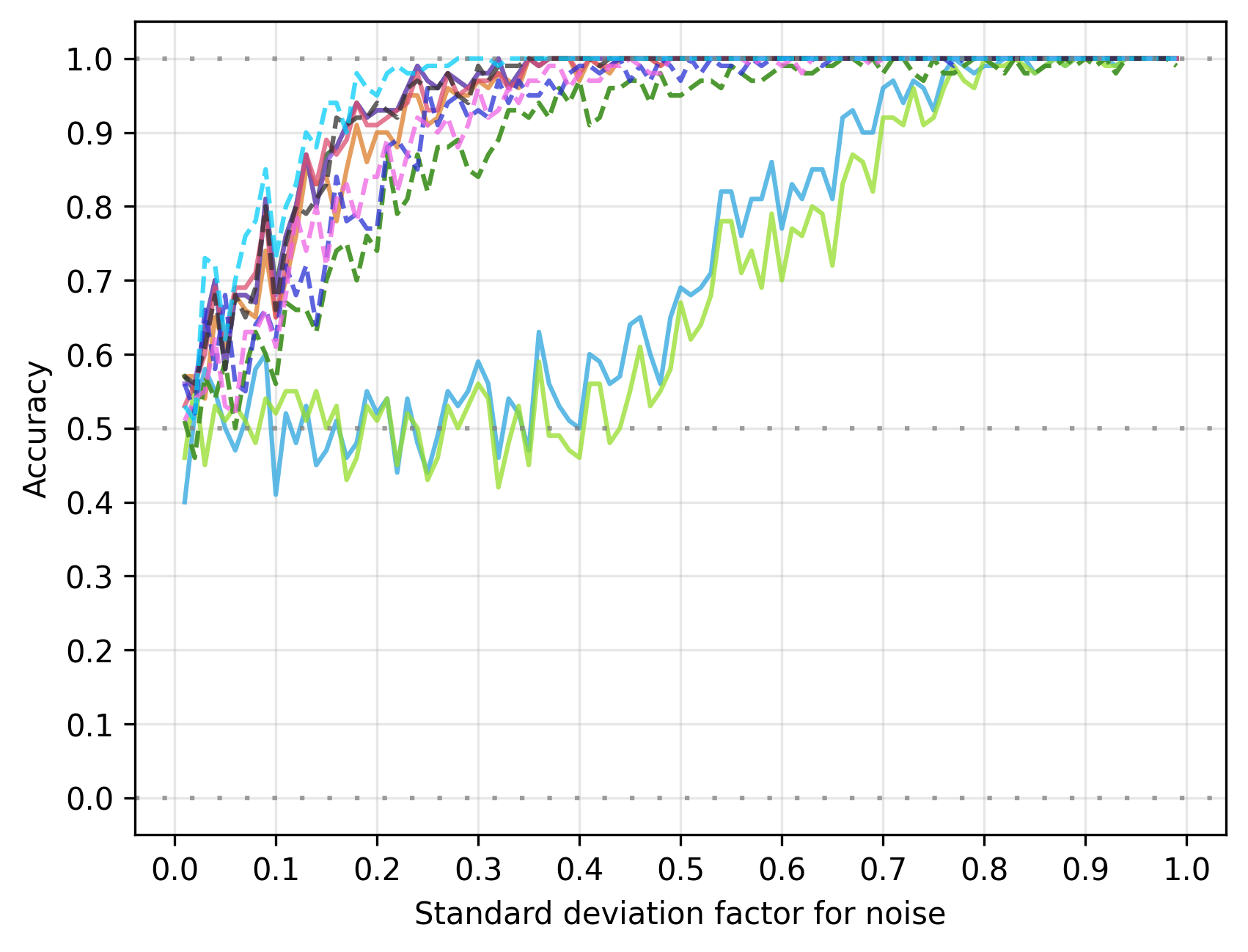}
\end{subfigure}%
\begin{subfigure}{.5\textwidth}
  \centering
  \includegraphics[scale=0.5]{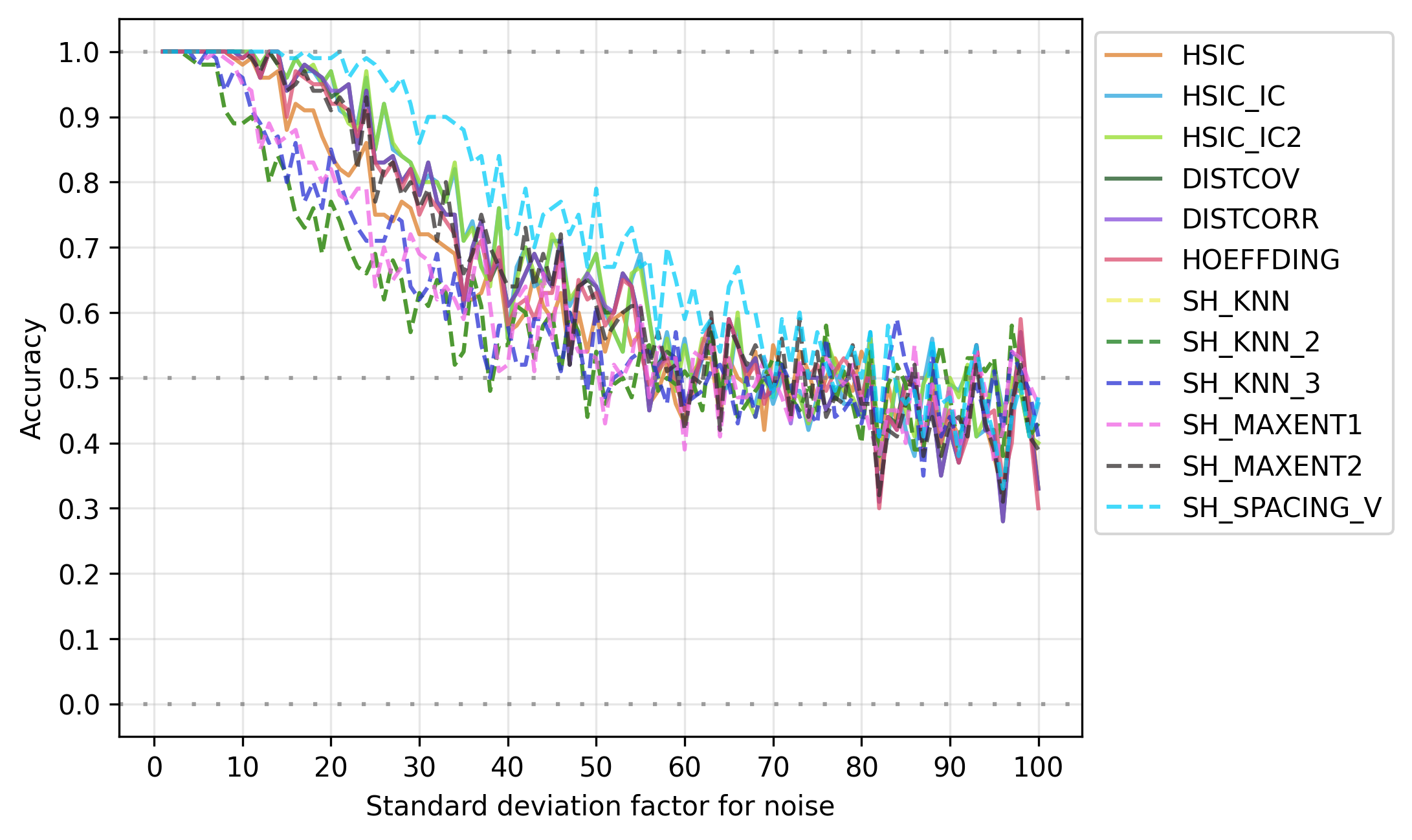}
\end{subfigure}
\caption{RESIT \& different noise levels \& coupled estimation \& $Y = \mathcal{N}+\mathcal{U}$}
\label{fig:20}
\end{figure}

Again, reading individual results is not necessary as we provide an summary table in the summary section
(\cref{summarytable2}).\\
\cref{fig:19} is the only case which remains unchanged performance wise ($Y= \mathcal{N} + \mathcal{N}$). 
\cref{fig:20} shows the linear model with $Y= \mathcal{N} + \mathcal{U}$.
All estimators reach now an accuracy close to 100\% at some interval $i \in [0.8; 5]$.
The accuracy in the $[0.01;1]$ interval climbs now faster for all estimators and drops
more slowly for $i \in [1;100]$.

\newpage

\begin{figure}[h]
\centering
\begin{subfigure}{.5\textwidth}
  \centering
  \includegraphics[scale=0.5]{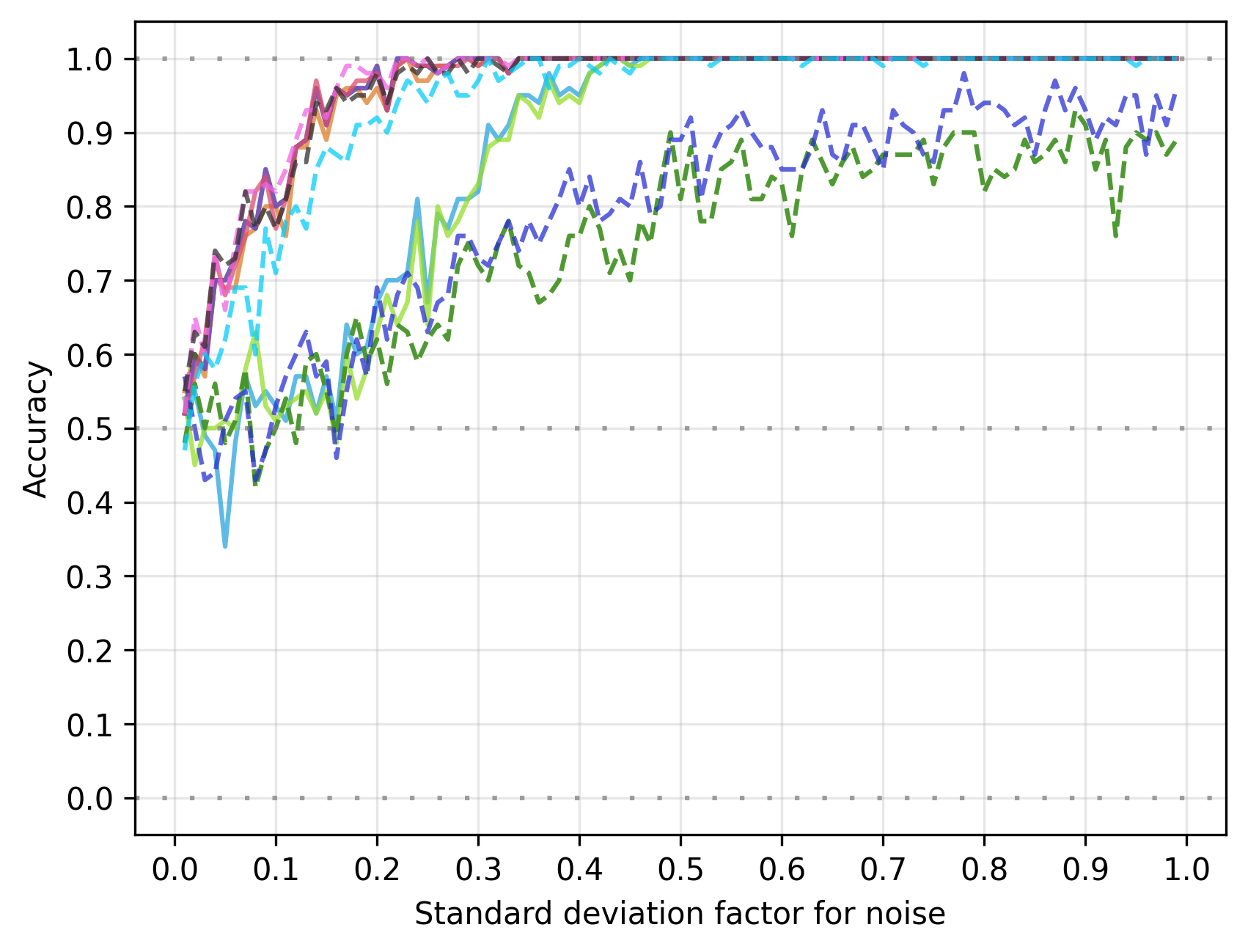}
\end{subfigure}%
\begin{subfigure}{.5\textwidth}
  \centering
  \includegraphics[scale=0.5]{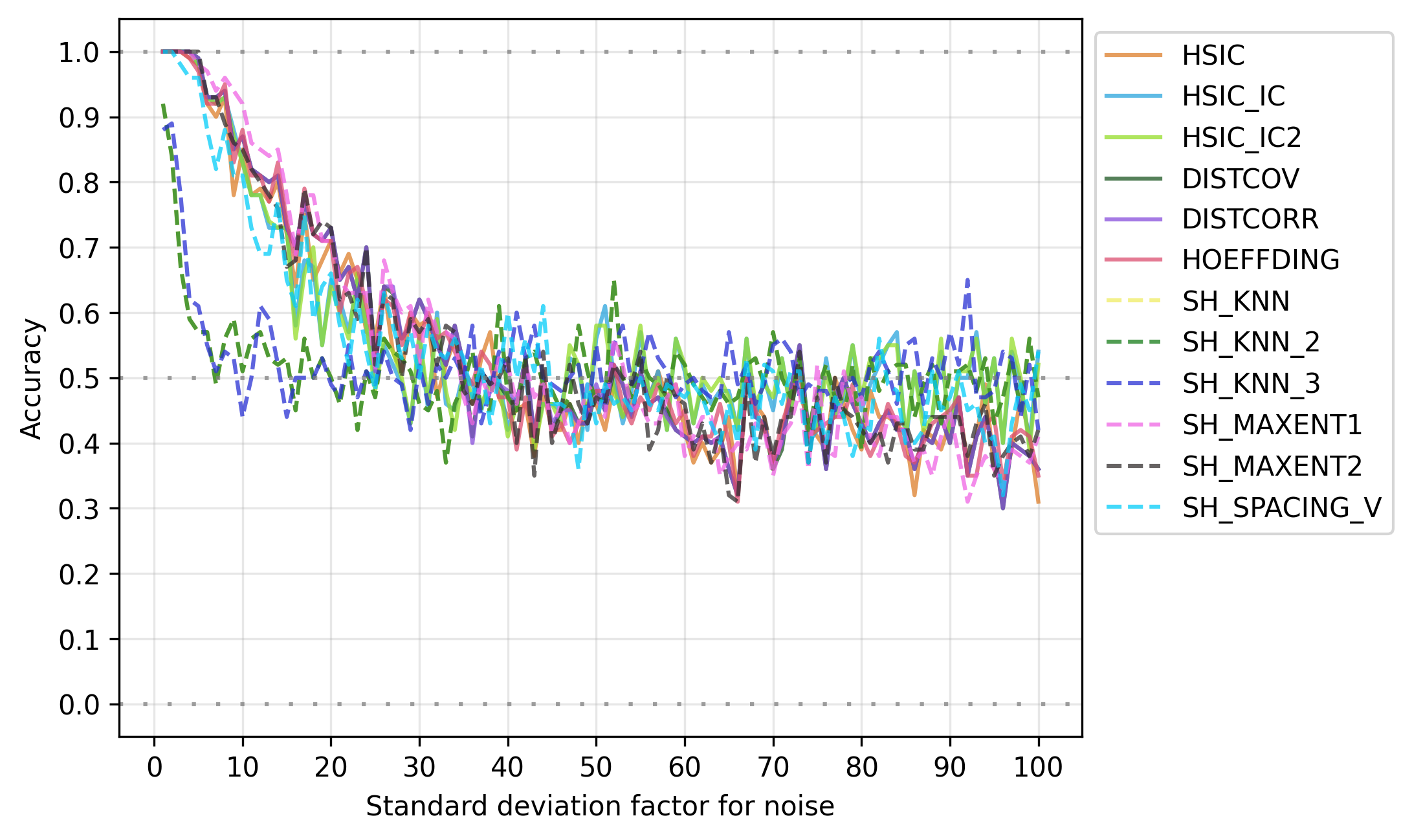}
\end{subfigure}
\caption{RESIT \& different noise levels \& coupled estimation \& $Y = \mathcal{N}+\mathcal{L}$}
\label{fig:21}

\centering
\begin{subfigure}{.5\textwidth}
  \centering
  \includegraphics[scale=0.5]{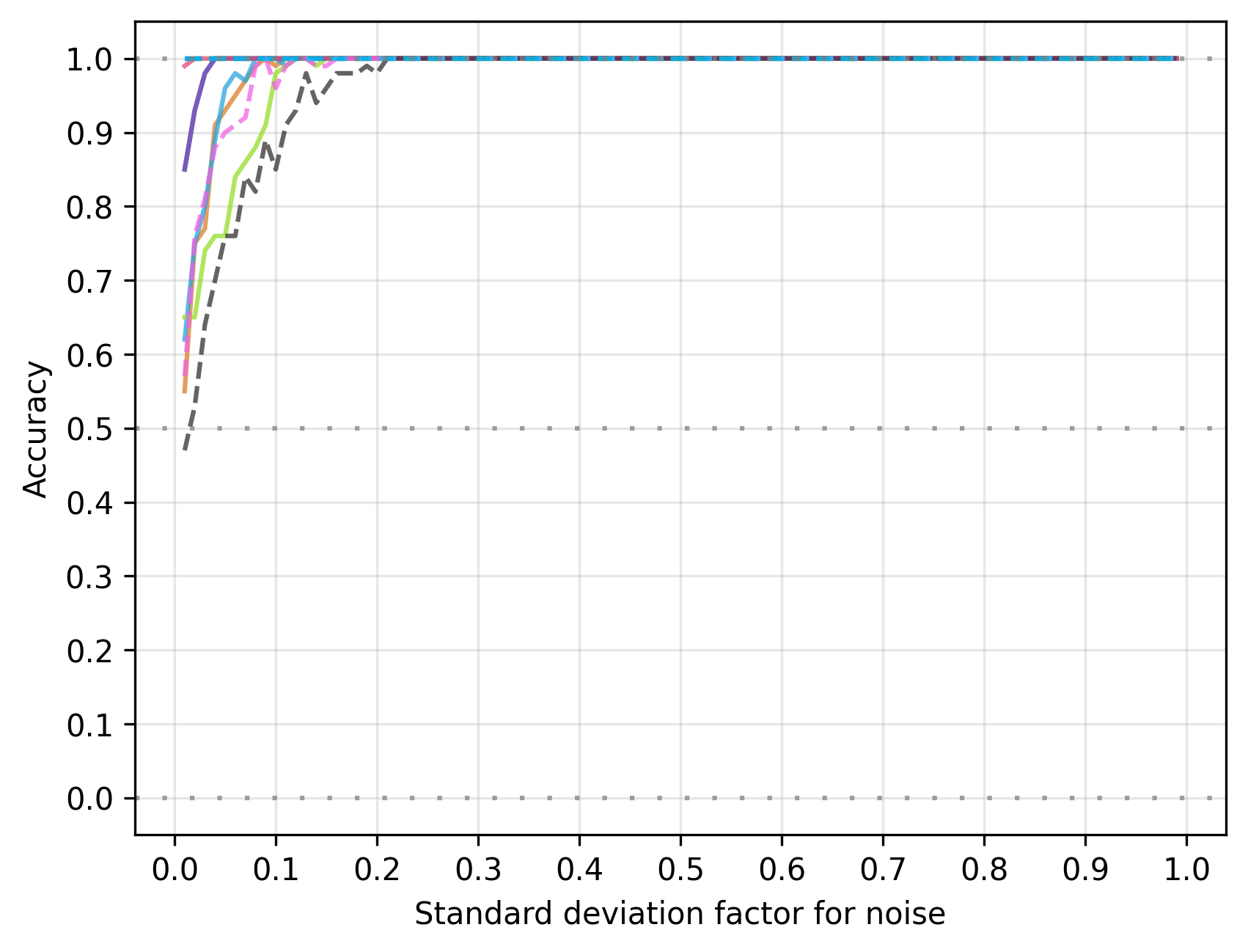}
\end{subfigure}%
\begin{subfigure}{.5\textwidth}
  \centering
  \includegraphics[scale=0.5]{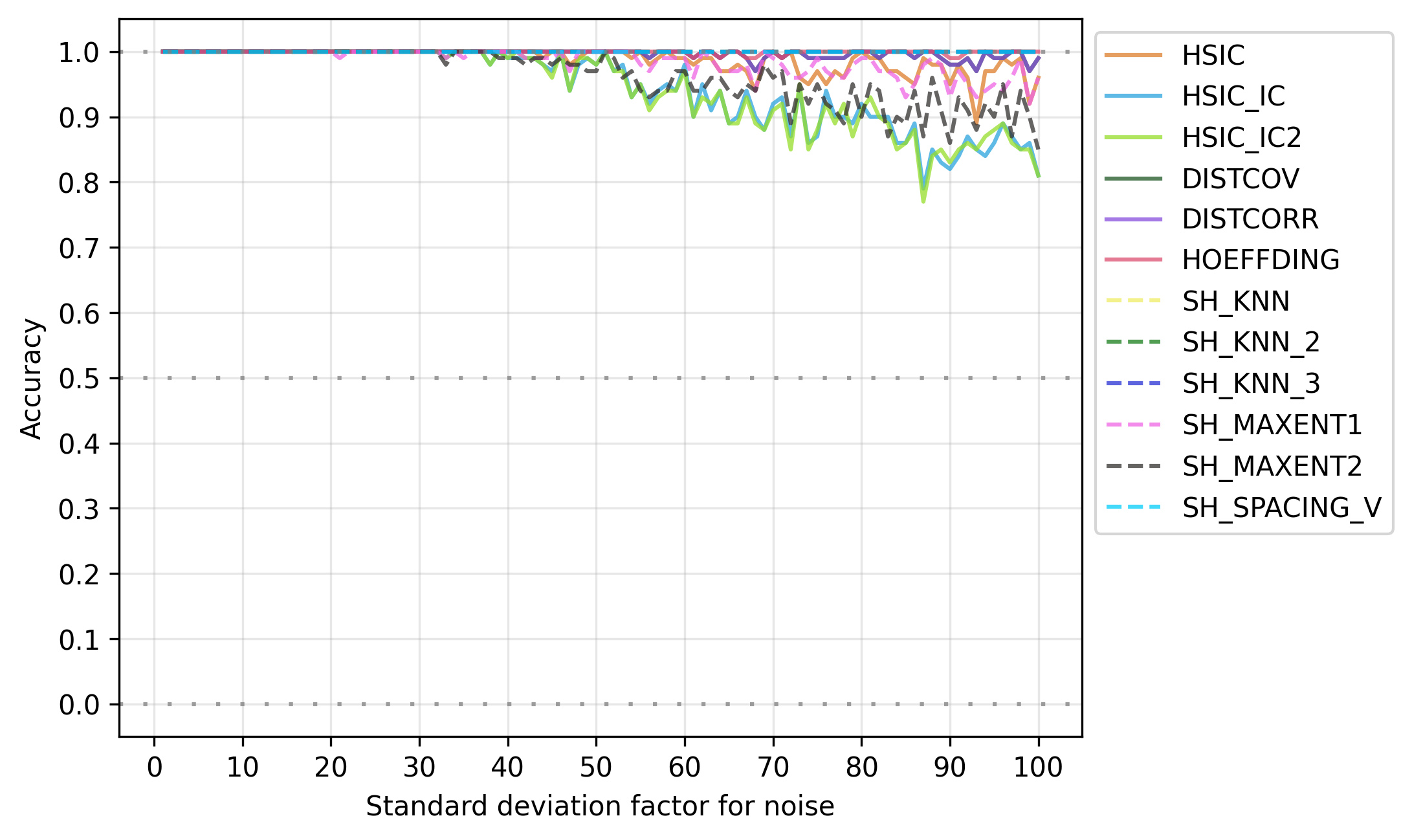}
\end{subfigure}
\caption{RESIT \& different noise levels \& coupled estimation \& $Y = \mathcal{N}^3+\mathcal{N}$}
\label{fig:22}
\end{figure}

\cref{fig:21} shows the linear model $Y= \mathcal{N} + \mathcal{L}$. All estimators
except for the three Shannon kNN estimators now reach an accuracy close to 100\%
for $i \in[0.42; 5]$. The three kNN estimators achieve better performance than 
in the decoupled estimation but remain worse than the other estimators.
\cref{fig:22} shows the non-linear model $Y= \mathcal{N}^3 + \mathcal{N}$. Here all estimators perform very good
with $i \in [0.15; 50]$ having an accuracy close to 100\%. With $i < 0.1$ most estimators drop fast below 90\% 
accuracy. For $i > 50$ most estimators remain above 90\% accuracy, except SH\_MAXENT2, HSIC\_IC and HSIC\_IC2 which
still remain above 80\% accuracy.

\newpage

\begin{figure}[h]
\centering
\begin{subfigure}{.5\textwidth}
  \centering
  \includegraphics[scale=0.5]{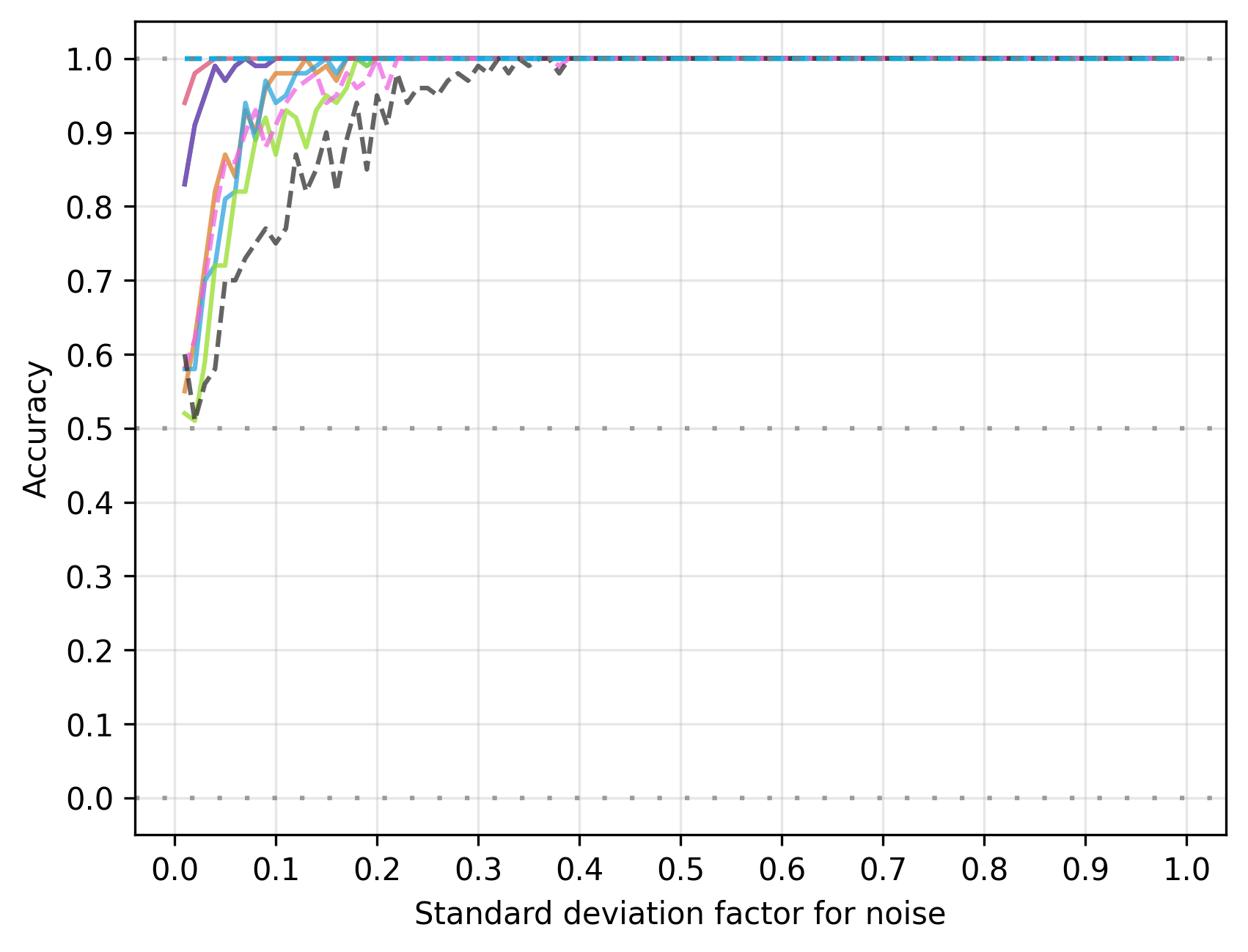}
\end{subfigure}%
\begin{subfigure}{.5\textwidth}
  \centering
  \includegraphics[scale=0.5]{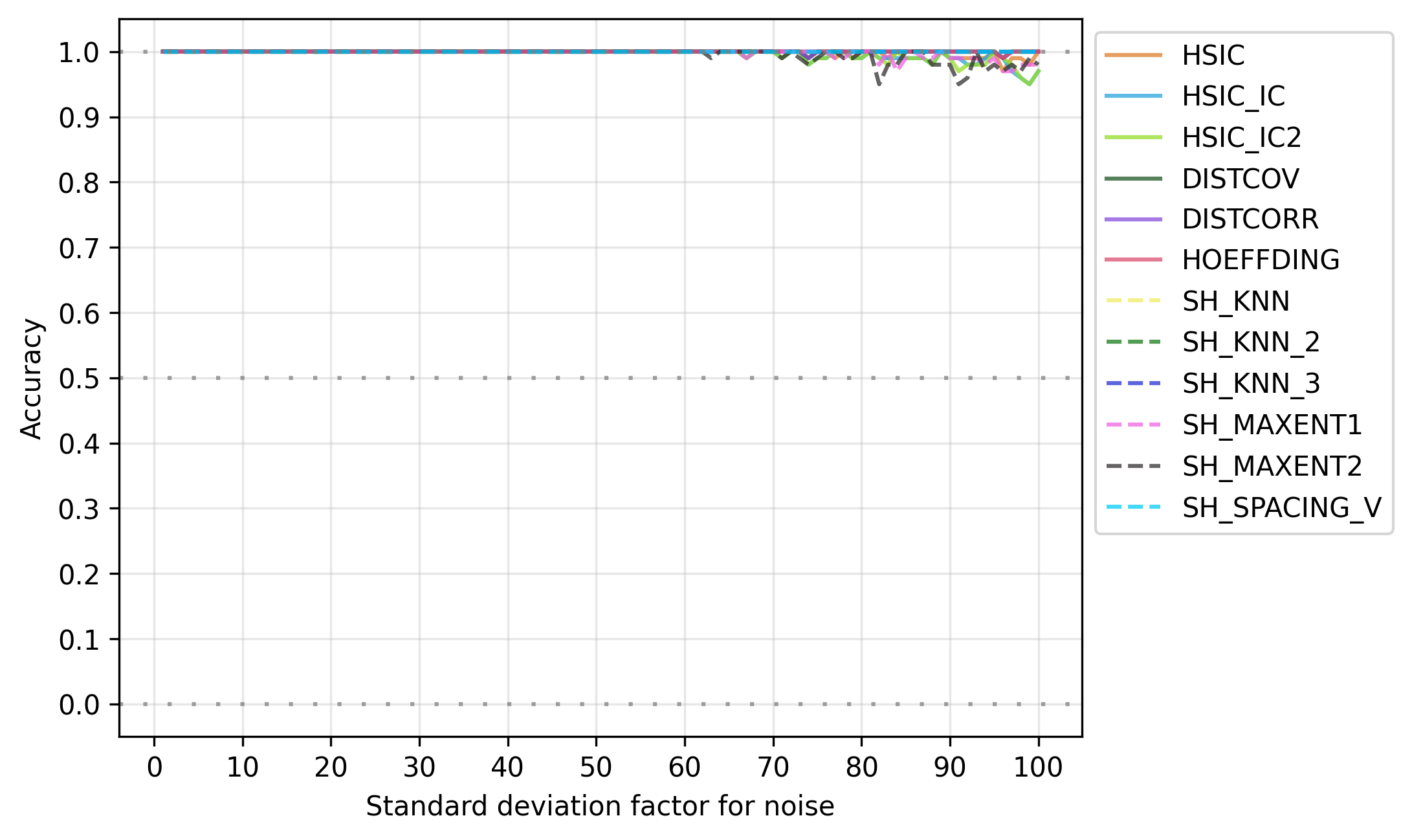}
\end{subfigure}
\caption{RESIT \& different noise levels \& coupled estimation \& $Y = \mathcal{N}^3+\mathcal{U}$}
\label{fig:23}

\centering
\begin{subfigure}{.5\textwidth}
  \centering
  \includegraphics[scale=0.5]{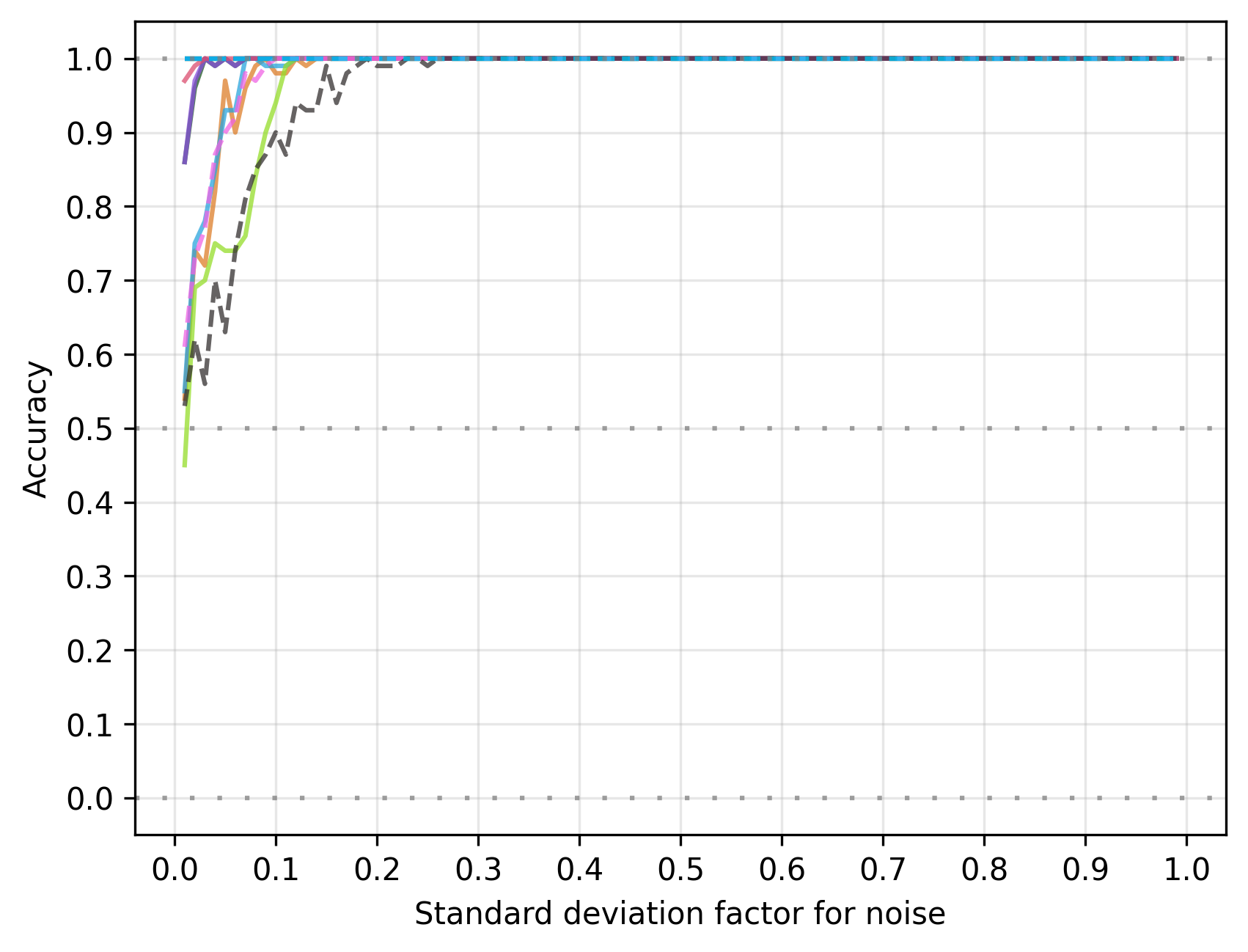}
\end{subfigure}%
\begin{subfigure}{.5\textwidth}
  \centering
  \includegraphics[scale=0.5]{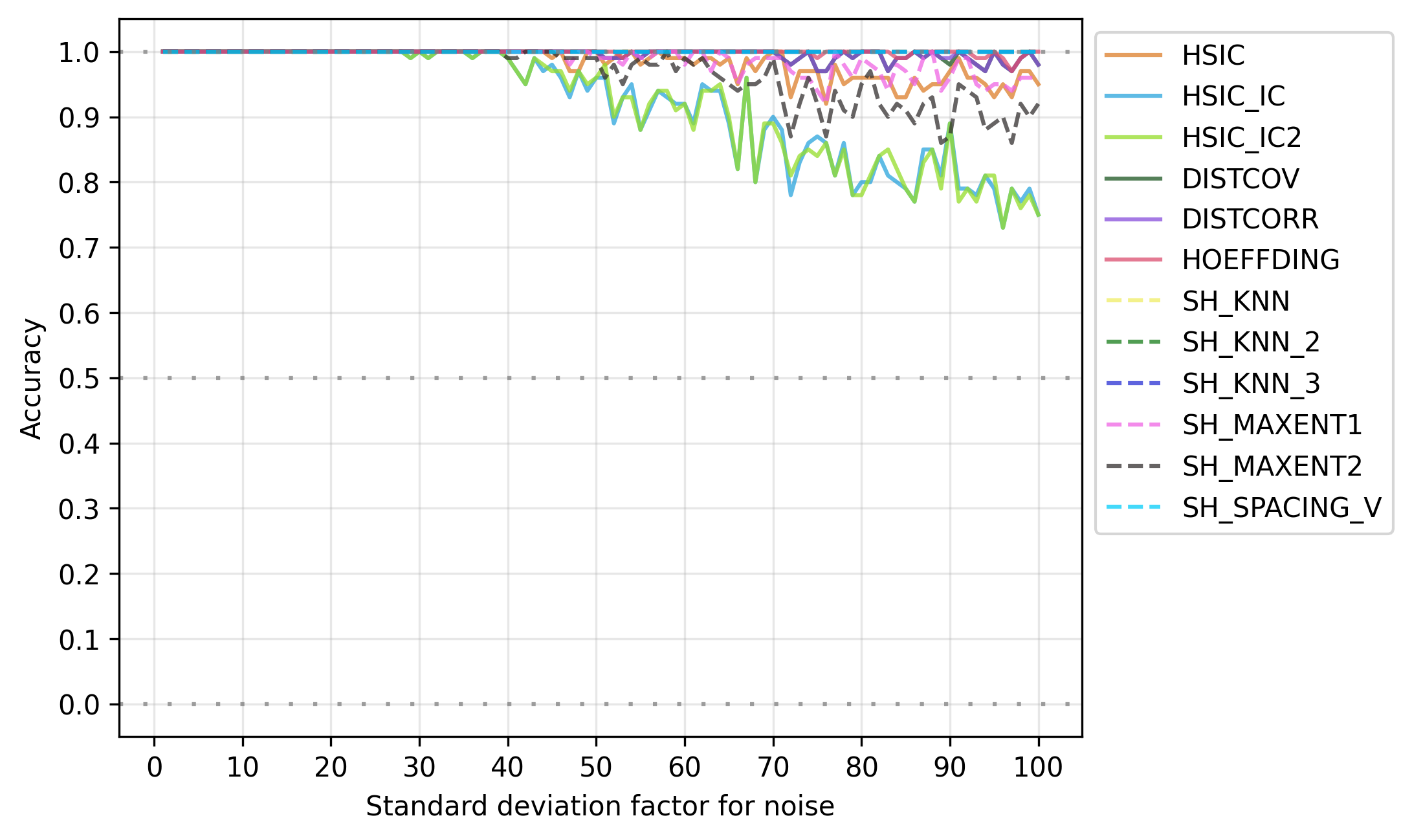}
\end{subfigure}
\caption{RESIT \& different noise levels \& coupled estimation \& $Y = \mathcal{N}^3+\mathcal{L}$}
\label{fig:24}
\end{figure}

\cref{fig:23} shows the non-linear model $Y= \mathcal{N}^3 + \mathcal{U}$. This result is similar as the previous one.
With $i \in [0.2; 100]$ we have accuracy close to 100\% for all estimators. SH\_MAXENT1 and SH\_MAXENT2 now perform
almost as good as other estimators for $i < 0.4$.
\cref{fig:24} shows the non-linear model $Y= \mathcal{N}^3 + \mathcal{L}$. For $i \in [0.17;40]$ all estimators
are close to or at 100\% accuracy. SH\_MAXENT1 and SH\_MAXENT2 perform as good as other estimators for $i < 0.2$.
and $i > 60$. The accuracy for HSIC\_IC and HSIC\_IC2 still drops fairly far after $i >50$ compared to the rest.

\newpage

\begin{figure}[h]
\centering
\begin{subfigure}{.5\textwidth}
  \centering
  \includegraphics[scale=0.5]{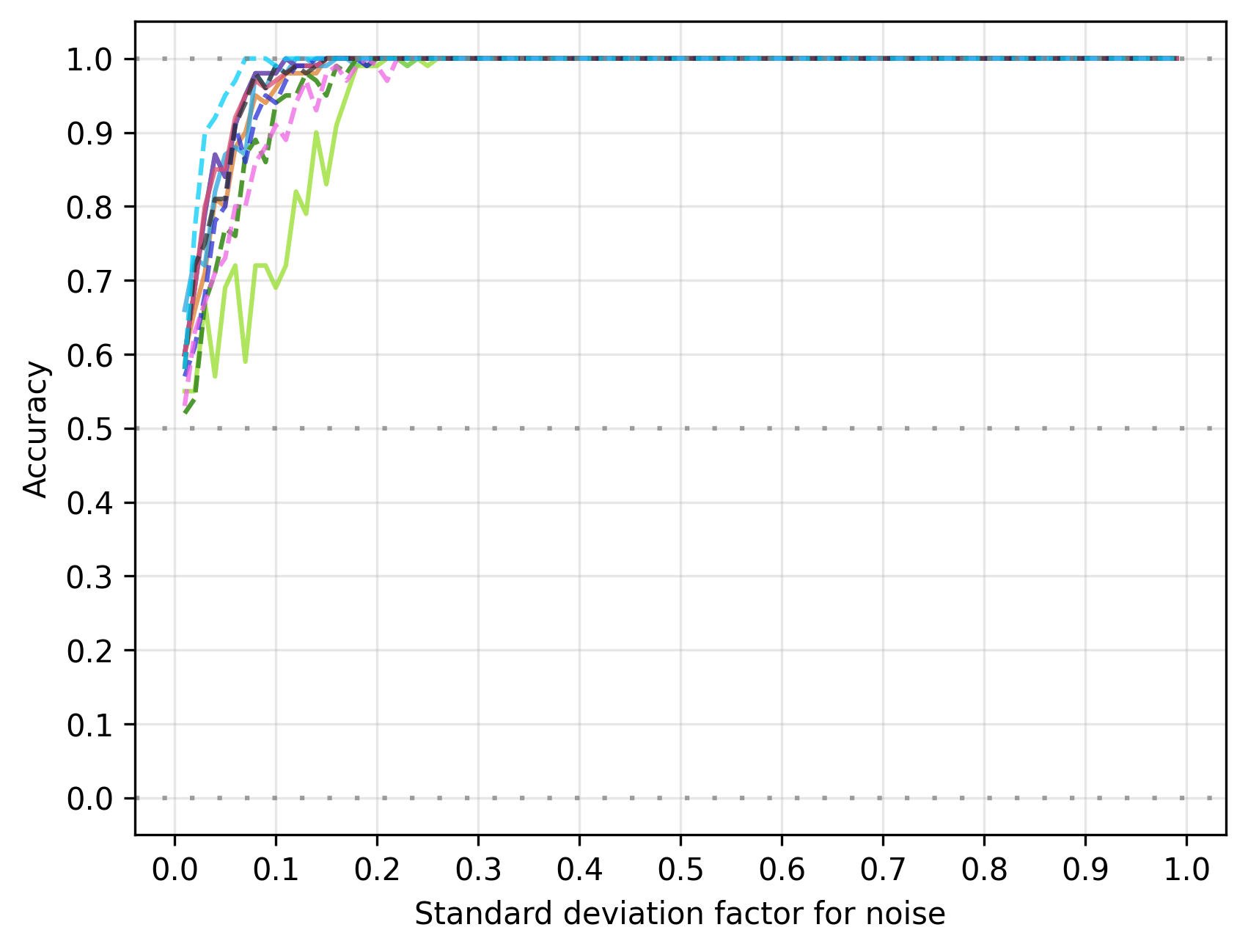}
\end{subfigure}%
\begin{subfigure}{.5\textwidth}
  \centering
  \includegraphics[scale=0.5]{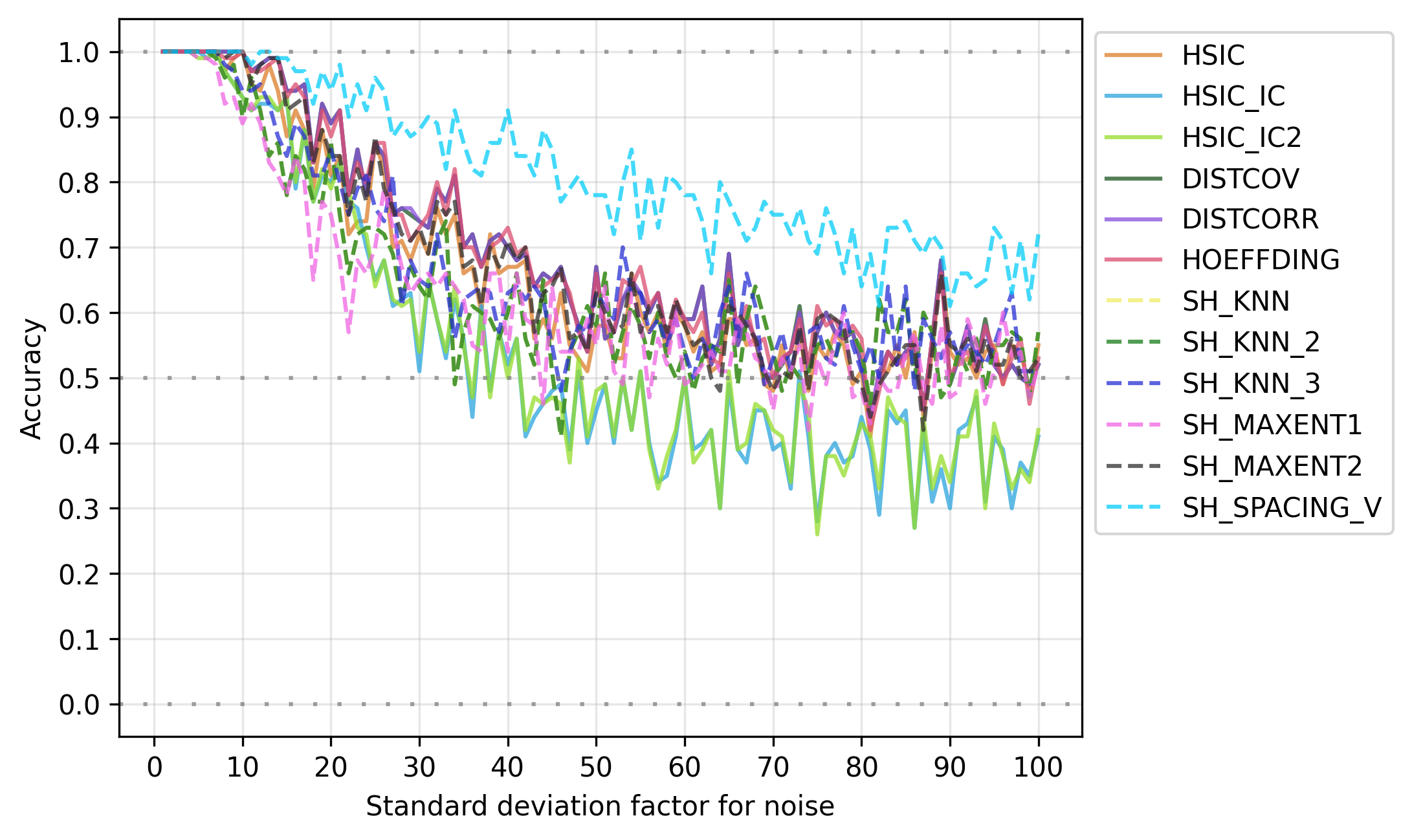}
\end{subfigure}
\caption{RESIT \& different noise levels \& coupled estimation \& $Y = \mathcal{U}+\mathcal{U}$}
\label{fig:25}

\centering
\begin{subfigure}{.5\textwidth}
  \centering
  \includegraphics[scale=0.5]{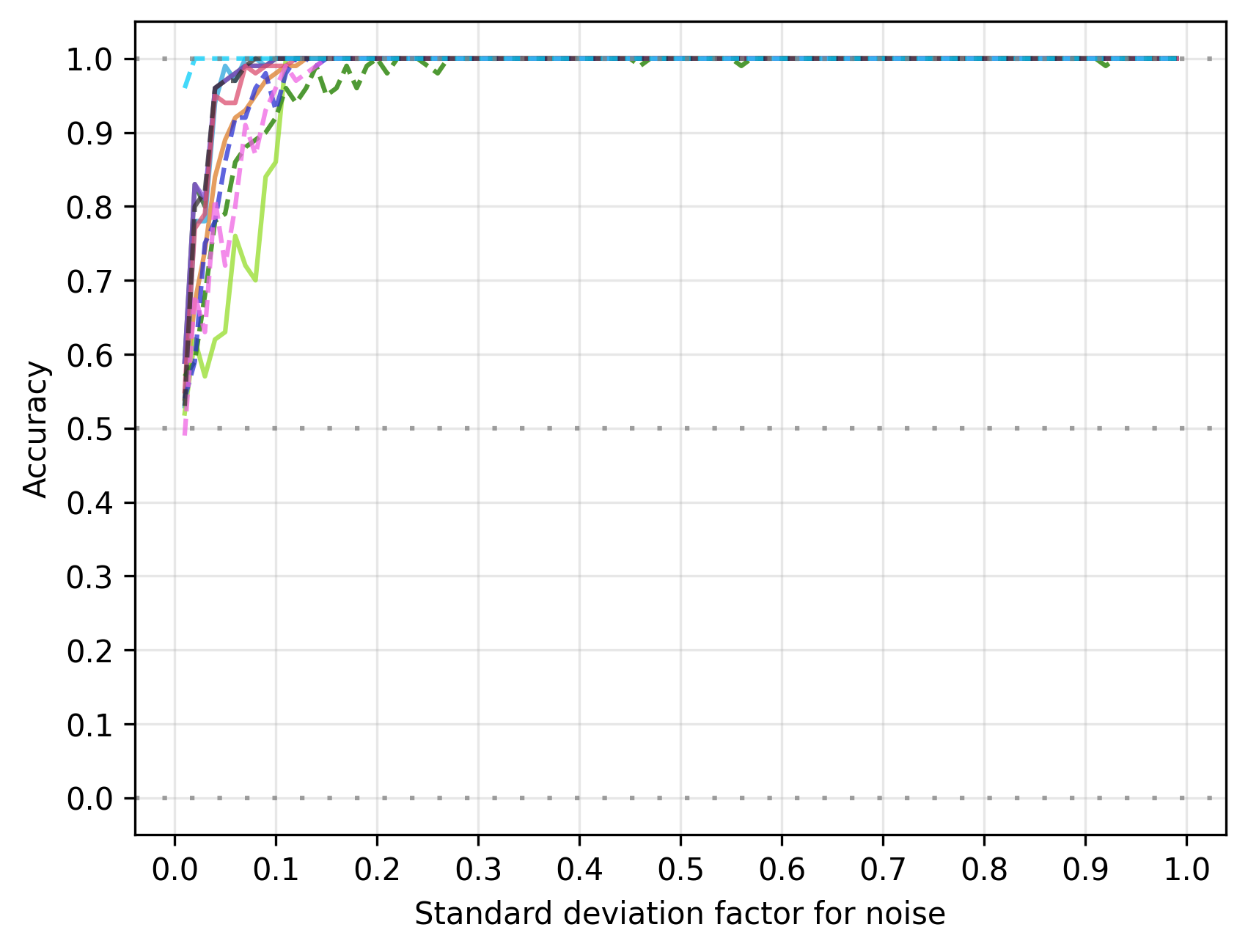}
\end{subfigure}%
\begin{subfigure}{.5\textwidth}
  \centering
  \includegraphics[scale=0.5]{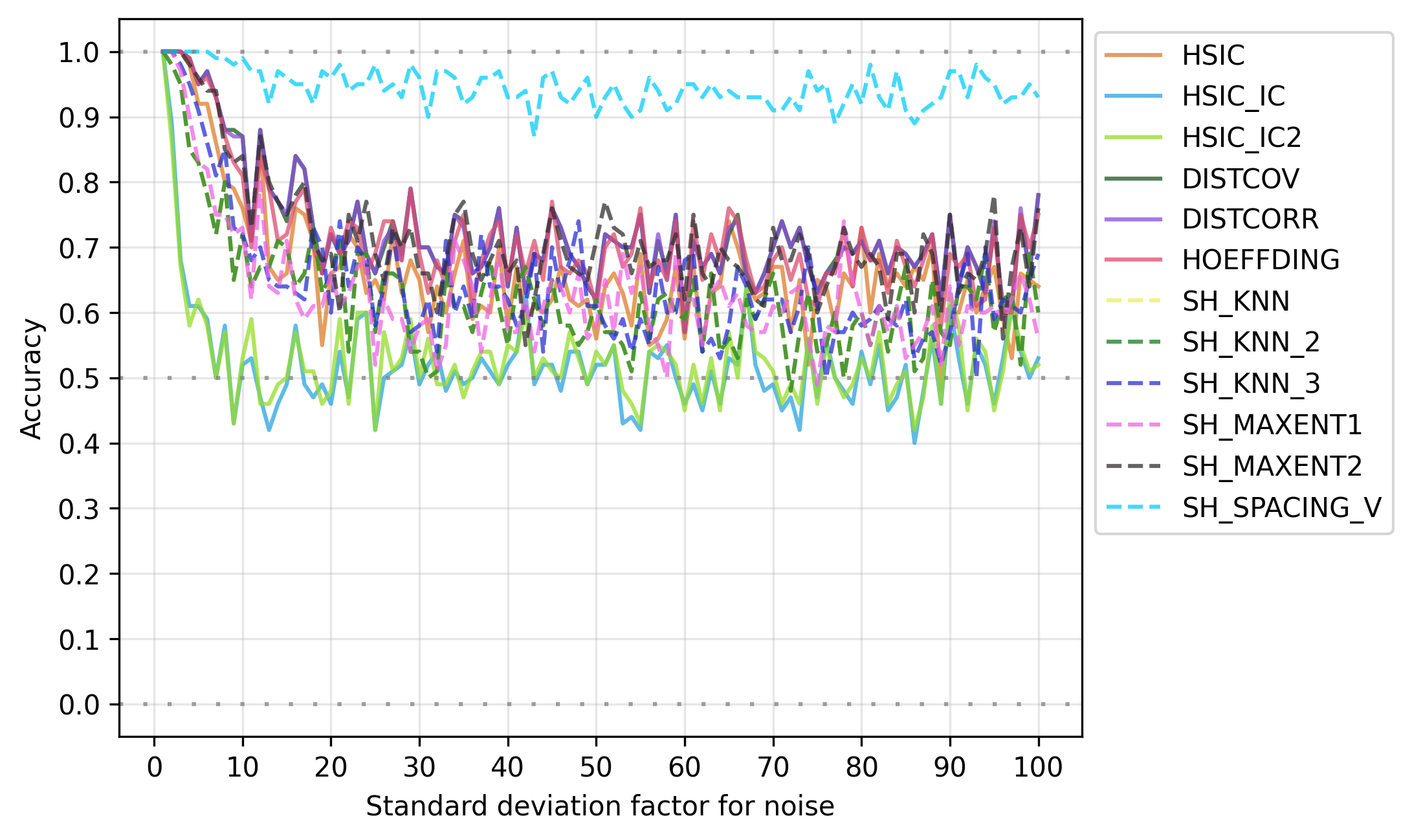}
\end{subfigure}
\caption{RESIT \& different noise levels \& coupled estimation \& $Y = \mathcal{U}+\mathcal{N}$}
\label{fig:26}
\end{figure}

\cref{fig:25} shows the linear model $Y= \mathcal{U} + \mathcal{U}$. This shows some interesting changes.
In the decoupled estimation HSIC\_IC and HSIC\_IC2 never reached a consistent range with accuracy above 90\% but
in the coupled estimation they do now. All estimators have an accuracy of 100\% for $i \in [0.25;5]$.
Interestingly, SH\_SPACING\_V does not converge to 50\%. HSIC\_IC and HSIC\_IC2 do clearly fall below
40\% for $i > \sim 85$. (Note that 0\% accuracy means that the wrong direction ($Y \to X$) is consistently identified.)
\cref{fig:26} shows the linear model $Y= \mathcal{U} + \mathcal{N}$. For $i \in [0.2;1]$ all estimators
are close to or at 100\% accuracy. In the decoupled estimation only two estimators were close to 100\% accuracy in
that interval. After $i = 1$ most estimators converge towards $\sim$ 60\% accuracy. HSIC\_IC and HSIC\_IC2 completely
drop to unidentifiability. SH\_SPACING\_V remains above 90\% accuracy for $i \in [1;100]$.

\newpage

\begin{figure}[h]
\centering
\begin{subfigure}{.5\textwidth}
  \centering
  \includegraphics[scale=0.5]{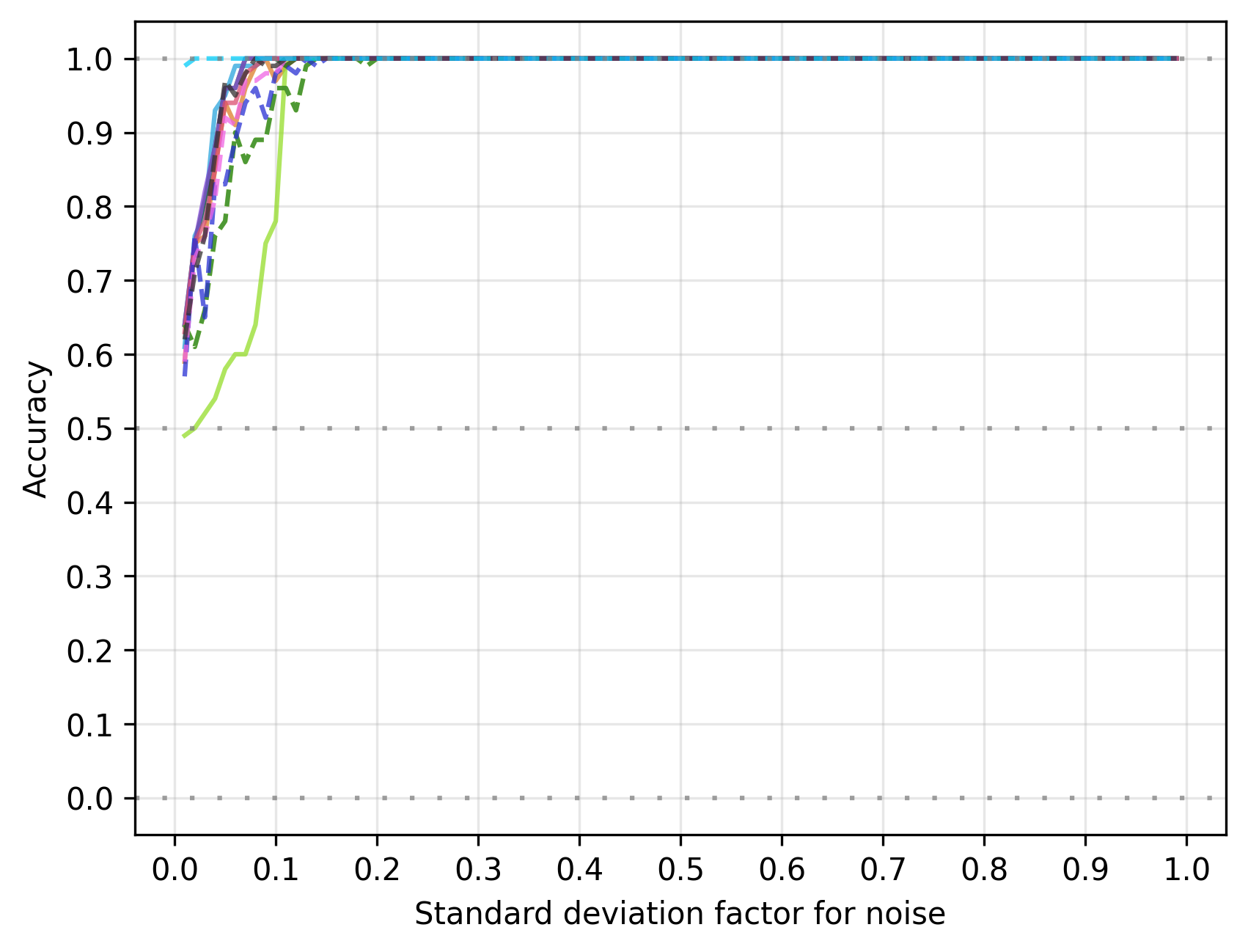}
\end{subfigure}%
\begin{subfigure}{.5\textwidth}
  \centering
  \includegraphics[scale=0.5]{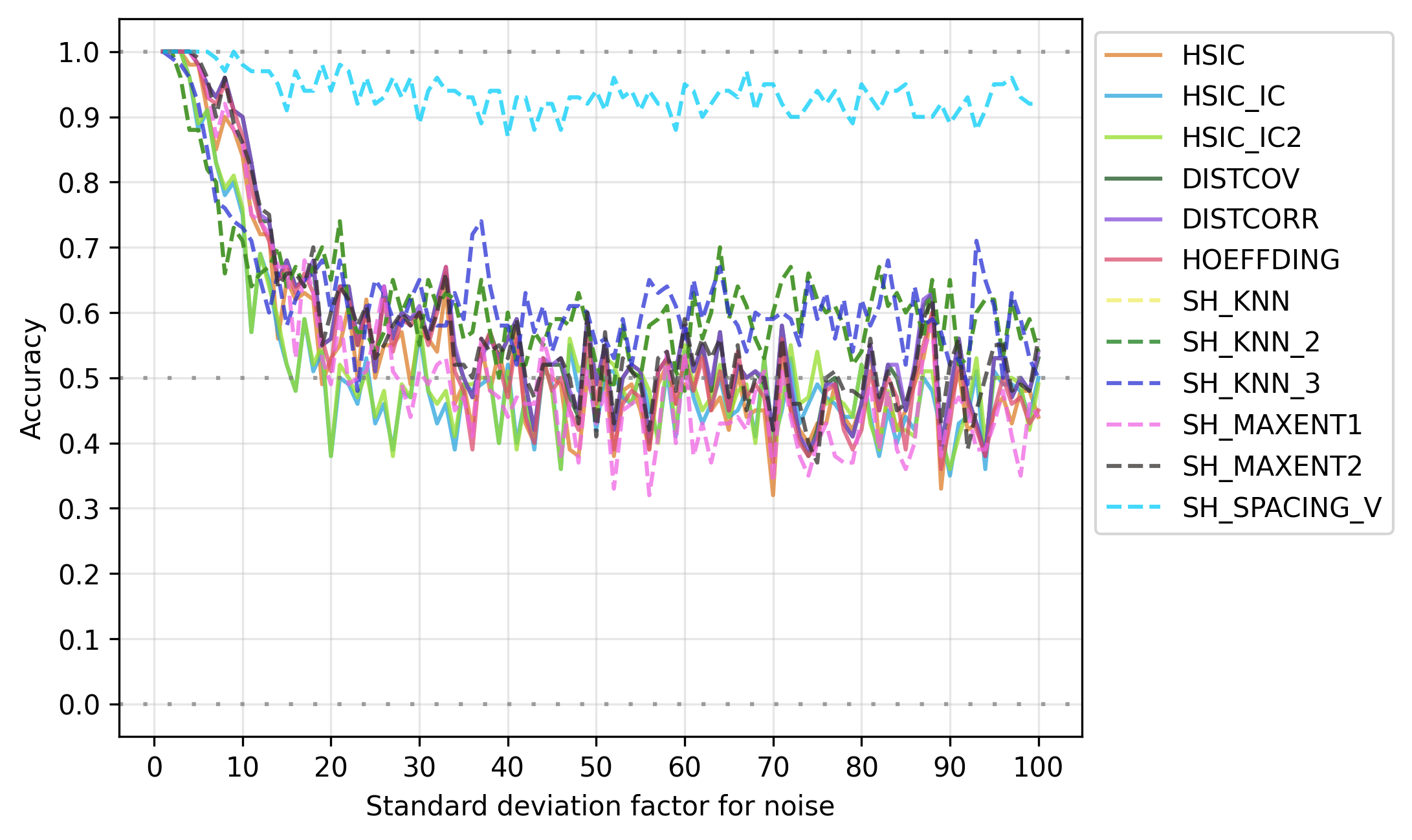}
\end{subfigure}
\caption{RESIT \& different noise levels \& coupled estimation \& $Y = \mathcal{U}+\mathcal{L}$}
\label{fig:27}

\centering
\begin{subfigure}{.5\textwidth}
  \centering
  \includegraphics[scale=0.5]{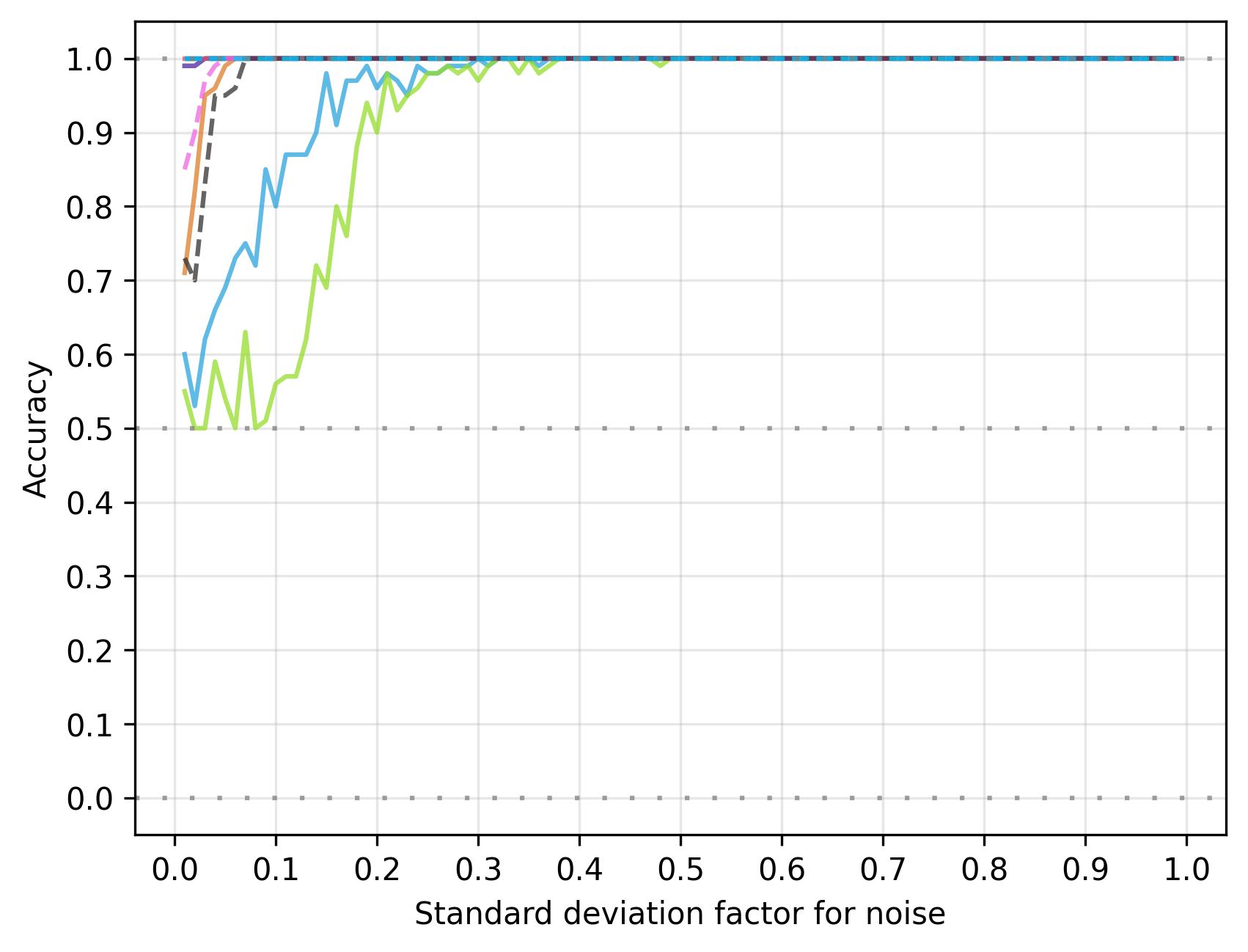}
\end{subfigure}%
\begin{subfigure}{.5\textwidth}
  \centering
  \includegraphics[scale=0.5]{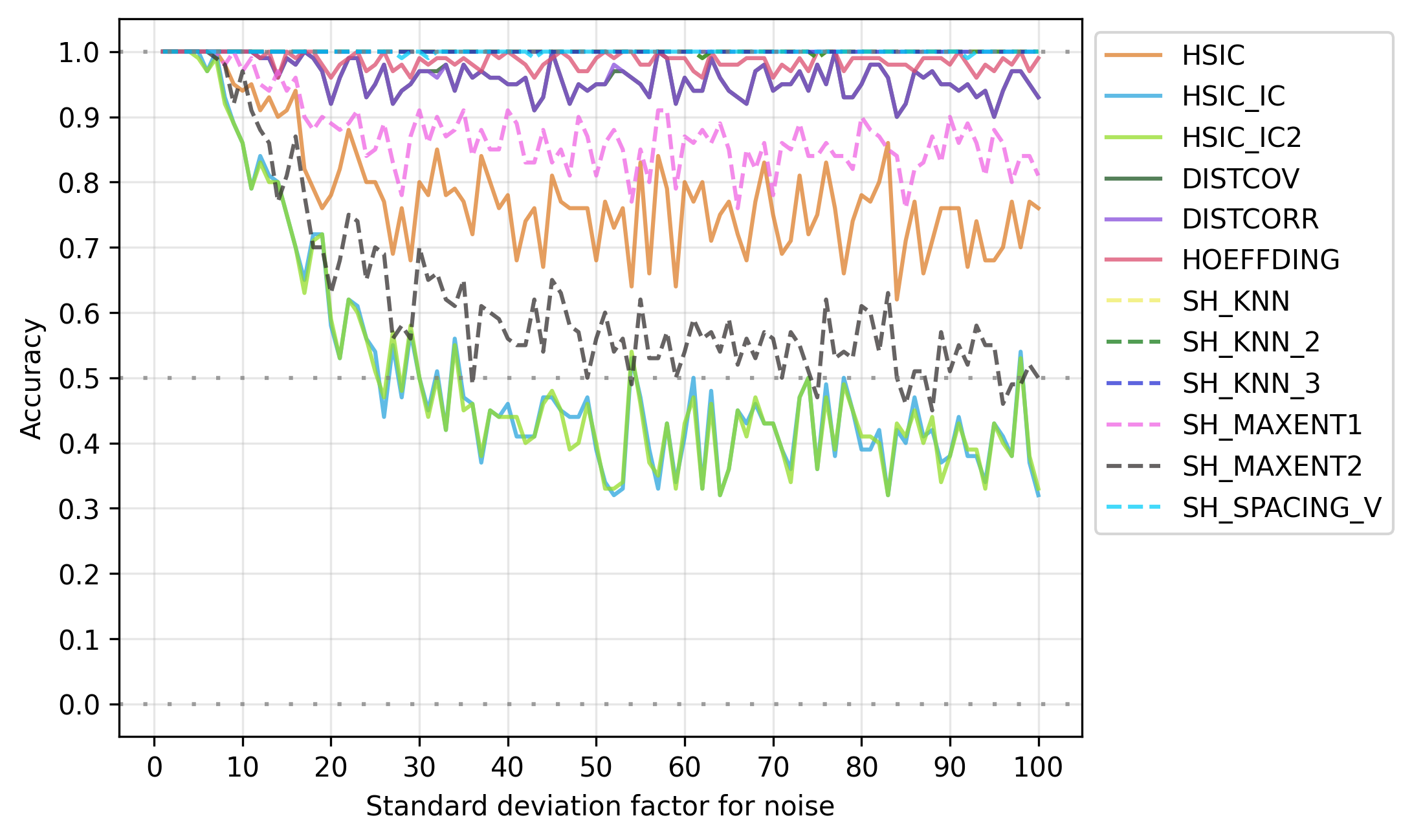}
\end{subfigure}
\caption{RESIT \& different noise levels \& coupled estimation \& $Y = \mathcal{U}^3+\mathcal{U}$}
\label{fig:28}
\end{figure}

\cref{fig:27} shows the linear model $Y= \mathcal{U} + \mathcal{L}$. This case has similar results as the previous one.
For $i \in [0.1;3]$ all estimators are close to or at 100\% accuracy. SH\_SPANNING\_V remains again over 90\% accuracy
for $i \in [1;100]$. All other converge towards unidentifiability while the three Shannon kNN estimators converge
towards 60\% accuracy.
\cref{fig:28} shows the non-linear model $Y= \mathcal{U}^3 + \mathcal{U}$. HSIC\_IC and HSIC\_IC2
now perform better and are close to 100\% accuracy for $i \in [0.3;5]$ while others have highest
accuracy for $i \in [0.05;5]$. For $i$ larger than 5, results look similar to the decoupled
estimation but overall the performance increased by an additive $5-15$\% for the various estimators.
The exception again is HSIC\_IC and HSIC\_IC2 which lie clearly below 50\% accuracy now.

\newpage

\begin{figure}[h]
\centering
\begin{subfigure}{.5\textwidth}
  \centering
  \includegraphics[scale=0.5]{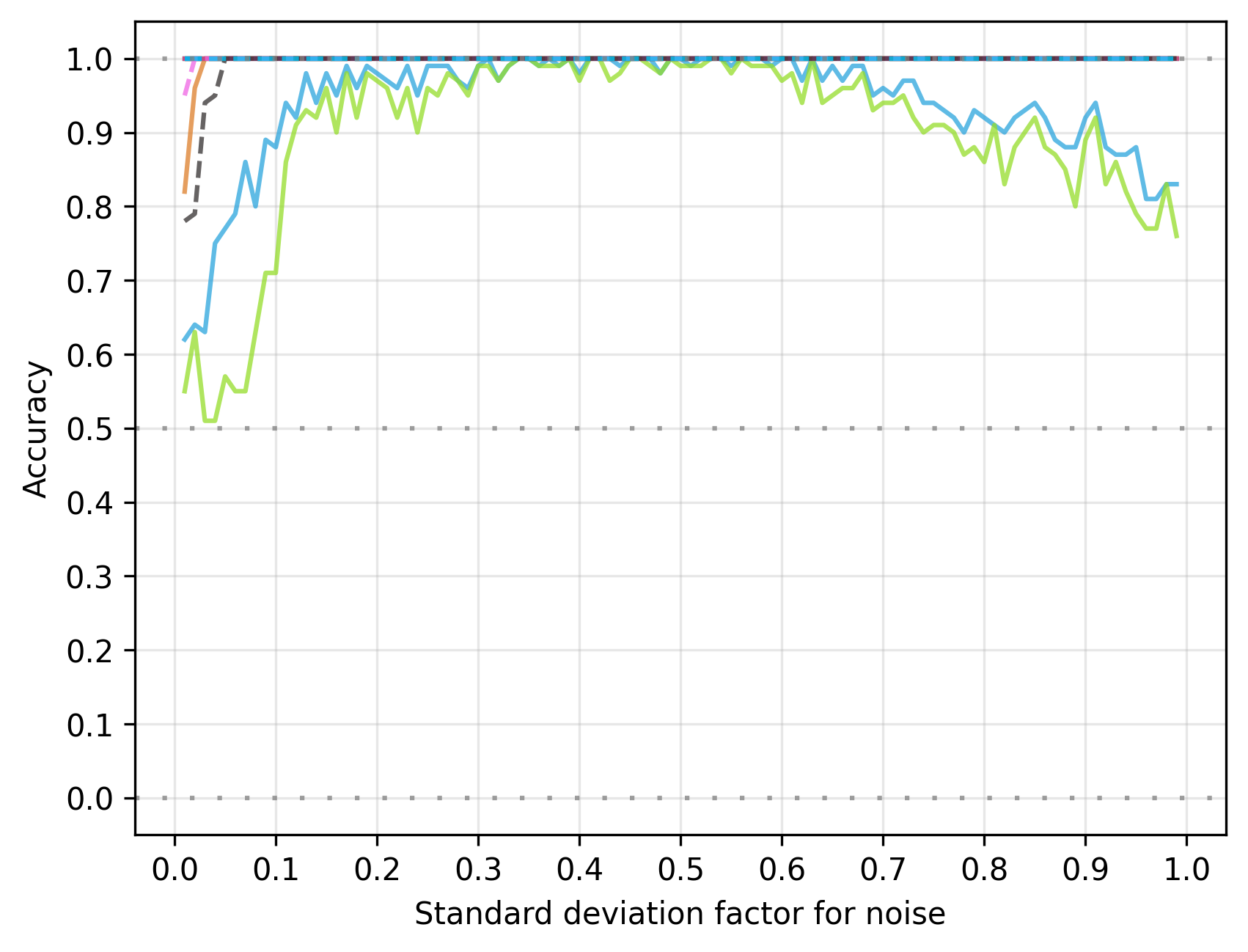}
\end{subfigure}%
\begin{subfigure}{.5\textwidth}
  \centering
  \includegraphics[scale=0.5]{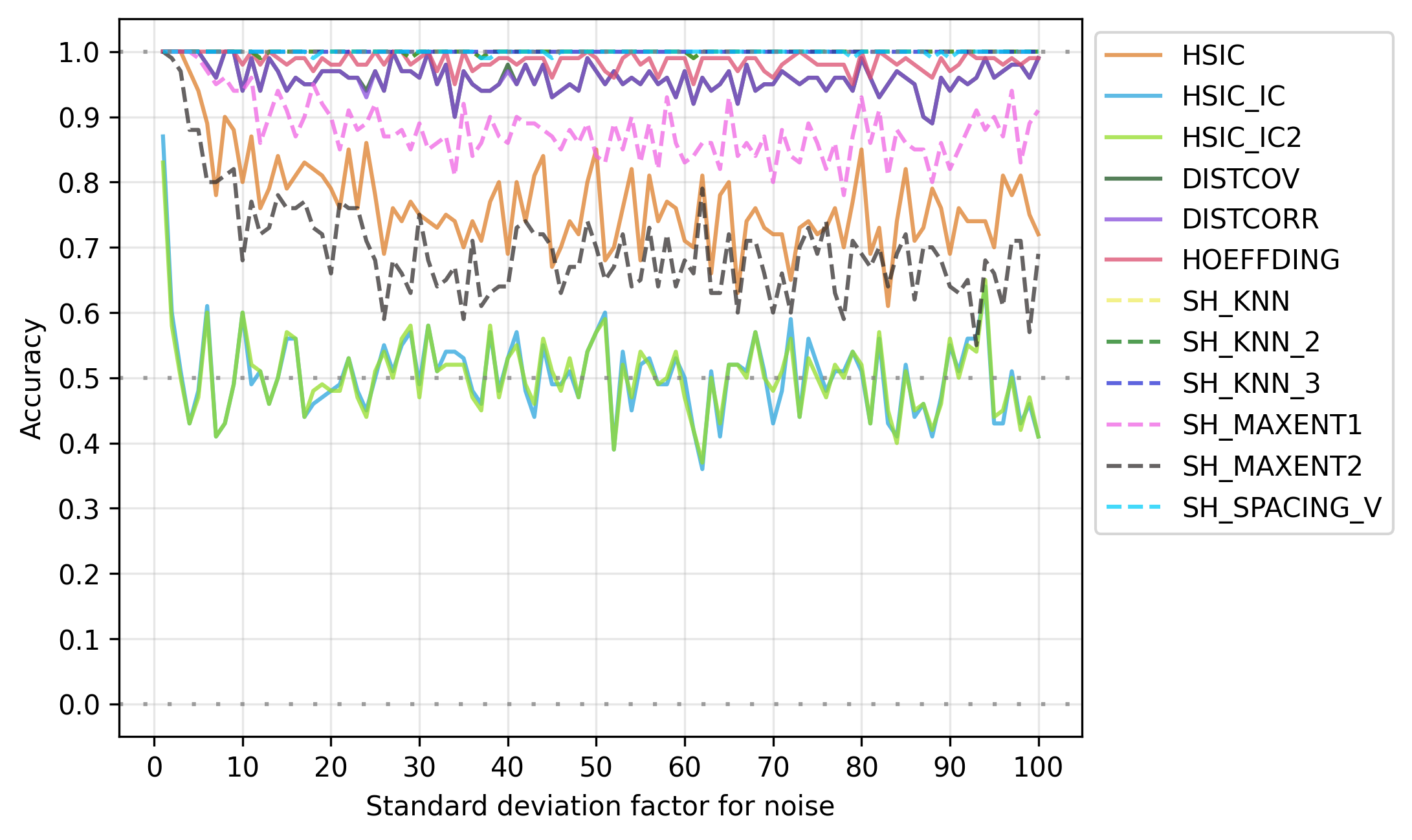}
\end{subfigure}
\caption{RESIT \& different noise levels \& coupled estimation \& $Y = \mathcal{U}^3+\mathcal{N}$}
\label{fig:29}

\centering
\begin{subfigure}{.5\textwidth}
  \centering
  \includegraphics[scale=0.5]{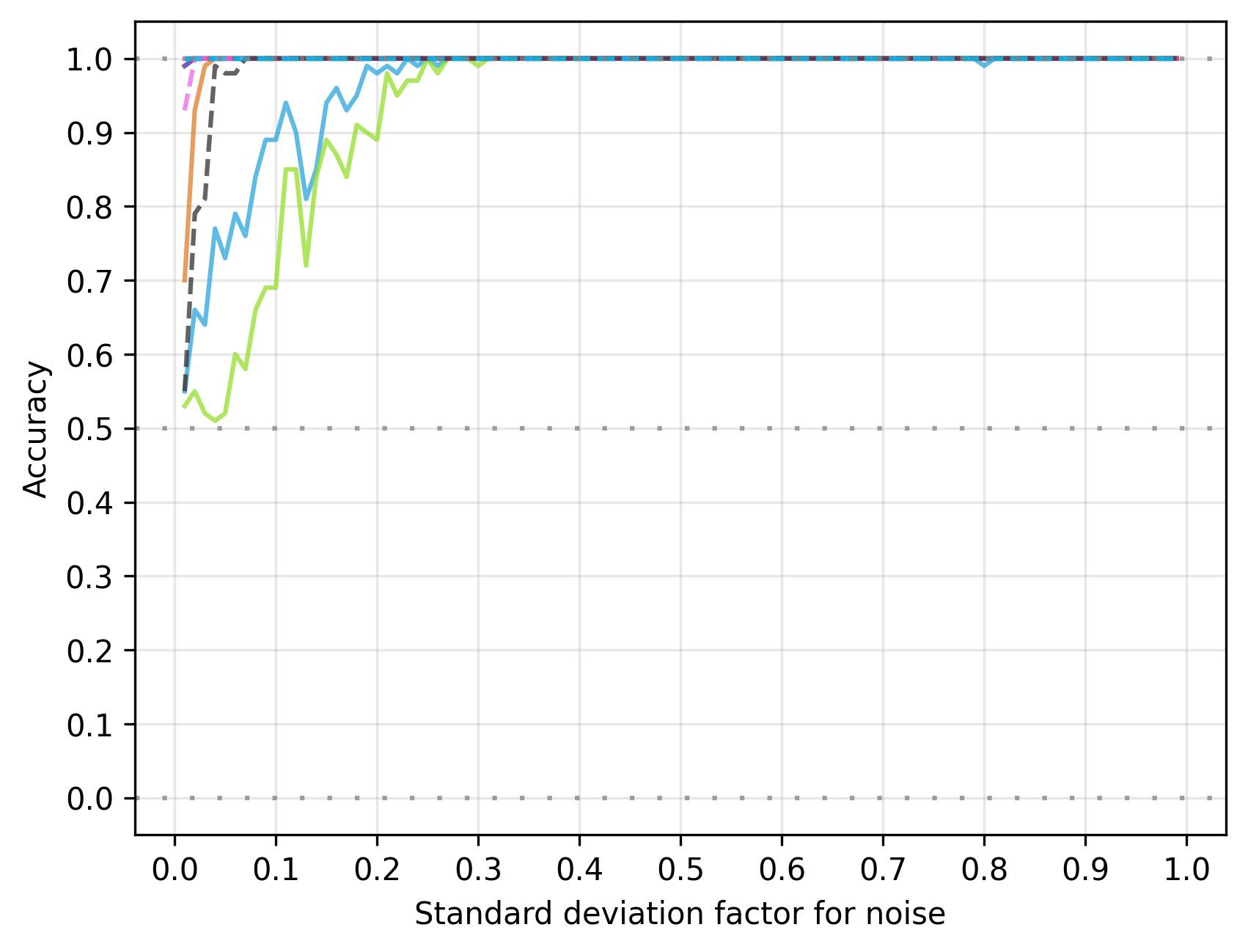}
\end{subfigure}%
\begin{subfigure}{.5\textwidth}
  \centering
  \includegraphics[scale=0.5]{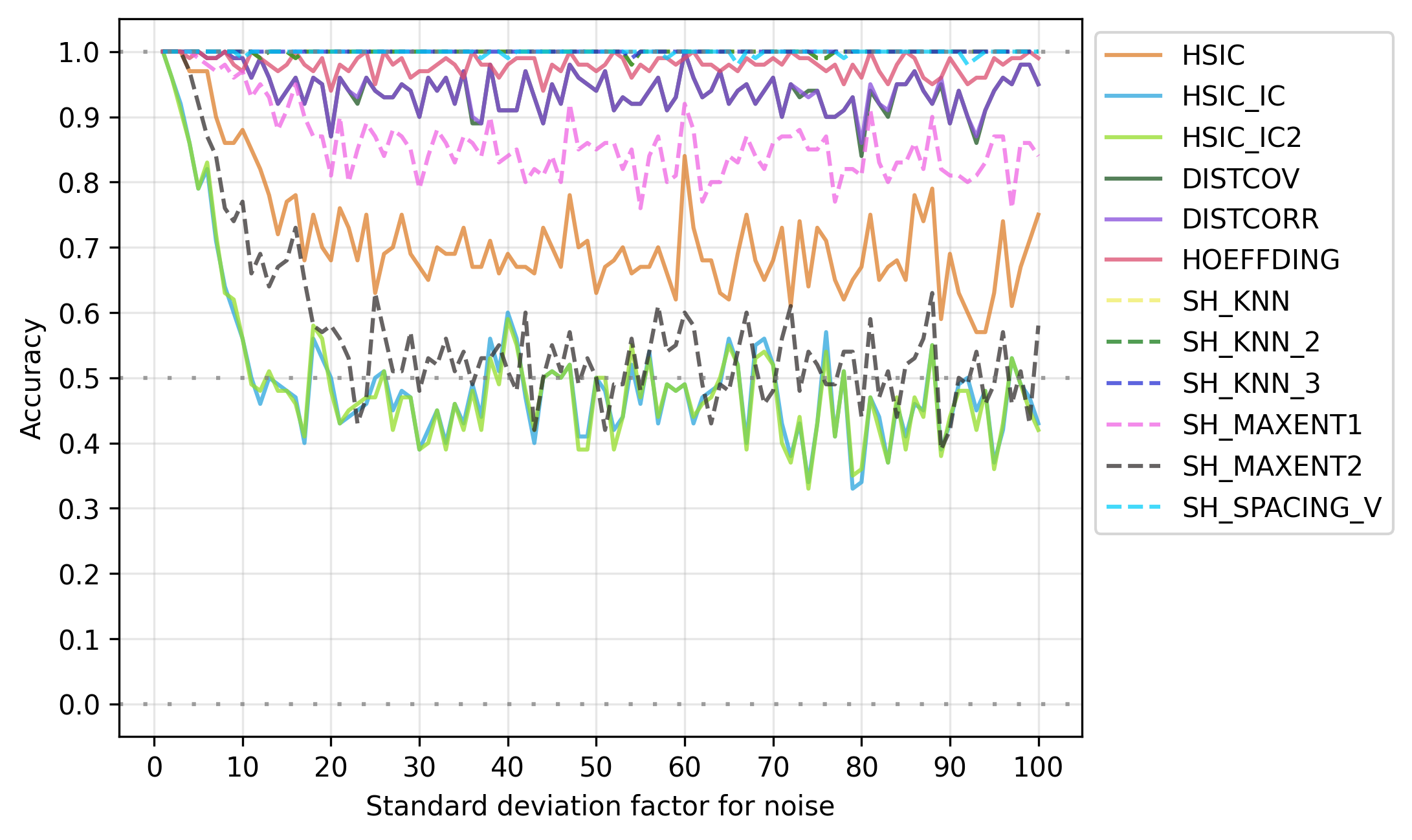}
\end{subfigure}
\caption{RESIT \& different noise levels \& coupled estimation \& $Y = \mathcal{U}^3+\mathcal{L}$}
\label{fig:30}
\end{figure}

\cref{fig:29} shows the non-linear model $Y= \mathcal{U}^3 + \mathcal{N}$. In the decoupled estimation
HSIC\_IC and HSIC\_IC2 never reached identifiability. This is now different for the coupled estimation:
for $i \in [0.1;0.75]$ they are both above 90\% accuracy but begin to drop already after $i=0.75$ towards
unidentifiability. SH\_SPACING\_V and the three Shannon kNN estimators have perfect accuracy for $i \in [0.01;100]$.
All other estimators have their accuracy increased additively by $5-20$\% for $i$ larger than 1.
\cref{fig:30} shows the non-linear model $Y= \mathcal{U}^3 + \mathcal{L}$. This is similar to the previous case.
SH\_SPACING\_V and the three Shannon kNN estimators are at 100\% accuracy for $i \in [0.01;100]$.
All others reach 100\% accuracy for $i \in [0.25;1]$ and for $i > 1$ their accuracy have increased
by an additive $5-20$\%.

\newpage

\begin{figure}[h]
\centering
\begin{subfigure}{.5\textwidth}
  \centering
  \includegraphics[scale=0.5]{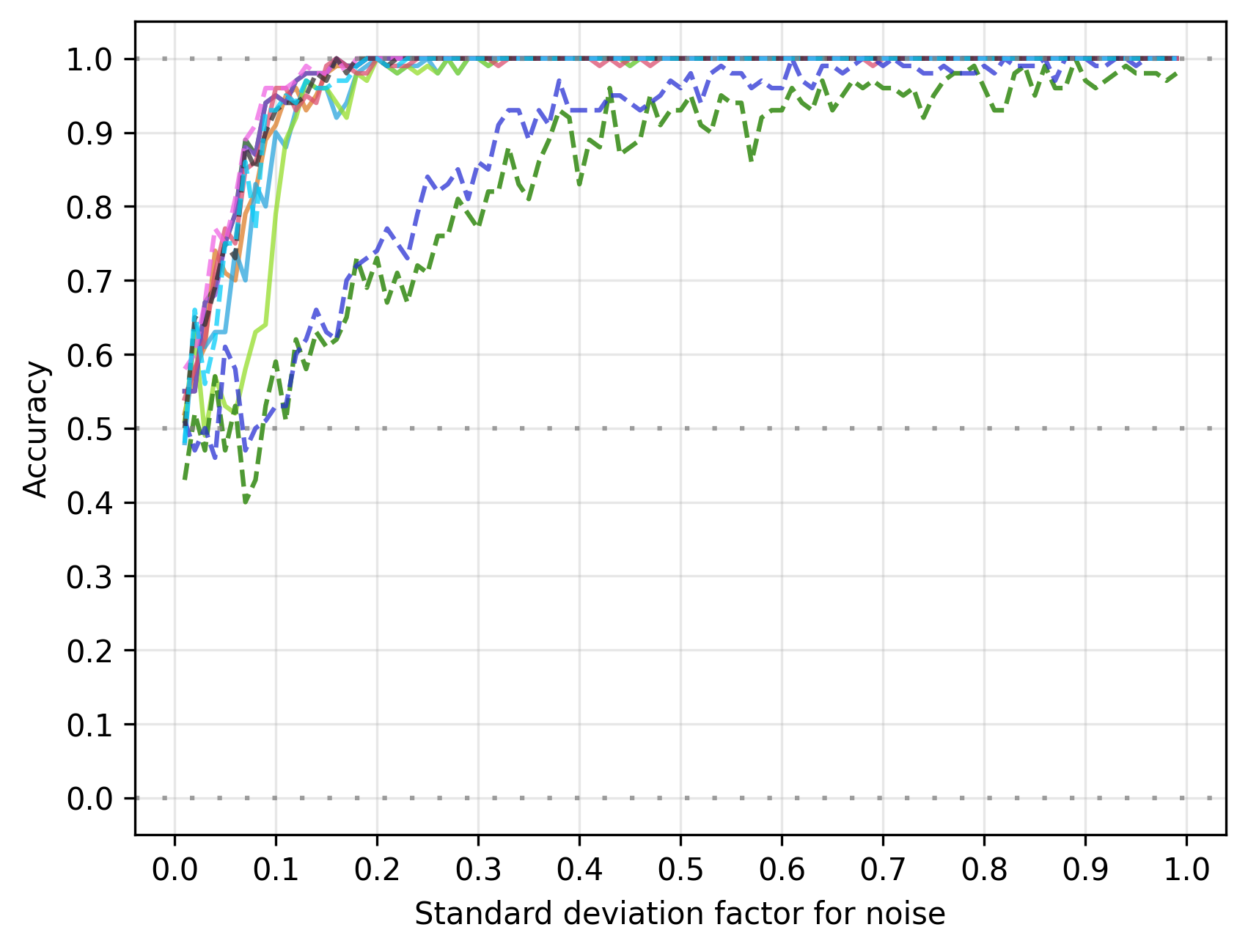}
\end{subfigure}%
\begin{subfigure}{.5\textwidth}
  \centering
  \includegraphics[scale=0.5]{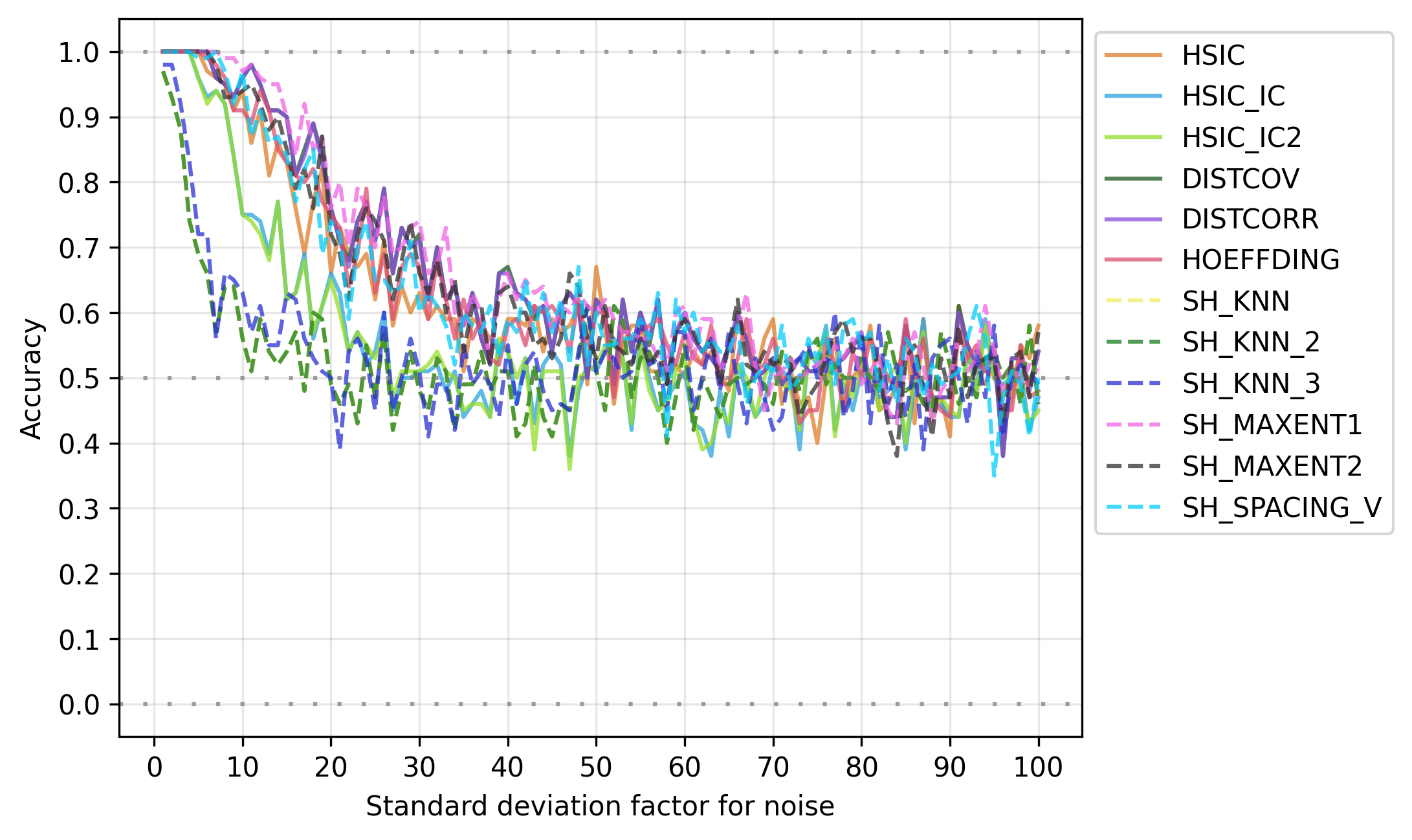}
\end{subfigure}
\caption{RESIT \& different noise levels \& coupled estimation \& $Y = \mathcal{L}+\mathcal{L}$}
\label{fig:31}

\centering
\begin{subfigure}{.5\textwidth}
  \centering
  \includegraphics[scale=0.5]{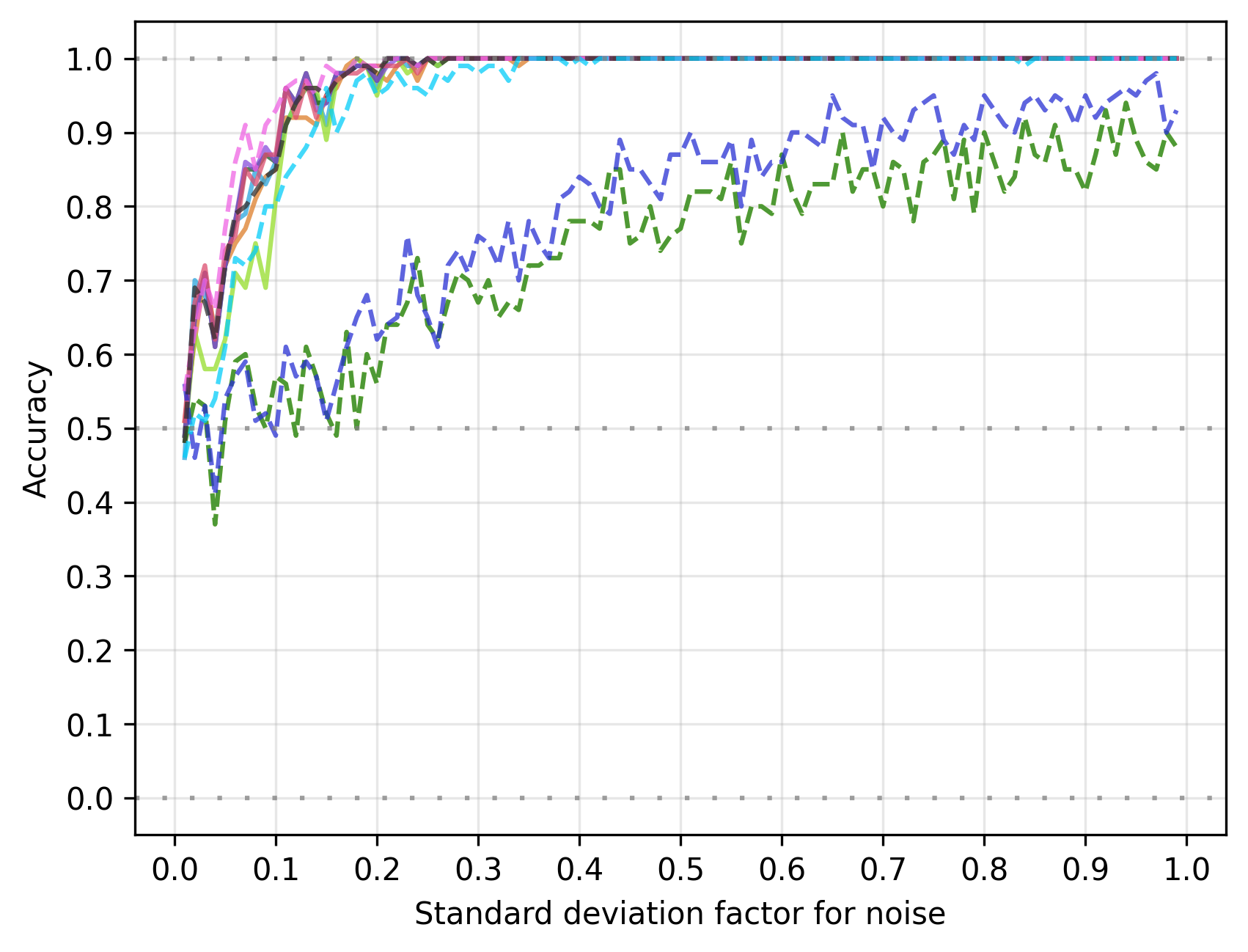}
\end{subfigure}%
\begin{subfigure}{.5\textwidth}
  \centering
  \includegraphics[scale=0.5]{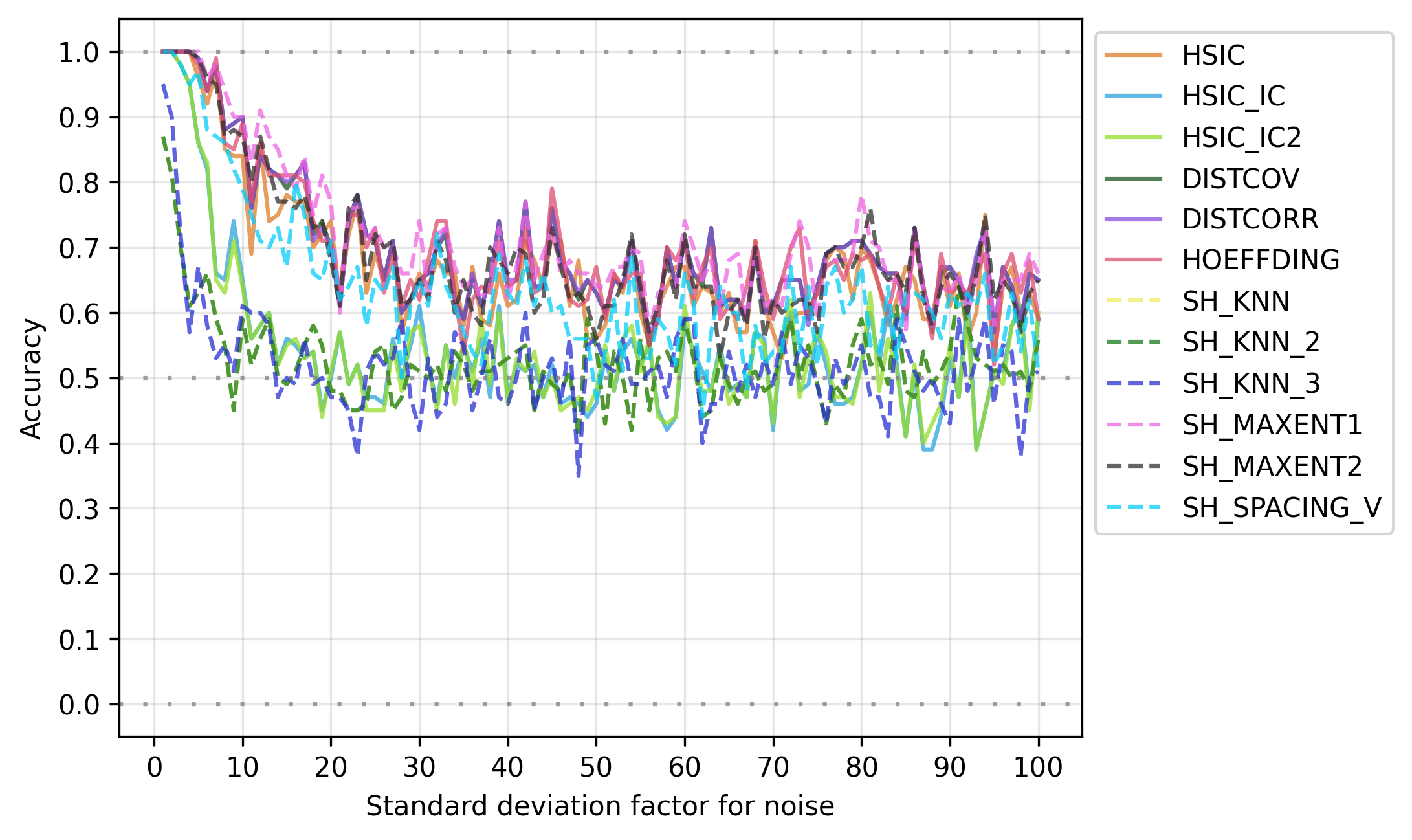}
\end{subfigure}
\caption{RESIT \& different noise levels \& coupled estimation \& $Y = \mathcal{L}+\mathcal{N}$}
\label{fig:32}
\end{figure}

\cref{fig:31} shows the linear model $Y= \mathcal{L} + \mathcal{L}$. For $i \in [0.01;1]$ estimators
have improved significantly. The three Shannon kNN estimators still perform worse than other estimators, but they
do reach now accuracy above 90\% for $i \in [0.45; 1]$. All other estimators are close to or at 100\% for $i \in [0.2; 5]$.
\cref{fig:32} shows the linear model $Y= \mathcal{L} + \mathcal{N}$. This has a similar pattern as the previous one.
The three Shannon kNN estimators still perform worse than other estimators and they barely reach 90\% accuracy for
$i = 1$. All other estimators are now close to or at 100\% for $i \in [0.25; 5]$.

\newpage

\begin{figure}[h]
\centering
\begin{subfigure}{.5\textwidth}
  \centering
  \includegraphics[scale=0.5]{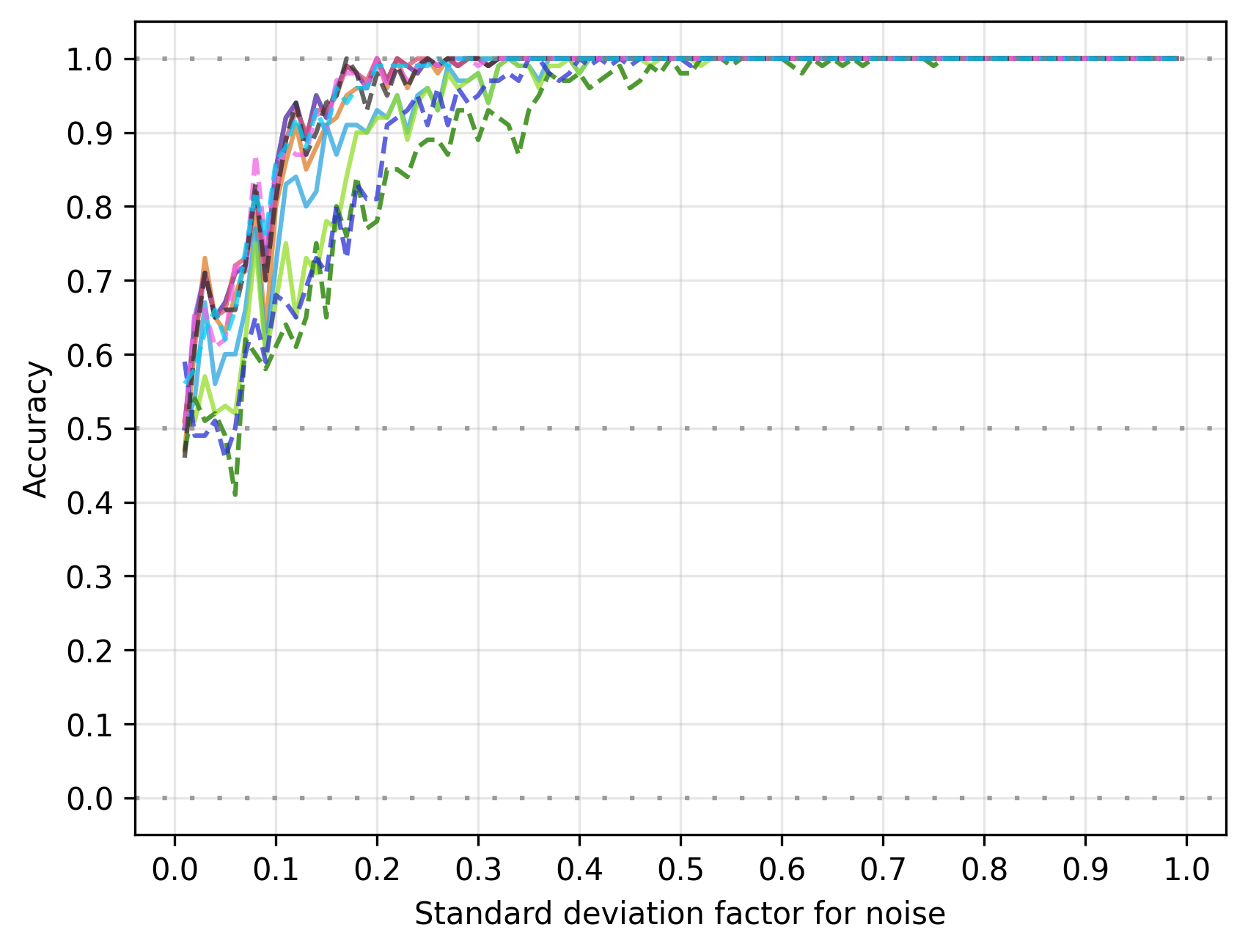}
\end{subfigure}%
\begin{subfigure}{.5\textwidth}
  \centering
  \includegraphics[scale=0.5]{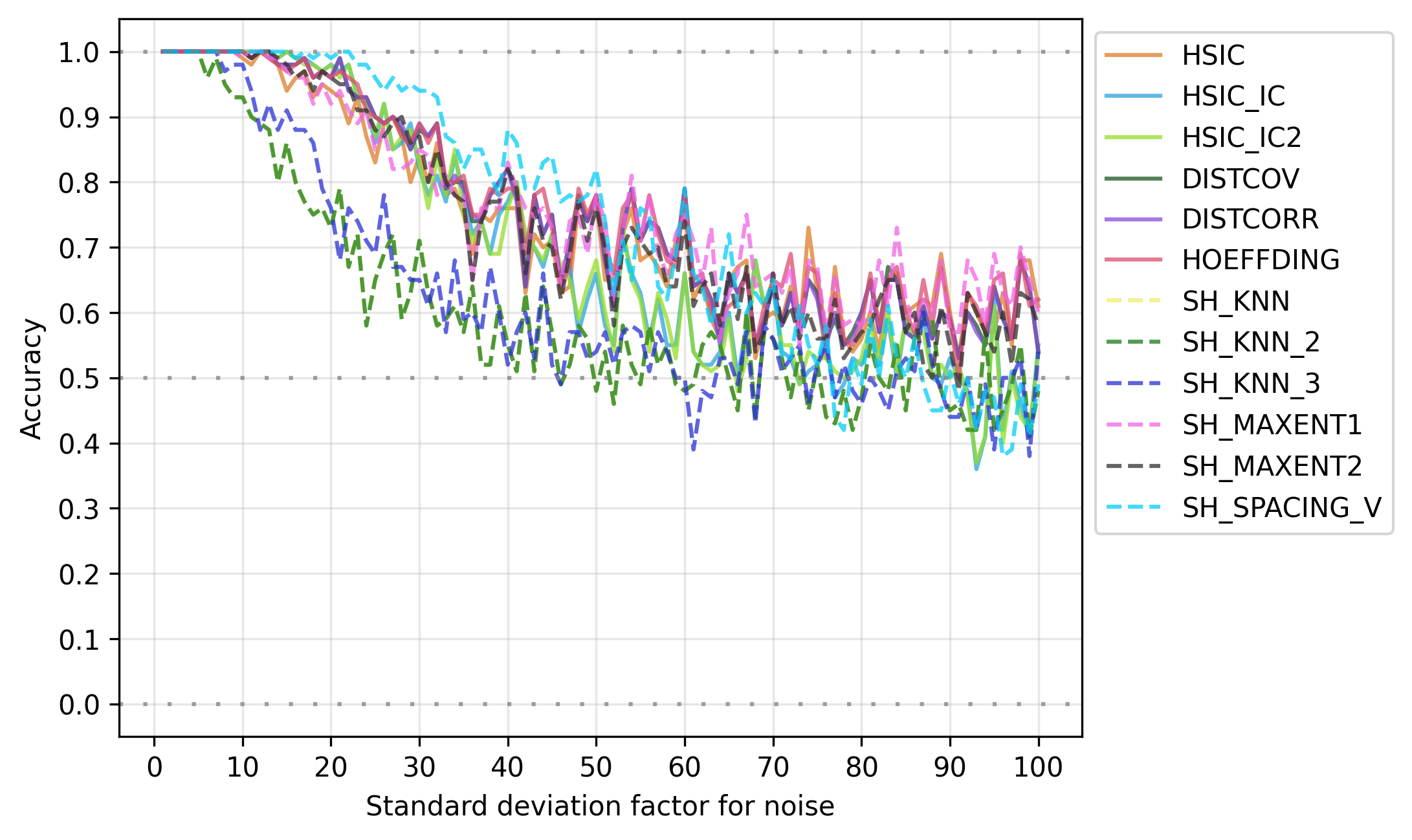}
\end{subfigure}
\caption{RESIT \& different noise levels \& coupled estimation \& $Y = \mathcal{L}+\mathcal{U}$}
\label{fig:33}

\centering
\begin{subfigure}{.5\textwidth}
  \centering
  \includegraphics[scale=0.5]{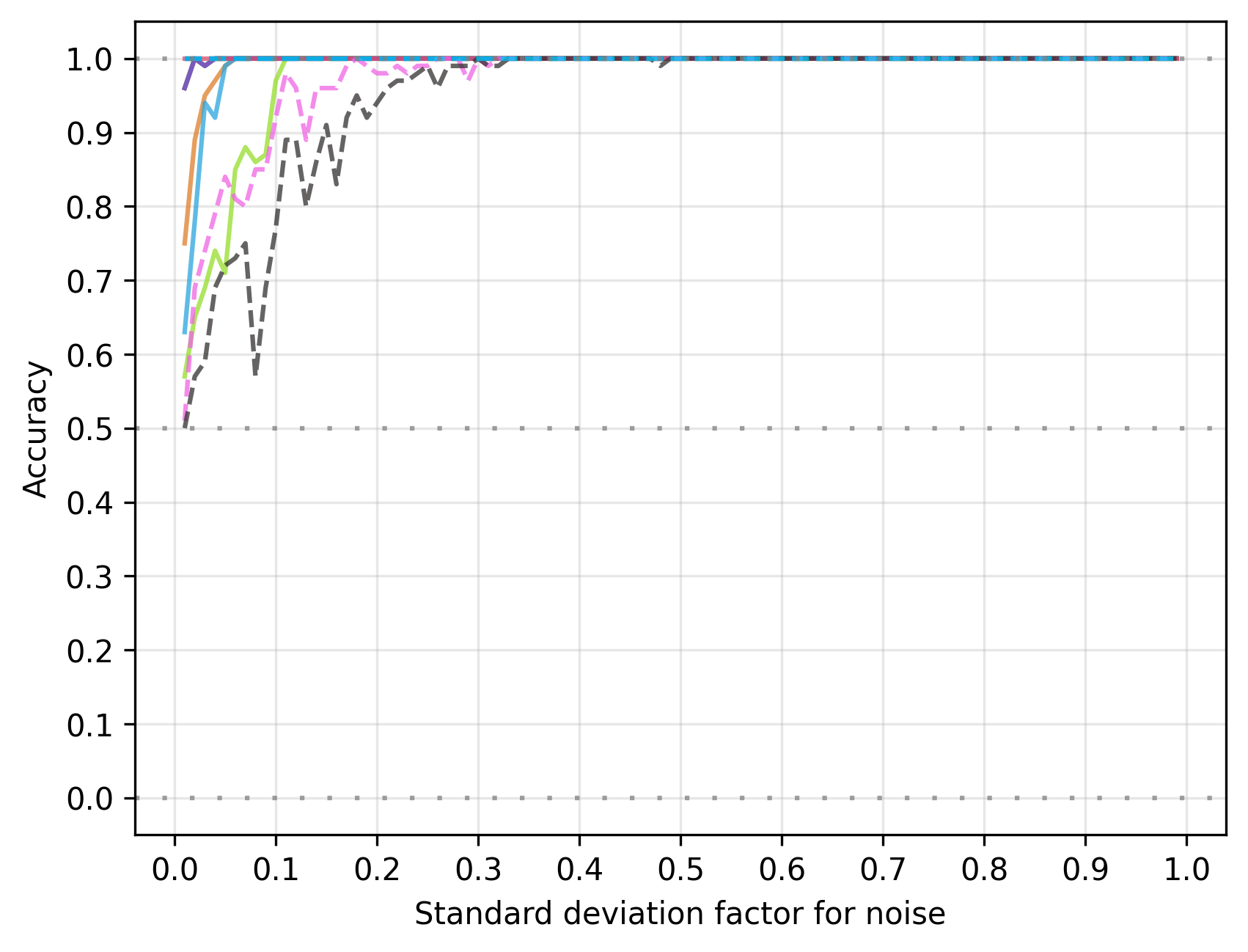}
\end{subfigure}%
\begin{subfigure}{.5\textwidth}
  \centering
  \includegraphics[scale=0.5]{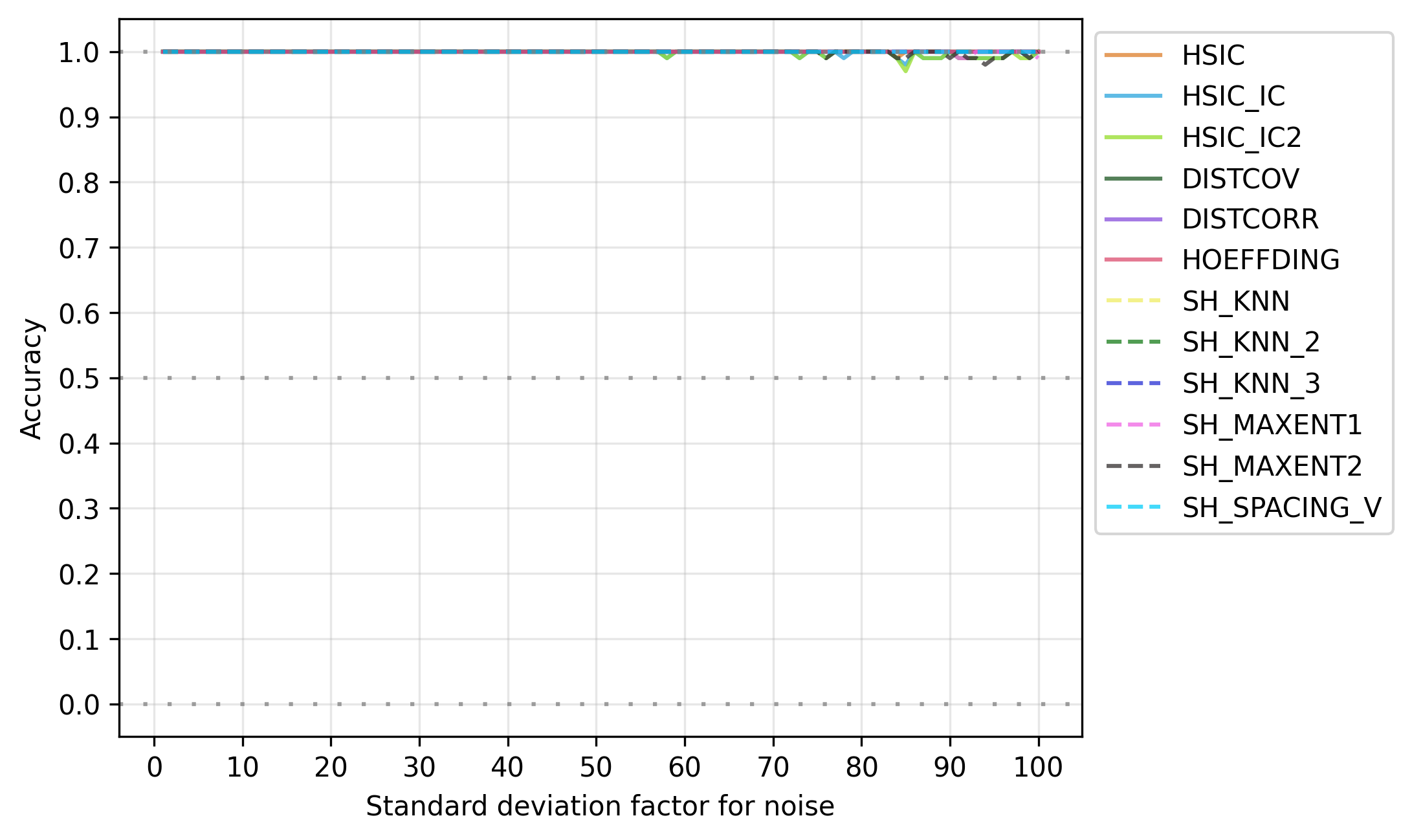}
\end{subfigure}
\caption{RESIT \& different noise levels \& coupled estimation \& $Y = \mathcal{L}^3+\mathcal{L}$}
\label{fig:34}
\end{figure}

\cref{fig:33} shows the linear model $Y= \mathcal{L} + \mathcal{U}$. While SH\_SPACING\_V performed the best of all
estimators in the decoupled estimation most estimators are now close to the performance of SH\_SPACING\_V or even better.
For $i \in [0.4;5]$ all estimators have an accuracy close to or at 100\%.
\cref{fig:34} shows the non-linear model $Y= \mathcal{L}^3 + \mathcal{L}$. For $i \in [0.2;100]$ all estimators
are now close to or at 100\% accuracy.

\newpage

\begin{figure}[h]
\centering
\begin{subfigure}{.5\textwidth}
  \centering
  \includegraphics[scale=0.5]{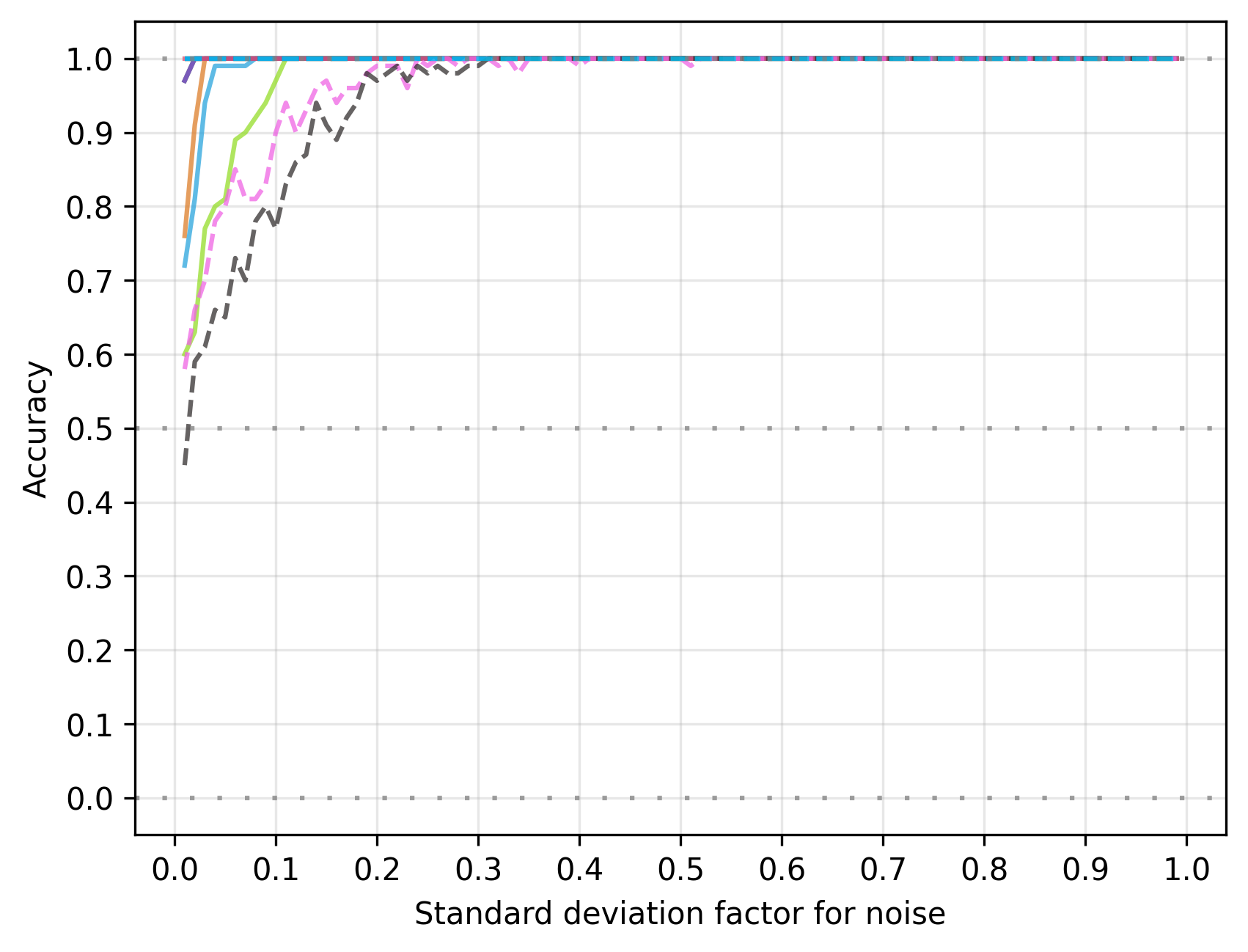}
\end{subfigure}%
\begin{subfigure}{.5\textwidth}
  \centering
  \includegraphics[scale=0.5]{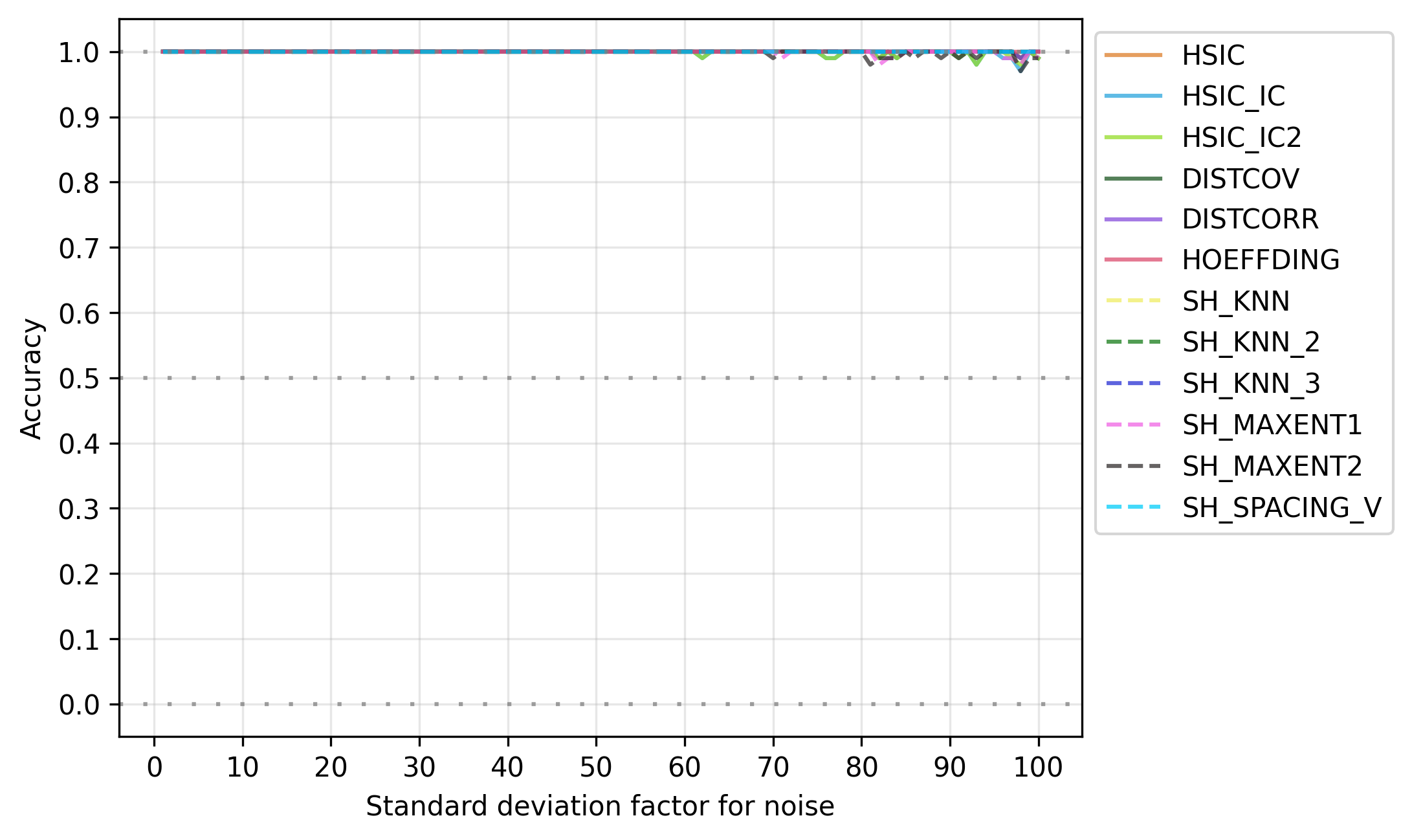}
\end{subfigure}
\caption{RESIT \& different noise levels \& coupled estimation \& $Y = \mathcal{L}^3+\mathcal{N}$}
\label{fig:35}

\centering
\begin{subfigure}{.5\textwidth}
  \centering
  \includegraphics[scale=0.5]{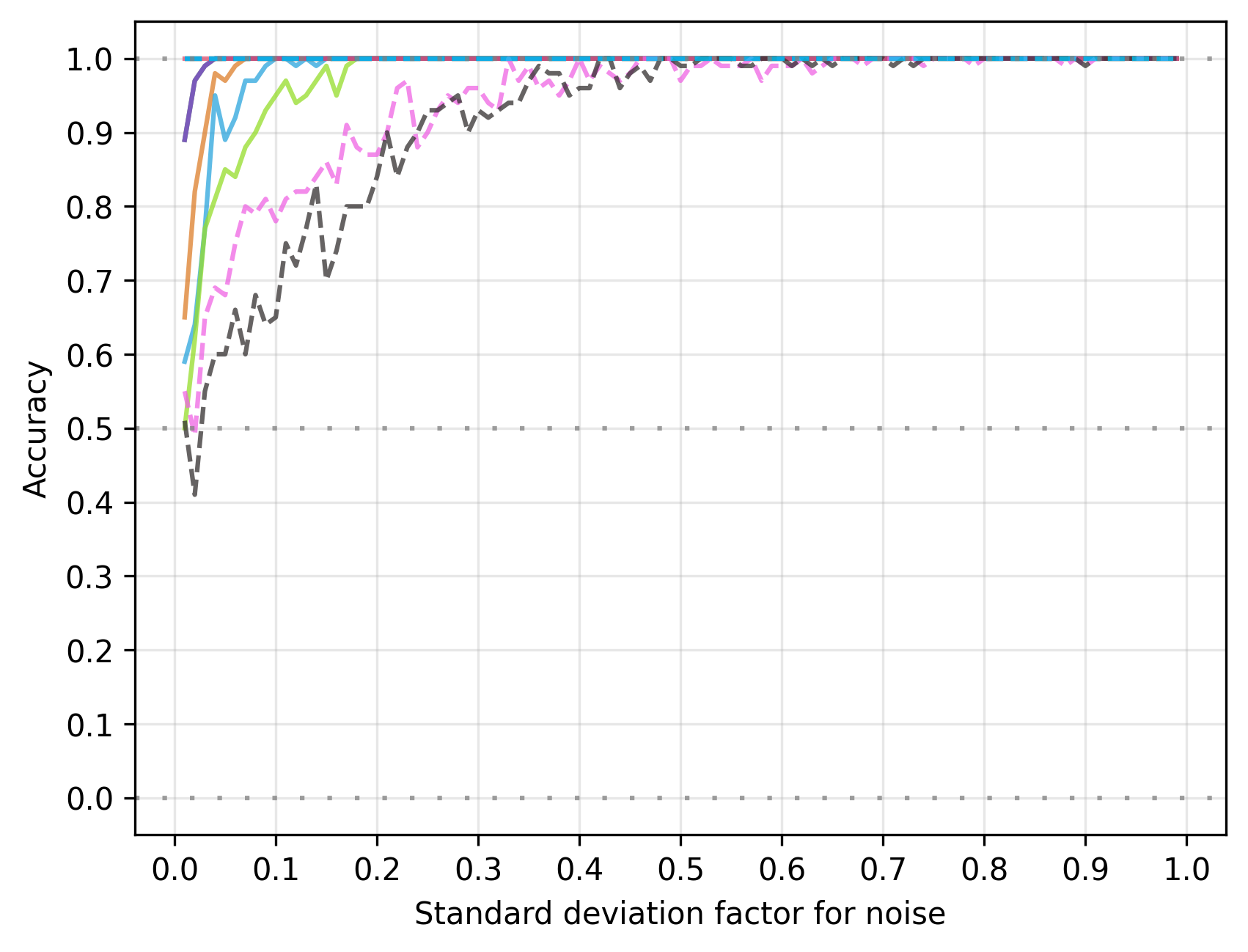}
\end{subfigure}%
\begin{subfigure}{.5\textwidth}
  \centering
  \includegraphics[scale=0.5]{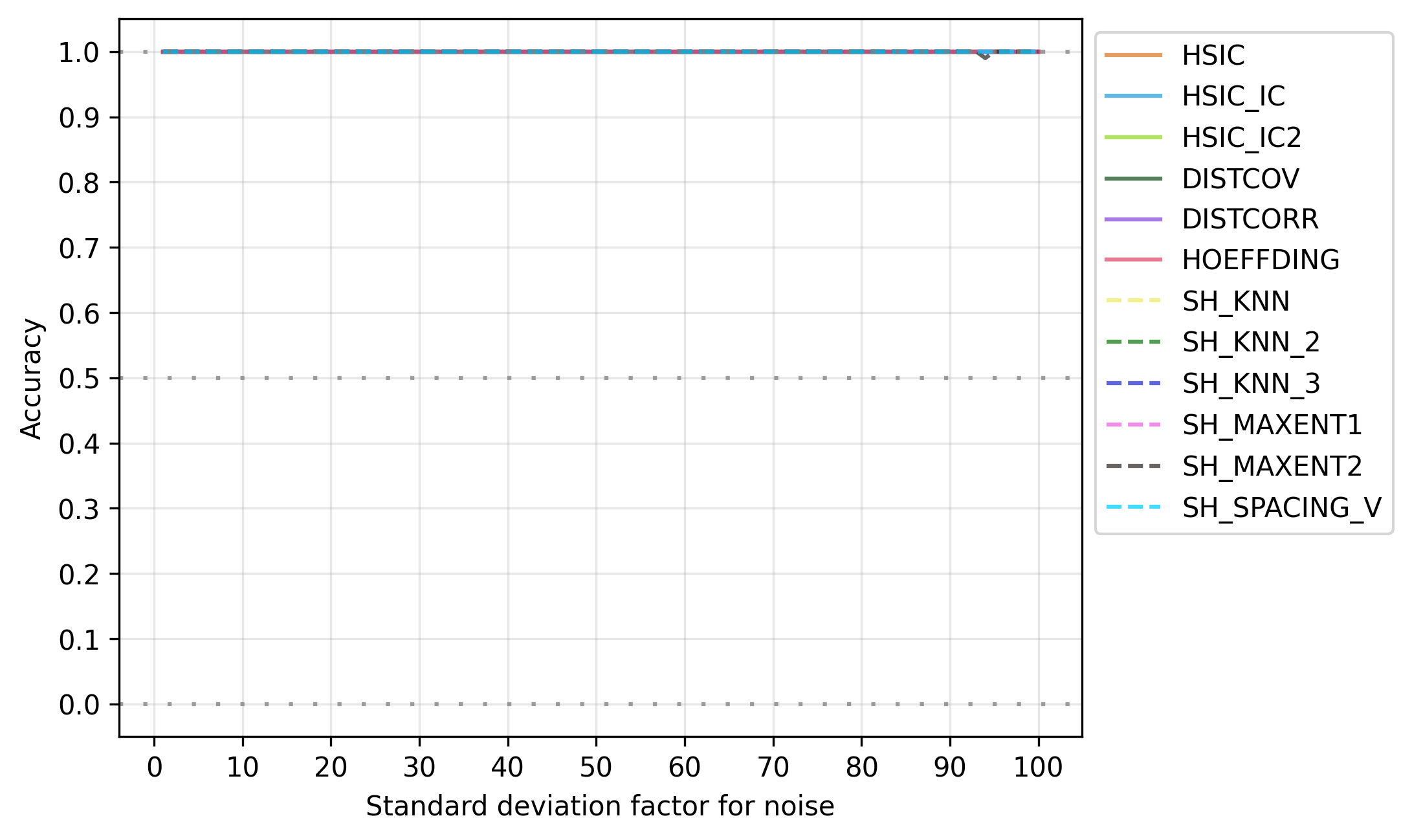}
\end{subfigure}
\caption{RESIT \& different noise levels \& coupled estimation \& $Y = \mathcal{L}^3+\mathcal{U}$}
\label{fig:36}
\end{figure}

\cref{fig:35} shows the non-linear model $Y= \mathcal{L}^3 + \mathcal{N}$. Same as \cref{fig:16}.
For $i \in [0.2;100]$ all estimators are now close to or at 100\% accuracy.
\cref{fig:36} shows the non-linear model $Y= \mathcal{L}^3 + \mathcal{U}$. Similar as the two previous results.
SH\_MAXENT1 and SH\_MAXENT2 do perform better now but are still worse than all other estimators which,
for $i \in [0.1;100]$, are now close to or at 100\% accuracy.

\subsubsection*{Summary and Conclusion}
\cref{summarytable2} follows the same scheme as \cref{summarytable1}.
Similar to the previous section, different noise levels do have an impact
on the identifiability performance of RESIT methods.
Overall, all estimators perform better than in the decoupled estimation case.
Some estimators show better performance in particular structural causal
models (e.g., SH\_SPACING\_V in \cref{fig:26}). Others do perform worse
than the rest of the used estimators (e.g., HSIC\_IC and HSIC\_IC2 in \cref{fig:20}).
For all non-linear equation models the SH\_SPACING\_V and the three Shannon
kNN estimators are always at or at least very close to 100\% accuracy
for $i \in [0.01; 100]$. SH\_SPACING\_V also keeps its good performance in the
linear equation models. For the independence measures HSIC, DISTCOV, DISTCORR and
HOEFFDING perform quite similar and good overall. Furthermore, for linear equation
models if the standard deviation of the noise term is smaller than the standard
deviation of the independent variable ($X$) then RESIT is strong for $i \in [\sim 0.2; 1]$.
In the opposite direction, when the standard deviation of the noise term is larger,
RESIT does not perform that well. Often it is only robust up to a factor $i = 10$ (e.g., \cref{fig:21}).
Note again that this results are also based on the assumption that in our bivariate
structure only one and exact one direction of cause and effect is present (namely $X\to Y$).
Therefore, without this assumption we cannot compare the estimates directly but rather need
to compare the estimate to a derived p-value given some value of alpha (e.g., commonly $\alpha = 0.05$).
The next section describes experiments and results without this assumption.

\begin{sidewaystable}[p]
\begin{center}
\resizebox{\textwidth}{!}{%
\begin{tabular}{|c|c|c|c|c|c|c|c|c|c|c|c|c|}
     \hline
     \textbf{Equation} & \textbf{HSIC} & \textbf{HISC\_IC} & \textbf{HSIC\_IC2} & \textbf{DISTCOV} & \textbf{DISTCORR} & \textbf{HOEFFDING} & \textbf{SH\_KNN} & \textbf{SH\_KNN\_2} & \textbf{SH\_KNN\_3} & \textbf{SH\_MAXENT1} &  \textbf{SH\_MAXENT2} & \textbf{SH\_SPACING\_V}\\\hline
     
     \textbf{$Y=\mathcal{N}+\mathcal{N}$} & & & & & & & & & & & & \\\hline
     
     \textbf{$Y=\mathcal{N}+\mathcal{U}$} & 0.17 - 18 & 0.65 - 26 & 0.7 - 26 & 0.16 - 23 & 0.16 - 23 & 0.16 - 25  & 0.32 - 12 & 0.32 - 12 & 0.24 - 12 & 0.23 - 12 & 0.15 - 22 & 0.13 - 33\\\hline
     
     \textbf{$Y=\mathcal{N}+\mathcal{L}$} & 0.13 - 8 & 0.31 - 7 & 0.33 - 7 & 0.13 - 7 & 0.13 - 8 & 0.13 - 8 & 0.76 - 1 & 0.76 - 1 & 0.51 - 1 & 0.12 - 10 & 0.13 - 7 & 0.17 - 5 \\\hline
     
     \textbf{$Y=\mathcal{N}^3+\mathcal{N}$} & 0.04 - 100 & 0.04 - 83 & 0.08 - 83 & 0.02 - 100 & 0.02 - 100 & 0.01 - 100 & 0.01 - 100 & 0.01 - 100 & 0.01 - 100 & 0.05 - 100 & 0.11 - 98 & 0.01 - 100 \\\hline
     
     \textbf{$Y=\mathcal{N}^3+\mathcal{U}$} & 0.08 - 100 & 0.06 - 100 & 0.08 - 100 & 0.02 - 100 & 0.02 - 100 & 0.01 - 100 & 0.01 - 100 & 0.01 - 100 & 0.01 - 100 & 0.06 - 100 & 0.16 - 100 & 0.01 - 100 \\\hline
     
     \textbf{$Y=\mathcal{N}^3+\mathcal{L}$} & 0.04 - 100 & 0.04 - 70 & 0.09 - 70 & 0.02 - 100 & 0.02 - 100 & 0.01 - 100 & 0.01 - 100 & 0.01 - 100 & 0.01 - 100 & 0.05 - 100 & 0.1 - 100 & 0.01 - 100\\\hline
     
     \textbf{$Y=\mathcal{U}+\mathcal{U}$} & 0.06 - 16 & 0.06 - 15 & 0.14 - 15 & 0.05 - 21 & 0.05 - 21 & 0.05 - 21 & 0.07 - 12 & 0.07 - 12 & 0.07 - 14 & 0.1 - 12 & 0.05 - 17 & 0.03 - 40 \\\hline
     
     \textbf{$Y=\mathcal{U}+\mathcal{N}$} & 0.05 - 6 & 0.04 - 3 & 0.1 - 3 & 0.04 - 7 & 0.04 - 7 & 0.04 - 7 & 0.08 - 4 & 0.08 - 4 & 0.05 - 5 & 0.06 - 4 & 0.03 - 7 & 0.01 - 100 \\\hline
     
     \textbf{$Y=\mathcal{U}+\mathcal{L}$} & 0.04 - 7 & 0.04 - 5 & 0.11 - 5 & 0.04 - 10 & 0.04 - 10 & 0.04 - 8 & 0.09 - 4 & 0.09 - 4 & 0.05 - 5 & 0.04 - 8 & 0.04 - 8 & 0.01 - 100 \\\hline
     
     \textbf{$Y=\mathcal{U}^3+\mathcal{U}$} & 0.03 - 16 & 0.14 - 13 & 0.17 - 13 & 0.01 - 100 & 0.01 - 100 & 0.01 - 100 & 0.01 - 100 & 0.01 - 100 & 0.01 - 100 & 0.02 - 90 & 0.04 - 12 & 0.01 - 100 \\\hline
     
     \textbf{$Y=\mathcal{U}^3+\mathcal{N}$} & 0.02 - 6 & 0.1 - 0.92 & 0.12 - 0.91 & 0.01 - 100 & 0.01 - 100 & 0.01 - 100 & 0.01 - 100 & 0.01 - 100 & 0.01 - 100 & 0.01 - 100 & 0.03 - 4 & 0.01 - 100 \\\hline
     
     \textbf{$Y=\mathcal{U}^3+\mathcal{L}$} & 0.03 - 7 & 0.1 - 4 & 0.17 - 4 & 0.01 - 100 & 0.01 - 100 & 0.01 - 100 & 0.01 - 100 & 0.01 - 100 & 0.01 - 100 & 0.01 - 88 & 0.04 - 5 & 0.01 - 100 \\\hline
     
     \textbf{$Y=\mathcal{L}+\mathcal{L}$} & 0.1 - 13 & 0.1 - 8 & 0.12 - 8 & 0.08 - 15 & 0.08 - 15 & 0.1 - 10 & 0.37 - 3 & 0.37 - 3 & 0.32 - 4 & 0.07 - 17 & 0.1 - 13 & 0.09 - 13 \\\hline
     
     \textbf{$Y=\mathcal{L}+\mathcal{N}$} & 0.1 - 7 & 0.1 - 4 & 0.1 - 4 & 0.1 - 7 & 0.1 - 7 & 0.1 - 7 & 0.61 - 1 & 0.61 - 1 & 0.37 - 3 & 0.07 - 13 & 0.1 - 7 & 0.14 - 6 \\\hline
     
     \textbf{$Y=\mathcal{L}+\mathcal{U}$} & 0.12 - 23 & 0.14 - 26 & 0.14 - 26 & 0.1 - 25 & 0.1 - 25 & 0.1 - 25 & 0.27 - 12 & 0.27 - 12 & 0.21 - 15 & 0.11 - 24 & 0.11 - 23 & 0.11 - 33 \\\hline
     
     \textbf{$Y=\mathcal{L}^3+\mathcal{L}$} & 0.02 - 100 & 0.03 - 100 & 0.09 - 100 & 0.01 - 100 & 0.01 - 100 & 0.01 - 100 & 0.01 - 100 & 0.01 - 100 & 0.01 - 100 & 0.1 - 100 & 0.15 - 100 & 0.01 - 100 \\\hline
     
     \textbf{$Y=\mathcal{L}^3+\mathcal{N}$} & 0.02 - 100 & 0.03 - 100 & 0.7 - 100 & 0.01 - 100 & 0.01 - 100 & 0.01 - 100 & 0.01 - 100 & 0.01 - 100 & 0.01 - 100 & 0.1 - 100 & 0.14 - 100 & 0.01 - 100 \\\hline
     
     \textbf{$Y=\mathcal{L}^3+\mathcal{U}$} & 0.04 - 100 & 0.05 - 100 & 0.07 - 100 & 0.01 - 100 & 0.01 - 100 & 0.01 - 100 & 0.01 - 100 & 0.01 - 100 & 0.01 - 100 & 0.17 - 100 & 0.21 - 100 & 0.01 - 100\\\hline
     
\end{tabular}}
\end{center}
\caption{Summary Table for RESIT \& different noise levels \& Coupled estimation. The numbers reflect the ranges of noise that allow identifiability with accuracy around 90\%.}
\label{summarytable2}
\end{sidewaystable}

\section{RESIT with Different Noise Levels without prior assumption}\label{res2}
Without assuming that exactly one direction has to be present in our bivariate model,
we must then account for four different possibilities instead:
\begin{enumerate}
    \item $X \to Y$ and $Y \cancel{\to} X$
    \item $X \cancel{\to} Y$ and $Y \to X$
    \item $X \to Y$ and $Y \to X$
    \item $X \cancel{\to} Y$ and $Y \cancel{\to} X$
\end{enumerate}
This means that we cannot compare estimates directly but have to estimate an $\alpha$ value
for the independence test when testing each direction and we can no longer use entropy estimators.

\subsection{Setup}
This setup is the same as in \cref{1Setup}. The only difference is a small modification
in the algorithm we use (\cref{algo1}).
More precisely, the first steps remain unchanged but the step number 5 needs to be adjusted.
Here, we perform independence test on the independent variable and the residuals and
compare the independence estimate to the alpha estimate with alpha = 0.05.
Furthermore, since the equation of our data is $Y = X + N_y$ only the outcome
$X \to Y$ and $Y \cancel{\to} X$ from our independence tests is the correct one and thus
we can formulate the following output function as the second step in our modified algorithm:

\begin{gather}
    dir = \begin{cases}
	  Correct & \text{if } I(X_{Test},Y_{res}) < \alpha (0.05) \land I(Y_{Test},X_{res}) > \alpha (0.05),\\
	  Incorrect & \text{otherwise }
	\end{cases}
\label{eqn:output}
\end{gather}\\

As before, Linear Regression is used. 
However, for simplicity we only used one independence test: Hilbert-Schmidt Independence Criterion 
with RBF Kernel (HSIC) \footnote{Source: https://github.com/amber0309/HSIC}. Here,
we use the gamma test to compare the HSIC estimate, that is with the alpha quantile 
of the gamma distribution with mean and variance of HSIC under independence hypothesis.
The HSIC estimator \parencite[]{gretton2005} is chosen because of its general good performance in previous
results.

\subsection{Execution}
The execution is exactly the same as in \cref{1Init}.

\subsection{Results}
In the following figures the y-axis shows the accuracy ($\frac{\text{\#successful tests}}{100}$) and 
the x-axis shows the range of the $i$ factor. Each figure contains two subfigures, the left with
$i \in \{0.01, 0.02, \dots, 1.00\}$ and the right figure with $i \in \{1, 2, \dots, 100\}$..
Differently from \cref{res1}, if HSIC is closer to 0 then we have \textbf{unidentifiability.} 
If plots are closer to accuracy $1$ then we have very good/consistent \textbf{identifiability}. 
The next subsection describes each figure individually and the  subsection thereafter 
provides a summary and a small conclusion. Additionally, we also directly cover
the decoupled estimation together with the coupled estimation approach. Lastly, the figures
apply some abbreviations which we shortly explain here:
\begin{itemize}
    \item GAU: Gaussian distribution.
    \item UNI: Uniform distribution.
    \item LAP: Laplace distribution.
    \item NL***: Non-linear variant.
\end{itemize}
Furthermore, if in a legend of a plot there are two of the above abbreviations concatenated with a "+", e.g.,
GAU+UNI, then this signifies that $X \sim \mathcal{N}$ and $N_y \sim \mathcal{U}$ (respecting order).
If an abbreviation stands alone, for example GAU, then both variables ($X, N_y$) are drawn from that distribution.

\newpage
\subsubsection*{Individual Analysis}
\begin{figure}[!h]
\centering
\begin{subfigure}{.5\textwidth}
  \centering
  \includegraphics[scale=0.5]{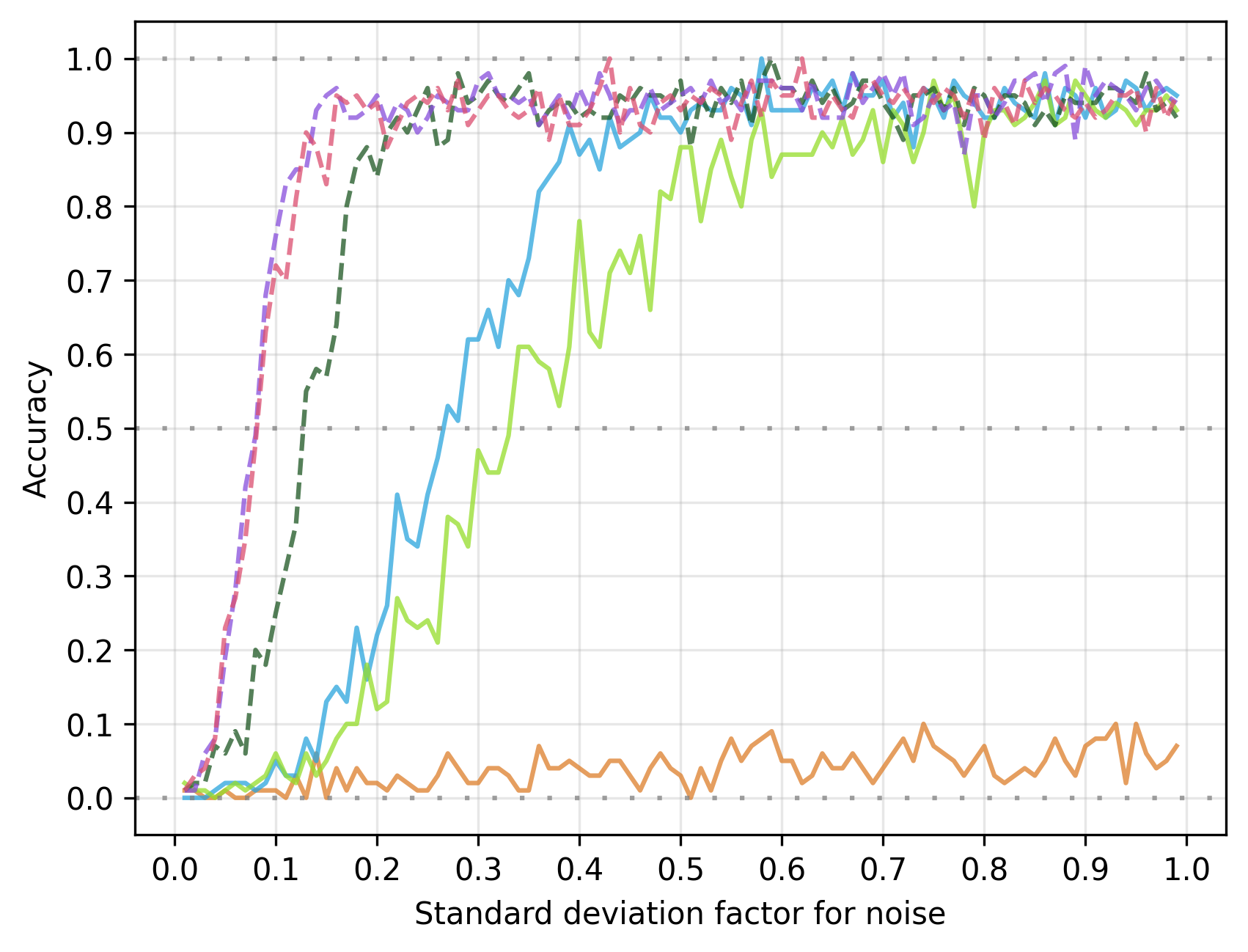}
\end{subfigure}%
\begin{subfigure}{.5\textwidth}
  \centering
  \includegraphics[scale=0.5]{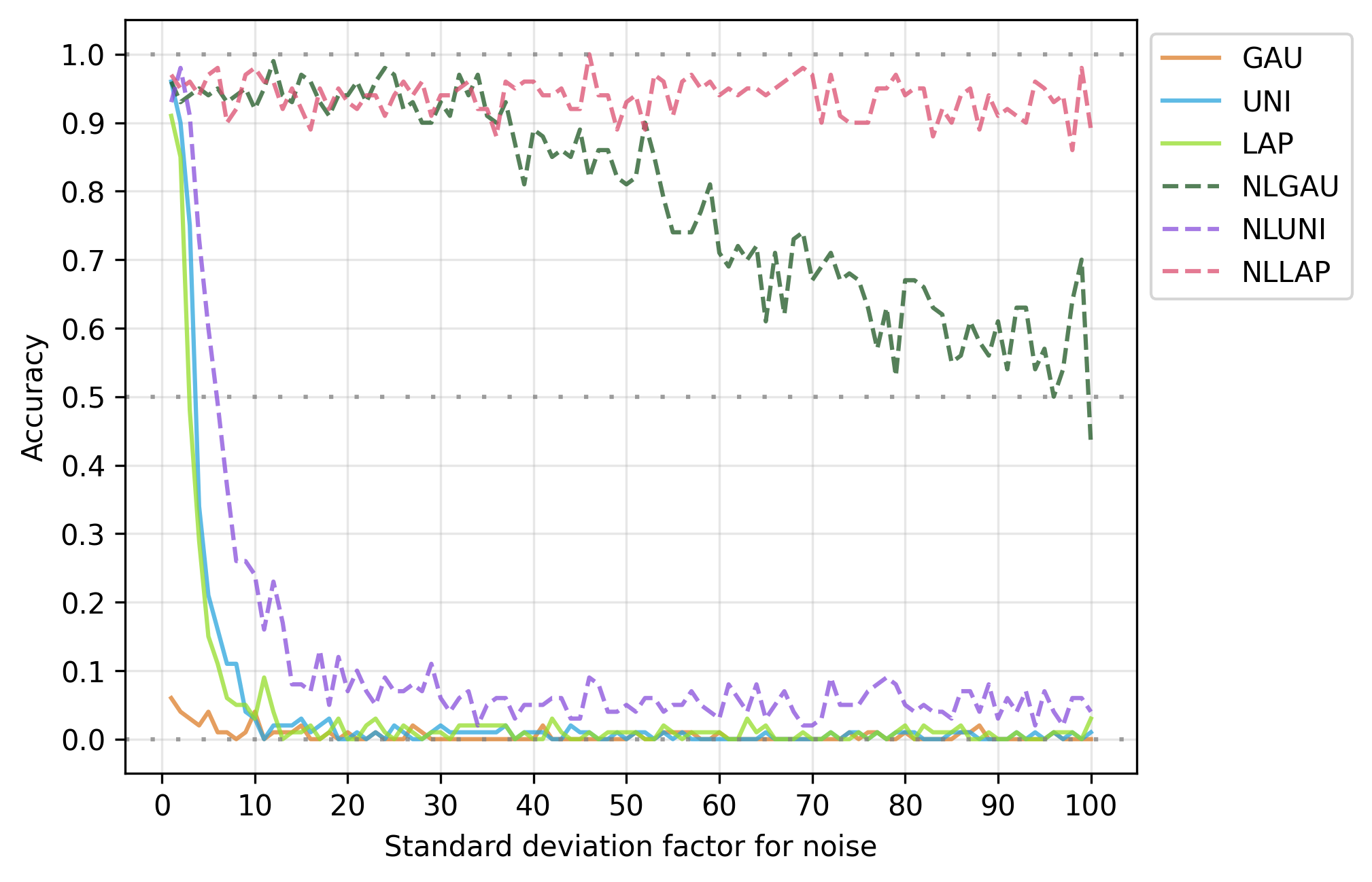}
\end{subfigure}
\caption{$Y = X + N_y$. Contains all cases where $X$ and $N_y$ are drawn from the same
    type of distribution. Dashed lines are non-linear cases. Decoupled estimation (80\% split).}
\label{fig:37}

\centering
\begin{subfigure}{.5\textwidth}
  \centering
  \includegraphics[scale=0.5]{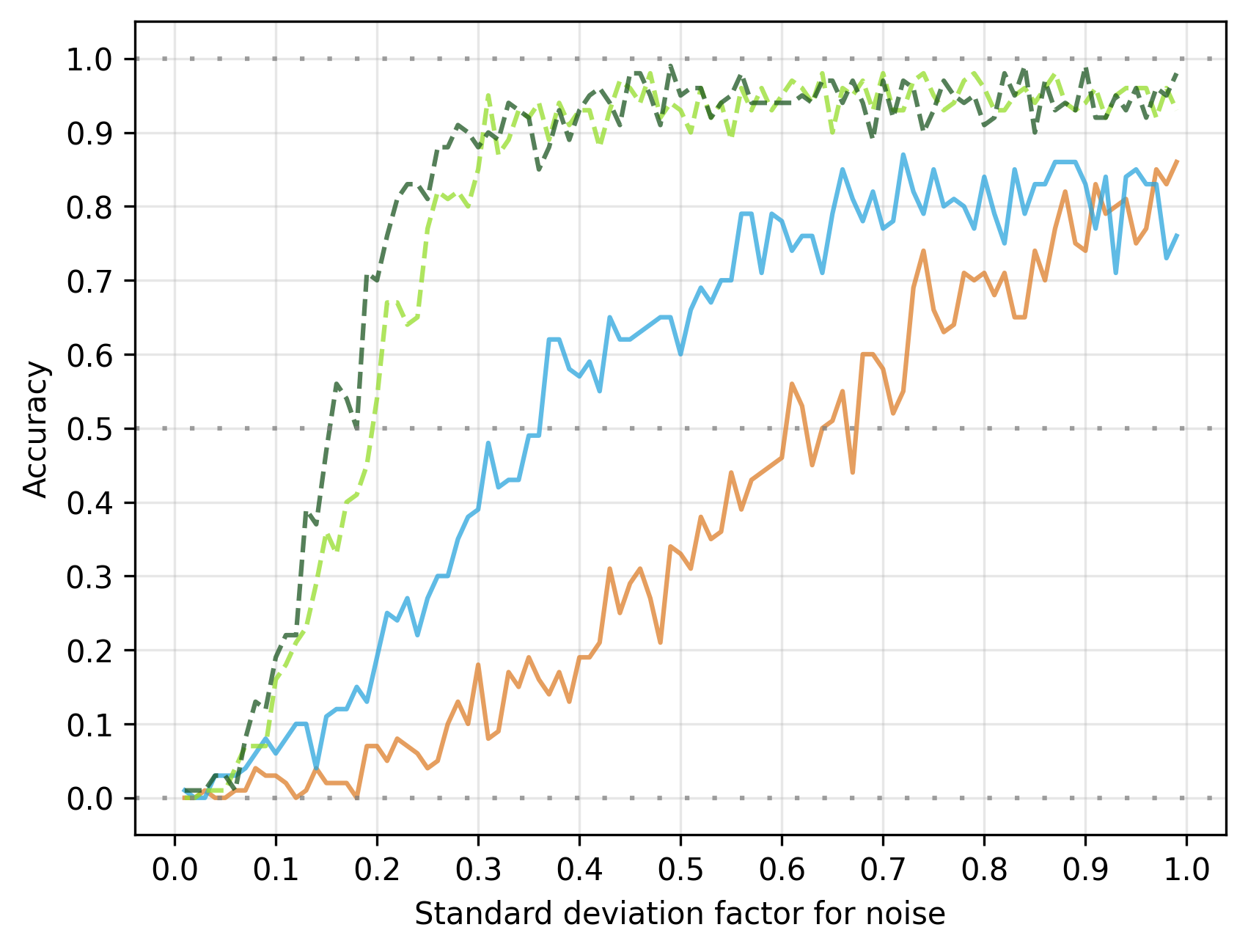}
\end{subfigure}%
\begin{subfigure}{.5\textwidth}
  \centering
  \includegraphics[scale=0.5]{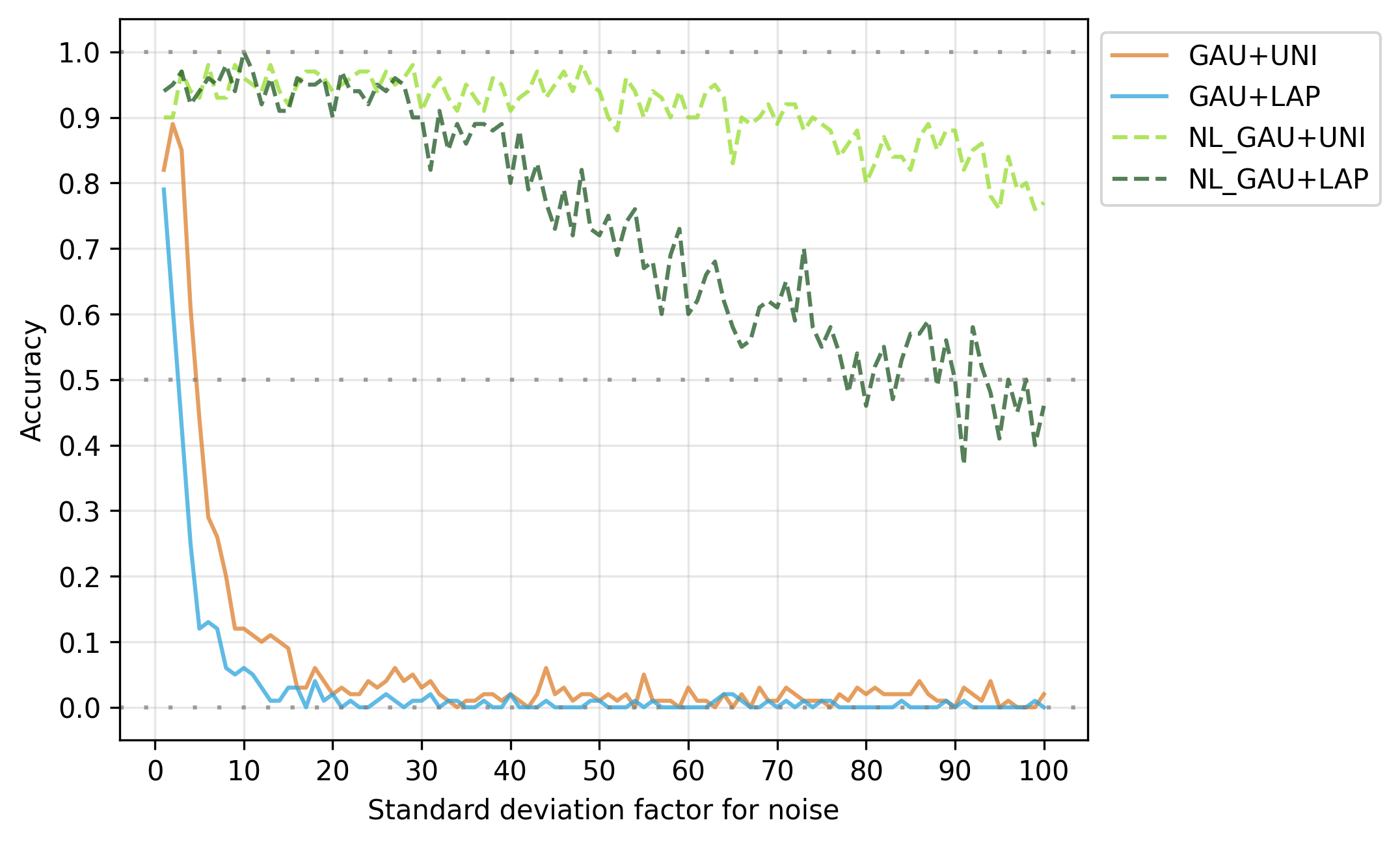}
\end{subfigure}
\caption{$Y = X + N_y$. Contains all cases where $X \sim \mathcal{N}$ and $N_y \cancel{\sim} \mathcal{N}$. 
Dashed lines are non-linear cases. Decoupled estimation (80\% split).}
\label{fig:38}
\end{figure}

\begin{figure}[!h]
\centering
\begin{subfigure}{.5\textwidth}
  \centering
  \includegraphics[scale=0.5]{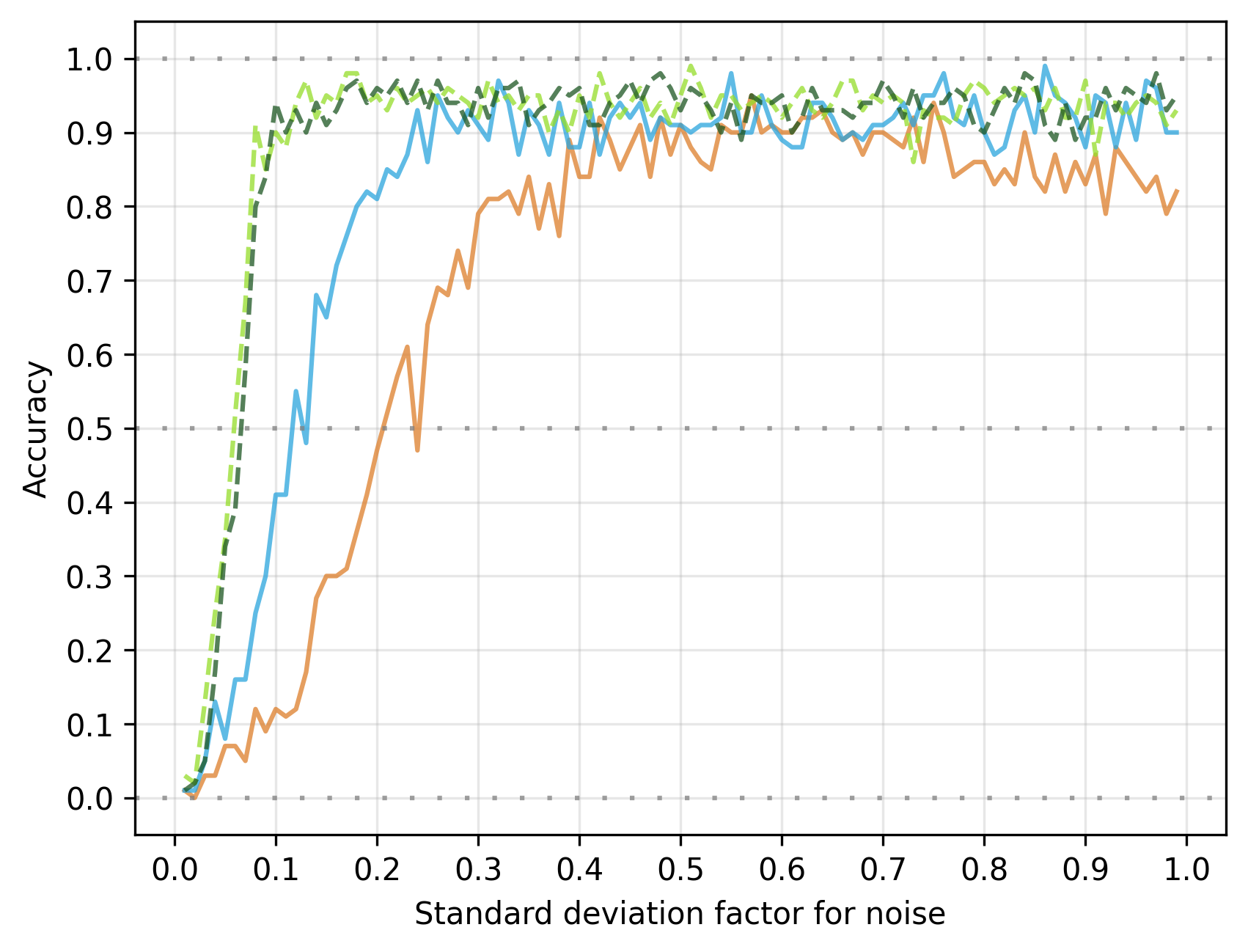}
\end{subfigure}%
\begin{subfigure}{.5\textwidth}
  \centering
  \includegraphics[scale=0.5]{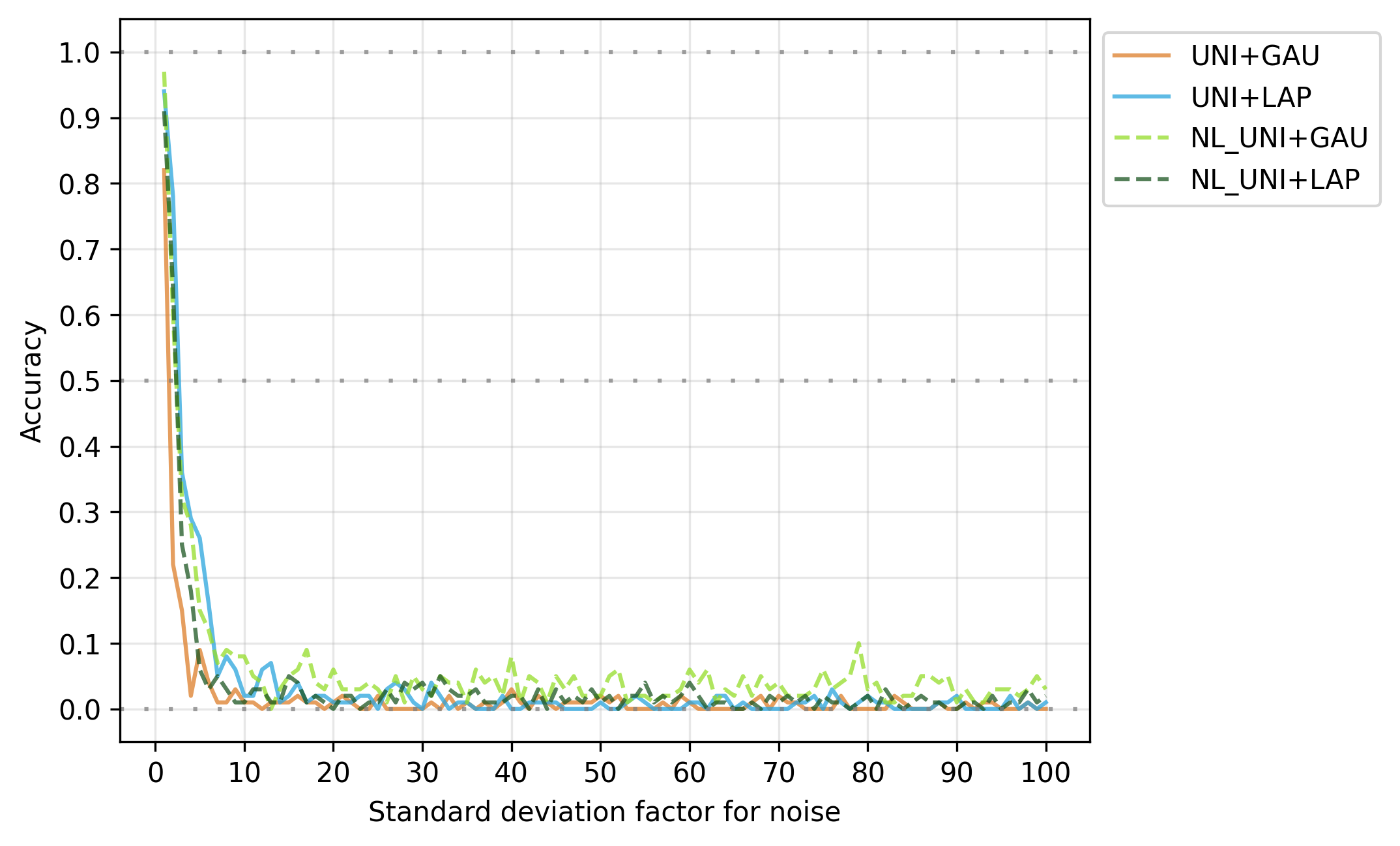}
\end{subfigure}
\caption{$Y = X + N_y$. Contains all cases where $X \sim \mathcal{U}$ and $N_y \cancel{\sim} \mathcal{U}$.  
Dashed lines are non-linear cases. Decoupled estimation (80\% split).}
\label{fig:39}

\centering
\begin{subfigure}{.5\textwidth}
  \centering
  \includegraphics[scale=0.5]{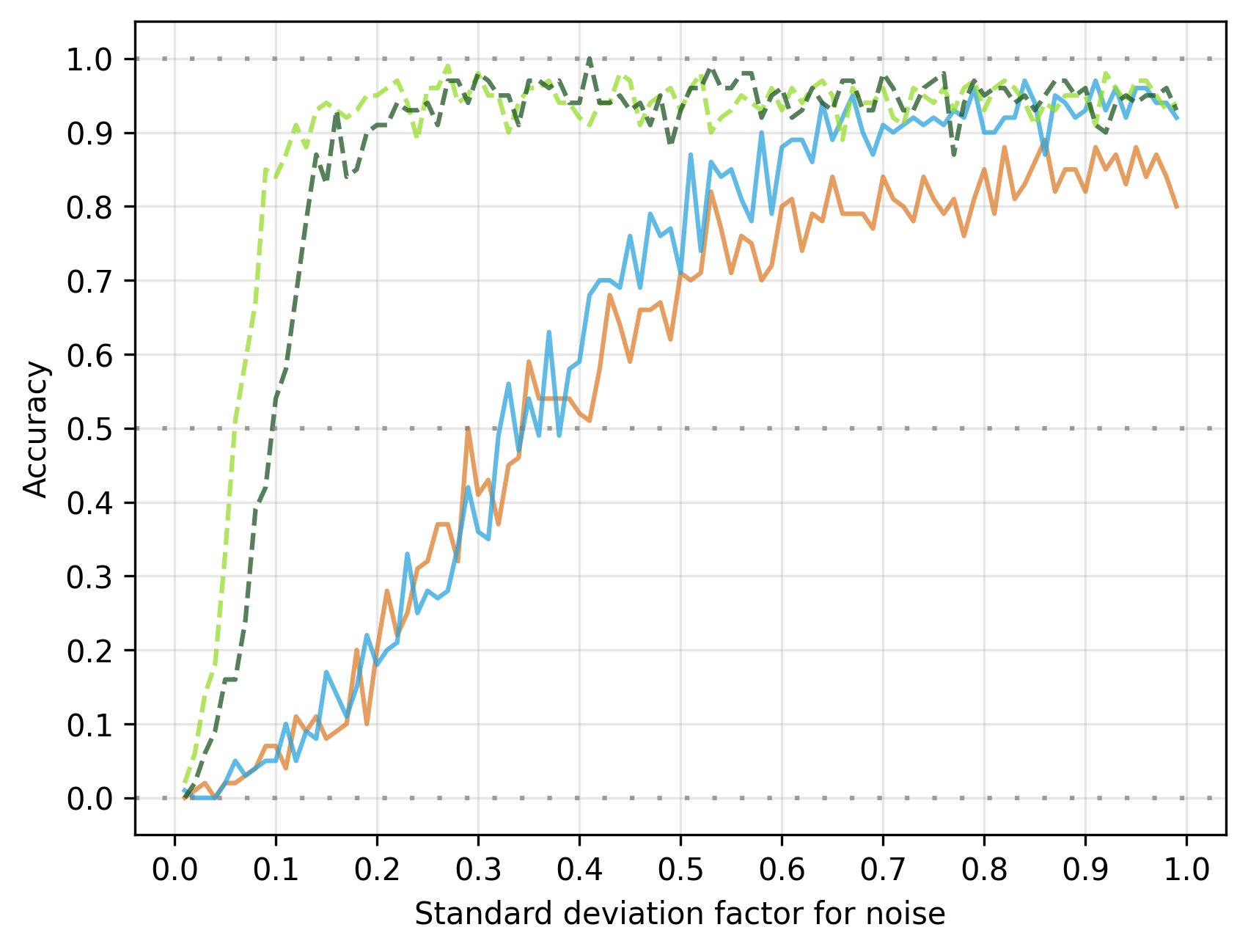}
\end{subfigure}%
\begin{subfigure}{.5\textwidth}
  \centering
  \includegraphics[scale=0.5]{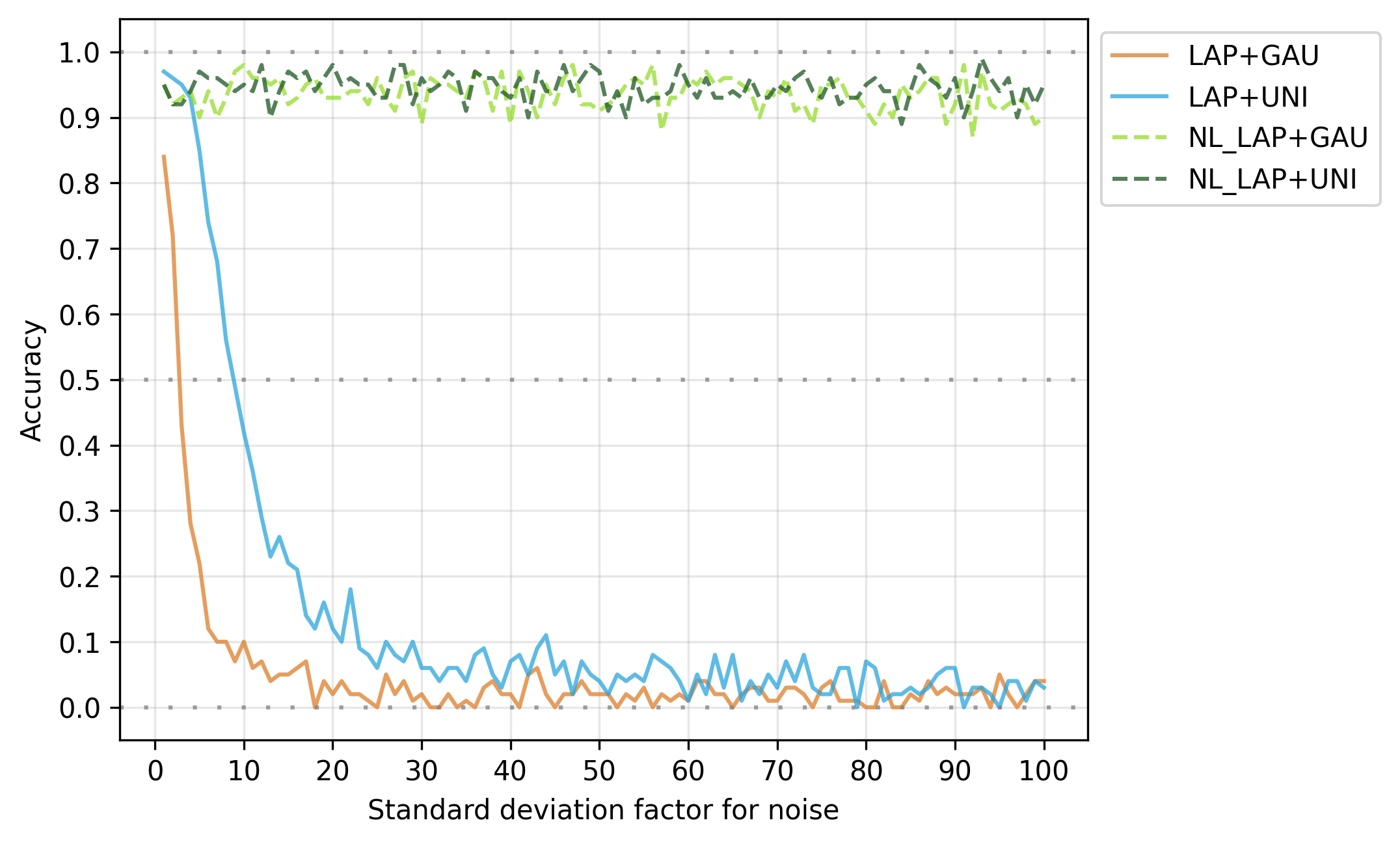}
\end{subfigure}
\caption{$Y = X + N_y$. Contains all cases where $X \sim \mathcal{L}$ and $N_y \cancel{\sim} \mathcal{L}$. 
Dashed lines are non-linear cases. Decoupled estimation (80\% split).}
\label{fig:40}
\end{figure}

Again, individual analysis can be skipped as we provide a summary in \cref{summarytable3}.
The following paragraphs describe \cref{fig:37} - \cref{fig:40} which show the performance
of using decoupled estimation with a split of 80\%/20\%.
\cref{fig:37} shows all cases where both $X$ and $N_y$ are drawn from the same distribution.
$Y = \mathcal{N} + \mathcal{N}$ is the only case which never
achieves identifiability with $i \in [0.01; 100]$. The non-linear cases are pretty robust
for $i \in [0.2; 1]$; NLLAP remains above 90\% accuracy for $i \in [0.3; 100]$ and
NLGAU remains over 90\% accuracy for $i \in [0.25; 35]$. However, NLUNI already drops
fast for $i > 2$ but remains above 90\% accuracy for $i \in [0.12;2]$. UNI and LAP
are over 80\% accuracy for $i \in [0.5; 2]$ but drop then fast after $i = 2$.
\cref{fig:38} shows all cases where $X \sim \mathcal{N}$ and $N_y \cancel{\sim} \mathcal{N}$. 
GAU+UNI only reaches an accuracy of 90\% when $i = 2$. GAU+LAP never reaches over 90\% accuracy 
but remains partially over 80\% for $i \in [0.6; 1]$. NL\_GAU+UNI has an accuracy over 90\% 
for $i \in [0.3;65]$ and NL\_GAU+LAP has also a good accuracy for $i \in [0.2; 20]$.

\newpage

\cref{fig:39} shows all cases where $X \sim \mathcal{U}$ and $N_y \cancel{\sim} \mathcal{U}$.
All cases drop fast to $<$ 10\% accuracy for $i > 1$. NL\_UNI+GAU and NL\_UNI+LAP are
over 90\% accuracy for $i \in [0.1; 1]$. UNI+LAP hangs around 90\% accuracy for $i \in [0.2; 1]$
and UNI+GAU never reaches a consistent accuracy above 90\%, only over 80\% for $i \in [0.33; 0.9]$.
\cref{fig:40} shows all cases where $X \sim \mathcal{L}$ and $N_y \cancel{\sim} \mathcal{L}$.
NL\_LAP+GAU performs the best here with accuracy above 90\% for $i \in [0.3; 100]$, followed by
NL\_LAP+UNI for $i \in [0.5; 100]$. LAP\_UNI only reaches 90\% accuracy around $i = 1$. LAP+GAU never
$>$ 90\%, only 80\% for $i \in [0.95; 1]$.

\newpage

\begin{figure}[h!]
\centering
\begin{subfigure}{.5\textwidth}
  \centering
  \includegraphics[scale=0.5]{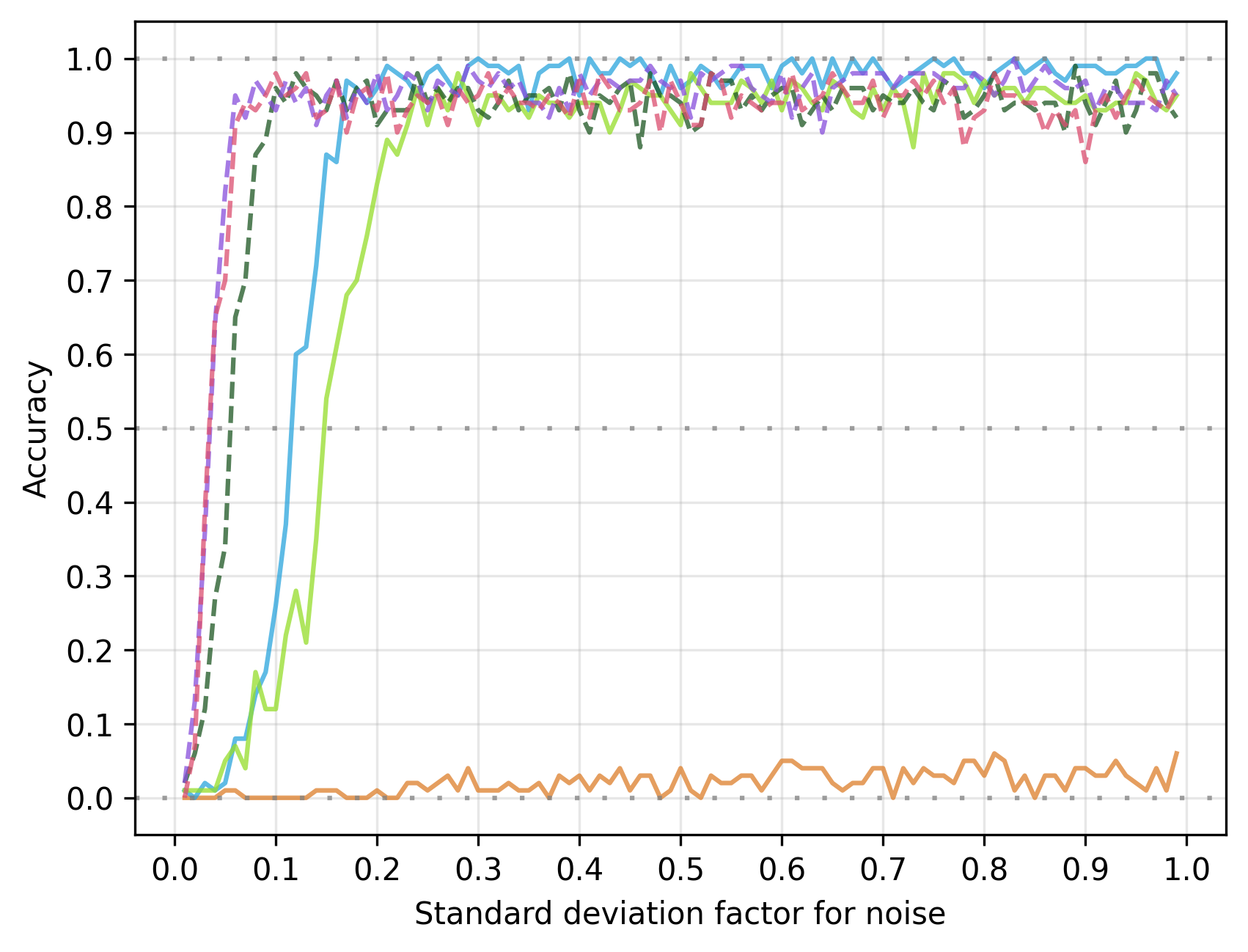}
\end{subfigure}%
\begin{subfigure}{.5\textwidth}
  \centering
  \includegraphics[scale=0.5]{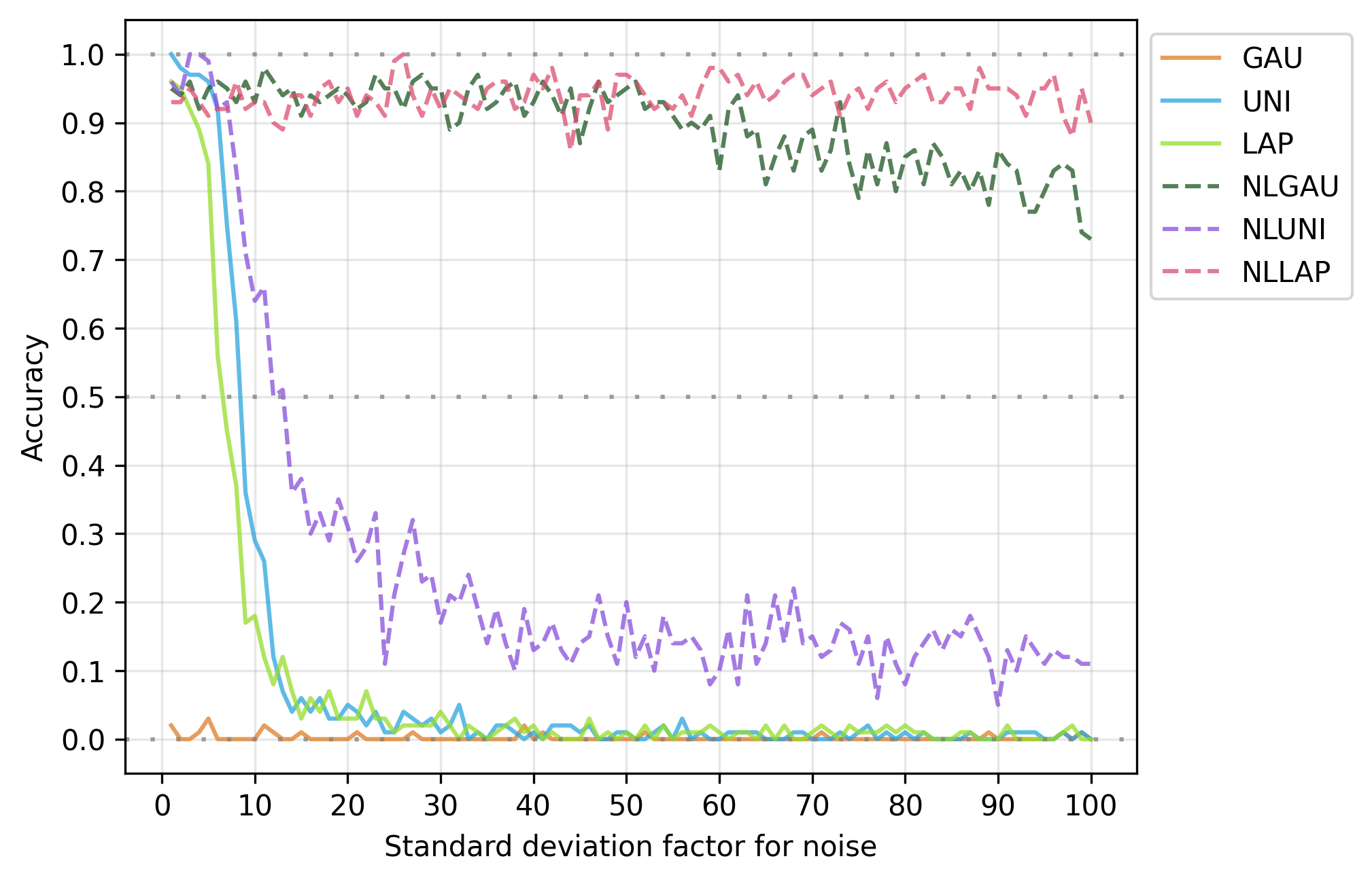}
\end{subfigure}
\caption{$Y = X + N_y$. Contains all cases where $X$ and $N_y$ are drawn from the same
type of distribution. Dashed lines are non-linear cases. Coupled estimation.}
\label{fig:41}

\centering
\begin{subfigure}{.5\textwidth}
  \centering
  \includegraphics[scale=0.5]{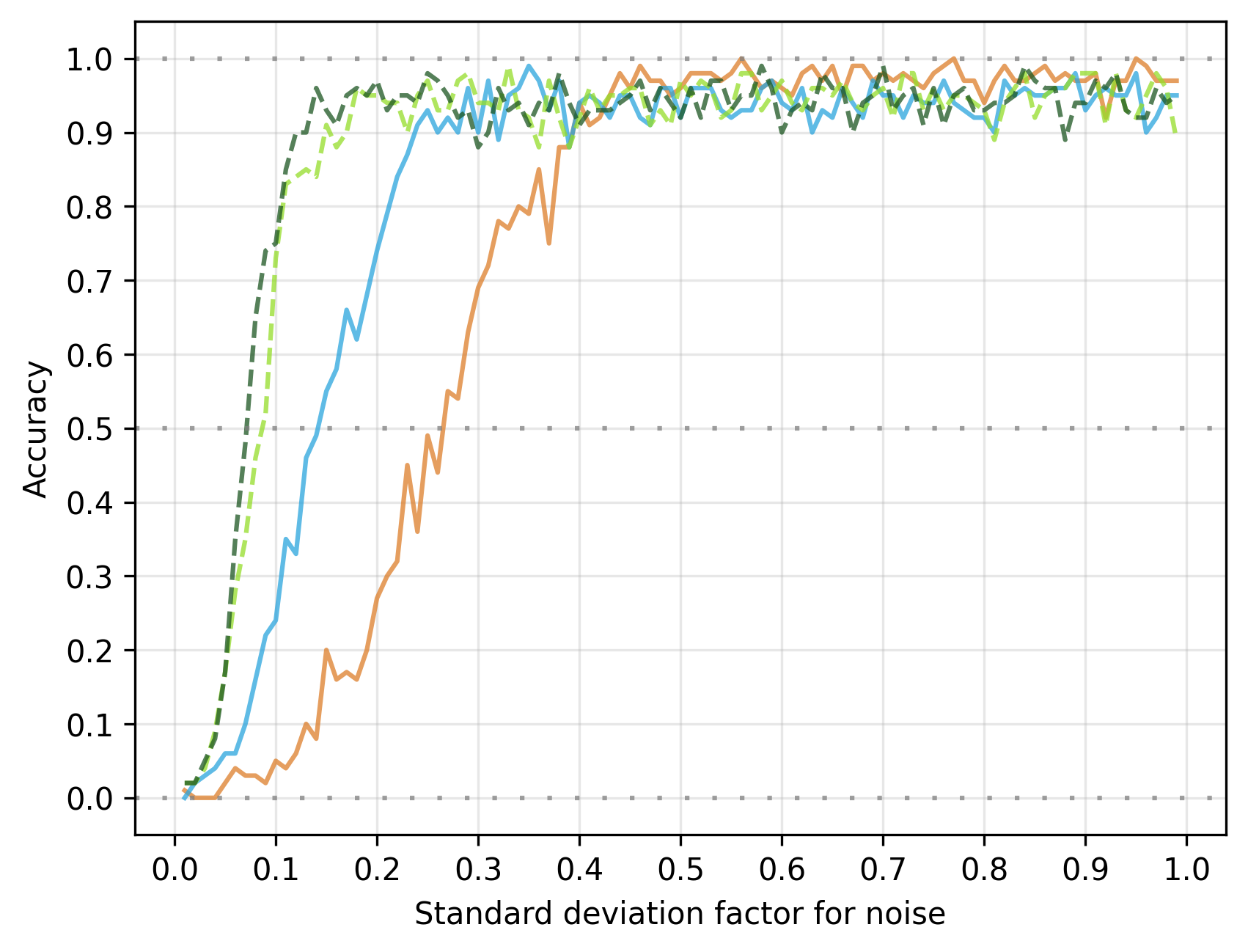}
\end{subfigure}%
\begin{subfigure}{.5\textwidth}
  \centering
  \includegraphics[scale=0.5]{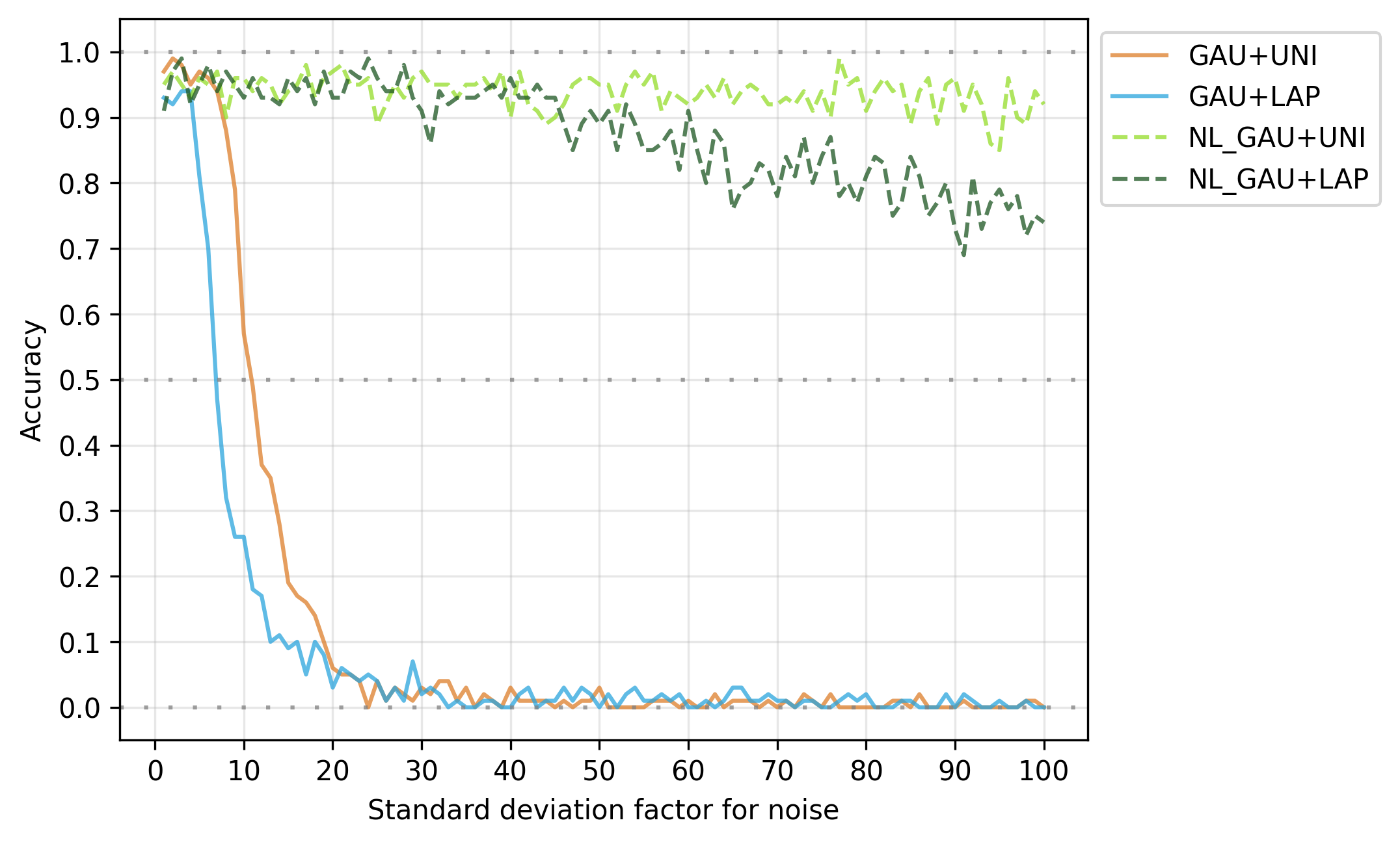}
\end{subfigure}
\caption{$Y = X + N_y$. Contains all cases where $X \sim \mathcal{N}$ and $N_y \cancel{\sim} \mathcal{N}$. 
Dashed lines are non-linear cases. Coupled estimation.}
\label{fig:42}
\end{figure}

\begin{figure}[h!]
\centering
\begin{subfigure}{.5\textwidth}
  \centering
  \includegraphics[scale=0.5]{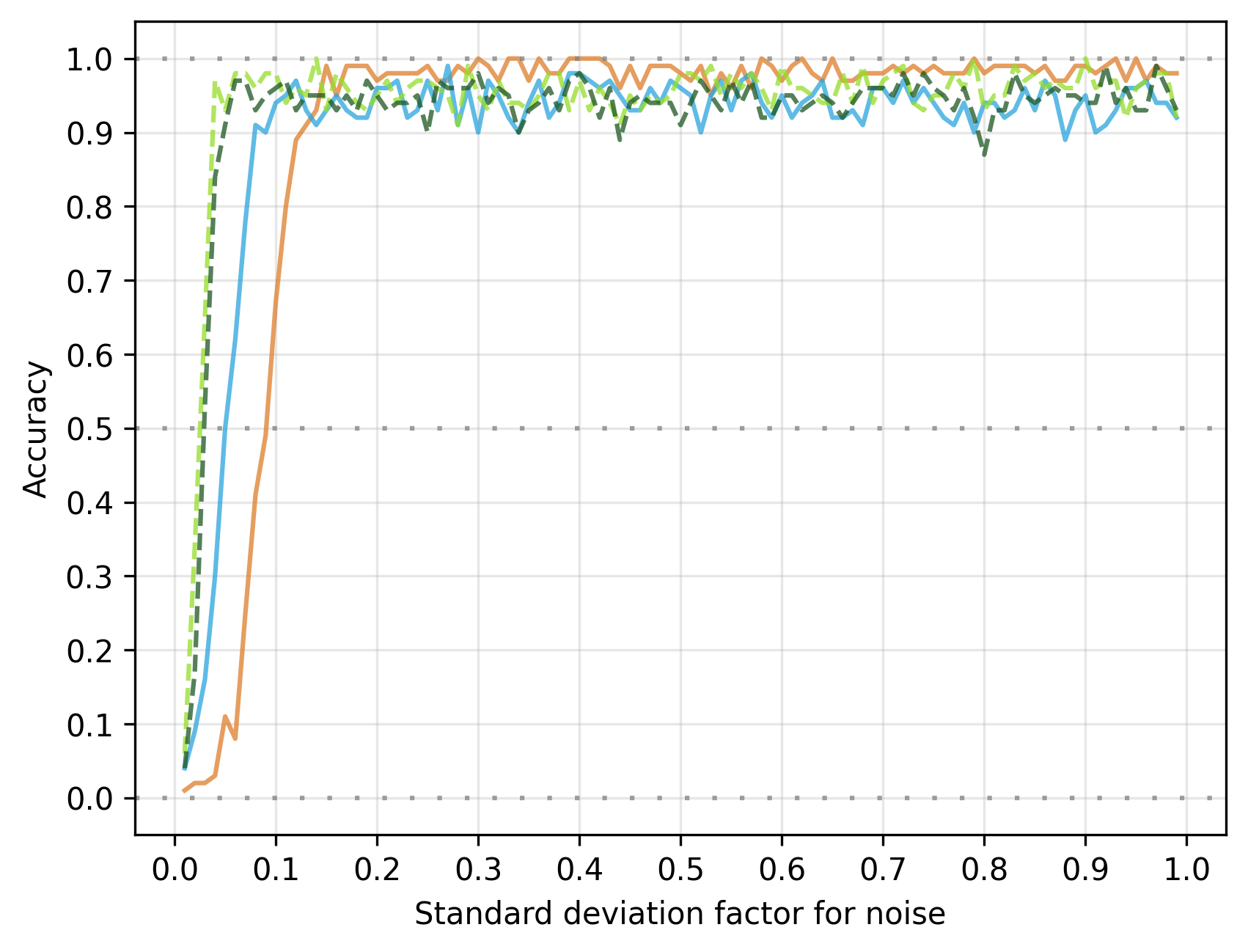}
\end{subfigure}%
\begin{subfigure}{.5\textwidth}
  \centering
  \includegraphics[scale=0.5]{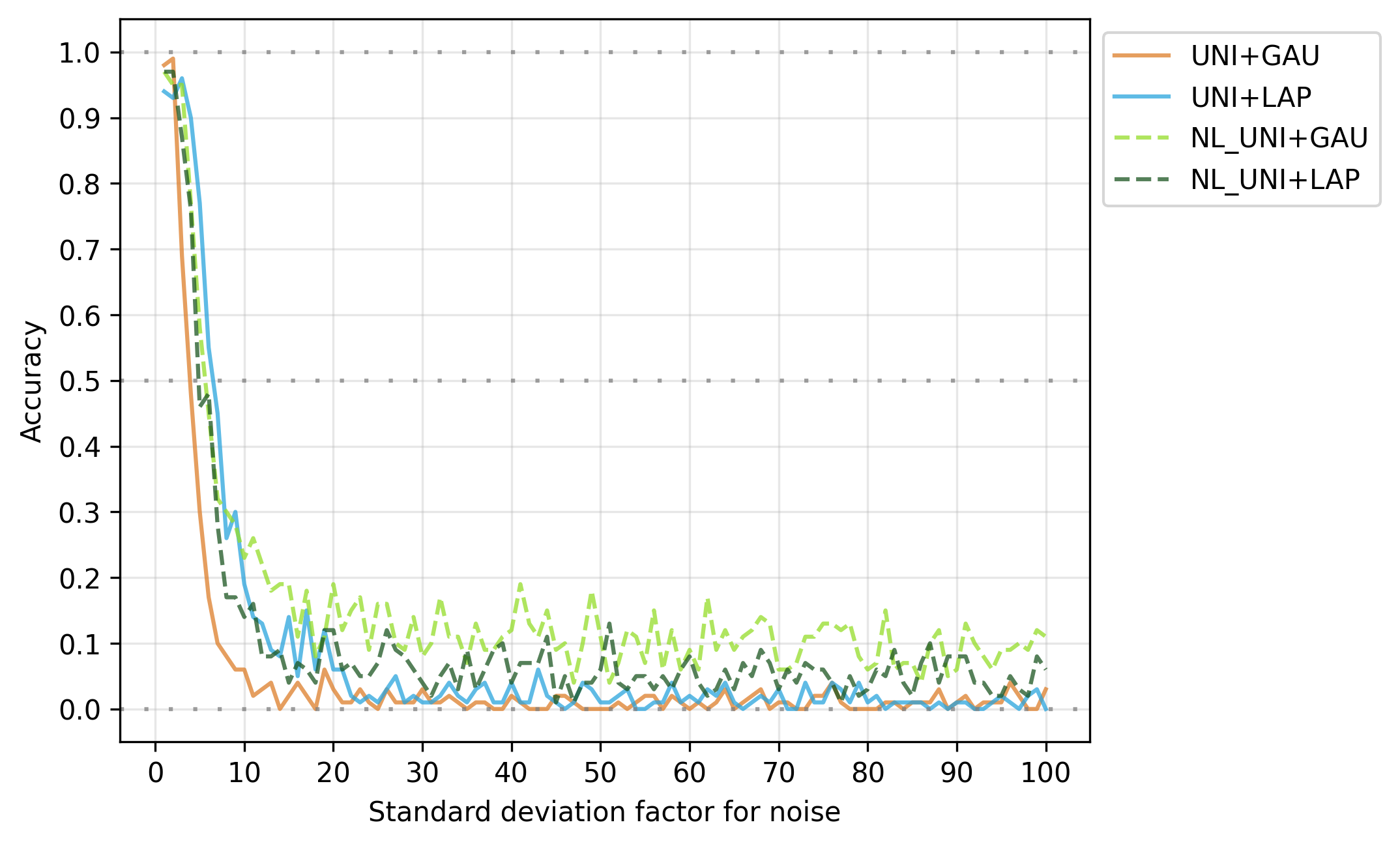}
\end{subfigure}
\caption{$Y = X + N_y$. Contains all cases where $X \sim \mathcal{U}$ and $N_y \cancel{\sim} \mathcal{U}$. 
Dashed lines are non-linear cases. Coupled estimation.}
\label{fig:43}

\centering
\begin{subfigure}{.5\textwidth}
  \centering
  \includegraphics[scale=0.5]{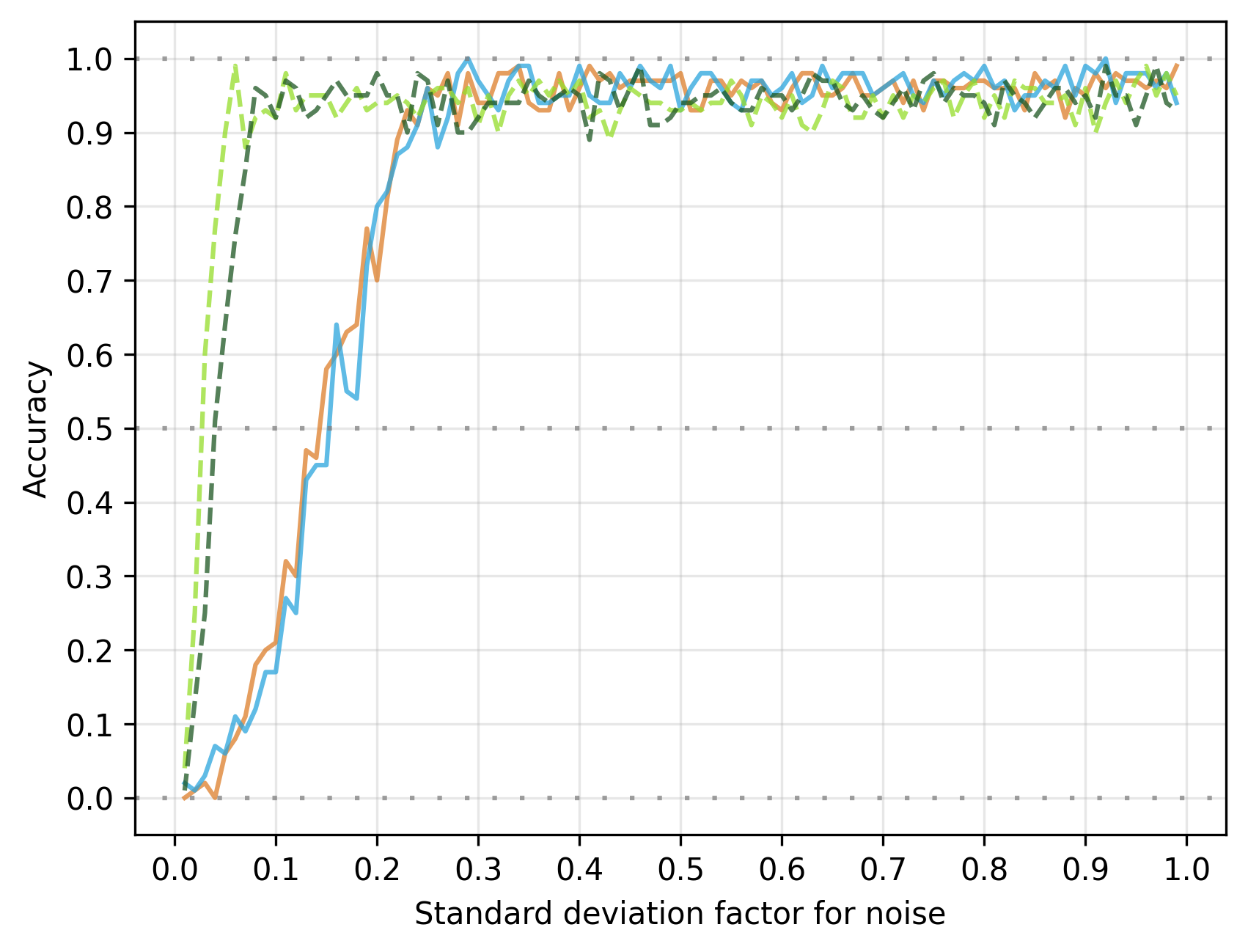}
\end{subfigure}%
\begin{subfigure}{.5\textwidth}
  \centering
  \includegraphics[scale=0.5]{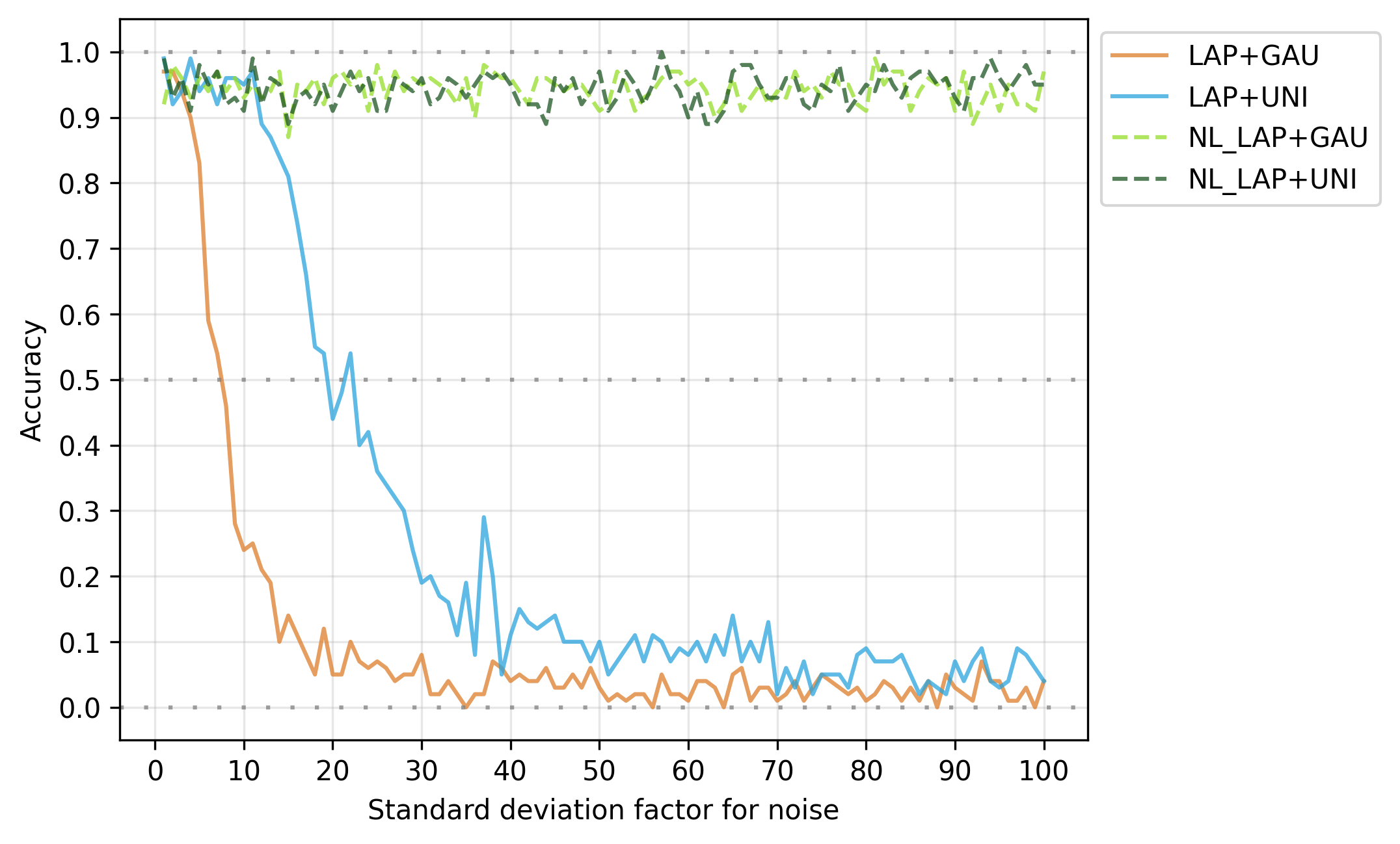}
\end{subfigure}
\caption{$Y = X + N_y$. Contains all cases where $X \sim \mathcal{L}$ and $N_y \cancel{\sim} \mathcal{L}$.
Dashed lines are non-linear cases. Coupled estimation.}
\label{fig:44}
\end{figure}

The next paragraphs describe \cref{fig:41} - \cref{fig:44} which show the performance
of using coupled estimation. Only differences to the decoupled estimation counter part will be
described.
\cref{fig:41} shows all cases where both $X$ and $N_y$ are drawn from the same distribution.
For $i < 1$ all cases (except Gaussian noise only) reach over 90\% accuracy earlier,
$i \in [0.08;1]$ for the non-linear equation models and $i \in [0.2;1]$ for the UNI and LAP models.
Furthermore, NLGAU drops more slowly for higher $i$.
\cref{fig:42} shows all cases where $X \sim \mathcal{N}$ and $N_y \cancel{\sim} \mathcal{N}$. 
For $i < 1$ all cases reach over 90\% accuracy earlier. Also both linear models now reach
an accuracy over 90\%, for $i \in [0.25;5]$ (GAU+LAP) and for $i \in [0.4;7]$ (GAU+UNI).
Both non-linear equation models drop more slowly now for higher $i$.

\newpage

\cref{fig:43} shows all cases where $X \sim \mathcal{U}$ and $N_y \cancel{\sim} \mathcal{U}$.
For $i < 1$ all cases reach over 90\% accuracy earlier. However, both linear equation models
show the strongest improvement and have now a consistent accuracy over 90\% for $i \in [0.1;1]$.
\cref{fig:44} shows all cases where $X \sim \mathcal{L}$ and $N_y \cancel{\sim} \mathcal{L}$.
For $i < 1$ all cases reach over 90\% accuracy earlier. Again, both linear equation models
show the strongest improvement and have now a consistent accuracy over 90\% for $i \in [0.25;1]$.

\subsubsection*{Summary and Conclusion}
\cref{summarytable3} follows the same scheme as \cref{summarytable1}. However, columns designate
decoupled and coupled estimation instead.
In general, non-linear cases are often better identifiable than the linear cases. Our results
show the differences quite well and further confirm that \textit{"nonlinearities in the data-generating
process are in fact a blessing rather than a curse"} - \citet{hoyer2008nonlinear}.
Without the prior assumption models can be well identifiable but are also affected by different
noise levels.
Also note that in the decoupled estimation we used only 200 samples in the independence test, whereas
in the coupled estimation we used the entire set consisting of 1000 samples. For independence tests
this can make a difference when considering computation time. Additionally, the changes in terms
of improvement between decoupled and coupled estimations are bigger in linear equation
models than in non-linear structural causal models. Therefore, for linear structural equations
we generally advise against decoupled estimation, and for non-linear structural equations decoupled estimation
can almost always be considered. Again, the biggest advantage of decoupled estimation is lesser computation
time, and if the difference of computation time between decoupled and coupled estimation is not significant,
then we suggest to go with coupled estimation.

\begin{table}[h]
\begin{center}
\begin{tabular}{|c|c|c|}
     \hline
     \textbf{Combinations} & \textbf{Decoupled} & \textbf{Coupled} \\\hline
     
     \textbf{GAU} & &  \\\hline
     
     \textbf{UNI} & 0.38 - 3 & 0.16 - 6 \\\hline
     
     \textbf{LAP} & 0.57 - 2 & 0.23 - 4 \\\hline
     
     \textbf{NLGAU} & 0.21 - 37 & 0.09 - 74 \\\hline
     
     \textbf{NLUNI} & 0.14 - 4 & 0.05 - 7 \\\hline
     
     \textbf{NLLAP} & 0.13 - 100 & 0.05 - 100 \\\hline
     
     \textbf{GAU+UNI} & & 0.39 - 7 \\\hline
     
     \textbf{GAU+LAP} & & 0.24 - 4 \\\hline
     
     \textbf{NL\_GAU+UNI} & 0.3 - 74 & 0.15 - 100 \\\hline
     
     \textbf{NL\_GAU+LAP} & 0.28 - 33 & 0.12 - 60 \\\hline
     
     \textbf{UNI+GAU} & 0.42 - 0.84 & 0.13 - 3 \\\hline
     
     \textbf{UNI+LAP} & 0.24 - 1 & 0.07 - 4 \\\hline
     
     \textbf{NL\_UNI+GAU} & 0.08 - 1 & 0.04 - 4 \\\hline
     
     \textbf{NL\_UNI+LAP} & 0.09 - 1 & 0.05 - 3 \\\hline
     
     \textbf{LAP+GAU} & & 0.23 - 4 \\\hline
     
     \textbf{LAP+UNI} & 0.64 - 4 & 0.24 - 12 \\\hline
     
     \textbf{NL\_LAP+GAU} & 0.12 - 100 & 0.05 - 100 \\\hline
     
     \textbf{NL\_LAP+UNI} & 0.15 - 100 & 0.07 - 100 \\\hline
     
\end{tabular}
\end{center}
\caption{Summary Table for RESIT \& different noise levels \& Coupled estimation. The numbers reflect the ranges of noise that allow identifiability with accuracy around 90\%.}
\label{summarytable3}
\end{table}
\newpage
\section{RESIT with different means}\label{res4}
One question which emerged was how different means for the variables in the equation
of additive noise models have an impact on the outcome of the RESIT method. 
In the previous experiments (\cref{res1} and \cref{res2}), we always used linear regression
which is computational wise very cheap and showed very good results in linear and non-linear
structural causal models. Many recent scientific papers state that any regression method can be used
and explicitly or implicitly indicate that means of 0 are assumed in the data (both artificial and natural data).
Although, from a mathematical point of view one can already guess that for linear equations there should be no difference
and for the non-linear equations identifiability might only be observable around 0 mean
for the $X$ variable. Nonetheless, we conducted a range of experiments to analyze how different means for the noise term have an impact on the outcome of RESIT.
Again, we have two scenarios, the first one with the prior assumption we made in \cref{res1} (that is
only one causal direction must be present in the bivariate case) and the second scenario without this assumption.

\subsection{Setup}
The setup here is a bit different than in \cref{1Setup}.
For all empirical tests we generate data following $X \to Y$ and generate linear as well as non-linear
cases for each test. Thus, we have the following structural causal model:
$$Y = \beta X + N_y$$ 
$$(Y = \beta X^3 + N_y \text{ for the non-linear case})$$ 
where
$$\beta = 1,$$ and
$$X \sim \begin{cases}
	    \mathcal{N}(\mu_X, 1)& \text{or} \\
	    \mathcal{U}(\mu_X-1, \mu_X+1)& \text{or}\\
	    \mathcal{L}(\mu_X, 1)
	 \end{cases}$$ and

$$N_y \sim \begin{cases}
            \mathcal{N}(\mu_N, 1)& \text{or} \\
            \mathcal{U}(\mu_N-1, \mu_N+1)& \text{or}\\
            \mathcal{L}(\mu_N, 1)
         \end{cases}$$ where $\mu_X, \mu_N \in \{-100, -90, -80, \dots, 100\}$. 
         Note that standard deviation (range between lower and upper bound in the uniform case) for both
         variables are the same. This equals the case where $i = 1$ which showed one of the best results
         in \cref{res1} and in \cref{res2} so we focus only on these cases for this experiment setup.
         The algorithm used is the same as in \cref{res1} and we also make use of linear regression
         and the 12 estimators introduced under \cref{1Setup}.\\

\subsection{Execution}
As previously mentioned, the means $\mu_X$ and $\mu_N$ are drawn from a set of 21 values:
$$\mu_X, \mu_N \in \{-100, -90, -80, \dots, 100\}$$
Thus we have 441 different combinations and for each combination a linear and a non-linear model of 
different combinations of distribution types have been tested
(18 in total, these are again the same as in \cref{1Init}).

\subsection{Results}
In the following figures the y-axis shows the accuracy ($\frac{\text{\#successful tests}}{100}$) and 
the x-axis shows all possible combinations of the means for the $X$ and $N_y$ variable, in the following
form: $\mu_X / \mu_N$.
If plots of the estimators are close to $0.5$ accuracy then this means that in 50\% 
of the tests the algorithm decided the correct direction (and vice versa 50\% the wrong direction) 
and thus indicates \textbf{unidentifiability}. If plots are closer to accuracy $1$ then we have very 
good/consistent \textbf{identifiability}. This time we will not cover each individual model because our results confirmed
that there is no change at all in terms of accuracy, and results are showing that if $\mu_X$ is significantly different from 0
then for non-linear cases the models becomes unidentifiable.

\subsubsection*{Individual analysis}
\begin{figure}[h]
\centering
\begin{subfigure}{.5\textwidth}
  \centering
  \includegraphics[scale=0.5]{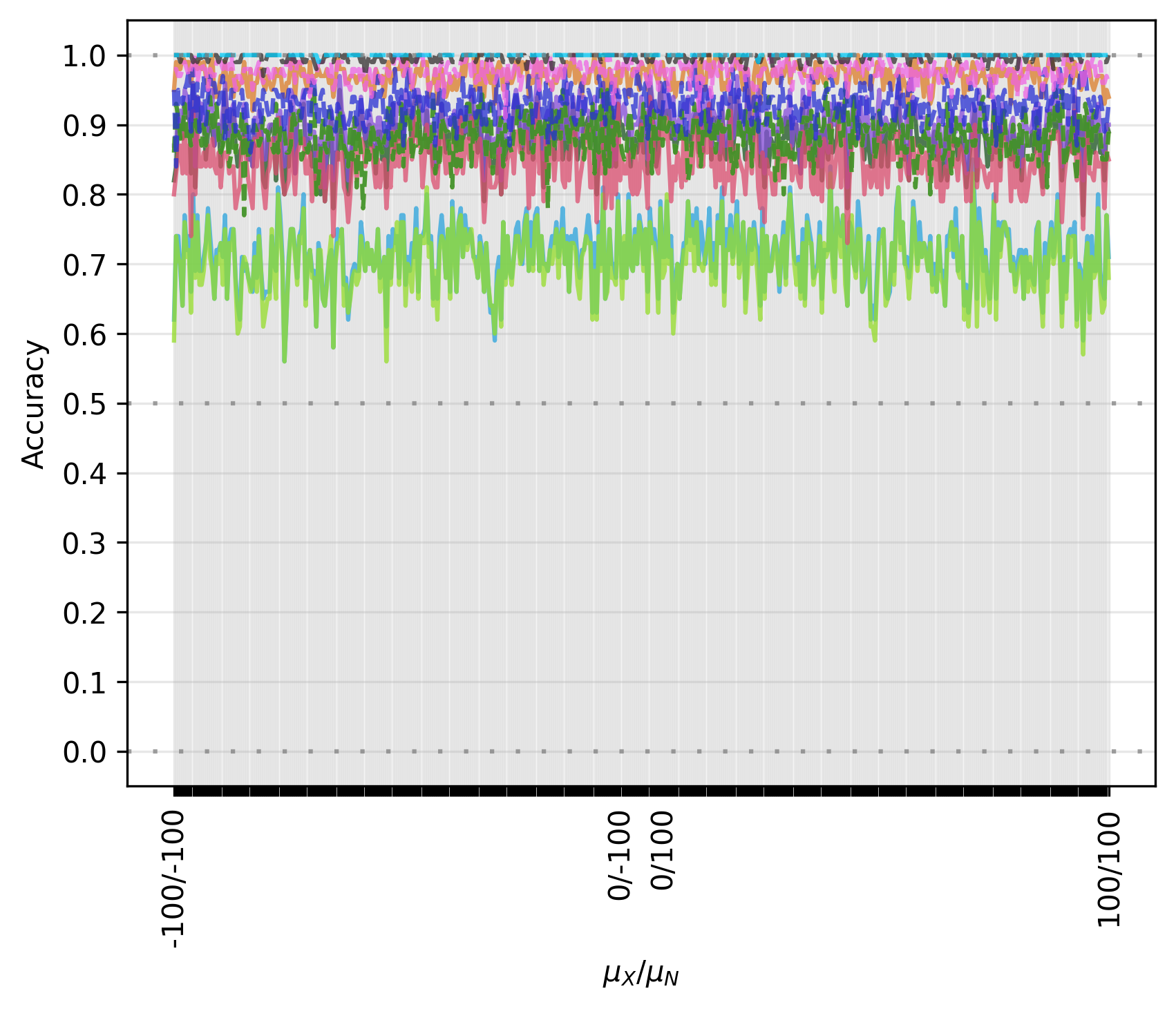}
\end{subfigure}%
\begin{subfigure}{.5\textwidth}
  \centering
  \includegraphics[scale=0.5]{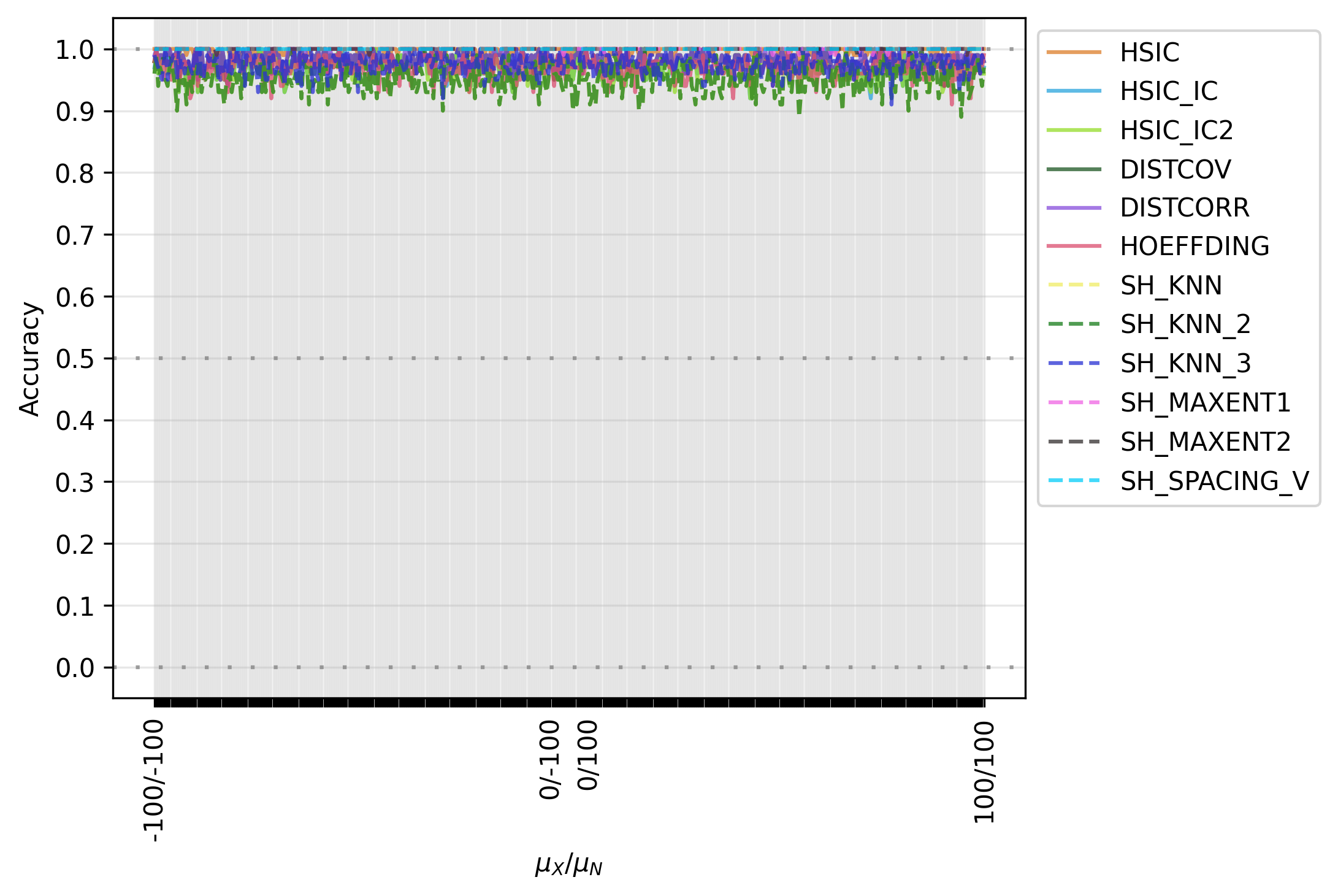}
\end{subfigure}
\caption{Left: $X \sim \mathcal{N}, N_y \sim \mathcal{U}$. Right: $X \sim \mathcal{U}, N_y \sim \mathcal{L}$.}
\label{fig:45}

\centering
\begin{subfigure}{.5\textwidth}
  \centering
  \includegraphics[scale=0.5]{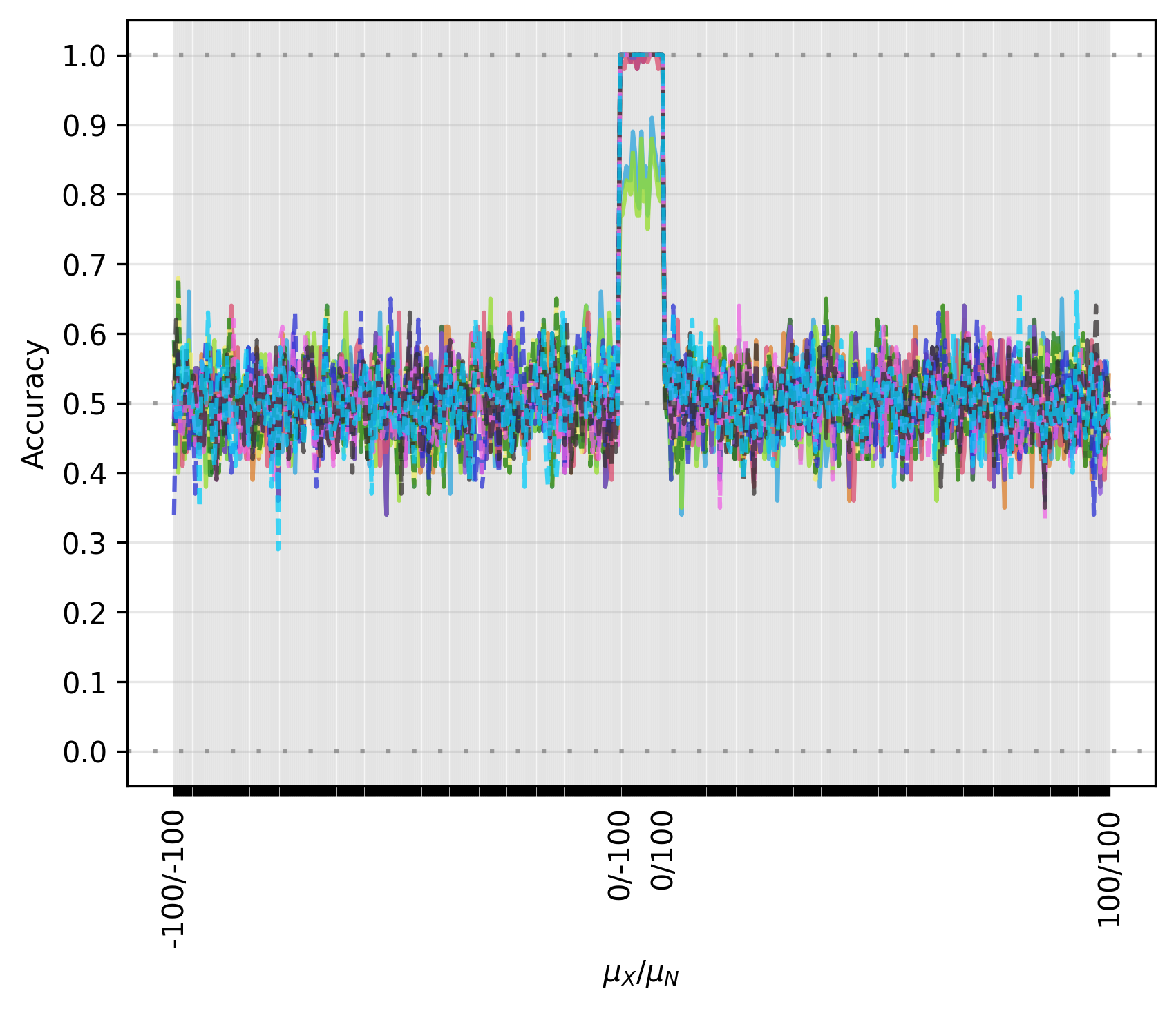}
\end{subfigure}%
\begin{subfigure}{.5\textwidth}
  \centering
  \includegraphics[scale=0.5]{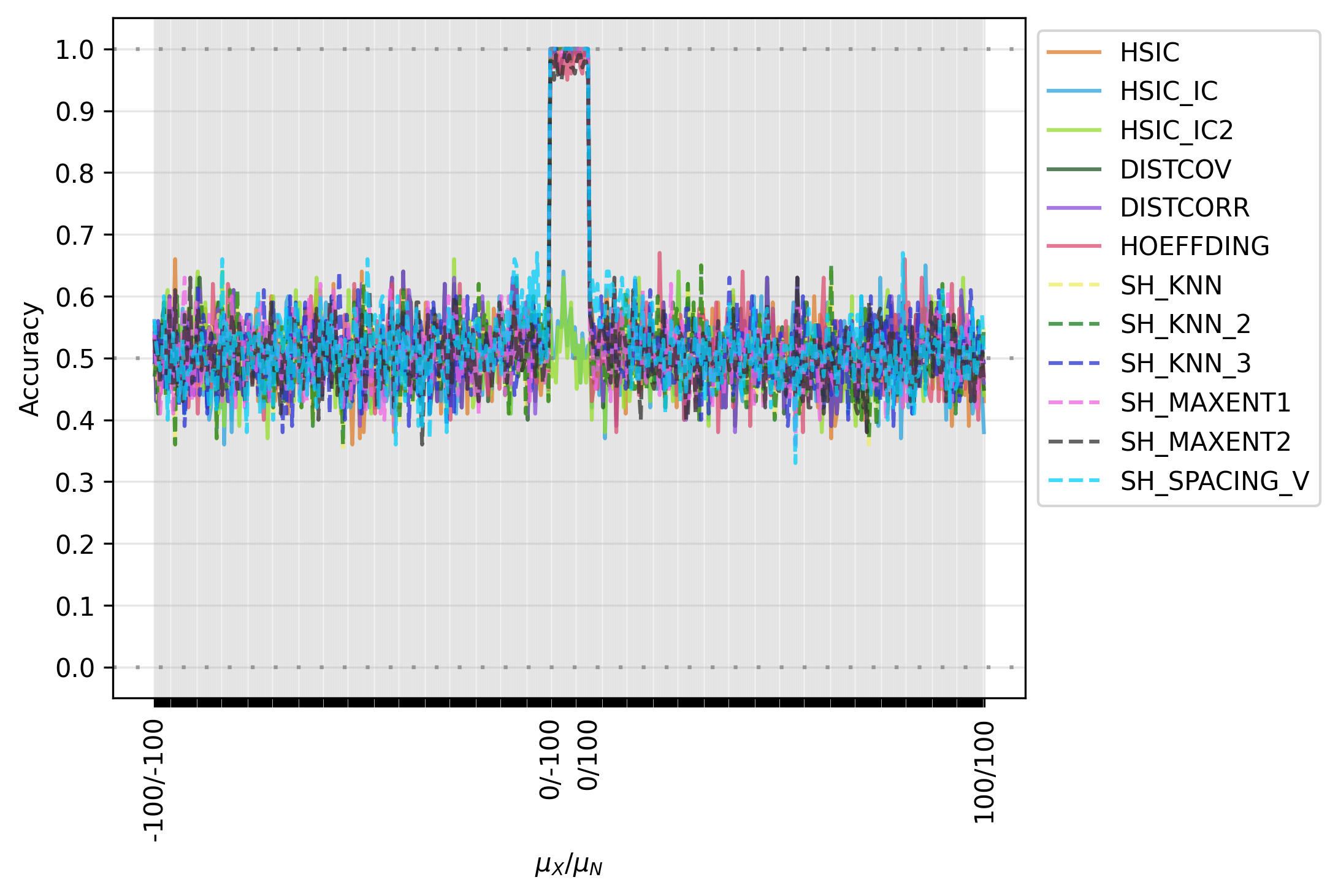}
\end{subfigure}
\caption{Left: $X \sim \mathcal{U}^3, N_y \sim \mathcal{U}$. Right: $X \sim \mathcal{U}^3, N_y \sim \mathcal{N}$.}
\label{fig:46}
\end{figure}

\cref{fig:45} contains two linear models: on the left we have $Y = \mathcal{N} + \mathcal{U}$
and on the right $Y = \mathcal{U} + \mathcal{L}$. Now if we compare them to \cref{fig:2} and
\cref{fig:9}, for $i = 1$, respectively, we can see that all estimators remain at the same accuracy.
\cref{fig:46} contains two non-linear models: on the left we have $Y = \mathcal{U}^3 + \mathcal{U}$
and on the right $Y = \mathcal{U}^3 + \mathcal{N}$. For both models if $\mu_X = 0 \land \mu_N \in [-100;
100]$ then we have identifiability and otherwise the models are unidentifiable as expected. However,
in the left model HSIC\_IC only reaches $\sim$ 80\% accuracy and for the right model HSIC\_IC does not
achieve identifiability at all, same as in \cref{fig:10} and \cref{fig:11}, respectively.\\

\subsubsection*{Summary and Conclusion}
As expected the causal relationship directions are recoverable for the linear case on the
entire test range for the means. However, for the non-linear case the relationship directions 
are only recoverable for $\mu_X = 0$. Furthermore, for the non-linear case the estimators HSIC\_IC
and HSIC\_IC2 perform badly, similar to the previous results (see \cref{summarytable1,summarytable2,summarytable3}).

\chapter{Experiments on Identification using Conditional Variances}\label{park}
As already mentioned in \cref{1Init} the problem of the identifiability 
of the linear Gaussian structural equation model has only been recently tackled
successfully. \citet{10.1093/biomet/ast043}, \citet{loh2013highdimensional}
and \citet{ghoshal2017learning} proved that Gaussian linear
structural equation models with equal or known error variances are identifiable,
and \citet{ghoshal2018learning} and \citet{chen2019causal} proved
that Gaussian linear structural equation models with unknown heterogeneous
error variances can be identifiable. In this chapter, we will discuss and test
the algorithm of \citet{park2019identifiability} and \citet{park2020identifiability}
on bivariate additive noise models,
which proved the identifiability of Gaussian structural equation models with 
both homogeneous and heterogeneous unknown error variances.

\section{Introduction}
Let
$$G_1: Y = \beta X + N_Y$$
$$G_2: X = \beta Y + N_X$$
$$G_3: X = N_X \text{ and } Y = N_Y$$
be the three possible structural equation models for any bivariate additive 
noise models (acyclic). For $G_1$ \citet{park2020identifiability} states
that if the error variance ratio satisfies $\frac{Var(Y)}{Var(X)} > (1 - \beta^2)$
then from the law of total variance we have the following two conditions:
$$(A) \text{ } Var(Y) = \mathbb{E}(Var(Y|X)) + Var(\mathbb{E}(Y|X)) = Var(Y) + \beta^2 Var(X) > Var(X)$$

\begin{align*}
(B) \text{ } \mathbb{E}(Var(X|Y)) &= Var(X)-Var(\mathbb{E}(X|Y)) \\
                                  &= Var(X) - \frac{\beta^2 Var(X)^2}{\beta^2Var(X)+Var(Y)} < Var(Y) = \mathbb{E}(var(Y|X))
\end{align*}
For $(A)$ the intuition here is that the level of uncertainty of $X$ is lower
than the level of uncertainty of $Y$ since $X$ has only one random source and
$Y$ has two random sources. For $(B)$ the intuition is that after eliminating
the other variable effect the level of uncertainty of $X$ is smaller than the
level of uncertainty of $Y$. This is because after eliminating the effect of $X$
the remaining part of $Y$ is $N_Y$ and when eliminating the effect of $Y$ on $X$
the remaining part is $N_X$ ($=X$) since $Y$ contains some information of $N_X$.
Therefore, one can order the observed variables and one can always recover
the true ordering as long as $\frac{Var(Y)}{Var(X)} > (1 - \beta^2)$, even if
the error variances are different.\\
In the same fashion, one can find the true ordering for $G_2$ as long as
$\frac{Var(X)}{Var(Y)} > (1 - \beta^2)$. For the last structural equation model
$G_3$, we have no guarantee which marginal or conditional variance is bigger
and therefore any ordering is considered correct.\\
Lastly, after determining the orderings for each graph, we simply need to
test for the presence of an edge by verifying the dependence relationships
between variables. For the graphs $G_1$ and $G_2$, the variables $X$ and $Y$ are
dependent and for the graph $G_3$ the variables are independent. Thus one can
recover the true graphs.

\subsection{Algorithm}
The algorithm used for the experiments in this work is composed
of Algorithm 2 and Algorithm 3 from \citet{park2020identifiability}.
As previously hinted, the algorithm is composed of two parts: 1) ordering and
2) conditional independence testing.

\subsubsection*{1) Ordering - \textit{Backward step wise selection}}
For the ordering step the paper \citet{park2020identifiability} proposes
two algorithms (Algorithm 1 and 2 from \citet{park2020identifiability}):\\\\
\textbf{Theorem 2 (Identifiability Conditions for ANMs)} \textit{Let $P(X)$ be 
generated from an Additive Noise Model with directed acyclic graph G and true
ordering $\pi$. Suppose that causal minimality holds. Then, G is uniquely
identifiable if either of the two following conditions is satisfied: For
any node $j= \pi_m \in V,k \in De(j)$, and $l \in An(j)$,}
$$(A) \text{ Forward step wise selection: } \sigma^2_j < \sigma^2_k + \mathbb{E}(Var(\mathbb{E}(X_k|X_{Pa(k)}) | X_{\pi_1},\dots,X_{\pi_{m-1}}))$$, or
$$(B) \text{ Backward step wise selection: } \sigma^2_j > \sigma^2_l - \mathbb{E}(Var(\mathbb{E}(X_l|X_{\pi_1}, \dots, X_{\pi_m} \text{\textbackslash} X_l) | X_{Pa(l)}))$$
For $(A)$ an additive noise model is identifiable if the conditional variance of a node $j$
is smaller than that of its descendant, De($j$), given the non-descendants, Nd($j$). This can
be understood that the variance of $N_j$ is overestimated owing to lack of parents.
For $(B)$ an additive noise model is identifiable if the conditional variance of a node
$j$ given its parents, Pa($j$), is bigger than that of its ancestor, An($j$), given
the union of its parents and any of its descendants.\\

In this work we used \textit{Backward step wise selection} (\cref{algo2}) as it was more convenient
to implement in Python. First we have our set $S$ which contains all nodes from 
the additive noise model. We iterate over this set $S$ and for each
node we calculate its conditional variance given all other remaining nodes in the set $S$. We select
the node with the highest conditional variance and append it to 
the ordering $\pi$ and also remove it from the set $S$. With the updated set $S$
we repeat the process of finding the node with the highest conditional variance,
append it to the ordering $\pi$ and remove it from $S$. This is repeated until
$S$ is empty. Lastly, the \textit{reverse} of the ordering $\pi$ is returned.
(The first node to be appended to the ordering is actually the last one in the ordering,
therefore "\textit{backward step wise selection}" is used).

\subsubsection*{2) Uncertainty Scoring}
The second step consists of the uncertainty scoring (\cref{algo3}).
Here, we iterate over the ordering $\pi$ and for each node $j$ we perform conditional independence tests
with each other node $l$ appearing before node $j$ in the ordering $\pi$. If any node $l$ is dependent on $j$, then
we add node $l$ to Pa($j$). Here, the first node in the ordering never has a parent and we start the iteration
at the second node in the ordering. For the conditional independence test, Fisher's z-transform
of the partial correlation is used. In the following sections we refer to this 
algorithm as Park algorithm or simply Park.

\begin{algorithm}
    \caption{Backward step-wise selection}\label{algo2}
    \begin{algorithmic}[1]
        \State \textbf{Input:} All variables from an ANM: $X = (x_1, x_2, \dots, x_n)$
        \State \textbf{Output:} Estimated ordering $\pi = (\pi_1, \pi_2, \dots, \pi_n)$ 
        \State
        \State \textbf{Procedure}
        \State Set $S = \{1,2, \dots, n\}$
        \State List $\pi = [\hspace{2mm}]$
        \State \textbf{for} $m = 1 \dots n \textbf{ do}$
        \State \hspace{5mm} \textbf{for} $j \in S$ \textbf{do}
        \State \hspace{10mm} Estimate the conditional variance $x_j$ given $\{x_1, \dots, x_n\}\textbackslash x_j$, $\sigma^2_{j|S\textbackslash j}$
        \State \hspace{5mm} \textbf{end}
        \State \hspace{5mm} Append $\pi_m = argmax_j \sigma^2_{j|S\textbackslash j}$ to $\pi$
        \State \hspace{5mm} Update $S = S\textbackslash \pi_m$
        \State \textbf{end}
        \State Reverse list $\pi$
    \end{algorithmic}
\end{algorithm}

\begin{algorithm}
    \caption{Uncertainty Scoring}\label{algo3}
    \begin{algorithmic}[1]
        \State \textbf{Input:} All variables from an ANM: $X = (x_1, x_2, \dots, x_n)$
        \State \textbf{Output:} Dictionary with estimated parents for all variables: $G = \{Pa(x_1): [\dots], Pa(x_2): [\dots], \dots, Pa(x_n): [\dots]\}$ 
        \State
        \State \textbf{Procedure}
        \State 1) Get ordering from backward step-wise selection:
        \State $\pi = (\pi_1, \pi_2, \dots, \pi_n)$
        \State
        \State 2) Parents estimation
        \State $G = \{\}$
        \State \textbf{for} $m = 2 \dots n \textbf{ do}$
        \State \hspace{5mm} $Pa(\pi_m) = [\text{  }]$
        \State \hspace{5mm} \textbf{for} $j = 1 \dots m - 1$ \textbf{do}
        \State \hspace{10mm} Conditional independence test between $\pi_m$ and $\pi_j$ given $\{\pi_1, \dots, \pi_{m-1}\} \textbackslash \pi_j$
        \State \hspace{10mm} If dependent, include $\pi_j$ into $Pa(\pi_m)$
        \State \hspace{5mm} \textbf{end}
        \State \hspace{5mm} Insert $Pa(\pi_m)$ into $G$
        \State \textbf{end} 
    \end{algorithmic}
\end{algorithm}

\section{Setup}
The setup is the same as in \cref{1Setup}.

\subsection{Execution}
The execution is the same as in \cref{1Init} except that we use the Park algorithm instead.
Here, when using the Park algorithm the output
will be a list of parent sets for all nodes in the additive noise model:
$$\text{Output: } [(Pa(X), Pa(Y)].$$ A test is then successful if and only
if $$(Pa(X) = \{\}) \land (Pa(Y) = \{X\}).$$

\subsection{Results}
In the following figures the y-axis shows the accuracy ($\frac{\text{\#successful tests}}{100}$) and 
the x-axis shows the range of the $i$ factor. Each figure contains two subfigures, the left with
$i \in \{0.01, 0.02, \dots, 1.00\}$ and the right figure with $i \in \{1, 2, \dots, 100\}$.
Differently from \cref{res1}, if HSIC is closer to 0 then we have \textbf{unidentifiability.} 
If plots are closer to accuracy $1$ then we have very good/consistent \textbf{identifiability}. 
The next subsection describes each figure individually and the following subsection thereafter 
provides a summary and a small conclusion.

\newpage
\subsubsection*{Individual Analysis}
\begin{figure}[!h]
\centering
\begin{subfigure}{.5\textwidth}
  \centering
  \includegraphics[scale=0.5]{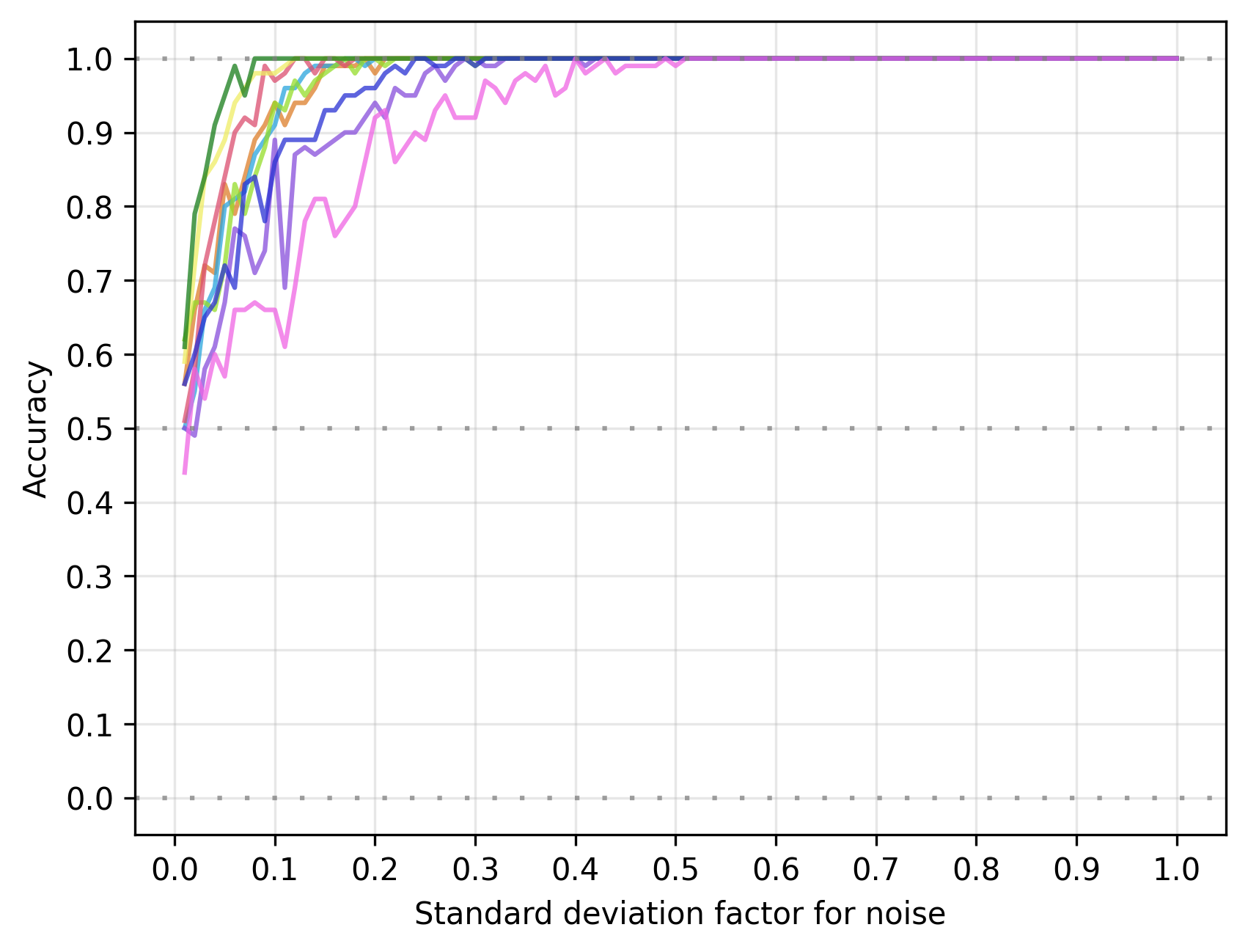}
\end{subfigure}%
\begin{subfigure}{.5\textwidth}
  \centering
  \includegraphics[scale=0.5]{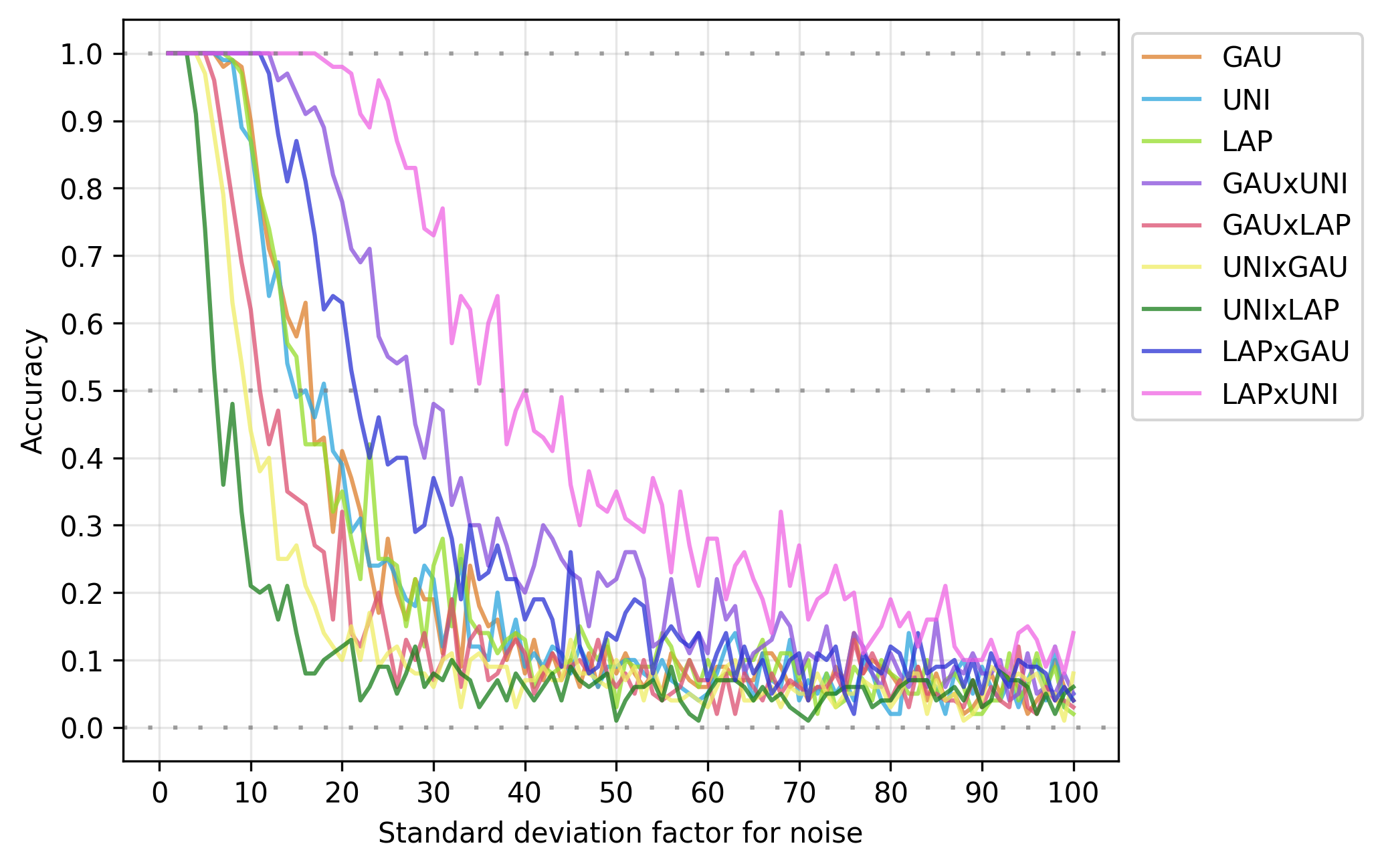}
\end{subfigure}
\caption{$Y = X + N_y$. Contains all linear cases.}
\label{fig:47}

\centering
\begin{subfigure}{.5\textwidth}
  \centering
  \includegraphics[scale=0.5]{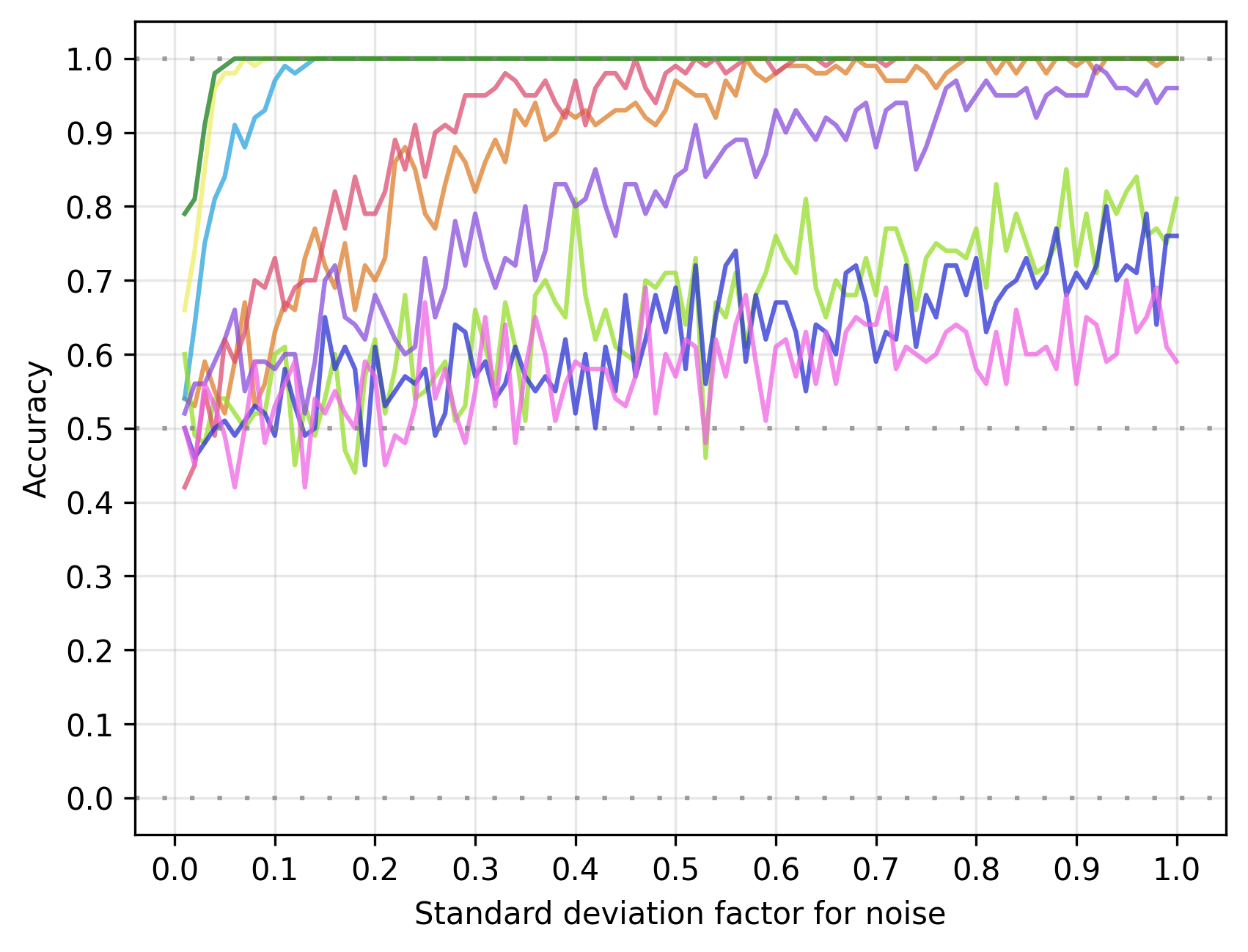}
\end{subfigure}%
\begin{subfigure}{.5\textwidth}
  \centering
  \includegraphics[scale=0.5]{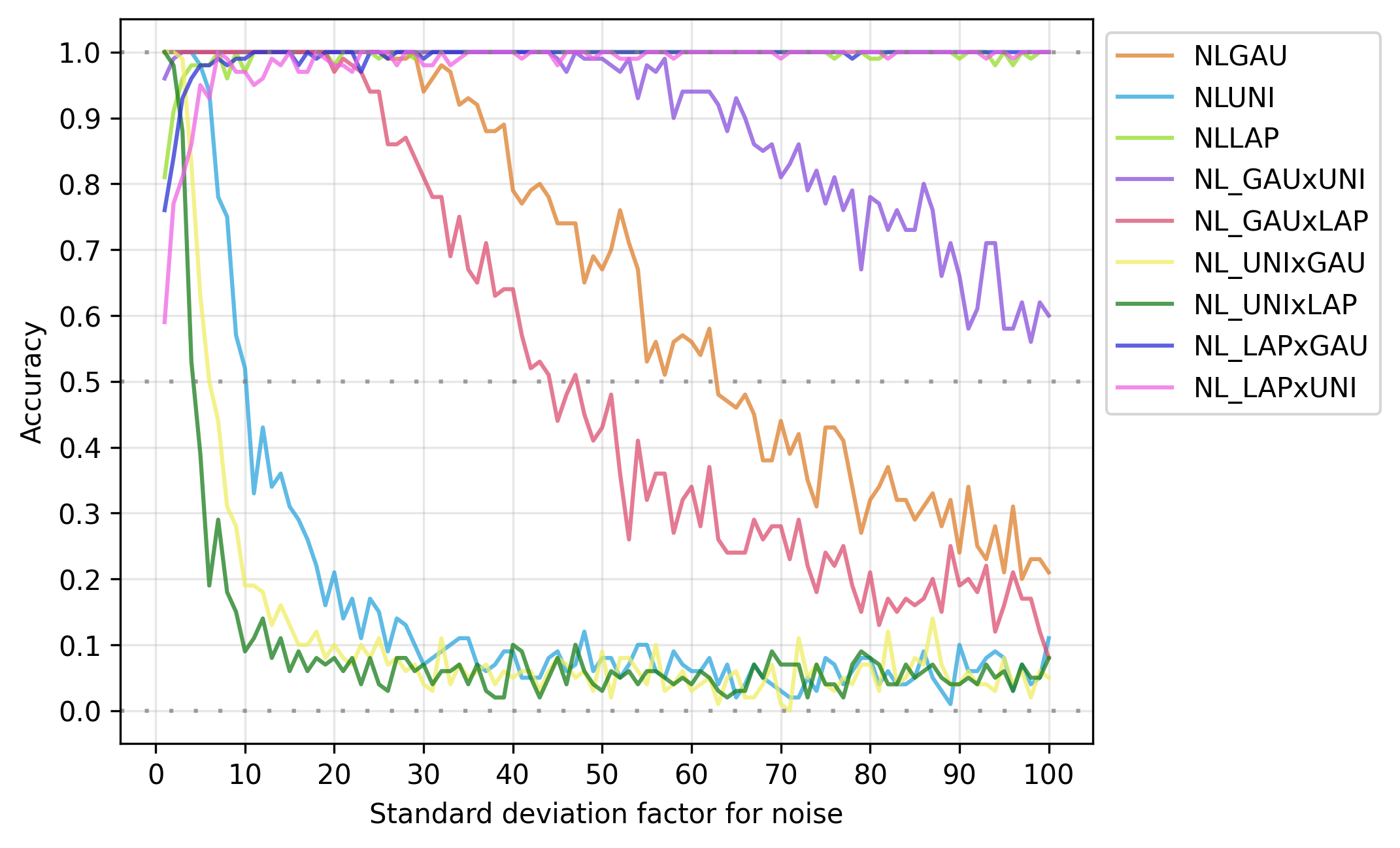}
\end{subfigure}
\caption{$Y = X^3 + N_y$. Contains all non-linear cases.}
\label{fig:48}
\end{figure}

\cref{fig:47} shows the results for all linear cases. For $i \in [0.5; 3]$ all 
linear structural equation models are always identifiable. For $i < 0.5$ all 
models slowly drop towards 50\% accuracy. For $i > 3$ all cases start dropping in terms of accuracy. 
Some do drop faster (UNIxLAP, UNIxGAU, GAUxLAP, GAU, UNI, LAP) and others more slowly (LAPxGAU, GAUxUNI, LAPxUNI).
For $i > 5$ the first case (UNIxLAP) is below 90\% accuracy and for $i > 25$ all cases are below
90\% accuracy.

\cref{fig:48} shows the results for non-linear cases. Here, the results are quite different
and varying. NL\_UNIxLAP,NL\_UNIxGAU and NLUNI have accuracy 100\% for $i \in [0.12; 1]$. After
$i = 1$ they drop fast below 50\% accuracy. NLGAU and NL\_GAUxLAP reach an accuracy over 90\%
for $i \in [0.35; 25]$ and slowly drop outside this range in terms of accuracy. NL\_GAUxUNI has accuracy over 90\%
for $i \in [0.77; 58]$ and slowly drops outside this range. The last three, NLLAP, NL\_LAPxGAU and NL\_LAPxUNI
only reach an accuracy higher than 90\% for $i \in [3; 100]$.

\subsubsection*{Summary and Conclusion}
\cref{summarytable4} follows the same scheme as \cref{summarytable3}.
Interestingly, for $i \in [0.01; 1]$ the linear cases perform better than the non-linear
cases and for $i \in [1; 100]$ several non-linear cases perform better than most linear cases,
which wasn't the case for RESIT. Furthermore, as the results show, after specific noise levels
the identifiability from the Park algorithm starts to suffer and thus not all graphs
can be successfully recovered. The reason for this lies within step 2, the conditional independence test.
If noise levels are significantly different then the independence test fails to capture the correlation
between the two nodes and therefore concludes that the nodes are independent (Type II Error).
However, for any given $i$, the ordering step always performs correctly.\footnote{A quick test in python shell, with $i = 57$,
$X \sim \mathcal{L}$ and $N_y \sim \mathcal{U}$ and 100 repetitions showed that in these runs the ordering was always correct but only in 35 runs (from the 100 repetitions) the independence tests were correct.}

\begin{table}[h]
\begin{center}
\begin{tabular}{|c|c|}
     \hline
     \textbf{Combinations} & \textbf{Coupled} \\\hline
     
     \textbf{GAU} & 0.08 - 10  \\\hline
     
     \textbf{UNI} & 0.1 - 8 \\\hline
     
     \textbf{LAP} & 0.1 - 10  \\\hline
     
     \textbf{GAU+UNI} & 0.16 - 10 \\\hline
     
     \textbf{GAU+LAP} & 0.05 - 6 \\\hline
     
     \textbf{UNI+GAU} & 0.04 - 5 \\\hline
     
     \textbf{UNI+LAP} & 0.03 - 3  \\\hline
     
     \textbf{LAP+GAU} & 0.14 - 13 \\\hline
     
     \textbf{LAP+UNI} & 0.19 - 26 \\\hline
     
     \textbf{NLGAU} & 0.33 - 37 \\\hline
     
     \textbf{NLUNI} & 0.05 - 6 \\\hline
     
     \textbf{NLLAP} & 2 - 100  \\\hline
     
     \textbf{NL\_GAU+UNI} & 0.52 - 67  \\\hline
     
     \textbf{NL\_GAU+LAP} & 0.23 - 25  \\\hline
     
     \textbf{NL\_UNI+GAU} & 0.04 - 4 \\\hline
     
     \textbf{NL\_UNI+LAP} & 0.03 - 3  \\\hline
     
     \textbf{NL\_LAP+GAU} & 4 - 100  \\\hline
     
     \textbf{NL\_LAP+UNI} & 5 - 100  \\\hline
     
\end{tabular}
\end{center}
\caption{Summary Table for Uncertainty Scoring \& different noise levels \& Coupled estimation. The numbers reflect the ranges of noise that allow identifiability with accuracy around 90\%.}
\label{summarytable4}
\end{table}

\chapter{Conclusion and Future Work}
The results from the experiments showed that two analyzed
causal discovery methods (RESIT and Uncertainty Scoring) are affected by different noise scales.
For significantly small noise levels in the disturbance term $N_y$ (almost
deterministic data), or significantly
high noise levels, these causal discovery methods fail to capture
the true causal relationship of the given structural equation model.
Recall that \textit{significantly} here depends on the model. For example, on some
models if the noise level was already twice larger then the methods failed to
determine the causal direction consistently. Other models remained identifiable
with 100 times higher noise levels (The range of different noise levels in the experiments
was quite exhaustive, and realistically speaking having noise levels
100 times higher then the potential cause variable is very rare\footnote{Additionally,
with very high noise levels the effect of the cause variable is very likely
negligible anyways}). 
This also shows different behavior for different distribution types
(e.g., Gaussian or Uniform). Therefore, if observed data differs significantly
in terms of variances scientists need to be careful when analyzing causal
relationships and in drawing conclusions.\\

In \cref{resit} we tested RESIT in the bi-variate case using several estimators (12 in total).
We compared decoupled estimation (splitting data into training and test data) and coupled estimation
(no splitting) in \cref{res1} and \cref{res3}. In \cref{res2} we omitted a particular assumption
which allowed us to use entropy estimators and therefore restricted our setting in using only
independence tests. All results from these sections are quite similar. In general, if the standard deviation
of the noise term is smaller than the standard deviation of the cause then models remained identifiable
as opposed to the case when the standard deviation of the noise term was bigger. For example, often when the standard deviation
of the noise term was only half of that of the cause, the model was still identifiable. However, in several
cases, if the standard deviation of the noise term was already twice larger than the standard deviation of the cause, then the model became unidentifiable. We also tested linear and non-linear models and our results show that non-linear models
were still identifiable in situations where a linear model were unidentifiable. For example, some non-linear
models, where the noise term's standard deviation was 100 times higher than that of the cause, were still perfectly
identifiable while their linear counterpart wasn't. Similar results can be observed in the Uncertainty Scoring method (\cref{park}).\\
Specific to RESIT, we also tested whether splitting data into training and test data was beneficial.
The reason to do this is to use only the test data in the estimators and therefore reduce the overall computation time of the algorithm. Hence, we analysed the identifiability performance between decoupled estimation and
coupled estimation. Our results show that there is a performance drop from coupled to decoupled estimation.
However, in linear cases, this drop is bigger than in non-linear cases. Also, our results show that decoupled
estimation for non-linear cases has almost no significant impact on the performance of RESIT. For linear
models we generally advise to use coupled estimation. In the end this is a trade-off between performance 
and computation speed and should be carefully selected depending on the individual experiment. Lastly,
for RESIT we used several estimators (6 independence estimators and 6 entropy estimators). Our results show differences in terms
of performance in these estimators. In our experiments Hilbert-Schmidt Independence Criterion with RBF Kernel
was the best independence estimator, and Shannon entropy estimator using Vasicek's spacing method was
the best entropy estimator.\\

\textbf{Future work.} In our experiments, we tested only two particular methods
and three different distribution types. Similar results are expected with other
methods for causal discovery for additive noise models, as the failing point
is the independence estimation (or entropy estimation) in the two used methods.
Therefore, methods relying on these estimations are generally prone to errors for some levels
of noise. Furthermore, this work does not formalize the effect of different
noise levels in ANM causal discovery methods but could be done in a future work.

In reality, observed data does not always strictly follow a certain distribution type. 
As there are many different combinations possible, it would be interesting
to generalize the impact of different noise levels on any distribution by using
the different properties an observed distribution exhibits.

\newpage
\section*{Acknowledgement}
I want to express my many thanks to Dr. Marharyta Aleksandrova. I enjoyed working on this topic a lot, and I am grateful
to have had Dr. Aleksandrova as my advisor. She is a very proactive and professional researcher, and I learned a lot from her. Furthermore, she gave me a lot of feedback on my thesis text, which helped me improve the quality of the thesis text a lot.

\newpage
\normalem
\printbibliography

@article{hoyer2008nonlinear,
  title={Nonlinear causal discovery with additive noise models},
  author={Hoyer, Patrik and Janzing, Dominik and Mooij, Joris M and Peters, Jonas and Sch{\"o}lkopf, Bernhard},
  journal={Advances in neural information processing systems},
  volume={21},
  pages={689--696},
  year={2009}
}

@misc{loh2013highdimensional,
      title={High-dimensional learning of linear causal networks via inverse covariance estimation},
      author={Po-Ling Loh and Peter Bühlmann},
      year={2013},
      eprint={1311.3492},
      archivePrefix={arXiv},
      primaryClass={stat.ML}
}

@article{10.1093/biomet/ast043,
    author = {Peters, J. and Bühlmann, P.},
    title = "{Identifiability of Gaussian structural equation models with equal error variances}",
    journal = {Biometrika},
    volume = {101},
    number = {1},
    pages = {219-228},
    year = {2013},
    month = {11},
    issn = {0006-3444},
    doi = {10.1093/biomet/ast043},
    url = {https://doi.org/10.1093/biomet/ast043},
    eprint = {https://academic.oup.com/biomet/article-pdf/101/1/219/17460568/ast043.pdf},
}

@misc{park2019identifiability,
      title={Identifiability of Gaussian Structural Equation Models with Homogeneous and Heterogeneous Error Variances}, 
      author={Gunwoong Park and Younghwan Kim},
      year={2019},
      eprint={1901.10134},
      archivePrefix={arXiv},
      primaryClass={stat.ML}
}

@article{shimizu2006linear,
  title={A linear non-Gaussian acyclic model for causal discovery.},
  author={Shimizu, Shohei and Hoyer, Patrik O and Hyv{\"a}rinen, Aapo and Kerminen, Antti and Jordan, Michael},
  journal={Journal of Machine Learning Research},
  volume={7},
  number={10},
  year={2006}
}

@misc{shimizu2014direct,
      title={A direct method for estimating a causal ordering in a linear non-Gaussian acyclic model}, 
      author={Shohei Shimizu and Aapo Hyvarinen and Yoshinobu Kawahara},
      year={2014},
      eprint={1408.2038},
      archivePrefix={arXiv},
      primaryClass={cs.LG}
}

@inproceedings{mooij2009regression,
  title={Regression by dependence minimization and its application to causal inference in additive noise models},
  author={Mooij, Joris and Janzing, Dominik and Peters, Jonas and Sch{\"o}lkopf, Bernhard},
  booktitle={Proceedings of the 26th annual international conference on machine learning},
  pages={745--752},
  year={2009}
}

@inproceedings{kpotufe2014consistency,
  title={Consistency of causal inference under the additive noise model},
  author={Kpotufe, Samory and Sgouritsa, Eleni and Janzing, Dominik and Sch{\"o}lkopf, Bernhard},
  booktitle={International Conference on Machine Learning},
  pages={478--486},
  year={2014},
  organization={PMLR}
}

@article{mooij2016distinguishing,
  title={Distinguishing cause from effect using observational data: methods and benchmarks},
  author={Mooij, Joris M and Peters, Jonas and Janzing, Dominik and Zscheischler, Jakob and Sch{\"o}lkopf, Bernhard},
  journal={The Journal of Machine Learning Research},
  volume={17},
  number={1},
  pages={1103--1204},
  year={2016},
  publisher={JMLR. org}
}

@misc{zhang2012identifiability,
      title={On the Identifiability of the Post-Nonlinear Causal Model}, 
      author={Kun Zhang and Aapo Hyvarinen},
      year={2012},
      eprint={1205.2599},
      archivePrefix={arXiv},
      primaryClass={stat.ML}
}

@ARTICLE{szabo14information,
  AUTHOR =       {Zolt{\'a}n Szab{\'o}},
  TITLE =        {Information Theoretical Estimators Toolbox},
  JOURNAL =      {Journal of Machine Learning Research},
  YEAR =         {2014},
  volume =       {15},
  pages =        {283-287},
}

@article{peters2014causal,
  title={Causal discovery with continuous additive noise models},
  author={Peters, Jonas and Mooij, Joris M and Janzing, Dominik and Sch{\"o}lkopf, Bernhard},
  year={2014}
}

@book{cover1999elements,
  title={Elements of information theory},
  author={Cover, Thomas M},
  year={1999},
  publisher={John Wiley \& Sons}
}

@article{ghoshal2017learning,
  title={Learning identifiable gaussian bayesian networks in polynomial time and sample complexity},
  author={Ghoshal, Asish and Honorio, Jean},
  journal={arXiv preprint arXiv:1703.01196},
  year={2017}
}

@inproceedings{ghoshal2018learning,
  title={Learning linear structural equation models in polynomial time and sample complexity},
  author={Ghoshal, Asish and Honorio, Jean},
  booktitle={International Conference on Artificial Intelligence and Statistics},
  pages={1466--1475},
  year={2018},
  organization={PMLR}
}

@article{chen2019causal,
  title={On causal discovery with an equal-variance assumption},
  author={Chen, Wenyu and Drton, Mathias and Wang, Y Samuel},
  journal={Biometrika},
  volume={106},
  number={4},
  pages={973--980},
  year={2019},
  publisher={Oxford University Press}
}

@article{park2020identifiability,
  title={Identifiability of Additive Noise Models Using Conditional Variances.},
  author={Park, Gunwoong},
  journal={Journal of Machine Learning Research},
  volume={21},
  number={75},
  pages={1--34},
  year={2020}
}

@article{judea2000causality,
  title={Causality: models, reasoning, and inference},
  author={Judea, Pearl},
  journal={Cambridge University Press. ISBN 0},
  volume={521},
  number={77362},
  pages={8},
  year={2000}
}

@book{spirtes2000causation,
  title={Causation, prediction, and search},
  author={Spirtes, Peter and Glymour, Clark N and Scheines, Richard and Heckerman, David},
  year={2000},
  publisher={MIT press}
}

@book{bookspirtes1993,
author = {Spirtes, Peter and Glymour, Clark and Scheines, Richard},
year = {1993},
month = {01},
pages = {},
title = {Causation, Prediction, and Search},
volume = {81},
isbn = {978-1-4612-7650-0},
journal = {Causation, Prediction, and Search},
doi = {10.1007/978-1-4612-2748-9}
}

@article{DBLP:journals/corr/abs-1301-3857,
  author    = {Nir Friedman and
               Iftach Nachman},
  title     = {Gaussian Process Networks},
  journal   = {CoRR},
  volume    = {abs/1301.3857},
  year      = {2013},
  url       = {http://arxiv.org/abs/1301.3857},
  archivePrefix = {arXiv},
  eprint    = {1301.3857},
  bibsource = {dblp computer science bibliography, https://dblp.org}
}

@inproceedings{kano2003causal,
  title={Causal inference using nonnormality},
  author={Kano, Yutaka and Shimizu, Shohei},
  booktitle={Proceedings of the international symposium on science of modeling, the 30th anniversary of the information criterion},
  pages={261--270},
  year={2003}
}

@article{wright1921correlation,
  title={Correlation and causation},
  author={Wright, Sewall},
  journal={J. agric. Res.},
  volume={20},
  pages={557--580},
  year={1921}
}

@inproceedings{sun2006causal,
  title={Causal inference by choosing graphs with most plausible Markov kernels},
  author={Sun, Xiaohai and Janzing, Dominik and Sch{\"o}lkopf, Bernhard},
  booktitle={Ninth International Symposium on Artificial Intelligence and Mathematics (AIMath 2006)},
  pages={1--11},
  year={2006}
}

@article{sun2008causal,
  title={Causal reasoning by evaluating the complexity of conditional densities with kernel methods},
  author={Sun, Xiaohai and Janzing, Dominik and Sch{\"o}lkopf, Bernhard},
  journal={Neurocomputing},
  volume={71},
  number={7-9},
  pages={1248--1256},
  year={2008},
  publisher={Elsevier}
}

@article{janzing2009telling,
  title={Telling cause from effect based on high-dimensional observations},
  author={Janzing, Dominik and Hoyer, Patrik O and Sch{\"o}lkopf, Bernhard},
  journal={arXiv preprint arXiv:0909.4386},
  year={2009}
}

@article{stegle2010probabilistic,
  title={Probabilistic latent variable models for distinguishing between cause and effect},
  author={Stegle, Oliver and Janzing, Dominik and Zhang, Kun and Mooij, Joris M and Sch{\"o}lkopf, Bernhard},
  journal={Advances in neural information processing systems},
  volume={23},
  pages={1687--1695},
  year={2010},
  publisher={Citeseer}
}

@article{daniusis2012inferring,
  title={Inferring deterministic causal relations},
  author={Daniusis, Povilas and Janzing, Dominik and Mooij, Joris and Zscheischler, Jakob and Steudel, Bastian and Zhang, Kun and Sch{\"o}lkopf, Bernhard},
  journal={arXiv preprint arXiv:1203.3475},
  year={2012}
}

@article{mooij2011causal,
  title={On causal discovery with cyclic additive noise model},
  author={Mooij, Joris M and Janzing, Dominik and Heskes, Tom and Sch{\"o}lkopf, Bernhard},
  year={2011},
  publisher={[Sl]: NIPS}
}

@article{janzing2012information,
  title={Information-geometric approach to inferring causal directions},
  author={Janzing, Dominik and Mooij, Joris and Zhang, Kun and Lemeire, Jan and Zscheischler, Jakob and Daniu{\v{s}}is, Povilas and Steudel, Bastian and Sch{\"o}lkopf, Bernhard},
  journal={Artificial Intelligence},
  volume={182},
  pages={1--31},
  year={2012},
  publisher={Elsevier}
}

@article{hyvarinen2013pairwise,
  title={Pairwise likelihood ratios for estimation of non-Gaussian structural equation models},
  author={Hyv{\"a}rinen, Aapo and Smith, Stephen M},
  journal={Journal of Machine Learning Research},
  volume={14},
  number={Jan},
  pages={111--152},
  year={2013}
}

@article{shimizu2014lingam,
  title={LiNGAM: Non-Gaussian methods for estimating causal structures},
  author={Shimizu, Shohei},
  journal={Behaviormetrika},
  volume={41},
  number={1},
  pages={65--98},
  year={2014},
  publisher={Springer}
}

@article{nowzohour2016score,
  title={Score-based causal learning in additive noise models},
  author={Nowzohour, Christopher and B{\"u}hlmann, Peter},
  journal={Statistics},
  volume={50},
  number={3},
  pages={471--485},
  year={2016},
  publisher={Taylor \& Francis}
}

@inproceedings{sgouritsa2015inference,
  title={Inference of cause and effect with unsupervised inverse regression},
  author={Sgouritsa, Eleni and Janzing, Dominik and Hennig, Philipp and Sch{\"o}lkopf, Bernhard},
  booktitle={Artificial intelligence and statistics},
  pages={847--855},
  year={2015},
  organization={PMLR}
}

@article{rebane,
  author    = {George Rebane and
               Judea Pearl},
  title     = {The Recovery of Causal Poly-Trees from Statistical Data},
  journal   = {CoRR},
  volume    = {abs/1304.2736},
  year      = {2013},
  url       = {http://arxiv.org/abs/1304.2736},
  archivePrefix = {arXiv},
  eprint    = {1304.2736},
  bibsource = {dblp computer science bibliography, https://dblp.org}
}

@inproceedings{gretton2005,
  title={Measuring statistical dependence with Hilbert-Schmidt norms},
  author={Gretton, Arthur and Bousquet, Olivier and Smola, Alex and Sch{\"o}lkopf, Bernhard},
  booktitle={International conference on algorithmic learning theory},
  pages={63--77},
  year={2005},
  organization={Springer}
}
\end{document}